\documentclass{article}

\usepackage{microtype}
\usepackage{graphicx}
\usepackage{subfigure}
\usepackage{booktabs} 

\usepackage[hyphens]{url}
\usepackage{hyperref}

\usepackage[accepted]{icml2024}

\usepackage{amsmath}
\usepackage{amssymb}
\usepackage{mathtools}
\usepackage{amsthm}

\usepackage[capitalize,noabbrev]{cleveref}

\theoremstyle{plain}

\theoremstyle{definition}

\theoremstyle{remark}

\usepackage[textsize=tiny]{todonotes}

\usepackage{tikz}
\usepackage{pgfplots}
\usepackage{listings}
\usepackage{float}
\usepackage{adjustbox}
\usepackage[T1]{fontenc}
\usepackage{bm}
\usepackage{siunitx}
\usepackage{capt-of}
\usepackage{enumitem}
\usepackage{titlesec}
\usepackage{empheq}
\allowdisplaybreaks

\newcommand*\widefbox[1]{\fbox{\hspace{2em}#1\hspace{2em}}}
\newcommand{\ord}[1]{$\mathcal{O}(#1)$}
\newcommand{\ordb}[1]{$\bm{\mathcal{O}(#1)}$}

\newcommand{\stoptocwriting}{%
  \addtocontents{toc}{\protect\setcounter{tocdepth}{-5}}}
\newcommand{\resumetocwriting}{%
  \addtocontents{toc}{\protect\setcounter{tocdepth}{\arabic{tocdepth}}}}

\pgfplotsset{compat=1.18}
\icmltitlerunning{Transformers Get Stable: An End-to-End Signal Propagation Theory for Language Models}

\begin{document}

\twocolumn[
\icmltitle{Transformers Get Stable: An End-to-End \\ Signal Propagation Theory for Language Models}



\icmlsetsymbol{equal}{*}

\begin{icmlauthorlist}
\icmlauthor{Akhil Kedia}{equal,samsung}
\icmlauthor{Mohd Abbas Zaidi}{equal,samsung}
\icmlauthor{Sushil Khyalia}{equal,cmu}
\icmlauthor{JungHo Jung}{samsung}
\icmlauthor{Harshith Goka}{samsung}
\icmlauthor{Haejun Lee}{samsung}
\end{icmlauthorlist}

\icmlaffiliation{samsung}{Language Intelligence Lab, Samsung Research, Seoul, South Korea.}
\icmlaffiliation{cmu}{Department of Computer Science, Carnegie Mellon University, Pittsburgh, Pennsylvania, USA (work done while at Samsung)}

\icmlcorrespondingauthor{Akhil Kedia}{akhil.kedia@samsung.com}

\icmlkeywords{Machine Learning, ICML}

\vskip 0.3in
]



\printAffiliationsAndNotice{\icmlEqualContribution} 

\begin{abstract}
In spite of their huge success, transformer models remain difficult to scale in depth. In this work, we develop a unified signal propagation theory and provide formulae that govern the moments of the forward and backward signal through the transformer model. Our framework can be used to understand and mitigate vanishing/exploding gradients, rank collapse, and instability associated with high attention scores. We also propose DeepScaleLM, an initialization and scaling scheme that conserves unit output/gradient moments throughout the model, enabling the training of very deep models with $1000$ layers. We find that transformer models could be much deeper -- our deep models with fewer parameters outperform shallow models in Language Modeling, Speech Translation, and Image Classification, across encoder-only, decoder-only and encoder-decoder variants, for both Pre-LN and Post-LN transformers, for multiple datasets and model sizes. These improvements also translate into improved performance on downstream Question Answering tasks and improved robustness for Image Classification.

\end{abstract}

\stoptocwriting
\section{Introduction}

Transformer models are extremely popular across different domains of machine learning, however, deep transformers are plagued with issues of gradient explosion/vanishing~\citep{gopher, NormFormerImprovedTransformerb, UsingDeepSpeedMegatrona, LayerNormalizationsResiduala, UsingDeepSpeedMegatrona, opt, ScalingVisionTransformers, palm, adamInstability, smallProxy} and rank collapse~\citep{DeepViTDeeperVision,noci2022signal} that adversely affect training stability. Proposed remedies include residual scaling, changing initialization or extra/modified layernorms~\citep{DSInit,preln,ReZeroAllYou,deepnet,ScalingVisionTransformers}. 


Theoretical analysis via signal propagation and kernel methods has led to an improved understanding of these issues. Several works in the signal propagation domain~\citep{glorot,NormalizationPropagation,xu2019understanding, AttentionNotAll, Catformer, AntiOversmoothingDeepVision} have analysed the propagation of moments for some components of deep transformers, but often make simplifying assumptions of IID inputs, uncorrelated outputs, ignoring effect of query/key initialization, simplifying non-linearity, etc. We observed break down of each of these assumptions with real world data, adversely affecting model stability. 



These issues highlight the need for a holistic theoretical framework that can fully explain signal propagation through transformer models with real data. In this work, we provide a framework to fully explain signal propagation through transformer models, by deriving closed-form expressions for the first and second-order moments (mean and variance) of the outputs and gradients of each of the components of the transformer model (Embeddings, FFN, ReLU/GeLU, LayerNorm, Dropout, Softmax, Single-Head Attention), Attention and FFN blocks, and through the entire model. Our derived equations are empirically verified within strict error bounds with real world data\footnote{Code: \href{https://github.com/akhilkedia/TranformersGetStable}{https://github.com/akhilkedia/TranformersGetStable}}.



We apply this framework to understand and mitigate instability issues with deep transformers -- vanishing/exploding gradients, rank collapse, and instability caused by high QK values. To harness the improved complexity of deeper models~\citep{montufar2014number,ExponentialExpressivity,ExpressivePowerDeepa}, we propose DeepScaleLM, a novel initialization scheme that augments residual/output scaling, and ensures the moments of outputs and gradients remain fully conserved throughout the model. DSLM enables us to break the depth barrier and train models with $100$s of layers which outperform shallow models for BERT, GPT, Encoder-Decoder models across text, vision and speech modalities.

\begin{table*}[t]
\caption{Signal propagation for forward and backward passes through components of a transformer (GeLU in \cref{proof: gelu}). The expressions here are illustrative simplification of full closed form formulae in \cref{appendix:section:summary_table,appendix:section:Moment_prop_components}.}
\begin{center}
\begin{adjustbox}{max width=\linewidth}
\begin{tabular}{l c c c c c} 
 \toprule
 \textbf{Component} & $\mathbf{\mu_{x_{\text{out}}}}$ & $\mathbf{\sigma^2_{x_{\text{out}}}}$ & $\mathbf{\sigma^2_{g_{\text{in}}}}$  & $\mathbf{r^l_{x_{\text{out}}}}$  & $\mathbf{r^l_{g_{\text{in}}}}$ \\ [2.0ex] 
 \midrule
    Embeddings &  0 & $ \sum\sigma^2_{w_{\text{embd}}} $ & - & $\dfrac{\pi^2}{18*\mathrm{log}(|V|)^2} + \dfrac{2}{9}$ & - \\ [2.5ex] 
    
    Linear ($d_{\text{in}} \rightarrow d_{\text{out}}$)&  0 & $d_{\text{in}}  \sigma_w^2  (\sigma_{x_{\text{in}}}^2 + \mu_{x_{\text{in}}}^2)$ & $d_{\text{out}}  \sigma_w^2  \sigma_{g_{\text{out}}}^2 $ & $\dfrac{r^l_{x_{\text{in}}} + \mu_{x_{\text{in}}}^2/\sigma^2_{x_{\text{in}}}}{1 + \mu_{x_{\text{in}}}^2/\sigma^2_{x_{\text{in}}}}$ & $r^l_{g_{\text{out}}}$ \\ [2.5ex] 
    
    ReLU &  $\dfrac{\sigma_{x_{\text{in}}}}{\sqrt{(2\pi)}}$ & $\dfrac{(\pi-1)}{(2\pi)}  \sigma_{x_{\text{in}}}^2$ & $\dfrac{1}{2}\sigma_{g_{\text{out}}}^2$ & $0.7r^l_{x_{\text{in}}} + 0.3{r^l_{x_{\text{in}}}}^2$ & $(\dfrac{1}{2} + \dfrac{\sin^{-1}{(r^l_{x_{\text{in}}})}}{\pi})\mathrm{r^l_{g_{\text{out}}}}$ \\ [2.5ex] 
    
    
    LayerNorm ($d$)&  0 & 1 & $\dfrac{\sigma^2_{g_{\text{out}}}}{\sigma^2_{x_{\text{in}}}}$ & $r^l_{x_{\text{in}}}$ & $r^l_{g_{\text{out}}}$ \\ [2.5ex]

    Dropout ($p$) &  $\mu_{x_{in}}$ & $\dfrac{\sigma_{x_{\text{in}}}^2 + p  \mu_{x_{\text{in}}}^2}{1-p}$ & $\dfrac{1}{1-p}\sigma_{g_{\text{out}}}^2$ &$\dfrac{r^l_{x_{\text{in}}}(1-p)}{1 + p  \mu_{x_{\text{in}}}^2/\sigma_{x_{\text{in}}}^2}$ &$(1-p)r^l_{g_{\text{out}}}$ \\ [2.5ex] 

    SHA-without V & 0 & $r^l_{x_{\text{in}}}\sigma^2_{x_{\text{in}}}$ & $r^l_{g_{{\text{out}}}}\sigma^2_{g_{{\text{out}}}}$ & $1$ & $1$ \\ [2.5ex]


    Softmax &  $\dfrac{1}{L}$ & $\dfrac{e^{(1 - r^d_{x_{\text{in}}})\sigma^2_{x_{\text{in}}}} - 1}{L^2}$ & $ \dfrac{e^{(1 - r^d_{x_{\text{in}}})\sigma_{x_{\text{in}}}^2}}{L^2}\sigma_{g_{\text{out}}}^2 $ & - & - \\ [2.5ex] 

    
\bottomrule
\end{tabular}
\end{adjustbox}
\end{center}
\label{table:moment_full_formulas1}
\end{table*}

\begin{table*}[t]
\caption{Moment Propagation through the blocks of a transformer layer. Exact closed forms / proofs are provided  in \cref{appendix:section:summary_table,appendix:section:Moment_prop_blocks}.}
\begin{center}
\begin{adjustbox}{max width=\linewidth}
\begin{tabular}{c c c c c} 
 \toprule
 \textbf{Component} & $\mathbf{\sigma^2_{x_{out}}}$ & $\mathbf{r^l_{x_{out}}}$ & $\mathbf{\sigma^2_{g_{in}}}$ & $\mathbf{r^l_{g_{in}}}$\\ [0.8ex] 
 \midrule
    Attention Block & $\dfrac{d^2\sigma_o^2\sigma_v^2\sigma^2_{x_{\text{in}}} *r^l_{x_{\text{in}}}}{(1-p)}$ & $ 1-p $  & $\dfrac{d^2\sigma_o^2\sigma_v^2*\sigma^2_{g_{\text{out}}}}{(1-p)}r^l_{g_{\text{out}}}$ & $1-p$ \\ [2.5ex] 
    FFN Block & $\dfrac{2d^2\sigma_{w_1}^2\sigma_{w_2}^2\sigma^2_{x_{\text{in}}}}{(1-p)}$ & $ (1-p)(\dfrac{1}{\pi} + \dfrac{r^l_{x_{\text{in}}}}{2} + (\dfrac{1}{2}-\dfrac{1}{\pi}){r^l_{x_{\text{in}}}}^2) $ & $\sigma^2_{x_{\text{out}}}*\sigma^2_{g_{\text{out}}}$ & $(1-p) (\frac{1}{2} + \frac{\sin^{-1}{(r^l_{x_{\text{in}}})}}{\pi})r^l_{g_{\text{out}}}$ \\ [0.8ex] 
\bottomrule
\end{tabular}
\end{adjustbox}
\end{center}
\label{table:block_moments}
\end{table*}

\section{Moments of Transformer Models}
\label{proposed_method}

\subsection{Moments of Transformer Components} 
Following an analysis similar to that of Xavier initialization~\citep{glorot}, we derive closed-form expressions for the mean and variance of the output and of the backpropagated gradient for all the components of the transformer model in \autoref{table:moment_full_formulas1}.

Here $\mu_{x_{\text{in}}}$, $\sigma_{x_{\text{in}}}^2$, $\mu_{x_{\text{out}}}$, $\sigma^2_{x_{\text{out}}}$ are the mean and variance of the input/outputs, $\sigma_{g_{\text{out}}}^2$, $\sigma^2_{g_{\text{in}}}$ are the variance of the gradient back-propagated to/from the component, and $r^l$, $r^d$ are the correlations across sequence length and hidden dimension. $p$ is the dropout probability, $L$ sequence length, $d_{\text{in}}, d_{\text{out}}$ input/output dimensions of Linear layer, $\sigma_w^2$, $\sigma^2_{w_{\text{embd}}}$ variances of the weights of the Linear layer and the Embeddings table. At the input side, $r^l_{x_{\text{in}}}$ originates from repeated tokens. For text, we estimate input correlation theoretically by assuming that input tokens follow Zipf~\citep{zipf} distribution. Detailed proofs are provided in \cref{appendix:section:Moment_prop_components}, and all assumptions are summarized in \cref{section:appendix:assumptions}.


\begin{figure*}[ht]
\begin{minipage}[ht]{0.22\linewidth}
    \begin{center}
    \includegraphics[page=1]{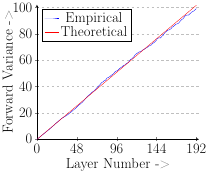}
    \caption{Pre-LN: Variance of forward signal increases linearly across layers $N$.}
    \label{fig:var_fw_pre}
    \end{center}
\end{minipage}\hfill
\begin{minipage}[ht]{0.23\linewidth}
    \begin{center}
    \includegraphics[page=2]{fig/all_figs.pdf}
    \caption{Pre-LN: Backward gradient variance increases hyperbolically across layers $N$.}
    \label{fig:var_bck_pre}
    \end{center}
\end{minipage}\hfill
\begin{minipage}[ht]{0.23\linewidth}
    \begin{center}
    \includegraphics[page=3]{fig/all_figs.pdf}
    \caption{Post-LN: Backward gradient variance vanishes exponentially (y-axis log-scale).}
    \label{fig:var_bck_post}
    \end{center}
\end{minipage}\hfill
\begin{minipage}[ht]{0.27\linewidth}
    \begin{center}
    \includegraphics[page=4]{fig/all_figs.pdf}
    \caption{DeepScaleLM: The variances remain conserved for both forward and backward pass.}
    \label{fig:var_moment}
    \end{center}
\end{minipage}
\end{figure*}

\subsection{Moments of Transformer Blocks}
Combining the expressions reported in \autoref{table:moment_full_formulas1}, we derive closed-form expressions for the moment transformation during the forward and backward pass of the transformer Attention and FFN blocks. The Attention block refers to the $Q,K,V$ projection, followed by Multi-Head Attention and Output-Projection Layer. The FFN block refers to the Linear layer followed by non-linearity (ReLU) and output Linear layer. \cref{table:block_moments} provides our derived equations for these, where $\sigma_v^2$, $\sigma_o^2$, $\sigma_{w_1}^2$, $\sigma_{w_2}^2$ are the variances for $V$ weights, Output-Projection weights, and weights of FFN block Linear layers, and $d$ is model the hidden size. These results show that considering correlation $r^l$, dropout $p$ and effects of non-linearity are crucial for correctly modelling signal propagation through Transformer blocks.

\subsection{Moments of Entire Transformer Model}
By repeatedly applying the expressions in \cref{table:block_moments} for each layer, we calculate the propagation of moments of outputs and gradients through the entire transformer model. We do this for both Pre-LN style transformers, in which the skip connection bypasses the LayerNorm, and for Post-LN style transformers, in which the Layernorm is applied before the skip-connection. The method is fully detailed in \cref{proof: vanilla preLN,proof: vanilla postLN}. \cref{fig:var_fw_pre,,fig:var_bck_pre,,fig:var_bck_post} provide the forward (left to right) and backward (right to left) signal propagation at initialization through the layers of a very deep $192$-layer model with Xavier initialization.

\subsection{Numerical Validation of Theoretical Results}
\label{numerical_verification}

We verify the theoretical formulae of transformer components and blocks by running simulations with real/synthetic data, (detailed in \cref{appendix:section:numerical_verification}, code released). Even at $99$ percentile, no error (other than SHA gradient $\sigma^2$) is larger than $10\%$, verifying our assumptions.

All our derivations are modality-agnostic. We verify our formulae for the entire transformer model using real textual MLM data, as shown in \cref{fig:var_fw_pre,,fig:var_bck_pre,,fig:var_bck_post}~(Reproducible using our released code), and using ImageNet data~(as shown in \cref{appendix:subsection:vision}). Our formulae predict the observed gradient and forward/backward norms with remarkable accuracy, with mean and median relative errors of $6.8\%$ and $5.2\%$ respectively, and an $R^2$ of $0.998$. We further verify that for model depths in range $[1-768]$, and model dimensions $[128-6096]$, the reported formulae are within $10\%$ error, even across $768$ layers of the transformer model.


\subsection{Validity of Theoretical Predictions even after Training}

\begin{figure}[h]
\begin{minipage}[t]{0.47\linewidth}
    \begin{center}
    \includegraphics[page=8]{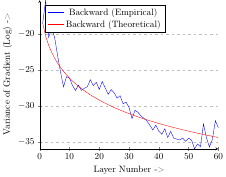}
    \caption{Backward gradient variance increases hyperbolically after $150$k train steps.}
    \label{fig:var_bck_30B_explode}
    \end{center}
\end{minipage}\hfill
\begin{minipage}[t]{0.47\linewidth}
    \begin{center}
    \includegraphics[page=9]{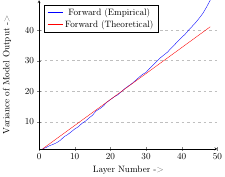}
    \caption{Linear growth in the forward pass for a $48$-layer after $100$k train steps.}
    \label{fig:var_fwd_48_explode}
    \end{center}
\end{minipage}
\end{figure}

Interestingly, our theoretical estimates hold approximately even after the models have been trained for a large number of steps. The model stays in the regime it is initialized with (as has also been shown in \citet{LearningOverparameterizedNeural, ExactComputationInfinitely, WideNeuralNetworks, EffectInitialConfiguration, FineGrainedAnalysisOptimization, qlora}), highlighting the importance of correct initialization. We analyze gradient explosion in a $30B$ parameter $64$-layer PreLN model (after $150$k training steps) and use our theory to predict the moments. Our hyperbolic estimation for the gradient explosion match closely with the observed moments as shown in \cref{fig:var_bck_30B_explode}. Similarly, forward growth in a $48$-layer $1024$-d PreLN model (after $100$k training steps) matches our linear estimations (\cref{fig:var_fwd_48_explode}). 

\section{Applications}

\subsection{Explaining Variance Explosion in Transformer}

Our approach theoretically proves the gradient vanishing/explosion (\cref{table:block_growth}) for both Pre-LN and Post-LN transformers.

\paragraph{Exploding Output and Gradient in Pre-LN} The forward output for Pre-LN transformer increases linearly with increasing depth $N$~(\cref{proof: vanilla preLN}) since each layer's output is directly added to the skip connection, as seen in \cref{fig:var_fw_pre}. For the backward pass, the gradient increases hyperbolically with increasing $N$, as seen in \cref{fig:var_bck_pre}. Intuitively, this is because the gradient increases in every layer when a block's gradient is added to the skip connection, and the fractional increase in gradient is inversely proportional to the forward variance~(which increases by $N$) because of LayerNorm.

\paragraph{Vanishing/Exploding Gradient in Post-LN} While layernorm solves the explosion in the forward pass of networks with residual connections~\citep{BatchNormalizationBiasesa}, it has the opposite impact on the gradient. As proved in \cref{proof: vanilla postLN}, the gradient in a Post-LN transformer grows/decays exponentially with the number of layers~(\cref{fig:var_bck_post}). 

Intuitively, the gradient is first transformed within the layer and then at the LayerNorm placed before the layer. The multiplicative factor is applied repeatedly, and causes gradient to vanish or explode exponentially, as was also observed in \citet{DeepInformationPropagation}. This explains why Post-LN models are more challenging to train than Pre-LN for deeper networks~\citep{deepnet, NormFormerImprovedTransformerb, LayerNormalizationsResiduala}.

\begin{table}[h]
\caption{Comparison of maximum theoretical forward pass and backward pass growth in variance for the entire transformer model across methods (See \cref{appendix:section:Moment_prop_model} for proofs). Here $\beta$ is the initial value of residual scaling for LayerScale.}
\begin{center}
\begin{adjustbox}{max width=1.0\linewidth}
\begin{tabular}{l c c c c c } 
 \toprule    
 \textbf{Method} & \multicolumn{2}{|c|}{\textbf{Post-LN}} & \multicolumn{3}{c}{\textbf{Pre-LN}}  \\
 & \multicolumn{1}{|c}{\textbf{Backward}} & \multicolumn{1}{c|}{\textbf{Sensitivity}} & \textbf{Forward} & \textbf{Backward} & \textbf{Sensitivity} \\ 
 \midrule
 Vanilla & \ord{c^{\pm N}} & \ordb{N} & \ord{N} & \ord{N} & \ordb{logN} \\ [0.8ex]
 DSInit & \ord{1} & \ord{N^{-1}} & \ord{1} & \ord{1} & \ord{N^{-1}}   \\ [0.8ex]
 LayerScale & \ord{1} & \ord{\beta N} & \ord{1} & \ord{1} & \ord{\beta N} \\ [0.8ex]
 DeepNet & \ord{1} & \ord{N^{-0.5}} & - & - & - \\ [0.8ex]
 \midrule
\textbf{DSLM (Ours)} & \ordb{1} & \ord{1}  & \textbf{1} & \ordb{1} & \ord{1}   \\ [0.8ex]
 \bottomrule
\end{tabular}
\end{adjustbox}
\end{center}
\label{table:block_growth}
\end{table}

\subsection{Explaining Higher Pruning of Deeper Layers}

\citet{unreasonableIneffectiveness} found that LLMs such as Llama-2-70B~\citep{llama2} have minimal degradation in performance on Question Answering tasks until almost half the deeper layers are removed -- suggesting that parameters in deeper layers are less effective in current LLMs. As we prove in \cref{proof: vanilla preLN}, the output of a Pre-LN transformer grows proportionally with depth (\cref{fig:var_fw_pre}). For an $80$-layer model like Llama-2, this implies the deeper layers will have a significantly reduced impact on changing the output.

\subsection{Explaining Impact of Large QK Values}
\label{QK_explosion}

In \citet{ScalingVisionTransformers}, the authors observed large QK values destabilized the training, and solved this empirically by adding a layernorm after attention scores. Unlike prior works~\citep{deepnet,noci2022signal}, note from our derivations of softmax(\cref{proof: softmax}) that the backwards gradients from Q/K are exponentially related to their variance, highlighting the critical significance of correct initialization of Q/K. For example, by initializing them to only $2$x the xavier values (all other initializations the same), backwards gradients exploded $10000$x through a $192$ layer model. Our theory explains these empirical observations, and suggests a simple initialization strategy to fix this problem, achieving the same variance on QK without the overhead of LayerNorm~(\cref{sec:DeepScaleLM}).

\subsection{Explaining and Mitigating Rank Collapse}
\label{rank_collapse} 

Similar to our work, \citet{noci2022signal} also analyze moment propagation through the transformer, and observed the rank collapse of the token's representations at initialization after just a few layers, i.e., all the token representations became the same ($r^l_x \approx 1$ after just $12$ layers) at initialization. This has also been reported in \citet{RevisitingOversmoothingBERT,DeepViTDeeperVision,AntiOversmoothingDeepVision,DeepTransformersShortcutsa, ReZeroAllYou, StabilizingTransformerTrainingc}, and suggested modifications such as adding a skip connection on attention scores, initializing Q/other weights to $0$, or normalizing all FFN weights. 

\begin{figure}[h]
\begin{minipage}[t]{0.48\linewidth}
    \begin{center}
    \includegraphics[page=5]{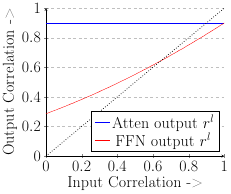}
    \caption{Forward $r^l_{x_{out}}$ for FFN and Attention blocks with $p=0.1$. FFN reduces $r^l_{x_{out}}$ for $r^l_{x_{in}}>0.65$, and attention always has $r^l_{x_{out}}<1$.}
    \label{fig:rank_theoretical}
    \end{center}
\end{minipage}\hfill
\begin{minipage}[t]{0.48\linewidth}
    \begin{center}
    \includegraphics[page=6]{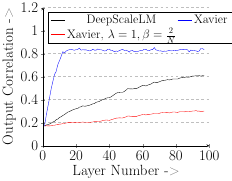}
    \caption{No rank collapse is observed with Xavier init and dropout. $r^l$ increases slower with $\beta^2=\frac{2}{N}$ or for DeepScaleLM.}
    \label{fig:rank_collapse}
    \end{center}
\end{minipage}
\end{figure}




Our theory suggests a very simple solution -- Dropout. As our closed form expressions show, both FFN block (because of ReLU) and dropout reduce the correlation (\cref{fig:rank_theoretical}). With dropout, our method shows that such a rank collapse will not occur, and $r^l_x$ will quickly reach a stable value $< 1$ (\cref{appendix:section:correlation_analysis}), as verified empirically in \cref{fig:rank_collapse}.
 

Alternatively, scaling the block output by $\beta = \frac{1}{\sqrt{N}}$, or equivalently initializing the weights very small in Post-LN will also prevent rank collapse, even without Dropout. For Pre-LN, $\lambda=1$ slows down increase in $r^l$ compared to $\lambda^2=1-\frac{1}{N}$ (but the same slowdown can be achieved by decreasing $\beta$). This highlights the criticality of correct initialization, dropout and scaling for deep transformer models, as well as the explainability power of our theoretical framework.



\subsection{DeepScaleLM: Enabling Deep Transformers }
\label{sec:DeepScaleLM}


We propose DeepScaleLM (DSLM), a new initialization / scaling scheme that alleviates the issues discussed above.

\paragraph{Residual/Skip-Connection Scaling} Let $\sigma^2_{\text{skip}}$, $\sigma^2_{\text{block}}$, $\sigma^2_{\text{model}}$ be the variances of the skip connection, the block, and the output of the final layer of the model, respectively. Let $\sigma^2_{\text{skip}} = \sigma^2_{\text{block}}$, and we scale them by scalars $\lambda$ and $\beta$ respectively. Then, as has been proven in numerous works (\cref{section:related:residual_scaling}), if $\lambda^2 + \beta^2 = 1$, this scaling will maintain the variance after addition of the residual.

\paragraph{Initialization} However while ensuring $\sigma^2_{\text{skip}} = \sigma^2_{\text{block}}$ (and equal to the variance of model input) has been done for ResNets (\cref{section:related:initialization}), it is difficult to achieve theoretically for transformers. By leveraging the equations in \cref{table:block_moments}, our theory provides us the tools to achieve this. We modify the initialization of the components of the transformer FFN and Attention blocks such that the variance of their output is $1$, as further detailed in \cref{sec: sup_pseudocode} -- 

\begin{enumerate}[itemsep=-0.15em, topsep=-0.0em]
    \item We set the variance of embedding weights as $\sigma_{e}^2 = \frac{1-p}{num_{\text{embd}}}$, where $num_{\text{embd}}$ is the number of embeddings types. As embeddings are followed by a dropout, this ensures the input variance to the model is $1$.
    \item  We set $\sigma_{w_2}^2 = \sigma_{w_1}^2 = \frac{1}{d}*\sqrt{\frac{1-p}{2}}$, to make the output of the FFN block $1$. 
    \item We iteratively calculate layer-by-layer $r^l_{x_{\text{in}}}$, $r^l_{x_{\text{out}}}$ using expressions from \cref{table:block_moments}, and calculate the initial variance of the attention block weights to make the output variance $1$.
\end{enumerate}


This initialization of transformer blocks, combined with the scaling of the skip connection and residual, and correct initialization of the embeddings, results $\sigma^2_{\text{model}}=1$, irrespective of the number of layer $N$. This initialization also preserves the backward gradient, as proved for Pre-LN and Post-LN, in \cref{proof: DeepScaleLM-static Pre-LN,,proof: DeepScaleLM-static Post-LN}. Empirically, we show the backward gradient being preserved for both Pre-LN and Post-LN even across 192 layers at initialization~(\cref{fig:var_moment}).

\begin{figure}[h]
\begin{minipage}[t]{0.5\linewidth}
    \begin{center}
    \includegraphics[page=10]{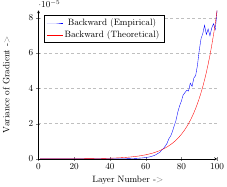}
    \caption{Gradient vanishes using $\lambda^2=0.9$ and $\beta^2=0.1$, after 50k training steps.}
    \label{fig:var_bck_vanish}
    \end{center}
\end{minipage}\hfill
\begin{minipage}[t]{0.45\linewidth}
    \begin{center}
    \includegraphics[page=11]{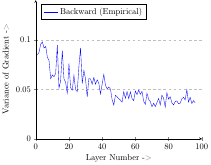}
    \caption{Gradient remains conserved using $\lambda^2=1-\frac{1}{N}$ and $\beta^2=\frac{1}{N}$, after 50k steps.}
    \label{fig:var_bck_vanish_not}
    \end{center}
\end{minipage}
\end{figure}

\paragraph{Choice of Scaling Parameters} While any choice of $\beta$ will work at initialization, higher values of $\beta$, for example $\beta^2=0.5$ causes gradients to vanish (\cref{fig:var_bck_vanish}, \cref{table: res1}). This is because covariance between residual and skip connection increases the forward variance, which causes normalization to decrease backward gradient~\citep{BatchNormalizationBiasesa}. 

Similar to other prior works (\cref{section:related:residual_scaling}), we use $\beta^2=\frac{k}{N}$ in all our experiments, where $k$ is some small constant. This enables us to bound the fall in gradient (\cref{proof: DeepScaleLM-static Pre-LN}) for Pre-LN. For Post-LN, $\beta^2 \leq \frac{k}{N^2}$ is theoretically required to bound the gradient (\cref{proof: DeepScaleLM Post-LN}). In practice, with $\beta^2=\frac{2}{N}$, even with $768$ layers, we empirically observed the final output variance from the model does not exceed $30$, and all our models converge. We hence use $\beta^2=\frac{k}{N}$ (\cref{fig:var_bck_vanish_not}), but a practitioner may choose $\beta^2=\frac{k}{N^\alpha}$, with $\alpha>1$ if more stability is required at the expense of performance/``sensitivity'' (Refer to discussion of relative strength in \cref{section:rel_strength} and comparison to prior works in \cref{sec:prior}). While the above analysis assumes positive covariance (which we always observed experimentally), negative covariance follows a similar reasoning, and will cause gradient explosion instead.

\paragraph{Preventing Rank Collapse} For DSLM, applying block equations iteratively shows that $r^l_x < 1-\frac{1}{e^2}$ after $N$ layers. 

\paragraph{Simpler Initialization} Another avenue to handle the covariance between residual and skip connection could be to set $\lambda^2 + \beta^2 < 1$. We therefore also consider a simpler initialization method(\cref{fig:pseudo_impl}), in which we modify the initialization of attention value and output matrices to be the same as those of FFN block. This decreases the "effective" $\beta$ of the attention block, but as the attention block has $2$x fewer params than FFN, this change in weightage seems reasonable. As we show in \cref{proof: DeepScaleLM Pre-LN,,proof: DeepScaleLM Post-LN} while variances are no longer unit at initialization, they are still bounded. This change does not impact performance significantly, as we show in \cref{table: ablation_sigma}. All further experiments in \cref{sec:res} used this simpler initialization.

\paragraph{Folding Scaling into Weights for Inference} The scaling parameters introduced here can be fully absorbed into the model checkpoint weights by recursively scaling layernorm gain and output linear weights, hence and do not require any changes to vanilla transformers inference code.

DeepScaleLM enables training deeper-narrower models with $100$s of layers, outperforming standard models across transformer variants, tasks and modalities.









\section{DeepScaleLM Results}
\label{sec:res}

\subsection{Improvements on Encoder-only Models (BERT)}
\label{results:DSLM}
\paragraph{Implementation Details} 
We test our method on the Masked Language Modelling task with the BERT~\citep{bert} model. Pile-CC dataset~\citep{pile} was used to pre-train our model. We use $k=2$ for $\beta$ while keeping all the original hyper-parameters of BERT the same, except for learning rate (LR). We find that higher LR is needed for our deeper-narrower models (similar to \citet{tensorprograms}). Hence, we search for LR for all the models. The training steps were decided based on Chinchilla~\citep{chinchilla}, at $6.6$B tokens. \cref{hparams_bert} provides all hyper-parameter details. For DSLM, model output was down-scaled by $\sqrt{d}$ before being passed to the LM-head. 

We train different language models with the same number of parameters and compute -- while increasing the depth ($N$), we reduce the hidden dimension $d$ keeping number of transformer parameters ($Nd^2$) constant. When changing from $12$-layer $1024$-d model to $192$-layer $256$-d model, compute negligibly increases by only $6.6\%$ when keeping $Nd^2$ constant (\cref{table:appendix:compute}), while the number of parameters decreases by $5-15\%$ due to decreased embedding parameters.

\paragraph{Evaluation Metrics} Pre-training Perplexity (exponential of pre-training test-set loss) is often used to measure MLM pre-training performance (RoBERTa~\citep{Roberta}, Megatron-LM~\citep{Megatron}, \citet{ScalingLawsModelArchitectures}, or similar variants in \citet{PPL1,PPL2}), and is well-correlated with downstream performance~\citep{PPL3}. We use the perplexity as reported by Megatron-LM \href{https://github.com/NVIDIA/Megatron-LM/blob/443ce9f3f98fdc5a53c6b480c6e21b79944d198e/megatron/training.py#L975}{here}. Calling this measure ``perplexity'' is a slight abuse of notation (as previous words which are masked are not available, and future words are). For downstream fine-tuning, we use accuracy while for Speech-to-Text translation, we use BLEU score.

\paragraph{Pre-Training Improvements} In \cref{table: res1}, we provide the results obtained on scaling model depth after applying DSLM  to Post-LN. Post-LN models often diverge while scaling model depth. DSLM stabilizes the training of Post-LN models, and even a $768$ layer Post-LN model (with $2300$ Linear and $768$ attention layers) converges.  

\begin{table}[H]
\caption{Performance (perplexity) of BERT models with different shapes. Deep-Thin models provide large improvements with fewer parameters.}
\begin{center}
\begin{adjustbox}{max width=0.95\linewidth}
\begin{tabular}{ccccc}
\toprule
    \textbf{Model N/D} & \textbf{12/1024} & \textbf{48/512} & \textbf{192/256} & \textbf{768/128} \vspace{-3pt} \\
    \textbf{\scriptsize{(\# Params)}} & \scriptsize{(\textbf{185M})} & \scriptsize{(\textbf{168M})} & \scriptsize{(\textbf{160M})} & \scriptsize{(\textbf{156M})} \\
    \midrule
    Baseline & 14.2 & 14.8 & 17.2 & diverge \\
    DSLM & 15.5 & 13.1 & \textbf{12.9} & 18.4 \\
    \midrule
    \textbf{Model N/D} & \textbf{24/1024} & \textbf{96/512} & \textbf{384/256} & - \vspace{-3pt}\\
    \textbf{\scriptsize{(\# Params)}} & \scriptsize{(\textbf{336M})} & \scriptsize{(\textbf{319M})} & \scriptsize{(\textbf{311M})} & - \\
    \midrule
    Baseline & 13.2 & diverge & diverge & - \\
    DSLM & 14.0 & \textbf{11.7} & 12.3 & - \\
    \bottomrule
\end{tabular}
\label{table: res1}
\end{adjustbox}
\end{center}
\end{table}

Our method is comparable to the baseline for shallow models but starts to outperform as the model gets deeper. Our $192$-layer model outperforms the vanilla $12$-layer, and our $96$ layer outperforms the vanilla $24$-layer model. The $160$M $192$-layer model outperforms the vanilla $24$-layer $336$M model with more than $2\times$ the params. 

Reading \cref{table: res1} vertically, we can compare the performance of our approach with the baseline as we vary the model depth ($N$) while keeping the hidden dimension ($d$) constant. The baseline models often diverge at larger depths. By stabilizing the training, DSLM allows training larger models with better performance, with consistent improvements at larger depths.


\paragraph{Pre-training Improvements for Pre-LN} We also applied DSLM to the deep Pre-LN models, trained for $3.3$B tokens. \cref{table: pre_deep} show that DSLM significantly improves the performance of the Pre-LN model across a range of model depths.


\begin{table}[h]
\caption{DSLM with Pre-LN Models.}
\begin{center}
\begin{adjustbox}{max width=0.85\linewidth}
\begin{tabular}{lcccc}
\toprule
    \textbf{Model N/D} & \textbf{12/512} & \textbf{96/512} & \textbf{192/256} & \textbf{768/128}  \\
    \midrule
    Baseline & 29.4 & 20.6 & 19.8 & 26.9 \\
    DSLM & \textbf{26.0} & \textbf{15.4} & \textbf{17.0} & \textbf{25.9} \\
    \bottomrule
\end{tabular}
\label{table: pre_deep}
\end{adjustbox}
\end{center}

\end{table}

\paragraph{Sustained Improvements after Longer Pre-training}
Due to compute limitations, our models were trained for Chinchilla optimal steps. To ensure reproducibility of our work (scripts provided in released code), and demonstrate sustained improvements for standard models, we trained the BERT-base model using public Wikipedia data for $64$B tokens ($30x$ chinchilla tokens). We train a $4x$ deeper, $10\%$ smaller model using DSLM ($N$/$d$ = $48$ / $384$). We finetune these models on the public RACE-M, RACE-H~\citep{race}, MNLI~\citep{mnli} and QQP~\footnote{\href{https://quoradata.quora.com/First-Quora-Dataset-Release-Question-Pairs}{Quora Question Pairs dataset}} datasets. As shown in \cref{table:finetune_stderr}, our model provides better pretraining performance which is translated into downstream Question-Answering tasks' performance across all datasets.

\begin{table}[ht]
\caption{BERT-base (trained for $64$B tokens) pre-training and finetuning results (mean accuracy across 5 runs with stderr).}
\begin{center}
\begin{adjustbox}{max width=0.75\linewidth}
\begin{tabular}{lll}
    \toprule
    \textbf{Dataset} & \textbf{Baseline} &   \textbf{DSLM} \\
    \midrule
    \multicolumn{3}{c}{\textit{Pretraining Performance}} \\
    Validation PPL & \multicolumn{1}{c}{8.3} & \multicolumn{1}{c}{\textbf{7.8}} \\
    \midrule
    \multicolumn{3}{c}{\textit{Finetuning Accuracy}} \\
    MNLI & 82.4 \small{$\pm$ 0.1} & \textbf{83.7} \small{$\pm$ 0.1} \\
    QQP & 90.8 \small{$\pm$ 0.03} & \textbf{91.1} \small{$\pm$ 0.05} \\
    RACE-Middle & 71.1 \small{$\pm$ 0.2} & \textbf{74.0} \small{$\pm$ 0.3} \\
    RACE-High & 63.7 \small{$\pm$ 0.1} & \textbf{65.7} \small{$\pm$ 0.2}\\
    \bottomrule
\end{tabular}
\end{adjustbox}
\end{center}
\label{table:finetune_stderr}

\end{table}

\paragraph{Downstream Low Rank Finetuning}
DSLM continues to outperform the baseline on finetuning for downstream tasks with Low Rank Adapters~\citep{lora}, as shown in \cref{table:qlora}. Following QLoRA~\citep{qlora}, we apply LoRA on all linear modules, with $r=32$, $\alpha=16$, and searched for LR. 

\begin{table}[ht]
\caption{Accuracy on MNLI after low rank finetuning using LoRA}
\begin{center}
\begin{adjustbox}{max width=0.95\linewidth}
\begin{tabular}{lccc}
    \toprule
    \textbf{Model} & \multicolumn{2}{|c|}{\textbf{Model Size}} & \textbf{Score (Accuracy)} \\
    & \multicolumn{1}{|c}{\textbf{Layers (N)}} & \multicolumn{1}{c|}{\textbf{Hidden Dim (d)}} & \\
    \midrule
    Baseline & 12 & 768 & 82.2 $\pm 0.1$ \\
    DSLM & 48 & 384 & \textbf{82.9} $\pm 0.1$ \\
    \bottomrule
\end{tabular}
\end{adjustbox}
\end{center}
\label{table:qlora}
\end{table}

\subsection{Improvements on Decoder-only Models (GPT)}
\label{results:DSLM_gpt}
We applied DSLM to the decoder-only GPT model, trained for $8$B tokens (slightly more than Chinchilla-optimal). Similar to BERT, increasing model depth by $4x$ with DSLM while keeping the parameters constant results in improved performance (\cref{table:gpt}).



\begin{table}[h]
\caption{Application of DSLM to Decoder-only model (GPT), while increasing model depth to 4x (token-level PPL).}
\begin{center}
\begin{adjustbox}{max width=\linewidth}
\begin{tabular}{lccccc}
    \toprule
    \textbf{Model} & \multicolumn{3}{|c|}{\textbf{Model Size}}  & \multicolumn{2}{c}{\textbf{LM Perplexity}} \\
    &  \multicolumn{1}{|c}{\textbf{Layers (N)}} &  \multicolumn{1}{c}{\textbf{Dim (d)}} & \multicolumn{1}{c|}{\textbf{Params}}  & \textbf{Pre-LN} & \textbf{Post-LN} \\
    \midrule
    Baseline & 12 & 1024 & \footnotesize{204M} & 11.6 & 12.7   \\
    DSLM & 12 & 1024 & \footnotesize{204M} &  11.5 & \textbf{11.5} \\
    DSLM & 48 & 512 & \footnotesize{178M} &  \textbf{11.2} & 11.7 \\
    \midrule
    Baseline & 24 & 1024 & \footnotesize{355M} & 10.4 & 11.6   \\
    DSLM & 24 & 1024 & \footnotesize{355M} &  10.2 & \textbf{10.5} \\
    DSLM & 96 & 512 & \footnotesize{329M} &  \textbf{10.1} & 10.6 \\
    \bottomrule
\end{tabular}
\label{table:gpt}
\end{adjustbox}
\end{center}
\end{table}

\subsection{Improvements on Speech (Encoder-Decoder)}
\label{results:DSLM_multimodal}

We apply DSLM on encoder/decoder style transformer for Speech-to-Text translation task. Applying our method to speech additionally requires handling the input embeddings. Instead of theoretical estimates as in the case of text inputs (\cref{proof: embeddings}), the moments for speech embedding were replaced by the empirically observed values. This input variance and correlation was observed as $2.2$ and $0.29$. 

The baseline was trained on the MuST-C~\citep{mustc} dataset using fairseq~\citep{ott2019fairseq}. Using DSLM, we successfully train $4$x deeper models which outperforms the $18$-layer ($12$-encoder, $6$-decoder layers) baseline with $9\%$ less parameters as seen in \cref{table:speech}.

\begin{table}[h]
\caption{Application of DSLM to Speech-to-Text translation. N\textsubscript{enc} and N\textsubscript{dec} refer to number of layers in the encoder and the decoder respectively. For models marked with *, maximum source sequence length was limited to $1024$ due to compute limitations, and longer examples were discarded for both train and test.}
\begin{center}
\begin{adjustbox}{max width=\linewidth}
\begin{tabular}{llccccc}
    \toprule
    \textbf{Model} & \multicolumn{1}{|c}{\textbf{Lang}} & \multicolumn{3}{|c}{\textbf{Model Size}}  & \multicolumn{1}{|c}{\textbf{BLEU}} \\
    & \multicolumn{1}{|c}{} &  \multicolumn{1}{|c}{\textbf{N\textsubscript{enc}, N\textsubscript{dec}}} &  \textbf{Dim (d)} & \textbf{Params}  & \multicolumn{1}{|c}{} \\
    \midrule
    Baseline Pre-LN & en$\rightarrow$de & 12,6 & 256 & \footnotesize{31.1M} & 24.9 \\
    DSLM Pre-LN & en$\rightarrow$de & 48,24 & 128 & \footnotesize{28.4M} & \textbf{25.6} \\
    \midrule
    Baseline Post-LN & en$\rightarrow$de & 12,6 & 256 & \footnotesize{31.1M} & 21.9   \\
    DSLM Post-LN & en$\rightarrow$de & 48,24 & 128 & \footnotesize{28.4M} & \textbf{23.8} \\
    \midrule
    Baseline Pre-LN* & en $\rightarrow$ es & 12,6 & 256 & \footnotesize{31.1M} & 21.61      \\
    DSLM Pre-LN* & en $\rightarrow$ es & 48,24 & 128 & \footnotesize{28.4M} &  \textbf{23.03} \\
    \midrule
    Baseline Pre-LN* & en $\rightarrow$ fr &  12,6 & 256 & \footnotesize{31.1M} & 23.74   \\
    DSLM Pre-LN* & en $\rightarrow$ fr & 48,24 & 128 & \footnotesize{28.4M} &  \textbf{26.30}  \\
    \bottomrule
\end{tabular}
\label{table:speech}
\end{adjustbox}
\end{center}
\end{table}

\subsection{Improvements on Vision Modality}
Similar to speech domain, applying our method to vision modality simply requires handling the input embedding (\cref{appendix:subsection:vision}). Using ImageNet-1k~\citep{Imagenet} data with ViT~\citep{vit} model, our method can also constrain the growth of moments in Vision Transformers, as we show in \cref{fig:appendix:vit_plot}.

We train our models on the Image Classification task using ViT baselines provided by \citet{vitbaseline}, and trained a $4$x deeper model with same params. The deeper DSLM model outperforms the baseline ViT both in both $90$ and $300$ epoch settings. The improvements also translate to improved robustness on ImageNet-v2~\citep{ImagenetV2}, ImageNet-R~\citep{ImagenetR} and ImageNet-Sketch~\citep{ImagenetSketch}.

\begin{table}[ht]
\caption{Applying DSLM to Image classification using ViT.}
\begin{center}
\begin{adjustbox}{max width=1.0\linewidth}
\begin{tabular}{lcccc}
    \toprule
    \textbf{Eval Set} & \multicolumn{2}{|c|}{\textbf{90-epoch}} & \multicolumn{2}{c}{\textbf{300-epoch}} \\
    & \multicolumn{1}{|c}{\textbf{Baseline}} &   \multicolumn{1}{c|}{\textbf{DSLM}} & \textbf{Baseline} &   \textbf{DSLM} \\
    \midrule
    ImageNet & 76.5 & \textbf{77.2} & 79.8 & \textbf{80.3}  \\
    ImageNet-Real & 83.2 & \textbf{83.8} & 85.4 & \textbf{85.5} \\
    ImageNet-v2 & 63.7 & \textbf{65.2} & 67.9 & \textbf{68.3}  \\
    ImageNet-R & 23.9 & \textbf{24.4} & 27.8 & \textbf{28.3} \\
    ImageNet-Sketch & 24.4 & \textbf{25.5} & 28.7 & \textbf{29.9} \\
    \bottomrule
\end{tabular}
\end{adjustbox}
\label{table:vision}
\end{center}

\end{table}


\subsection{Comparison with Prior Methods}
\label{sec:prior}

In \cref{table:comparison}, we compare DSLM with several prior methods for deep transformers. DSInit and DeepNet stabilize the model training at the expense of reduced ``sensitivity'' (\cref{section:rel_strength}) by  using smaller effective values of $\beta^2$, at \ord{N^{-2}} and \ord{N^{-1.5}} respectively. Interestingly, $96$-layer model diverges with DSInit, despite DSInit using a smaller $\beta$ asymptotically -- this is because the constants hidden in \ord{N^{-2}} are much larger for DSInit. Our method, by analysing signal propagation, sets constants exactly at $1$. 

Bamboo method is a vanilla Pre-LN transformer, which our method out-performs. SkipInit, ReZero, LayerScale and Value-Skipinit all initialize $\beta$ to zero/very small values -- this choice may slow down learning initially by reducing back-propagated gradients, and a learnable $\beta$ under-performs compared to fixed~(\cref{table: ablation_components}). Vanilla $\mu$P targets hyper-parameter transfer from thinner to wider models, and also diverges. Zero-initializing the output layers solves this divergence, but under-performs similar to SkipInit. \citet{noci2022signal} initializes Query and Key matrices to a large value, causing divergence~(\cref{QK_explosion}). ADMIN requires an extra profiling pass through the model, and more importantly, cannot stop vanishing gradients~(\cref{section:related:initialization}), causing the $192$-Layer model to diverge.

\begin{table}[h]
\caption{Comparison with prior methods for deep Transformers.}
\begin{center}
\begin{adjustbox}{max width=\linewidth}
\begin{tabular}{lcc}
\toprule
\textbf{Method} & \textbf{192/256} & \textbf{96/512} \\ 
\midrule
DSInit~\citep{DSInit} & 15.9 & diverge \\
ADMIN~\citep{UnderstandingDifficultyTraininga} & diverge & 25.2 \\
SkipInit~\citep{BatchNormalizationBiasesa} & 15.1 & 13.1 \\
ReZero~\citep{ReZeroAllYou} & diverge & diverge \\
LayerScale~\citep{GoingDeeperImagea} & 13.2 & 14.4 \\
$\mu$P-Tensor Programs V~\citep{tensorprograms} & diverge & diverge \\
DeepNorm~\citep{deepnet} & 14.4 & 13.4 \\
\citet{noci2022signal} & diverge & diverge \\
Bamboo~\citep{StudyTransformerConfigurationa} & 17.1 & diverge \\
Value-SkipInit~\citep{DeepTransformersShortcutsa} & 18.8 & 17.1 \\
\textbf{DeepScaleLM (ours)} & \textbf{12.9} & \textbf{11.7} \\
\bottomrule
\end{tabular}
\label{table:comparison}
\end{adjustbox}
\end{center}

\end{table}

\subsection{Analysis of DSLM}

\paragraph{Model Quantization} 
Similar to Unit Scaling~\citep{UnitScalingOutoftheBoxa}, conserving unit activations and gradients from our method results in models which lose much less performance when quantized (via direct casting) to FP8 precision compared to original models. We apply $8$-bit quantization to the $48$-Layer $512$-dim BERT baseline model and the model trained with DSLM. \cref{table: fp8} provides the performance corresponding to the full precision inference and FP8 inferences (corresponding to two different FP8 standards, E5M2 and E4M3). DSLM model can be compressed to $25\%$ of the original size with significantly lower performance loss.

\begin{table}[h]
\caption{Model performance on direct casting to FP8}
\begin{center}
\begin{adjustbox}{max width=0.82\linewidth}
\begin{tabular}{lccc}
\toprule
\textbf{Model} & \textbf{FP32} & \textbf{E5M2} & \textbf{E4M3} \\ 
\midrule
Baseline & 14.8 & 42.5 \footnotesize{($\Delta$ 27.7)} & 16.5 \footnotesize{($\Delta$ 1.7)} \\
DSLM & \textbf{13.1} & \textbf{21.4} \footnotesize{($\Delta$~~ 8.3)} & \textbf{13.9} \footnotesize{($\Delta$ 0.8)} \\
\bottomrule
\end{tabular}
\end{adjustbox}
\end{center}
\label{table: fp8}
\end{table}

\paragraph{Ablation of Residual Scaling} 
\label{sec:abl_var}
\cref{table: ablation_components} provides the results corresponding to the different components of our proposed DSLM scheme for training $96$-layer $512$-d model Post-LN model. The model fails to converge without the proposed residual scaling. $\beta$ may also be set as learnable (similar to BatchNorm ~\cite{batchnorm}), after initializing it with $\beta^2=\frac{2}{N}$. We find that this does not significantly impact performance, and $\beta$ remains within $[0.2-5]\times$ of its initialized values.

\begin{table}[h]
\caption{Ablation of various DeepScaleLM components.}
\begin{center}
\begin{adjustbox}{max width=0.8\linewidth}
\begin{tabular}{lc}
    \toprule
    \textbf{Model} & \textbf{Perf}\\ 
    \midrule
    Vanilla Xavier (with or w/o $\beta^2=0.5$) & diverge   \\
    DSLM-Init (with or w/o $\beta^2=0.5$) & diverge  \\
    DSLM-Init + $\beta^2=\frac{2}{N}$ (learnable $\beta$) & 12.2  \\
    DSLM-Init + $\beta^2=\frac{2}{N}$ (fixed $\beta$) & 11.7  \\
    \bottomrule
\end{tabular}
\label{table: ablation_components}
\end{adjustbox}
\end{center}
\end{table}

\paragraph{Ablation of Initialization} \cref{table: ablation_sigma} provides ablation results for our proposed initialization. All experiments in \cref{table: ablation_sigma} were conducted for the Pre-LN model with our proposed scaling ($\lambda, \beta$), since the Post-LN model diverged with Xavier initialization. Xavier initialization performs significantly worse for very deep models, due to higher QK initialization. BERT default initialization with $\sigma=0.02$ also performs worse. Finally, DSLM simpler initialization performs comparably to DSLM.

\begin{table}[h]
\caption{Ablation of the initializations.}
\begin{center}
\begin{adjustbox}{max width=0.75\linewidth}
\begin{tabular}{lcc}
    \toprule
    \textbf{Model} & \textbf{Model Size (N/d)} & \textbf{Perf}\\ 
    \midrule
    Xavier & 192/256~\footnotesize{(160M)} & 38.2 \\
    DSLM & 192/256~\footnotesize{(160M)} & \textbf{17.0}  \\
    DSLM (simple) & 192/256~\footnotesize{(160M)} & 17.9  \\
    \midrule
    Fixed $\sigma=0.02$ & 96/512~\footnotesize{(319M)} & 20.5   \\
    DSLM & 96/512~\footnotesize{(319M)} & \textbf{17.9}  \\
    \bottomrule
\end{tabular}
\label{table: ablation_sigma}
\end{adjustbox}
\end{center}
\end{table}



\paragraph{Compute} \cref{appendix:section:compute} provides detailed theoretical and wall-clock compute overheads for making models deeper. We observe that up to $200$ layers, the theoretical compute is within $6-7$\%  and wall-clock times is within $15\%$ of the original shallow model. While our $192$-layer $256$-d model requires $6$\% extra compute than the $12$-layer $1024$-d model, it manages to outperform the $24$-layer $1024$-d model, that has $62.5$\% more parameters, at equal wall-clock time and at equal number of tokens.

\paragraph{Discussion of Relative Strength}
\label{section:rel_strength}

In general, for a $\beta$ of the form $\beta^2 = \frac{k}{N^\alpha}$, we can choose from a wide range of values for the constant $k$ and exponent $\alpha$. There is an expressivity-trainability trade-off in training deep networks~\citep{EdgeOfChaos} -- having lower $\beta$ (smaller $k$ or higher $\alpha$) will result in networks where observed issues (forward growth or gradient explosion/vanishing) are mitigated, but they may converge slowly/sub-optimally.


\citet{Catformer} defines ``sensitivity'' as the variance of relative change in output for small perturbations in parameters, averaged across all parameters. If $\sigma^2_{\text{skip}}=1$, sensitivity can be shown to be mean across layers of $N*(1/\sigma^2_{\text{block}}) = N*\beta^2$. Mean is not robust to outliers, and hence we suggest median may provide a more robust measure. For e.g., for vanilla pre-LN, \citet{Catformer}’s definition gives sensitivity as \ord{log(N)}, whereas using median provides a more robust measure as \ord{1}. But only the first $N/10$ layers have \ord{log(N)} sensitivity, and the last $9N/10$ layers have \ord{1} sensitivity. We will use median in the discussion below.

In \cref{appendix:section:rel_strength}, we show that the fall in gradient for both pre-LN and post-LN for $\beta^2 = k/N^\alpha$ is \ord{e^{kN^{1-\alpha}}}. The sensitivity is hence $kN^{1-\alpha}$. For DSLM, we chose $\alpha=1$, that is the sweet spot on the stability-expressivity curve where both the gradient fall bound and sensitivity expressions become independent of model depth. For higher values of $\alpha$ such as $\alpha=2$~(DS-Init) and, $\alpha=1.5$~(DeepNet), the gradient becomes stable using but the model expressivity reduces with depth, as shown in \cref{table:block_growth}. Such models might not be able to extract better results when going deeper, as we indeed verify empirically in the comparison with prior works paragraph in \cref{sec:prior}.

\section{Related Works}
\label{section:rel_main}

For detailed discussion of prior works, refer to \cref{section:app_related}.

\paragraph{Initialization}
Several works~\citep{glorot, DelvingDeepRectifiers, CharacterizingSignalPropagationa, ExponentialExpressivity, DeepInformationPropagation} improved the initialization of ResNets/ReLU networks. These works do not consider transformers, and are unable to handle Softmax/Attention. Others, such as ADMIN~\citep{UnderstandingDifficultyTraininga}, \citet{AllYouNeed,verydeep} achieve unit variance for faster convergence by scaling the weights and/or outputs based on empirical profiling of a forward pass. \citet{UnitScalingOutoftheBoxa} also tries to achieve this, but does not completely handle correlation and non-zero mean of ReLU. We demonstrate that this profiling is unnecessary, and can instead be done theoretically in our work.

\paragraph{Signal Propagation}

Signal propagation in Neural Networks~\citep{rel_neal, rel_lecun} has a long history, such as for ResNets~\citep{DelvingDeepRectifiers, BatchNormalizationBiasesa, CharacterizingSignalPropagationa, DeepInformationPropagation, NormalisationDead, ProxyNormalizingActivationsMatcha, ScalingResNetsLargedeptha, SelfNormalizingNeuralNetworksb, Shattered}, and for transformers in~\citep{xu2019understanding, AttentionNotAll, Catformer, noci2022signal, RapidTrainingDeep, DeepTransformersShortcutsa, RevisitingOversmoothingBERT, AntiOversmoothingDeepVision}. Our work considers previously often neglected effects of dropout, input correlation, activation non-linearity, and $QK$ initialization, providing closed forms with verifiable correctness of signal propagation. This allows us to constrain the output and gradient to almost exactly unit variance.

\paragraph{Moment Control \& Residual Scaling}

Bounded gradients have been shown to results in better/faster convergence~\citep{PowerNormRethinkingBatcha, BlockNormalizedGradientMethoda, LargeBatchTrainingb, LargeBatchOptimizationb, LayerNormalizationsResiduala, NormFormerImprovedTransformerb, ImpactActivationFunction}. Different scaling schemes for residual networks ($\lambda$ for skip connections and $\beta$ for residual output) have been explored by prior works, such as $\lambda^2 + \beta^2=1$ for ResNets~\citep{Shattered, Inceptionv4, HowStartTraining, HowInitializeYour, fixup, NormalisationDead}. Learnable $\beta \approx 0$ was used in SkipInit~\citep{BatchNormalizationBiasesa}, ReZero~\citep{ReZeroAllYou}, LayerScale~\citep{GoingDeeperImagea}, Value-SkipInit~\citep{DeepTransformersShortcutsa}. Others proposed $\beta^2=$\ord{\frac{1}{N}}, where $N$ is max/current layer was used in \citet{HowInitializeYour, CharacterizingSignalPropagationa, ScalingResNetsLargedeptha, StabilizeDeepResNet, DeepTransformersShortcutsa, noci2022signal, BatchNormalizationBiasesa, UnderstandingDifficultyTraininga, verydeep, Catformer, UnitScalingOutoftheBoxa}, while DSInit~\citep{DSInit}, T-Fixup~\citep{Tfixup}, DeepNorm~\citep{deepnet} used $\beta^2 < $\ord{\frac{1}{N}}. However, the optimal initialization/scaling can vary based on data/model characteristics~\citep{StabilizeDeepResNet,ScalingResNetsLargedeptha}. Our contribution goes beyond providing an optimal scaling scheme -- our theory enables informed choices about these initialization/scaling schemes based on their expressivity-trainability trade-off. Some works such as DeepNet, ADMIN show performance improvements by making the model deeper, but much larger. In this work, we explore a stricter setting of keeping transformer parameters and compute constant while making the model deeper.

\paragraph{Other Network modifications for Deep Networks} 
Architectural modifications such as \citet{StabilizingTransformerTrainingc,DeepViTDeeperVision,NormFormerImprovedTransformerb} can only stabilize the model later during training and not at initialization. They are orthogonal to our approach, and our equations can be easily extended to cover these. 
\section{Conclusion}

We theoretically derive closed forms for the growth of variances for forward and backward pass through individual transformer components as well as the entire transformer model. These formulae enable us to identify and solve the key reasons for vanishing/exploding gradients and rank collapse in very deep transformers. Via scaling and correct initialization, we also enable training very deep transformers with $1000$ layers. Our experiments suggest that deeper transformers should be explored -- using our method, models with $100$s of layers outperform larger standard models across multiple modalities, tasks, and transformer variants.


\section*{Acknowledgements}

We would like to thank Dr. Kangwook Lee and Dr. Joohyung Lee of Samsung Research, Seoul, Korea, for their guidance and leadership. We would also like to thank all the reviewers for their valuable feedback and suggestions, which helped greatly improve the paper.

\section*{Impact Statement}

This paper presents work whose goal is to advance the field of Machine Learning. There are many potential societal consequences of our work, some which we feel must be specifically highlighted here. Using crawled web data for pre-training language models is questionable, something which society has yet to finalize its views on. Language modelling in particular suffers from hallucinations, and may be used for misinformation.

\nocite{*}

\bibliography{example_paper}

\begin{thebibliography}{131}
\providecommand{\natexlab}[1]{#1}
\providecommand{\url}[1]{\texttt{#1}}
\expandafter\ifx\csname urlstyle\endcsname\relax
  \providecommand{\doi}[1]{doi: #1}\else
  \providecommand{\doi}{doi: \begingroup \urlstyle{rm}\Url}\fi

\bibitem[Anil et~al.(2019)Anil, Lucas, and Grosse]{SortingOutLipschitza}
Anil, C., Lucas, J., and Grosse, R.~B.
\newblock Sorting out lipschitz function approximation.
\newblock In Chaudhuri, K. and Salakhutdinov, R. (eds.), \emph{Proceedings of the 36th International Conference on Machine Learning, {ICML} 2019, 9-15 June 2019, Long Beach, California, {USA}}, volume~97 of \emph{Proceedings of Machine Learning Research}, pp.\  291--301. {PMLR}, 2019.
\newblock URL \url{http://proceedings.mlr.press/v97/anil19a.html}.

\bibitem[Arora et~al.(2019{\natexlab{a}})Arora, Du, Hu, Li, Salakhutdinov, and Wang]{ExactComputationInfinitely}
Arora, S., Du, S.~S., Hu, W., Li, Z., Salakhutdinov, R., and Wang, R.
\newblock On exact computation with an infinitely wide neural net.
\newblock In Wallach, H.~M., Larochelle, H., Beygelzimer, A., d'Alch{\'{e}}{-}Buc, F., Fox, E.~B., and Garnett, R. (eds.), \emph{Advances in Neural Information Processing Systems 32: Annual Conference on Neural Information Processing Systems 2019, NeurIPS 2019, December 8-14, 2019, Vancouver, BC, Canada}, pp.\  8139--8148, 2019{\natexlab{a}}.
\newblock URL \url{https://proceedings.neurips.cc/paper/2019/hash/dbc4d84bfcfe2284ba11beffb853a8c4-Abstract.html}.

\bibitem[Arora et~al.(2019{\natexlab{b}})Arora, Du, Hu, Li, and Wang]{FineGrainedAnalysisOptimization}
Arora, S., Du, S.~S., Hu, W., Li, Z., and Wang, R.
\newblock Fine-grained analysis of optimization and generalization for overparameterized two-layer neural networks.
\newblock In Chaudhuri, K. and Salakhutdinov, R. (eds.), \emph{Proceedings of the 36th International Conference on Machine Learning, {ICML} 2019, 9-15 June 2019, Long Beach, California, {USA}}, volume~97 of \emph{Proceedings of Machine Learning Research}, pp.\  322--332. {PMLR}, 2019{\natexlab{b}}.
\newblock URL \url{http://proceedings.mlr.press/v97/arora19a.html}.

\bibitem[Arpit et~al.(2016)Arpit, Zhou, Kota, and Govindaraju]{NormalizationPropagation}
Arpit, D., Zhou, Y., Kota, B.~U., and Govindaraju, V.
\newblock Normalization propagation: {A} parametric technique for removing internal covariate shift in deep networks.
\newblock In Balcan, M. and Weinberger, K.~Q. (eds.), \emph{Proceedings of the 33nd International Conference on Machine Learning, {ICML} 2016, New York City, NY, USA, June 19-24, 2016}, volume~48 of \emph{{JMLR} Workshop and Conference Proceedings}, pp.\  1168--1176. JMLR.org, 2016.
\newblock URL \url{http://proceedings.mlr.press/v48/arpitb16.html}.

\bibitem[Arpit et~al.(2019)Arpit, Campos, and Bengio]{HowInitializeYour}
Arpit, D., Campos, V., and Bengio, Y.
\newblock How to initialize your network? robust initialization for weightnorm i\& resnets.
\newblock In Wallach, H.~M., Larochelle, H., Beygelzimer, A., d'Alch{\'{e}}{-}Buc, F., Fox, E.~B., and Garnett, R. (eds.), \emph{Advances in Neural Information Processing Systems 32: Annual Conference on Neural Information Processing Systems 2019, NeurIPS 2019, December 8-14, 2019, Vancouver, BC, Canada}, pp.\  10900--10909, 2019.
\newblock URL \url{https://proceedings.neurips.cc/paper/2019/hash/e520f70ac3930490458892665cda6620-Abstract.html}.

\bibitem[Bachlechner et~al.(2021)Bachlechner, Majumder, Mao, Cottrell, and McAuley]{ReZeroAllYou}
Bachlechner, T., Majumder, B.~P., Mao, H.~H., Cottrell, G., and McAuley, J.~J.
\newblock Rezero is all you need: fast convergence at large depth.
\newblock In de~Campos, C.~P., Maathuis, M.~H., and Quaeghebeur, E. (eds.), \emph{Proceedings of the Thirty-Seventh Conference on Uncertainty in Artificial Intelligence, {UAI} 2021, Virtual Event, 27-30 July 2021}, volume 161 of \emph{Proceedings of Machine Learning Research}, pp.\  1352--1361. {AUAI} Press, 2021.
\newblock URL \url{https://proceedings.mlr.press/v161/bachlechner21a.html}.

\bibitem[Balduzzi et~al.(2017)Balduzzi, Frean, Leary, Lewis, Ma, and McWilliams]{Shattered}
Balduzzi, D., Frean, M., Leary, L., Lewis, J.~P., Ma, K.~W., and McWilliams, B.
\newblock The shattered gradients problem: If resnets are the answer, then what is the question?
\newblock In Precup, D. and Teh, Y.~W. (eds.), \emph{Proceedings of the 34th International Conference on Machine Learning, {ICML} 2017, Sydney, NSW, Australia, 6-11 August 2017}, volume~70 of \emph{Proceedings of Machine Learning Research}, pp.\  342--350. {PMLR}, 2017.
\newblock URL \url{http://proceedings.mlr.press/v70/balduzzi17b.html}.

\bibitem[Beyer et~al.(2022)Beyer, Zhai, and Kolesnikov]{vitbaseline}
Beyer, L., Zhai, X., and Kolesnikov, A.
\newblock Better plain vit baselines for imagenet-1k, 2022.

\bibitem[Bingham \& Miikkulainen(2023)Bingham and Miikkulainen]{binghamAutoInitAnalyticSignalpreserving2023}
Bingham, G. and Miikkulainen, R.
\newblock {{AutoInit}}: Analytic signal-preserving weight initialization for neural networks.
\newblock In \emph{Proceedings of the {{Thirty-Seventh AAAI Conference}} on {{Artificial Intelligence}} and {{Thirty-Fifth Conference}} on {{Innovative Applications}} of {{Artificial Intelligence}} and {{Thirteenth Symposium}} on {{Educational Advances}} in {{Artificial Intelligence}}}, volume~37 of \emph{{{AAAI}}'23/{{IAAI}}'23/{{EAAI}}'23}, pp.\  6823--6833. AAAI Press, 2023.
\newblock ISBN 978-1-57735-880-0.
\newblock \doi{10.1609/aaai.v37i6.25836}.

\bibitem[Blake et~al.(2023)Blake, Orr, and Luschi]{UnitScalingOutoftheBoxa}
Blake, C., Orr, D., and Luschi, C.
\newblock Unit scaling: Out-of-the-box low-precision training.
\newblock In Krause, A., Brunskill, E., Cho, K., Engelhardt, B., Sabato, S., and Scarlett, J. (eds.), \emph{International Conference on Machine Learning, {ICML} 2023, 23-29 July 2023, Honolulu, Hawaii, {USA}}, volume 202 of \emph{Proceedings of Machine Learning Research}, pp.\  2548--2576. {PMLR}, 2023.
\newblock URL \url{https://proceedings.mlr.press/v202/blake23a.html}.

\bibitem[Bordelon et~al.(2023)Bordelon, Noci, Li, Hanin, and Pehlevan]{bordelonDepthwiseHyperparameterTransfer2023}
Bordelon, B., Noci, L., Li, M.~B., Hanin, B., and Pehlevan, C.
\newblock Depthwise {{Hyperparameter Transfer}} in {{Residual Networks}}: {{Dynamics}} and {{Scaling Limit}}.
\newblock In \emph{The {{Twelfth International Conference}} on {{Learning Representations}}}, 2023.

\bibitem[Brock et~al.(2021{\natexlab{a}})Brock, De, and Smith]{CharacterizingSignalPropagationa}
Brock, A., De, S., and Smith, S.~L.
\newblock Characterizing signal propagation to close the performance gap in unnormalized resnets.
\newblock In \emph{9th International Conference on Learning Representations, {ICLR} 2021, Virtual Event, Austria, May 3-7, 2021}. OpenReview.net, 2021{\natexlab{a}}.
\newblock URL \url{https://openreview.net/forum?id=IX3Nnir2omJ}.

\bibitem[Brock et~al.(2021{\natexlab{b}})Brock, De, Smith, and Simonyan]{brockHighPerformanceLargeScaleImage2021}
Brock, A., De, S., Smith, S.~L., and Simonyan, K.
\newblock High-performance large-scale image recognition without normalization.
\newblock In Meila, M. and Zhang, T. (eds.), \emph{Proceedings of the 38th International Conference on Machine Learning, {ICML} 2021, 18-24 July 2021, Virtual Event}, volume 139 of \emph{Proceedings of Machine Learning Research}, pp.\  1059--1071. {PMLR}, 2021{\natexlab{b}}.
\newblock URL \url{http://proceedings.mlr.press/v139/brock21a.html}.

\bibitem[Chen et~al.(2020)Chen, Wei, Huang, Ding, and Li]{SimpleDeepGraph}
Chen, M., Wei, Z., Huang, Z., Ding, B., and Li, Y.
\newblock Simple and deep graph convolutional networks.
\newblock In \emph{Proceedings of the 37th International Conference on Machine Learning, {ICML} 2020, 13-18 July 2020, Virtual Event}, volume 119 of \emph{Proceedings of Machine Learning Research}, pp.\  1725--1735. {PMLR}, 2020.
\newblock URL \url{http://proceedings.mlr.press/v119/chen20v.html}.

\bibitem[Chowdhery et~al.(2023)Chowdhery, Narang, Devlin, Bosma, Mishra, Roberts, Barham, Chung, Sutton, Gehrmann, Schuh, Shi, Tsvyashchenko, Maynez, Rao, Barnes, Tay, Shazeer, Prabhakaran, Reif, Du, Hutchinson, Pope, Bradbury, Austin, Isard, Gur{-}Ari, Yin, Duke, Levskaya, Ghemawat, Dev, Michalewski, Garcia, Misra, Robinson, Fedus, Zhou, Ippolito, Luan, Lim, Zoph, Spiridonov, Sepassi, Dohan, Agrawal, Omernick, Dai, Pillai, Pellat, Lewkowycz, Moreira, Child, Polozov, Lee, Zhou, Wang, Saeta, Diaz, Firat, Catasta, Wei, Meier{-}Hellstern, Eck, Dean, Petrov, and Fiedel]{palm}
Chowdhery, A., Narang, S., Devlin, J., Bosma, M., Mishra, G., Roberts, A., Barham, P., Chung, H.~W., Sutton, C., Gehrmann, S., Schuh, P., Shi, K., Tsvyashchenko, S., Maynez, J., Rao, A., Barnes, P., Tay, Y., Shazeer, N., Prabhakaran, V., Reif, E., Du, N., Hutchinson, B., Pope, R., Bradbury, J., Austin, J., Isard, M., Gur{-}Ari, G., Yin, P., Duke, T., Levskaya, A., Ghemawat, S., Dev, S., Michalewski, H., Garcia, X., Misra, V., Robinson, K., Fedus, L., Zhou, D., Ippolito, D., Luan, D., Lim, H., Zoph, B., Spiridonov, A., Sepassi, R., Dohan, D., Agrawal, S., Omernick, M., Dai, A.~M., Pillai, T.~S., Pellat, M., Lewkowycz, A., Moreira, E., Child, R., Polozov, O., Lee, K., Zhou, Z., Wang, X., Saeta, B., Diaz, M., Firat, O., Catasta, M., Wei, J., Meier{-}Hellstern, K., Eck, D., Dean, J., Petrov, S., and Fiedel, N.
\newblock Palm: Scaling language modeling with pathways.
\newblock \emph{J. Mach. Learn. Res.}, 24:\penalty0 240:1--240:113, 2023.
\newblock URL \url{http://jmlr.org/papers/v24/22-1144.html}.

\bibitem[Clark et~al.(2020)Clark, Luong, Le, and Manning]{electra}
Clark, K., Luong, M., Le, Q.~V., and Manning, C.~D.
\newblock {ELECTRA:} pre-training text encoders as discriminators rather than generators.
\newblock In \emph{8th International Conference on Learning Representations, {ICLR} 2020, Addis Ababa, Ethiopia, April 26-30, 2020}. OpenReview.net, 2020.
\newblock URL \url{https://openreview.net/forum?id=r1xMH1BtvB}.

\bibitem[Dasoulas et~al.(2021)Dasoulas, Scaman, and Virmaux]{LipschitzNormalizationSelfAttentionb}
Dasoulas, G., Scaman, K., and Virmaux, A.
\newblock Lipschitz normalization for self-attention layers with application to graph neural networks.
\newblock In Meila, M. and Zhang, T. (eds.), \emph{Proceedings of the 38th International Conference on Machine Learning, {ICML} 2021, 18-24 July 2021, Virtual Event}, volume 139 of \emph{Proceedings of Machine Learning Research}, pp.\  2456--2466. {PMLR}, 2021.
\newblock URL \url{http://proceedings.mlr.press/v139/dasoulas21a.html}.

\bibitem[Daunizeau(2017)]{SemianalyticalApproximationsStatisticala}
Daunizeau, J.
\newblock Semi-analytical approximations to statistical moments of sigmoid and softmax mappings of normal variables, 2017.
\newblock URL \url{https://arxiv.org/abs/1703.00091}.

\bibitem[Davis et~al.(2021)Davis, Gu, Choromanski, Dao, R{\'{e}}, Finn, and Liang]{Catformer}
Davis, J.~Q., Gu, A., Choromanski, K., Dao, T., R{\'{e}}, C., Finn, C., and Liang, P.
\newblock Catformer: Designing stable transformers via sensitivity analysis.
\newblock In Meila, M. and Zhang, T. (eds.), \emph{Proceedings of the 38th International Conference on Machine Learning, {ICML} 2021, 18-24 July 2021, Virtual Event}, volume 139 of \emph{Proceedings of Machine Learning Research}, pp.\  2489--2499. {PMLR}, 2021.
\newblock URL \url{http://proceedings.mlr.press/v139/davis21a.html}.

\bibitem[De \& Smith(2020)De and Smith]{BatchNormalizationBiasesa}
De, S. and Smith, S.~L.
\newblock Batch normalization biases residual blocks towards the identity function in deep networks.
\newblock In Larochelle, H., Ranzato, M., Hadsell, R., Balcan, M., and Lin, H. (eds.), \emph{Advances in Neural Information Processing Systems 33: Annual Conference on Neural Information Processing Systems 2020, NeurIPS 2020, December 6-12, 2020, virtual}, 2020.
\newblock URL \url{https://proceedings.neurips.cc/paper/2020/hash/e6b738eca0e6792ba8a9cbcba6c1881d-Abstract.html}.

\bibitem[Dehghani et~al.(2023)Dehghani, Djolonga, Mustafa, Padlewski, Heek, Gilmer, Steiner, Caron, Geirhos, Alabdulmohsin, Jenatton, Beyer, Tschannen, Arnab, Wang, Ruiz, Minderer, Puigcerver, Evci, Kumar, van Steenkiste, Elsayed, Mahendran, Yu, Oliver, Huot, Bastings, Collier, Gritsenko, Birodkar, Vasconcelos, Tay, Mensink, Kolesnikov, Pavetic, Tran, Kipf, Lucic, Zhai, Keysers, Harmsen, and Houlsby]{ScalingVisionTransformers}
Dehghani, M., Djolonga, J., Mustafa, B., Padlewski, P., Heek, J., Gilmer, J., Steiner, A.~P., Caron, M., Geirhos, R., Alabdulmohsin, I., Jenatton, R., Beyer, L., Tschannen, M., Arnab, A., Wang, X., Ruiz, C.~R., Minderer, M., Puigcerver, J., Evci, U., Kumar, M., van Steenkiste, S., Elsayed, G.~F., Mahendran, A., Yu, F., Oliver, A., Huot, F., Bastings, J., Collier, M., Gritsenko, A.~A., Birodkar, V., Vasconcelos, C.~N., Tay, Y., Mensink, T., Kolesnikov, A., Pavetic, F., Tran, D., Kipf, T., Lucic, M., Zhai, X., Keysers, D., Harmsen, J.~J., and Houlsby, N.
\newblock Scaling vision transformers to 22 billion parameters.
\newblock In Krause, A., Brunskill, E., Cho, K., Engelhardt, B., Sabato, S., and Scarlett, J. (eds.), \emph{International Conference on Machine Learning, {ICML} 2023, 23-29 July 2023, Honolulu, Hawaii, {USA}}, volume 202 of \emph{Proceedings of Machine Learning Research}, pp.\  7480--7512. {PMLR}, 2023.
\newblock URL \url{https://proceedings.mlr.press/v202/dehghani23a.html}.

\bibitem[Deshpande \& Narasimhan(2020)Deshpande and Narasimhan]{deshpandeGuidingAttentionSelfSupervised2020}
Deshpande, A. and Narasimhan, K.
\newblock Guiding attention for self-supervised learning with transformers.
\newblock In \emph{Findings of the Association for Computational Linguistics: EMNLP 2020}, pp.\  4676--4686, Online, 2020. Association for Computational Linguistics.
\newblock \doi{10.18653/v1/2020.findings-emnlp.419}.
\newblock URL \url{https://aclanthology.org/2020.findings-emnlp.419}.

\bibitem[Dettmers \& Zettlemoyer(2023)Dettmers and Zettlemoyer]{Case4bitPrecision}
Dettmers, T. and Zettlemoyer, L.
\newblock The case for 4-bit precision: k-bit inference scaling laws.
\newblock In Krause, A., Brunskill, E., Cho, K., Engelhardt, B., Sabato, S., and Scarlett, J. (eds.), \emph{International Conference on Machine Learning, {ICML} 2023, 23-29 July 2023, Honolulu, Hawaii, {USA}}, volume 202 of \emph{Proceedings of Machine Learning Research}, pp.\  7750--7774. {PMLR}, 2023.
\newblock URL \url{https://proceedings.mlr.press/v202/dettmers23a.html}.

\bibitem[Dettmers et~al.(2023)Dettmers, Pagnoni, Holtzman, and Zettlemoyer]{qlora}
Dettmers, T., Pagnoni, A., Holtzman, A., and Zettlemoyer, L.
\newblock {QL}o{RA}: Efficient finetuning of quantized {LLM}s.
\newblock In \emph{Thirty-seventh Conference on Neural Information Processing Systems}, 2023.
\newblock URL \url{https://openreview.net/forum?id=OUIFPHEgJU}.

\bibitem[Devlin et~al.(2019)Devlin, Chang, Lee, and Toutanova]{bert}
Devlin, J., Chang, M.-W., Lee, K., and Toutanova, K.
\newblock {BERT}: Pre-training of deep bidirectional transformers for language understanding.
\newblock In \emph{Proceedings of the 2019 Conference of the North {A}merican Chapter of the Association for Computational Linguistics: Human Language Technologies, Volume 1 (Long and Short Papers)}, pp.\  4171--4186, Minneapolis, Minnesota, 2019. Association for Computational Linguistics.
\newblock \doi{10.18653/v1/N19-1423}.
\newblock URL \url{https://aclanthology.org/N19-1423}.

\bibitem[Di~Gangi et~al.(2019)Di~Gangi, Cattoni, Bentivogli, Negri, and Turchi]{mustc}
Di~Gangi, M.~A., Cattoni, R., Bentivogli, L., Negri, M., and Turchi, M.
\newblock {M}u{ST}-{C}: a {M}ultilingual {S}peech {T}ranslation {C}orpus.
\newblock In \emph{Proceedings of the 2019 Conference of the North {A}merican Chapter of the Association for Computational Linguistics: Human Language Technologies, Volume 1 (Long and Short Papers)}, pp.\  2012--2017, Minneapolis, Minnesota, 2019. Association for Computational Linguistics.
\newblock \doi{10.18653/v1/N19-1202}.
\newblock URL \url{https://aclanthology.org/N19-1202}.

\bibitem[Dinan et~al.(2023)Dinan, Yaida, and Zhang]{EffectiveTheoryTransformers}
Dinan, E., Yaida, S., and Zhang, S.
\newblock Effective {{Theory}} of {{Transformers}} at {{Initialization}}, 2023.
\newblock URL \url{https://arxiv.org/abs/2304.02034}.

\bibitem[Dong et~al.(2021)Dong, Cordonnier, and Loukas]{AttentionNotAll}
Dong, Y., Cordonnier, J., and Loukas, A.
\newblock Attention is not all you need: pure attention loses rank doubly exponentially with depth.
\newblock In Meila, M. and Zhang, T. (eds.), \emph{Proceedings of the 38th International Conference on Machine Learning, {ICML} 2021, 18-24 July 2021, Virtual Event}, volume 139 of \emph{Proceedings of Machine Learning Research}, pp.\  2793--2803. {PMLR}, 2021.
\newblock URL \url{http://proceedings.mlr.press/v139/dong21a.html}.

\bibitem[Dosovitskiy et~al.(2021)Dosovitskiy, Beyer, Kolesnikov, Weissenborn, Zhai, Unterthiner, Dehghani, Minderer, Heigold, Gelly, Uszkoreit, and Houlsby]{vit}
Dosovitskiy, A., Beyer, L., Kolesnikov, A., Weissenborn, D., Zhai, X., Unterthiner, T., Dehghani, M., Minderer, M., Heigold, G., Gelly, S., Uszkoreit, J., and Houlsby, N.
\newblock An image is worth 16x16 words: Transformers for image recognition at scale.
\newblock In \emph{9th International Conference on Learning Representations, {ICLR} 2021, Virtual Event, Austria, May 3-7, 2021}. OpenReview.net, 2021.
\newblock URL \url{https://openreview.net/forum?id=YicbFdNTTy}.

\bibitem[Galanti(2022)]{ImplicitBiasMinimala}
Galanti, T.
\newblock A note on the implicit bias towards minimal depth of deep neural networks.
\newblock \emph{ArXiv preprint}, abs/2202.09028, 2022.
\newblock URL \url{https://arxiv.org/abs/2202.09028}.

\bibitem[Gao et~al.(2021)Gao, Biderman, Black, Golding, Hoppe, Foster, Phang, He, Thite, Nabeshima, Presser, and Leahy]{pile}
Gao, L., Biderman, S., Black, S., Golding, L., Hoppe, T., Foster, C., Phang, J., He, H., Thite, A., Nabeshima, N., Presser, S., and Leahy, C.
\newblock The pile: An 800gb dataset of diverse text for language modeling.
\newblock \emph{ArXiv preprint}, abs/2101.00027, 2021.
\newblock URL \url{https://arxiv.org/abs/2101.00027}.

\bibitem[Geiping \& Goldstein(2023)Geiping and Goldstein]{PPL3}
Geiping, J. and Goldstein, T.
\newblock Cramming: Training a language model on a single {GPU} in one day.
\newblock In Krause, A., Brunskill, E., Cho, K., Engelhardt, B., Sabato, S., and Scarlett, J. (eds.), \emph{International Conference on Machine Learning, {ICML} 2023, 23-29 July 2023, Honolulu, Hawaii, {USA}}, volume 202 of \emph{Proceedings of Machine Learning Research}, pp.\  11117--11143. {PMLR}, 2023.
\newblock URL \url{https://proceedings.mlr.press/v202/geiping23a.html}.

\bibitem[Glorot \& Bengio(2010)Glorot and Bengio]{glorot}
Glorot, X. and Bengio, Y.
\newblock Understanding the difficulty of training deep feedforward neural networks.
\newblock In Teh, Y.~W. and Titterington, D.~M. (eds.), \emph{Proceedings of the Thirteenth International Conference on Artificial Intelligence and Statistics, {AISTATS} 2010, Chia Laguna Resort, Sardinia, Italy, May 13-15, 2010}, volume~9 of \emph{{JMLR} Proceedings}, pp.\  249--256. JMLR.org, 2010.
\newblock URL \url{http://proceedings.mlr.press/v9/glorot10a.html}.

\bibitem[Gromov et~al.(2024)Gromov, Tirumala, Shapourian, Glorioso, and Roberts]{unreasonableIneffectiveness}
Gromov, A., Tirumala, K., Shapourian, H., Glorioso, P., and Roberts, D.~A.
\newblock The unreasonable ineffectiveness of the deeper layers.
\newblock \emph{ArXiv preprint}, abs/2403.17887, 2024.
\newblock URL \url{https://arxiv.org/abs/2403.17887}.

\bibitem[Hanin \& Rolnick(2018)Hanin and Rolnick]{HowStartTraining}
Hanin, B. and Rolnick, D.
\newblock How to start training: The effect of initialization and architecture.
\newblock In Bengio, S., Wallach, H.~M., Larochelle, H., Grauman, K., Cesa{-}Bianchi, N., and Garnett, R. (eds.), \emph{Advances in Neural Information Processing Systems 31: Annual Conference on Neural Information Processing Systems 2018, NeurIPS 2018, December 3-8, 2018, Montr{\'{e}}al, Canada}, pp.\  569--579, 2018.
\newblock URL \url{https://proceedings.neurips.cc/paper/2018/hash/d81f9c1be2e08964bf9f24b15f0e4900-Abstract.html}.

\bibitem[Hayou et~al.(2019)Hayou, Doucet, and Rousseau]{ImpactActivationFunction}
Hayou, S., Doucet, A., and Rousseau, J.
\newblock On the impact of the activation function on deep neural networks training.
\newblock In Chaudhuri, K. and Salakhutdinov, R. (eds.), \emph{Proceedings of the 36th International Conference on Machine Learning, {ICML} 2019, 9-15 June 2019, Long Beach, California, {USA}}, volume~97 of \emph{Proceedings of Machine Learning Research}, pp.\  2672--2680. {PMLR}, 2019.
\newblock URL \url{http://proceedings.mlr.press/v97/hayou19a.html}.

\bibitem[He \& Hofmann(2024)He and Hofmann]{SimplifyingTransformerBlocks}
He, B. and Hofmann, T.
\newblock Simplifying transformer blocks.
\newblock In \emph{The Twelfth International Conference on Learning Representations}, 2024.
\newblock URL \url{https://openreview.net/forum?id=RtDok9eS3s}.

\bibitem[He et~al.(2023)He, Martens, Zhang, Botev, Brock, Smith, and Teh]{DeepTransformersShortcutsa}
He, B., Martens, J., Zhang, G., Botev, A., Brock, A., Smith, S.~L., and Teh, Y.~W.
\newblock Deep transformers without shortcuts: Modifying self-attention for faithful signal propagation.
\newblock In \emph{The Eleventh International Conference on Learning Representations, {ICLR} 2023, Kigali, Rwanda, May 1-5, 2023}. OpenReview.net, 2023.
\newblock URL \url{https://openreview.net/pdf?id=NPrsUQgMjKK}.

\bibitem[He et~al.(2015)He, Zhang, Ren, and Sun]{DelvingDeepRectifiers}
He, K., Zhang, X., Ren, S., and Sun, J.
\newblock Delving deep into rectifiers: Surpassing human-level performance on imagenet classification.
\newblock In \emph{2015 {IEEE} International Conference on Computer Vision, {ICCV} 2015, Santiago, Chile, December 7-13, 2015}, pp.\  1026--1034. {IEEE} Computer Society, 2015.
\newblock \doi{10.1109/ICCV.2015.123}.
\newblock URL \url{https://doi.org/10.1109/ICCV.2015.123}.

\bibitem[He et~al.(2021{\natexlab{a}})He, Liu, Gao, and Chen]{Deberta}
He, P., Liu, X., Gao, J., and Chen, W.
\newblock Deberta: decoding-enhanced bert with disentangled attention.
\newblock In \emph{9th International Conference on Learning Representations, {ICLR} 2021, Virtual Event, Austria, May 3-7, 2021}. OpenReview.net, 2021{\natexlab{a}}.
\newblock URL \url{https://openreview.net/forum?id=XPZIaotutsD}.

\bibitem[He et~al.(2021{\natexlab{b}})He, Ravula, Kanagal, and Ainslie]{RealFormerTransformerLikes}
He, R., Ravula, A., Kanagal, B., and Ainslie, J.
\newblock {R}eal{F}ormer: Transformer likes residual attention.
\newblock In \emph{Findings of the Association for Computational Linguistics: ACL-IJCNLP 2021}, pp.\  929--943, Online, 2021{\natexlab{b}}. Association for Computational Linguistics.
\newblock \doi{10.18653/v1/2021.findings-acl.81}.
\newblock URL \url{https://aclanthology.org/2021.findings-acl.81}.

\bibitem[Hendrycks et~al.(2021)Hendrycks, Basart, Mu, Kadavath, Wang, Dorundo, Desai, Zhu, Parajuli, Guo, Song, Steinhardt, and Gilmer]{ImagenetR}
Hendrycks, D., Basart, S., Mu, N., Kadavath, S., Wang, F., Dorundo, E., Desai, R., Zhu, T., Parajuli, S., Guo, M., Song, D., Steinhardt, J., and Gilmer, J.
\newblock The many faces of robustness: {A} critical analysis of out-of-distribution generalization.
\newblock In \emph{2021 {IEEE/CVF} International Conference on Computer Vision, {ICCV} 2021, Montreal, QC, Canada, October 10-17, 2021}, pp.\  8320--8329. {IEEE}, 2021.
\newblock \doi{10.1109/ICCV48922.2021.00823}.
\newblock URL \url{https://doi.org/10.1109/ICCV48922.2021.00823}.

\bibitem[Hoedt et~al.(2022)Hoedt, Hochreiter, and Klambauer]{NormalisationDead}
Hoedt, P.-J., Hochreiter, S., and Klambauer, G.
\newblock Normalisation is dead, long live normalisation!
\newblock In \emph{ICLR Blog Track}, 2022.
\newblock URL \url{https://iclr-blog-track.github.io/2022/03/25/unnormalized-resnets/}.
\newblock https://iclr-blog-track.github.io/2022/03/25/unnormalized-resnets/.

\bibitem[Hoffmann et~al.(2022)Hoffmann, Borgeaud, Mensch, Buchatskaya, Cai, Rutherford, de~Las~Casas, Hendricks, Welbl, Clark, Hennigan, Noland, Millican, van~den Driessche, Damoc, Guy, Osindero, Simonyan, Elsen, Vinyals, Rae, and Sifre]{chinchilla}
Hoffmann, J., Borgeaud, S., Mensch, A., Buchatskaya, E., Cai, T., Rutherford, E., de~Las~Casas, D., Hendricks, L.~A., Welbl, J., Clark, A., Hennigan, T., Noland, E., Millican, K., van~den Driessche, G., Damoc, B., Guy, A., Osindero, S., Simonyan, K., Elsen, E., Vinyals, O., Rae, J.~W., and Sifre, L.
\newblock An empirical analysis of compute-optimal large language model training.
\newblock In Koyejo, S., Mohamed, S., Agarwal, A., Belgrave, D., Cho, K., and Oh, A. (eds.), \emph{Advances in Neural Information Processing Systems 35: Annual Conference on Neural Information Processing Systems 2022, NeurIPS 2022, New Orleans, LA, USA, November 28 - December 9, 2022}, 2022.
\newblock URL \url{http://papers.nips.cc/paper\_files/paper/2022/hash/c1e2faff6f588870935f114ebe04a3e5-Abstract-Conference.html}.

\bibitem[Hu et~al.(2022)Hu, Shen, Wallis, Allen{-}Zhu, Li, Wang, Wang, and Chen]{lora}
Hu, E.~J., Shen, Y., Wallis, P., Allen{-}Zhu, Z., Li, Y., Wang, S., Wang, L., and Chen, W.
\newblock Lora: Low-rank adaptation of large language models.
\newblock In \emph{The Tenth International Conference on Learning Representations, {ICLR} 2022, Virtual Event, April 25-29, 2022}. OpenReview.net, 2022.
\newblock URL \url{https://openreview.net/forum?id=nZeVKeeFYf9}.

\bibitem[Huang et~al.(2020{\natexlab{a}})Huang, P{\'{e}}rez, Ba, and Volkovs]{Tfixup}
Huang, X.~S., P{\'{e}}rez, F., Ba, J., and Volkovs, M.
\newblock Improving transformer optimization through better initialization.
\newblock In \emph{Proceedings of the 37th International Conference on Machine Learning, {ICML} 2020, 13-18 July 2020, Virtual Event}, volume 119 of \emph{Proceedings of Machine Learning Research}, pp.\  4475--4483. {PMLR}, 2020{\natexlab{a}}.
\newblock URL \url{http://proceedings.mlr.press/v119/huang20f.html}.

\bibitem[Huang et~al.(2020{\natexlab{b}})Huang, P{\'{e}}rez, Ba, and Volkovs]{huang2020improving}
Huang, X.~S., P{\'{e}}rez, F., Ba, J., and Volkovs, M.
\newblock Improving transformer optimization through better initialization.
\newblock In \emph{Proceedings of the 37th International Conference on Machine Learning, {ICML} 2020, 13-18 July 2020, Virtual Event}, volume 119 of \emph{Proceedings of Machine Learning Research}, pp.\  4475--4483. {PMLR}, 2020{\natexlab{b}}.
\newblock URL \url{http://proceedings.mlr.press/v119/huang20f.html}.

\bibitem[Ioffe \& Szegedy(2015)Ioffe and Szegedy]{batchnorm}
Ioffe, S. and Szegedy, C.
\newblock Batch normalization: Accelerating deep network training by reducing internal covariate shift.
\newblock In Bach, F.~R. and Blei, D.~M. (eds.), \emph{Proceedings of the 32nd International Conference on Machine Learning, {ICML} 2015, Lille, France, 6-11 July 2015}, volume~37 of \emph{{JMLR} Workshop and Conference Proceedings}, pp.\  448--456. JMLR.org, 2015.
\newblock URL \url{http://proceedings.mlr.press/v37/ioffe15.html}.

\bibitem[Jesus et~al.(2021)Jesus, Antunes, {da Costa}, Dorogovtsev, Mendes, and Aguiar]{EffectInitialConfiguration}
Jesus, R.~J., Antunes, M.~L., {da Costa}, R.~A., Dorogovtsev, S.~N., Mendes, J. F.~F., and Aguiar, R.~L.
\newblock Effect of {{Initial Configuration}} of {{Weights}} on {{Training}} and {{Function}} of {{Artificial Neural Networks}}.
\newblock \emph{Mathematics}, 9\penalty0 (18):\penalty0 2246, 2021.
\newblock ISSN 2227-7390.
\newblock \doi{10/gsshxg}.

\bibitem[Karras et~al.(2023)Karras, Aittala, Lehtinen, Hellsten, Aila, and Laine]{karrasAnalyzingImprovingTraining2024}
Karras, T., Aittala, M., Lehtinen, J., Hellsten, J., Aila, T., and Laine, S.
\newblock Analyzing and {{Improving}} the {{Training Dynamics}} of {{Diffusion Models}}, 2023.
\newblock URL \url{https://arxiv.org/abs/2312.02696}.

\bibitem[Kim et~al.(2021)Kim, Papamakarios, and Mnih]{LipschitzConstantSelfAttention}
Kim, H., Papamakarios, G., and Mnih, A.
\newblock The lipschitz constant of self-attention.
\newblock In Meila, M. and Zhang, T. (eds.), \emph{Proceedings of the 38th International Conference on Machine Learning, {ICML} 2021, 18-24 July 2021, Virtual Event}, volume 139 of \emph{Proceedings of Machine Learning Research}, pp.\  5562--5571. {PMLR}, 2021.
\newblock URL \url{http://proceedings.mlr.press/v139/kim21i.html}.

\bibitem[Kingsley(1935)]{zipf}
Kingsley, Z.~G.
\newblock \emph{The psycho-biology of language: an introduction to dynamic philology}.
\newblock Houghton Mifflin, 1935.

\bibitem[Klambauer et~al.(2017)Klambauer, Unterthiner, Mayr, and Hochreiter]{SelfNormalizingNeuralNetworksb}
Klambauer, G., Unterthiner, T., Mayr, A., and Hochreiter, S.
\newblock Self-normalizing neural networks.
\newblock In Guyon, I., von Luxburg, U., Bengio, S., Wallach, H.~M., Fergus, R., Vishwanathan, S. V.~N., and Garnett, R. (eds.), \emph{Advances in Neural Information Processing Systems 30: Annual Conference on Neural Information Processing Systems 2017, December 4-9, 2017, Long Beach, CA, {USA}}, pp.\  971--980, 2017.
\newblock URL \url{https://proceedings.neurips.cc/paper/2017/hash/5d44ee6f2c3f71b73125876103c8f6c4-Abstract.html}.

\bibitem[Korotkov \& Korotkov(2020)Korotkov and Korotkov]{integrals_book}
Korotkov, N.~E. and Korotkov, A.~N.
\newblock \emph{Integrals related to the error function}.
\newblock Chapman \& Hall/CRC, Philadelphia, PA, 2020.
\newblock ISBN 9780367408206.
\newblock URL \url{https://www.taylorfrancis.com/books/mono/10.1201/9780367809232/integrals-related-error-function-nikolai-korotkov-alexander-korotkov}.

\bibitem[Labatie et~al.(2021)Labatie, Masters, Eaton{-}Rosen, and Luschi]{ProxyNormalizingActivationsMatcha}
Labatie, A., Masters, D., Eaton{-}Rosen, Z., and Luschi, C.
\newblock Proxy-normalizing activations to match batch normalization while removing batch dependence.
\newblock In Ranzato, M., Beygelzimer, A., Dauphin, Y.~N., Liang, P., and Vaughan, J.~W. (eds.), \emph{Advances in Neural Information Processing Systems 34: Annual Conference on Neural Information Processing Systems 2021, NeurIPS 2021, December 6-14, 2021, virtual}, pp.\  16990--17006, 2021.
\newblock URL \url{https://proceedings.neurips.cc/paper/2021/hash/8d2a5f7d4afa5d0530789d3066945330-Abstract.html}.

\bibitem[Lai et~al.(2017)Lai, Xie, Liu, Yang, and Hovy]{race}
Lai, G., Xie, Q., Liu, H., Yang, Y., and Hovy, E.
\newblock {RACE}: Large-scale {R}e{A}ding comprehension dataset from examinations.
\newblock In \emph{Proceedings of the 2017 Conference on Empirical Methods in Natural Language Processing}, pp.\  785--794, Copenhagen, Denmark, 2017. Association for Computational Linguistics.
\newblock \doi{10.18653/v1/D17-1082}.
\newblock URL \url{https://aclanthology.org/D17-1082}.

\bibitem[LeCun et~al.(1996)LeCun, Bottou, Orr, and M{\"{u}}ller]{rel_lecun}
LeCun, Y., Bottou, L., Orr, G.~B., and M{\"{u}}ller, K.
\newblock Effiicient backprop.
\newblock In Orr, G.~B. and M{\"{u}}ller, K. (eds.), \emph{Neural Networks: Tricks of the Trade}, volume 1524 of \emph{Lecture Notes in Computer Science}, pp.\  9--50. Springer, 1996.
\newblock \doi{10.1007/3-540-49430-8\_2}.
\newblock URL \url{https://doi.org/10.1007/3-540-49430-8\_2}.

\bibitem[Lee et~al.(2019)Lee, Xiao, Schoenholz, Bahri, Novak, Sohl{-}Dickstein, and Pennington]{WideNeuralNetworks}
Lee, J., Xiao, L., Schoenholz, S.~S., Bahri, Y., Novak, R., Sohl{-}Dickstein, J., and Pennington, J.
\newblock Wide neural networks of any depth evolve as linear models under gradient descent.
\newblock In Wallach, H.~M., Larochelle, H., Beygelzimer, A., d'Alch{\'{e}}{-}Buc, F., Fox, E.~B., and Garnett, R. (eds.), \emph{Advances in Neural Information Processing Systems 32: Annual Conference on Neural Information Processing Systems 2019, NeurIPS 2019, December 8-14, 2019, Vancouver, BC, Canada}, pp.\  8570--8581, 2019.
\newblock URL \url{https://proceedings.neurips.cc/paper/2019/hash/0d1a9651497a38d8b1c3871c84528bd4-Abstract.html}.

\bibitem[Levine et~al.(2020)Levine, Wies, Sharir, Bata, and Shashua]{DepthtoWidthInterplaySelfAttention}
Levine, Y., Wies, N., Sharir, O., Bata, H., and Shashua, A.
\newblock Limits to depth efficiencies of self-attention.
\newblock In Larochelle, H., Ranzato, M., Hadsell, R., Balcan, M., and Lin, H. (eds.), \emph{Advances in Neural Information Processing Systems 33: Annual Conference on Neural Information Processing Systems 2020, NeurIPS 2020, December 6-12, 2020, virtual}, 2020.
\newblock URL \url{https://proceedings.neurips.cc/paper/2020/hash/ff4dfdf5904e920ce52b48c1cef97829-Abstract.html}.

\bibitem[Li et~al.(2022)Li, Efros, and Pathak]{liUnderstandingCollapseNonContrastive2022}
Li, A.~C., Efros, A.~A., and Pathak, D.
\newblock Understanding {{Collapse}} in {{Non-Contrastive Siamese Representation Learning}}, 2022.
\newblock URL \url{https://arxiv.org/abs/2209.15007}.

\bibitem[Li \& Liang(2018)Li and Liang]{LearningOverparameterizedNeural}
Li, Y. and Liang, Y.
\newblock Learning overparameterized neural networks via stochastic gradient descent on structured data.
\newblock In Bengio, S., Wallach, H.~M., Larochelle, H., Grauman, K., Cesa{-}Bianchi, N., and Garnett, R. (eds.), \emph{Advances in Neural Information Processing Systems 31: Annual Conference on Neural Information Processing Systems 2018, NeurIPS 2018, December 3-8, 2018, Montr{\'{e}}al, Canada}, pp.\  8168--8177, 2018.
\newblock URL \url{https://proceedings.neurips.cc/paper/2018/hash/54fe976ba170c19ebae453679b362263-Abstract.html}.

\bibitem[Lin et~al.(2021)Lin, Sekar, and Fanti]{linWhySpectralNormalization2021}
Lin, Z., Sekar, V., and Fanti, G.
\newblock Why spectral normalization stabilizes gans: Analysis and improvements.
\newblock In Ranzato, M., Beygelzimer, A., Dauphin, Y.~N., Liang, P., and Vaughan, J.~W. (eds.), \emph{Advances in Neural Information Processing Systems 34: Annual Conference on Neural Information Processing Systems 2021, NeurIPS 2021, December 6-14, 2021, virtual}, pp.\  9625--9638, 2021.
\newblock URL \url{https://proceedings.neurips.cc/paper/2021/hash/4ffb0d2ba92f664c2281970110a2e071-Abstract.html}.

\bibitem[Liu et~al.(2019{\natexlab{a}})Liu, Gao, Liu, and Lei]{SelfAdaptiveScalingLearnableb}
Liu, F., Gao, M., Liu, Y., and Lei, K.
\newblock Self-adaptive scaling for learnable residual structure.
\newblock In \emph{Proceedings of the 23rd Conference on Computational Natural Language Learning (CoNLL)}, pp.\  862--870, Hong Kong, China, 2019{\natexlab{a}}. Association for Computational Linguistics.
\newblock \doi{10.18653/v1/K19-1080}.
\newblock URL \url{https://aclanthology.org/K19-1080}.

\bibitem[Liu et~al.(2020{\natexlab{a}})Liu, Liu, Gao, Chen, and Han]{UnderstandingDifficultyTraininga}
Liu, L., Liu, X., Gao, J., Chen, W., and Han, J.
\newblock Understanding the difficulty of training transformers.
\newblock In \emph{Proceedings of the 2020 Conference on Empirical Methods in Natural Language Processing (EMNLP)}, pp.\  5747--5763, Online, 2020{\natexlab{a}}. Association for Computational Linguistics.
\newblock \doi{10.18653/v1/2020.emnlp-main.463}.
\newblock URL \url{https://aclanthology.org/2020.emnlp-main.463}.

\bibitem[Liu et~al.(2020{\natexlab{b}})Liu, Duh, Liu, and Gao]{verydeep}
Liu, X., Duh, K., Liu, L., and Gao, J.
\newblock Very deep transformers for neural machine translation.
\newblock \emph{ArXiv preprint}, abs/2008.07772, 2020{\natexlab{b}}.
\newblock URL \url{https://arxiv.org/abs/2008.07772}.

\bibitem[Liu et~al.(2019{\natexlab{b}})Liu, Ott, Goyal, Du, Joshi, Chen, Levy, Lewis, Zettlemoyer, and Stoyanov]{Roberta}
Liu, Y., Ott, M., Goyal, N., Du, J., Joshi, M., Chen, D., Levy, O., Lewis, M., Zettlemoyer, L., and Stoyanov, V.
\newblock Roberta: {A} robustly optimized {BERT} pretraining approach.
\newblock \emph{ArXiv preprint}, abs/1907.11692, 2019{\natexlab{b}}.
\newblock URL \url{https://arxiv.org/abs/1907.11692}.

\bibitem[Lo(2013)]{LognormalApprox}
Lo, C.~F.
\newblock {{WKB}} approximation for the sum of two correlated lognormal random variables.
\newblock \emph{Applied Mathematical Sciences}, 7:\penalty0 6355--6367, 2013.
\newblock ISSN 13147552.
\newblock \doi{10.12988/ams.2013.39511}.
\newblock URL \url{http://www.m-hikari.com/ams/ams-2013/ams-125-128-2013/39511.html}.

\bibitem[Lu et~al.(2023)Lu, Zhu, Han, Zhao, Namee, and Tan]{PPL2}
Lu, J., Zhu, D., Han, W., Zhao, R., Namee, B.~M., and Tan, F.
\newblock What makes pre-trained language models better zero-shot learners?
\newblock In Rogers, A., Boyd{-}Graber, J.~L., and Okazaki, N. (eds.), \emph{Proceedings of the 61st Annual Meeting of the Association for Computational Linguistics (Volume 1: Long Papers), {ACL} 2023, Toronto, Canada, July 9-14, 2023}, pp.\  2288--2303. Association for Computational Linguistics, 2023.
\newblock \doi{10.18653/V1/2023.ACL-LONG.128}.
\newblock URL \url{https://doi.org/10.18653/v1/2023.acl-long.128}.

\bibitem[Lu et~al.(2017)Lu, Pu, Wang, Hu, and Wang]{ExpressivePowerNeuralc}
Lu, Z., Pu, H., Wang, F., Hu, Z., and Wang, L.
\newblock The expressive power of neural networks: {A} view from the width.
\newblock In Guyon, I., von Luxburg, U., Bengio, S., Wallach, H.~M., Fergus, R., Vishwanathan, S. V.~N., and Garnett, R. (eds.), \emph{Advances in Neural Information Processing Systems 30: Annual Conference on Neural Information Processing Systems 2017, December 4-9, 2017, Long Beach, CA, {USA}}, pp.\  6231--6239, 2017.
\newblock URL \url{https://proceedings.neurips.cc/paper/2017/hash/32cbf687880eb1674a07bf717761dd3a-Abstract.html}.

\bibitem[Marion et~al.(2022)Marion, Fermanian, Biau, and Vert]{ScalingResNetsLargedeptha}
Marion, P., Fermanian, A., Biau, G., and Vert, J.
\newblock Scaling resnets in the large-depth regime.
\newblock \emph{ArXiv preprint}, abs/2206.06929, 2022.
\newblock URL \url{https://arxiv.org/abs/2206.06929}.

\bibitem[Martens et~al.(2021)Martens, Ballard, Desjardins, Swirszcz, Dalibard, Sohl{-}Dickstein, and Schoenholz]{RapidTrainingDeep}
Martens, J., Ballard, A., Desjardins, G., Swirszcz, G., Dalibard, V., Sohl{-}Dickstein, J., and Schoenholz, S.~S.
\newblock Rapid training of deep neural networks without skip connections or normalization layers using deep kernel shaping.
\newblock \emph{ArXiv preprint}, abs/2110.01765, 2021.
\newblock URL \url{https://arxiv.org/abs/2110.01765}.

\bibitem[Micikevicius et~al.(2022)Micikevicius, Stosic, Burgess, Cornea, Dubey, Grisenthwaite, Ha, Heinecke, Judd, Kamalu, Mellempudi, Oberman, Shoeybi, Siu, and Wu]{FP8}
Micikevicius, P., Stosic, D., Burgess, N., Cornea, M., Dubey, P., Grisenthwaite, R., Ha, S., Heinecke, A., Judd, P., Kamalu, J., Mellempudi, N., Oberman, S.~F., Shoeybi, M., Siu, M.~Y., and Wu, H.
\newblock {FP8} formats for deep learning.
\newblock \emph{ArXiv preprint}, abs/2209.05433, 2022.
\newblock URL \url{https://arxiv.org/abs/2209.05433}.

\bibitem[Mishkin \& Matas(2016)Mishkin and Matas]{AllYouNeed}
Mishkin, D. and Matas, J.
\newblock All you need is a good init.
\newblock In Bengio, Y. and LeCun, Y. (eds.), \emph{4th International Conference on Learning Representations, {ICLR} 2016, San Juan, Puerto Rico, May 2-4, 2016, Conference Track Proceedings}, 2016.
\newblock URL \url{http://arxiv.org/abs/1511.06422}.

\bibitem[Molybog et~al.(2023)Molybog, Albert, Chen, DeVito, Esiobu, Goyal, Koura, Narang, Poulton, Silva, Tang, Liskovich, Xu, Zhang, Kambadur, Roller, and Zhang]{adamInstability}
Molybog, I., Albert, P., Chen, M., DeVito, Z., Esiobu, D., Goyal, N., Koura, P.~S., Narang, S., Poulton, A., Silva, R., Tang, B., Liskovich, D., Xu, P., Zhang, Y., Kambadur, M., Roller, S., and Zhang, S.
\newblock A theory on adam instability in large-scale machine learning.
\newblock \emph{ArXiv preprint}, abs/2304.09871, 2023.
\newblock URL \url{https://arxiv.org/abs/2304.09871}.

\bibitem[Mont{\'{u}}far et~al.(2014)Mont{\'{u}}far, Pascanu, Cho, and Bengio]{montufar2014number}
Mont{\'{u}}far, G.~F., Pascanu, R., Cho, K., and Bengio, Y.
\newblock On the number of linear regions of deep neural networks.
\newblock In Ghahramani, Z., Welling, M., Cortes, C., Lawrence, N.~D., and Weinberger, K.~Q. (eds.), \emph{Advances in Neural Information Processing Systems 27: Annual Conference on Neural Information Processing Systems 2014, December 8-13 2014, Montreal, Quebec, Canada}, pp.\  2924--2932, 2014.
\newblock URL \url{https://proceedings.neurips.cc/paper/2014/hash/109d2dd3608f669ca17920c511c2a41e-Abstract.html}.

\bibitem[Neal(1995)]{rel_neal}
Neal, R.~M.
\newblock \emph{Bayesian learning for neural networks}.
\newblock PhD thesis, University of Toronto, Canada, 1995.
\newblock URL \url{https://librarysearch.library.utoronto.ca/permalink/01UTORONTO\_INST/14bjeso/alma991106438365706196}.

\bibitem[Ng \& Geller(1969)Ng and Geller]{TableIntegralsErrora}
Ng, E.~W. and Geller, M.
\newblock A table of integrals of the {{Error}} functions.
\newblock \emph{Journal of Research of the National Bureau of Standards, Section B: Mathematical Sciences}, 73B\penalty0 (1):\penalty0 1, 1969.
\newblock ISSN 0098-8979.
\newblock \doi{10/gdtk9p}.
\newblock URL \url{https://nvlpubs.nist.gov/nistpubs/jres/73B/jresv73Bn1p1\_A1b.pdf}.

\bibitem[Nguyen \& Salazar(2019)Nguyen and Salazar]{TransformersTearsImprovinga}
Nguyen, T.~Q. and Salazar, J.
\newblock Transformers without tears: Improving the normalization of self-attention.
\newblock In \emph{Proceedings of the 16th International Conference on Spoken Language Translation}, Hong Kong, 2019. Association for Computational Linguistics.
\newblock URL \url{https://aclanthology.org/2019.iwslt-1.17}.

\bibitem[Noci et~al.(2022)Noci, Anagnostidis, Biggio, Orvieto, Singh, and Lucchi]{noci2022signal}
Noci, L., Anagnostidis, S., Biggio, L., Orvieto, A., Singh, S.~P., and Lucchi, A.
\newblock Signal propagation in transformers: Theoretical perspectives and the role of rank collapse.
\newblock In \emph{NeurIPS}, 2022.
\newblock URL \url{http://papers.nips.cc/paper\\_files/paper/2022/hash/ae0cba715b60c4052359b3d52a2cff7f-Abstract-Conference.html}.

\bibitem[Noci et~al.(2023)Noci, Li, Li, He, Hofmann, Maddison, and Roy]{ShapedTransformerAttention}
Noci, L., Li, C., Li, M.~B., He, B., Hofmann, T., Maddison, C.~J., and Roy, D.
\newblock The shaped transformer: Attention models in the infinite depth-and-width limit.
\newblock In Oh, A., Naumann, T., Globerson, A., Saenko, K., Hardt, M., and Levine, S. (eds.), \emph{Advances in Neural Information Processing Systems 36: Annual Conference on Neural Information Processing Systems 2023, NeurIPS 2023, New Orleans, LA, USA, December 10 - 16, 2023}, 2023.
\newblock URL \url{http://papers.nips.cc/paper\_files/paper/2023/hash/aa31dc84098add7dd2ffdd20646f2043-Abstract-Conference.html}.

\bibitem[Ott et~al.(2019)Ott, Edunov, Baevski, Fan, Gross, Ng, Grangier, and Auli]{ott2019fairseq}
Ott, M., Edunov, S., Baevski, A., Fan, A., Gross, S., Ng, N., Grangier, D., and Auli, M.
\newblock fairseq: A fast, extensible toolkit for sequence modeling.
\newblock In \emph{Proceedings of the 2019 Conference of the North {A}merican Chapter of the Association for Computational Linguistics (Demonstrations)}, pp.\  48--53, Minneapolis, Minnesota, 2019. Association for Computational Linguistics.
\newblock \doi{10.18653/v1/N19-4009}.
\newblock URL \url{https://aclanthology.org/N19-4009}.

\bibitem[Peer et~al.(2022)Peer, Keulen, Stabinger, Piater, and Rodr{\'{\i}}guez{-}S{\'{a}}nchez]{ImprovingTrainabilityDeep}
Peer, D., Keulen, B., Stabinger, S., Piater, J.~H., and Rodr{\'{\i}}guez{-}S{\'{a}}nchez, A.~J.
\newblock Improving the trainability of deep neural networks through layerwise batch-entropy regularization.
\newblock \emph{Trans. Mach. Learn. Res.}, 2022, 2022.
\newblock URL \url{https://openreview.net/forum?id=LJohl5DnZf}.

\bibitem[Poole et~al.(2016)Poole, Lahiri, Raghu, Sohl{-}Dickstein, and Ganguli]{ExponentialExpressivity}
Poole, B., Lahiri, S., Raghu, M., Sohl{-}Dickstein, J., and Ganguli, S.
\newblock Exponential expressivity in deep neural networks through transient chaos.
\newblock In Lee, D.~D., Sugiyama, M., von Luxburg, U., Guyon, I., and Garnett, R. (eds.), \emph{Advances in Neural Information Processing Systems 29: Annual Conference on Neural Information Processing Systems 2016, December 5-10, 2016, Barcelona, Spain}, pp.\  3360--3368, 2016.
\newblock URL \url{https://proceedings.neurips.cc/paper/2016/hash/148510031349642de5ca0c544f31b2ef-Abstract.html}.

\bibitem[Qi et~al.(2023)Qi, Wang, Chen, Shi, and Zhang]{qiLipsFormerIntroducingLipschitz2023}
Qi, X., Wang, J., Chen, Y., Shi, Y., and Zhang, L.
\newblock Lipsformer: Introducing lipschitz continuity to vision transformers.
\newblock In \emph{The Eleventh International Conference on Learning Representations, {ICLR} 2023, Kigali, Rwanda, May 1-5, 2023}. OpenReview.net, 2023.
\newblock URL \url{https://openreview.net/pdf?id=cHf1DcCwcH3}.

\bibitem[Rae et~al.(2021)Rae, Borgeaud, Cai, Millican, Hoffmann, Song, Aslanides, Henderson, Ring, Young, Rutherford, Hennigan, Menick, Cassirer, Powell, van~den Driessche, Hendricks, Rauh, Huang, Glaese, Welbl, Dathathri, Huang, Uesato, Mellor, Higgins, Creswell, McAleese, Wu, Elsen, Jayakumar, Buchatskaya, Budden, Sutherland, Simonyan, Paganini, Sifre, Martens, Li, Kuncoro, Nematzadeh, Gribovskaya, Donato, Lazaridou, Mensch, Lespiau, Tsimpoukelli, Grigorev, Fritz, Sottiaux, Pajarskas, Pohlen, Gong, Toyama, de~Masson~d'Autume, Li, Terzi, Mikulik, Babuschkin, Clark, de~Las~Casas, Guy, Jones, Bradbury, Johnson, Hechtman, Weidinger, Gabriel, Isaac, Lockhart, Osindero, Rimell, Dyer, Vinyals, Ayoub, Stanway, Bennett, Hassabis, Kavukcuoglu, and Irving]{gopher}
Rae, J.~W., Borgeaud, S., Cai, T., Millican, K., Hoffmann, J., Song, H.~F., Aslanides, J., Henderson, S., Ring, R., Young, S., Rutherford, E., Hennigan, T., Menick, J., Cassirer, A., Powell, R., van~den Driessche, G., Hendricks, L.~A., Rauh, M., Huang, P., Glaese, A., Welbl, J., Dathathri, S., Huang, S., Uesato, J., Mellor, J., Higgins, I., Creswell, A., McAleese, N., Wu, A., Elsen, E., Jayakumar, S.~M., Buchatskaya, E., Budden, D., Sutherland, E., Simonyan, K., Paganini, M., Sifre, L., Martens, L., Li, X.~L., Kuncoro, A., Nematzadeh, A., Gribovskaya, E., Donato, D., Lazaridou, A., Mensch, A., Lespiau, J., Tsimpoukelli, M., Grigorev, N., Fritz, D., Sottiaux, T., Pajarskas, M., Pohlen, T., Gong, Z., Toyama, D., de~Masson~d'Autume, C., Li, Y., Terzi, T., Mikulik, V., Babuschkin, I., Clark, A., de~Las~Casas, D., Guy, A., Jones, C., Bradbury, J., Johnson, M.~J., Hechtman, B.~A., Weidinger, L., Gabriel, I., Isaac, W., Lockhart, E., Osindero, S., Rimell, L., Dyer, C., Vinyals, O., Ayoub, K., Stanway, J., Bennett,
  L., Hassabis, D., Kavukcuoglu, K., and Irving, G.
\newblock Scaling language models: Methods, analysis {\&} insights from training gopher.
\newblock \emph{ArXiv preprint}, abs/2112.11446, 2021.
\newblock URL \url{https://arxiv.org/abs/2112.11446}.

\bibitem[Raghu et~al.(2017)Raghu, Poole, Kleinberg, Ganguli, and Sohl{-}Dickstein]{ExpressivePowerDeepa}
Raghu, M., Poole, B., Kleinberg, J.~M., Ganguli, S., and Sohl{-}Dickstein, J.
\newblock On the expressive power of deep neural networks.
\newblock In Precup, D. and Teh, Y.~W. (eds.), \emph{Proceedings of the 34th International Conference on Machine Learning, {ICML} 2017, Sydney, NSW, Australia, 6-11 August 2017}, volume~70 of \emph{Proceedings of Machine Learning Research}, pp.\  2847--2854. {PMLR}, 2017.
\newblock URL \url{http://proceedings.mlr.press/v70/raghu17a.html}.

\bibitem[Recht et~al.(2019)Recht, Roelofs, Schmidt, and Shankar]{ImagenetV2}
Recht, B., Roelofs, R., Schmidt, L., and Shankar, V.
\newblock Do imagenet classifiers generalize to imagenet?
\newblock In Chaudhuri, K. and Salakhutdinov, R. (eds.), \emph{Proceedings of the 36th International Conference on Machine Learning, {ICML} 2019, 9-15 June 2019, Long Beach, California, {USA}}, volume~97 of \emph{Proceedings of Machine Learning Research}, pp.\  5389--5400. {PMLR}, 2019.
\newblock URL \url{http://proceedings.mlr.press/v97/recht19a.html}.

\bibitem[Roberts et~al.(2021)Roberts, Yaida, and Hanin]{robertsPrinciplesDeepLearning2022}
Roberts, D.~A., Yaida, S., and Hanin, B.
\newblock The principles of deep learning theory.
\newblock \emph{ArXiv preprint}, abs/2106.10165, 2021.
\newblock URL \url{https://arxiv.org/abs/2106.10165}.

\bibitem[Rong et~al.(2020)Rong, Huang, Xu, and Huang]{DropEdgeDeepGraph}
Rong, Y., Huang, W., Xu, T., and Huang, J.
\newblock Dropedge: Towards deep graph convolutional networks on node classification.
\newblock In \emph{8th International Conference on Learning Representations, {ICLR} 2020, Addis Ababa, Ethiopia, April 26-30, 2020}. OpenReview.net, 2020.
\newblock URL \url{https://openreview.net/forum?id=Hkx1qkrKPr}.

\bibitem[Russakovsky et~al.(2015)Russakovsky, Deng, Su, Krause, Satheesh, Ma, Huang, Karpathy, Khosla, Bernstein, Berg, and Fei{-}Fei]{Imagenet}
Russakovsky, O., Deng, J., Su, H., Krause, J., Satheesh, S., Ma, S., Huang, Z., Karpathy, A., Khosla, A., Bernstein, M.~S., Berg, A.~C., and Fei{-}Fei, L.
\newblock Imagenet large scale visual recognition challenge.
\newblock \emph{Int. J. Comput. Vis.}, 115\penalty0 (3):\penalty0 211--252, 2015.
\newblock \doi{10.1007/S11263-015-0816-Y}.
\newblock URL \url{https://doi.org/10.1007/s11263-015-0816-y}.

\bibitem[Salazar et~al.(2020)Salazar, Liang, Nguyen, and Kirchhoff]{PPL1}
Salazar, J., Liang, D., Nguyen, T.~Q., and Kirchhoff, K.
\newblock Masked language model scoring.
\newblock In \emph{Proceedings of the 58th Annual Meeting of the Association for Computational Linguistics}, pp.\  2699--2712, Online, 2020. Association for Computational Linguistics.
\newblock \doi{10.18653/v1/2020.acl-main.240}.
\newblock URL \url{https://aclanthology.org/2020.acl-main.240}.

\bibitem[Schoenholz et~al.(2017)Schoenholz, Gilmer, Ganguli, and Sohl{-}Dickstein]{DeepInformationPropagation}
Schoenholz, S.~S., Gilmer, J., Ganguli, S., and Sohl{-}Dickstein, J.
\newblock Deep information propagation.
\newblock In \emph{5th International Conference on Learning Representations, {ICLR} 2017, Toulon, France, April 24-26, 2017, Conference Track Proceedings}. OpenReview.net, 2017.
\newblock URL \url{https://openreview.net/forum?id=H1W1UN9gg}.

\bibitem[Shao et~al.(2020)Shao, Hu, Wang, Xue, and Raj]{NormalizationIndispensableTraininga}
Shao, J., Hu, K., Wang, C., Xue, X., and Raj, B.
\newblock Is normalization indispensable for training deep neural networks?
\newblock In \emph{Proceedings of the 34th {{International Conference}} on {{Neural Information Processing Systems}}}, {{NIPS}}'20, pp.\  13434--13444, Red Hook, NY, USA, 2020. Curran Associates Inc.
\newblock ISBN 978-1-71382-954-6.

\bibitem[Shen et~al.(2020)Shen, Yao, Gholami, Mahoney, and Keutzer]{PowerNormRethinkingBatcha}
Shen, S., Yao, Z., Gholami, A., Mahoney, M.~W., and Keutzer, K.
\newblock Powernorm: Rethinking batch normalization in transformers.
\newblock In \emph{Proceedings of the 37th International Conference on Machine Learning, {ICML} 2020, 13-18 July 2020, Virtual Event}, volume 119 of \emph{Proceedings of Machine Learning Research}, pp.\  8741--8751. {PMLR}, 2020.
\newblock URL \url{http://proceedings.mlr.press/v119/shen20e.html}.

\bibitem[Shi et~al.(2022)Shi, Gao, Xu, Liang, Li, Kong, Lee, and Kwok]{RevisitingOversmoothingBERT}
Shi, H., Gao, J., Xu, H., Liang, X., Li, Z., Kong, L., Lee, S. M.~S., and Kwok, J.~T.
\newblock Revisiting over-smoothing in {BERT} from the perspective of graph.
\newblock In \emph{The Tenth International Conference on Learning Representations, {ICLR} 2022, Virtual Event, April 25-29, 2022}. OpenReview.net, 2022.
\newblock URL \url{https://openreview.net/forum?id=dUV91uaXm3}.

\bibitem[Shleifer et~al.(2021)Shleifer, Weston, and Ott]{NormFormerImprovedTransformerb}
Shleifer, S., Weston, J., and Ott, M.
\newblock Normformer: Improved transformer pretraining with extra normalization.
\newblock \emph{ArXiv preprint}, abs/2110.09456, 2021.
\newblock URL \url{https://arxiv.org/abs/2110.09456}.

\bibitem[Shoeybi et~al.(2019)Shoeybi, Patwary, Puri, LeGresley, Casper, and Catanzaro]{Megatron}
Shoeybi, M., Patwary, M., Puri, R., LeGresley, P., Casper, J., and Catanzaro, B.
\newblock Megatron-lm: Training multi-billion parameter language models using model parallelism.
\newblock \emph{ArXiv preprint}, abs/1909.08053, 2019.
\newblock URL \url{https://arxiv.org/abs/1909.08053}.

\bibitem[Smith et~al.(2022)Smith, Patwary, Norick, LeGresley, Rajbhandari, Casper, Liu, Prabhumoye, Zerveas, Korthikanti, Zheng, Child, Aminabadi, Bernauer, Song, Shoeybi, He, Houston, Tiwary, and Catanzaro]{UsingDeepSpeedMegatrona}
Smith, S., Patwary, M., Norick, B., LeGresley, P., Rajbhandari, S., Casper, J., Liu, Z., Prabhumoye, S., Zerveas, G., Korthikanti, V., Zheng, E., Child, R., Aminabadi, R.~Y., Bernauer, J., Song, X., Shoeybi, M., He, Y., Houston, M., Tiwary, S., and Catanzaro, B.
\newblock Using deepspeed and megatron to train megatron-turing {NLG} 530b, {A} large-scale generative language model.
\newblock \emph{ArXiv preprint}, abs/2201.11990, 2022.
\newblock URL \url{https://arxiv.org/abs/2201.11990}.

\bibitem[Szegedy et~al.(2017)Szegedy, Ioffe, Vanhoucke, and Alemi]{Inceptionv4}
Szegedy, C., Ioffe, S., Vanhoucke, V., and Alemi, A.~A.
\newblock Inception-v4, inception-resnet and the impact of residual connections on learning.
\newblock In Singh, S.~P. and Markovitch, S. (eds.), \emph{Proceedings of the Thirty-First {AAAI} Conference on Artificial Intelligence, February 4-9, 2017, San Francisco, California, {USA}}, pp.\  4278--4284. {AAAI} Press, 2017.
\newblock URL \url{http://aaai.org/ocs/index.php/AAAI/AAAI17/paper/view/14806}.

\bibitem[Takase et~al.(2022)Takase, Kiyono, Kobayashi, and Suzuki]{LayerNormalizationsResiduala}
Takase, S., Kiyono, S., Kobayashi, S., and Suzuki, J.
\newblock On layer normalizations and residual connections in transformers.
\newblock \emph{ArXiv preprint}, abs/2206.00330, 2022.
\newblock URL \url{https://arxiv.org/abs/2206.00330}.

\bibitem[Tay et~al.(2022)Tay, Dehghani, Rao, Fedus, Abnar, Chung, Narang, Yogatama, Vaswani, and Metzler]{ScaleEfficientlyInsights}
Tay, Y., Dehghani, M., Rao, J., Fedus, W., Abnar, S., Chung, H.~W., Narang, S., Yogatama, D., Vaswani, A., and Metzler, D.
\newblock Scale efficiently: Insights from pretraining and finetuning transformers.
\newblock In \emph{The Tenth International Conference on Learning Representations, {ICLR} 2022, Virtual Event, April 25-29, 2022}. OpenReview.net, 2022.
\newblock URL \url{https://openreview.net/forum?id=f2OYVDyfIB}.

\bibitem[Tay et~al.(2023)Tay, Dehghani, Abnar, Chung, Fedus, Rao, Narang, Tran, Yogatama, and Metzler]{ScalingLawsModelArchitectures}
Tay, Y., Dehghani, M., Abnar, S., Chung, H.~W., Fedus, W., Rao, J., Narang, S., Tran, V.~Q., Yogatama, D., and Metzler, D.
\newblock Scaling laws vs model architectures: How does inductive bias influence scaling?
\newblock In Bouamor, H., Pino, J., and Bali, K. (eds.), \emph{Findings of the Association for Computational Linguistics: {EMNLP} 2023, Singapore, December 6-10, 2023}, pp.\  12342--12364. Association for Computational Linguistics, 2023.
\newblock URL \url{https://aclanthology.org/2023.findings-emnlp.825}.

\bibitem[Touvron et~al.(2021{\natexlab{a}})Touvron, Cord, Douze, Massa, Sablayrolles, and J{\'{e}}gou]{deit}
Touvron, H., Cord, M., Douze, M., Massa, F., Sablayrolles, A., and J{\'{e}}gou, H.
\newblock Training data-efficient image transformers {\&} distillation through attention.
\newblock In Meila, M. and Zhang, T. (eds.), \emph{Proceedings of the 38th International Conference on Machine Learning, {ICML} 2021, 18-24 July 2021, Virtual Event}, volume 139 of \emph{Proceedings of Machine Learning Research}, pp.\  10347--10357. {PMLR}, 2021{\natexlab{a}}.
\newblock URL \url{http://proceedings.mlr.press/v139/touvron21a.html}.

\bibitem[Touvron et~al.(2021{\natexlab{b}})Touvron, Cord, Sablayrolles, Synnaeve, and J{\'{e}}gou]{GoingDeeperImagea}
Touvron, H., Cord, M., Sablayrolles, A., Synnaeve, G., and J{\'{e}}gou, H.
\newblock Going deeper with image transformers.
\newblock In \emph{2021 {IEEE/CVF} International Conference on Computer Vision, {ICCV} 2021, Montreal, QC, Canada, October 10-17, 2021}, pp.\  32--42. {IEEE}, 2021{\natexlab{b}}.
\newblock \doi{10.1109/ICCV48922.2021.00010}.
\newblock URL \url{https://doi.org/10.1109/ICCV48922.2021.00010}.

\bibitem[Touvron et~al.(2023)Touvron, Martin, Stone, Albert, Almahairi, Babaei, Bashlykov, Batra, Bhargava, Bhosale, Bikel, Blecher, Canton{-}Ferrer, Chen, Cucurull, Esiobu, Fernandes, Fu, Fu, Fuller, Gao, Goswami, Goyal, Hartshorn, Hosseini, Hou, Inan, Kardas, Kerkez, Khabsa, Kloumann, Korenev, Koura, Lachaux, Lavril, Lee, Liskovich, Lu, Mao, Martinet, Mihaylov, Mishra, Molybog, Nie, Poulton, Reizenstein, Rungta, Saladi, Schelten, Silva, Smith, Subramanian, Tan, Tang, Taylor, Williams, Kuan, Xu, Yan, Zarov, Zhang, Fan, Kambadur, Narang, Rodriguez, Stojnic, Edunov, and Scialom]{llama2}
Touvron, H., Martin, L., Stone, K., Albert, P., Almahairi, A., Babaei, Y., Bashlykov, N., Batra, S., Bhargava, P., Bhosale, S., Bikel, D., Blecher, L., Canton{-}Ferrer, C., Chen, M., Cucurull, G., Esiobu, D., Fernandes, J., Fu, J., Fu, W., Fuller, B., Gao, C., Goswami, V., Goyal, N., Hartshorn, A., Hosseini, S., Hou, R., Inan, H., Kardas, M., Kerkez, V., Khabsa, M., Kloumann, I., Korenev, A., Koura, P.~S., Lachaux, M., Lavril, T., Lee, J., Liskovich, D., Lu, Y., Mao, Y., Martinet, X., Mihaylov, T., Mishra, P., Molybog, I., Nie, Y., Poulton, A., Reizenstein, J., Rungta, R., Saladi, K., Schelten, A., Silva, R., Smith, E.~M., Subramanian, R., Tan, X.~E., Tang, B., Taylor, R., Williams, A., Kuan, J.~X., Xu, P., Yan, Z., Zarov, I., Zhang, Y., Fan, A., Kambadur, M., Narang, S., Rodriguez, A., Stojnic, R., Edunov, S., and Scialom, T.
\newblock Llama 2: Open foundation and fine-tuned chat models.
\newblock \emph{ArXiv preprint}, abs/2307.09288, 2023.
\newblock URL \url{https://arxiv.org/abs/2307.09288}.

\bibitem[Trockman \& Kolter(2023)Trockman and Kolter]{trockmanMimeticInitializationSelfAttention2023}
Trockman, A. and Kolter, J.~Z.
\newblock Mimetic initialization of self-attention layers.
\newblock In Krause, A., Brunskill, E., Cho, K., Engelhardt, B., Sabato, S., and Scarlett, J. (eds.), \emph{International Conference on Machine Learning, {ICML} 2023, 23-29 July 2023, Honolulu, Hawaii, {USA}}, volume 202 of \emph{Proceedings of Machine Learning Research}, pp.\  34456--34468. {PMLR}, 2023.
\newblock URL \url{https://proceedings.mlr.press/v202/trockman23a.html}.

\bibitem[Vaswani et~al.(2017)Vaswani, Shazeer, Parmar, Uszkoreit, Jones, Gomez, Kaiser, and Polosukhin]{transformer}
Vaswani, A., Shazeer, N., Parmar, N., Uszkoreit, J., Jones, L., Gomez, A.~N., Kaiser, L., and Polosukhin, I.
\newblock Attention is all you need.
\newblock In Guyon, I., von Luxburg, U., Bengio, S., Wallach, H.~M., Fergus, R., Vishwanathan, S. V.~N., and Garnett, R. (eds.), \emph{Advances in Neural Information Processing Systems 30: Annual Conference on Neural Information Processing Systems 2017, December 4-9, 2017, Long Beach, CA, {USA}}, pp.\  5998--6008, 2017.
\newblock URL \url{https://proceedings.neurips.cc/paper/2017/hash/3f5ee243547dee91fbd053c1c4a845aa-Abstract.html}.

\bibitem[Wang et~al.(2019)Wang, Ge, Lipton, and Xing]{ImagenetSketch}
Wang, H., Ge, S., Lipton, Z.~C., and Xing, E.~P.
\newblock Learning robust global representations by penalizing local predictive power.
\newblock In Wallach, H.~M., Larochelle, H., Beygelzimer, A., d'Alch{\'{e}}{-}Buc, F., Fox, E.~B., and Garnett, R. (eds.), \emph{Advances in Neural Information Processing Systems 32: Annual Conference on Neural Information Processing Systems 2019, NeurIPS 2019, December 8-14, 2019, Vancouver, BC, Canada}, pp.\  10506--10518, 2019.
\newblock URL \url{https://proceedings.neurips.cc/paper/2019/hash/3eefceb8087e964f89c2d59e8a249915-Abstract.html}.

\bibitem[Wang et~al.(2023)Wang, Ma, Huang, Dong, Wang, Peng, Wu, Bajaj, Singhal, Benhaim, Patra, Liu, Chaudhary, Song, and Wei]{subln}
Wang, H., Ma, S., Huang, S., Dong, L., Wang, W., Peng, Z., Wu, Y., Bajaj, P., Singhal, S., Benhaim, A., Patra, B., Liu, Z., Chaudhary, V., Song, X., and Wei, F.
\newblock Magneto: {A} foundation transformer.
\newblock In Krause, A., Brunskill, E., Cho, K., Engelhardt, B., Sabato, S., and Scarlett, J. (eds.), \emph{International Conference on Machine Learning, {ICML} 2023, 23-29 July 2023, Honolulu, Hawaii, {USA}}, volume 202 of \emph{Proceedings of Machine Learning Research}, pp.\  36077--36092. {PMLR}, 2023.
\newblock URL \url{https://proceedings.mlr.press/v202/wang23u.html}.

\bibitem[Wang et~al.(2024)Wang, Ma, Dong, Huang, Zhang, and Wei]{deepnet}
Wang, H., Ma, S., Dong, L., Huang, S., Zhang, D., and Wei, F.
\newblock Deepnet: Scaling transformers to 1,000 layers.
\newblock \emph{IEEE Transactions on Pattern Analysis and Machine Intelligence}, 2024.

\bibitem[Wang et~al.(2022)Wang, Zheng, Chen, and Wang]{AntiOversmoothingDeepVision}
Wang, P., Zheng, W., Chen, T., and Wang, Z.
\newblock Anti-oversmoothing in deep vision transformers via the fourier domain analysis: From theory to practice.
\newblock In \emph{The Tenth International Conference on Learning Representations, {ICLR} 2022, Virtual Event, April 25-29, 2022}. OpenReview.net, 2022.
\newblock URL \url{https://openreview.net/forum?id=O476oWmiNNp}.

\bibitem[Williams et~al.(2018)Williams, Nangia, and Bowman]{mnli}
Williams, A., Nangia, N., and Bowman, S.
\newblock A broad-coverage challenge corpus for sentence understanding through inference.
\newblock In \emph{Proceedings of the 2018 Conference of the North {A}merican Chapter of the Association for Computational Linguistics: Human Language Technologies, Volume 1 (Long Papers)}, pp.\  1112--1122, New Orleans, Louisiana, 2018. Association for Computational Linguistics.
\newblock \doi{10.18653/v1/N18-1101}.
\newblock URL \url{https://aclanthology.org/N18-1101}.

\bibitem[Wortsman et~al.(2024)Wortsman, Liu, Xiao, Everett, Alemi, Adlam, Co-Reyes, Gur, Kumar, Novak, Pennington, Sohl-Dickstein, Xu, Lee, Gilmer, and Kornblith]{smallProxy}
Wortsman, M., Liu, P.~J., Xiao, L., Everett, K.~E., Alemi, A.~A., Adlam, B., Co-Reyes, J.~D., Gur, I., Kumar, A., Novak, R., Pennington, J., Sohl-Dickstein, J., Xu, K., Lee, J., Gilmer, J., and Kornblith, S.
\newblock Small-scale proxies for large-scale transformer training instabilities.
\newblock In \emph{The Twelfth International Conference on Learning Representations}, 2024.
\newblock URL \url{https://openreview.net/forum?id=d8w0pmvXbZ}.

\bibitem[Xiong et~al.(2020)Xiong, Yang, He, Zheng, Zheng, Xing, Zhang, Lan, Wang, and Liu]{preln}
Xiong, R., Yang, Y., He, D., Zheng, K., Zheng, S., Xing, C., Zhang, H., Lan, Y., Wang, L., and Liu, T.
\newblock On layer normalization in the transformer architecture.
\newblock In \emph{Proceedings of the 37th International Conference on Machine Learning, {ICML} 2020, 13-18 July 2020, Virtual Event}, volume 119 of \emph{Proceedings of Machine Learning Research}, pp.\  10524--10533. {PMLR}, 2020.
\newblock URL \url{http://proceedings.mlr.press/v119/xiong20b.html}.

\bibitem[Xu et~al.(2019)Xu, Sun, Zhang, Zhao, and Lin]{xu2019understanding}
Xu, J., Sun, X., Zhang, Z., Zhao, G., and Lin, J.
\newblock Understanding and improving layer normalization.
\newblock In Wallach, H.~M., Larochelle, H., Beygelzimer, A., d'Alch{\'{e}}{-}Buc, F., Fox, E.~B., and Garnett, R. (eds.), \emph{Advances in Neural Information Processing Systems 32: Annual Conference on Neural Information Processing Systems 2019, NeurIPS 2019, December 8-14, 2019, Vancouver, BC, Canada}, pp.\  4383--4393, 2019.
\newblock URL \url{https://proceedings.neurips.cc/paper/2019/hash/2f4fe03d77724a7217006e5d16728874-Abstract.html}.

\bibitem[Xue et~al.(2023)Xue, Chen, Sun, Ren, Zheng, He, Chen, Jiang, and You]{StudyTransformerConfigurationa}
Xue, F., Chen, J., Sun, A., Ren, X., Zheng, Z., He, X., Chen, Y., Jiang, X., and You, Y.
\newblock A study on transformer configuration and training objective.
\newblock In Krause, A., Brunskill, E., Cho, K., Engelhardt, B., Sabato, S., and Scarlett, J. (eds.), \emph{International Conference on Machine Learning, {ICML} 2023, 23-29 July 2023, Honolulu, Hawaii, {USA}}, volume 202 of \emph{Proceedings of Machine Learning Research}, pp.\  38913--38925. {PMLR}, 2023.
\newblock URL \url{https://proceedings.mlr.press/v202/xue23b.html}.

\bibitem[Yang \& Schoenholz(2017)Yang and Schoenholz]{EdgeOfChaos}
Yang, G. and Schoenholz, S.~S.
\newblock Mean field residual networks: On the edge of chaos.
\newblock In Guyon, I., von Luxburg, U., Bengio, S., Wallach, H.~M., Fergus, R., Vishwanathan, S. V.~N., and Garnett, R. (eds.), \emph{Advances in Neural Information Processing Systems 30: Annual Conference on Neural Information Processing Systems 2017, December 4-9, 2017, Long Beach, CA, {USA}}, pp.\  7103--7114, 2017.
\newblock URL \url{https://proceedings.neurips.cc/paper/2017/hash/81c650caac28cdefce4de5ddc18befa0-Abstract.html}.

\bibitem[Yang et~al.(2021)Yang, Hu, Babuschkin, Sidor, Liu, Farhi, Ryder, Pachocki, Chen, and Gao]{tensorprograms}
Yang, G., Hu, E.~J., Babuschkin, I., Sidor, S., Liu, X., Farhi, D., Ryder, N., Pachocki, J., Chen, W., and Gao, J.
\newblock Tuning large neural networks via zero-shot hyperparameter transfer.
\newblock In Ranzato, M., Beygelzimer, A., Dauphin, Y.~N., Liang, P., and Vaughan, J.~W. (eds.), \emph{Advances in Neural Information Processing Systems 34: Annual Conference on Neural Information Processing Systems 2021, NeurIPS 2021, December 6-14, 2021, virtual}, pp.\  17084--17097, 2021.
\newblock URL \url{https://proceedings.neurips.cc/paper/2021/hash/8df7c2e3c3c3be098ef7b382bd2c37ba-Abstract.html}.

\bibitem[Yang et~al.(2024)Yang, Yu, Zhu, and Hayou]{tensorprogramsvi}
Yang, G., Yu, D., Zhu, C., and Hayou, S.
\newblock Tensor programs {VI}: Feature learning in infinite depth neural networks.
\newblock In \emph{The Twelfth International Conference on Learning Representations}, 2024.
\newblock URL \url{https://openreview.net/forum?id=17pVDnpwwl}.

\bibitem[You et~al.(2017)You, Gitman, and Ginsburg]{LargeBatchTrainingb}
You, Y., Gitman, I., and Ginsburg, B.
\newblock Scaling {SGD} batch size to 32k for imagenet training.
\newblock \emph{ArXiv preprint}, abs/1708.03888, 2017.
\newblock URL \url{https://arxiv.org/abs/1708.03888}.

\bibitem[You et~al.(2020)You, Li, Reddi, Hseu, Kumar, Bhojanapalli, Song, Demmel, Keutzer, and Hsieh]{LargeBatchOptimizationb}
You, Y., Li, J., Reddi, S.~J., Hseu, J., Kumar, S., Bhojanapalli, S., Song, X., Demmel, J., Keutzer, K., and Hsieh, C.
\newblock Large batch optimization for deep learning: Training {BERT} in 76 minutes.
\newblock In \emph{8th International Conference on Learning Representations, {ICLR} 2020, Addis Ababa, Ethiopia, April 26-30, 2020}. OpenReview.net, 2020.
\newblock URL \url{https://openreview.net/forum?id=Syx4wnEtvH}.

\bibitem[Yu et~al.(2017)Yu, Lin, Salakhutdinov, and Carbonell]{BlockNormalizedGradientMethoda}
Yu, A.~W., Lin, Q., Salakhutdinov, R., and Carbonell, J.~G.
\newblock Normalized gradient with adaptive stepsize method for deep neural network training.
\newblock \emph{ArXiv preprint}, abs/1707.04822, 2017.
\newblock URL \url{https://arxiv.org/abs/1707.04822}.

\bibitem[Zhai et~al.(2023)Zhai, Likhomanenko, Littwin, Busbridge, Ramapuram, Zhang, Gu, and Susskind]{StabilizingTransformerTrainingc}
Zhai, S., Likhomanenko, T., Littwin, E., Busbridge, D., Ramapuram, J., Zhang, Y., Gu, J., and Susskind, J.~M.
\newblock Stabilizing transformer training by preventing attention entropy collapse.
\newblock In Krause, A., Brunskill, E., Cho, K., Engelhardt, B., Sabato, S., and Scarlett, J. (eds.), \emph{International Conference on Machine Learning, {ICML} 2023, 23-29 July 2023, Honolulu, Hawaii, {USA}}, volume 202 of \emph{Proceedings of Machine Learning Research}, pp.\  40770--40803. {PMLR}, 2023.
\newblock URL \url{https://proceedings.mlr.press/v202/zhai23a.html}.

\bibitem[Zhang et~al.(2019{\natexlab{a}})Zhang, Titov, and Sennrich]{DSInit}
Zhang, B., Titov, I., and Sennrich, R.
\newblock Improving deep transformer with depth-scaled initialization and merged attention.
\newblock In \emph{Proceedings of the 2019 Conference on Empirical Methods in Natural Language Processing and the 9th International Joint Conference on Natural Language Processing (EMNLP-IJCNLP)}, pp.\  898--909, Hong Kong, China, 2019{\natexlab{a}}. Association for Computational Linguistics.
\newblock \doi{10.18653/v1/D19-1083}.
\newblock URL \url{https://aclanthology.org/D19-1083}.

\bibitem[Zhang et~al.(2022{\natexlab{a}})Zhang, Botev, and Martens]{DeepLearningShortcuts}
Zhang, G., Botev, A., and Martens, J.
\newblock Deep learning without shortcuts: Shaping the kernel with tailored rectifiers.
\newblock In \emph{The Tenth International Conference on Learning Representations, {ICLR} 2022, Virtual Event, April 25-29, 2022}. OpenReview.net, 2022{\natexlab{a}}.
\newblock URL \url{https://openreview.net/forum?id=U0k7XNTiFEq}.

\bibitem[Zhang et~al.(2019{\natexlab{b}})Zhang, Dauphin, and Ma]{fixup}
Zhang, H., Dauphin, Y.~N., and Ma, T.
\newblock Fixup initialization: Residual learning without normalization.
\newblock In \emph{7th International Conference on Learning Representations, {ICLR} 2019, New Orleans, LA, USA, May 6-9, 2019}. OpenReview.net, 2019{\natexlab{b}}.
\newblock URL \url{https://openreview.net/forum?id=H1gsz30cKX}.

\bibitem[Zhang et~al.(2022{\natexlab{b}})Zhang, Yu, Yi, Chen, and Liu]{StabilizeDeepResNet}
Zhang, H., Yu, D., Yi, M., Chen, W., and Liu, T.
\newblock Stabilize deep resnet with a sharp scaling factor {\(\tau\)}.
\newblock \emph{Mach. Learn.}, 111\penalty0 (9):\penalty0 3359--3392, 2022{\natexlab{b}}.
\newblock \doi{10.1007/S10994-022-06192-X}.
\newblock URL \url{https://doi.org/10.1007/s10994-022-06192-x}.

\bibitem[Zhang et~al.(2022{\natexlab{c}})Zhang, Roller, Goyal, Artetxe, Chen, Chen, Dewan, Diab, Li, Lin, Mihaylov, Ott, Shleifer, Shuster, Simig, Koura, Sridhar, Wang, and Zettlemoyer]{opt}
Zhang, S., Roller, S., Goyal, N., Artetxe, M., Chen, M., Chen, S., Dewan, C., Diab, M.~T., Li, X., Lin, X.~V., Mihaylov, T., Ott, M., Shleifer, S., Shuster, K., Simig, D., Koura, P.~S., Sridhar, A., Wang, T., and Zettlemoyer, L.
\newblock {OPT:} open pre-trained transformer language models.
\newblock \emph{ArXiv preprint}, abs/2205.01068, 2022{\natexlab{c}}.
\newblock URL \url{https://arxiv.org/abs/2205.01068}.

\bibitem[Zhao et~al.(2023)Zhao, Ma, Zhang, Deng, and Wei]{AreMoreLayersa}
Zhao, H., Ma, S., Zhang, D., Deng, Z., and Wei, F.
\newblock Are more layers beneficial to graph transformers?
\newblock In \emph{The Eleventh International Conference on Learning Representations, {ICLR} 2023, Kigali, Rwanda, May 1-5, 2023}. OpenReview.net, 2023.
\newblock URL \url{https://openreview.net/pdf?id=uagC-X9XMi8}.

\bibitem[Zhou et~al.(2021)Zhou, Kang, Jin, Yang, Lian, Hou, and Feng]{DeepViTDeeperVision}
Zhou, D., Kang, B., Jin, X., Yang, L., Lian, X., Hou, Q., and Feng, J.
\newblock Deepvit: Towards deeper vision transformer.
\newblock \emph{ArXiv preprint}, abs/2103.11886, 2021.
\newblock URL \url{https://arxiv.org/abs/2103.11886}.

\bibitem[Zhu et~al.(2021)Zhu, Ni, Xu, Kong, Huang, and Goldstein]{zhuGradInitLearningInitialize2021}
Zhu, C., Ni, R., Xu, Z., Kong, K., Huang, W.~R., and Goldstein, T.
\newblock Gradinit: Learning to initialize neural networks for stable and efficient training.
\newblock In Ranzato, M., Beygelzimer, A., Dauphin, Y.~N., Liang, P., and Vaughan, J.~W. (eds.), \emph{Advances in Neural Information Processing Systems 34: Annual Conference on Neural Information Processing Systems 2021, NeurIPS 2021, December 6-14, 2021, virtual}, pp.\  16410--16422, 2021.
\newblock URL \url{https://proceedings.neurips.cc/paper/2021/hash/88ae6372cfdc5df69a976e893f4d554b-Abstract.html}.

\end{thebibliography}
\bibliographystyle{icml2024}

\newpage
\appendix
\onecolumn
\resumetocwriting
\tableofcontents

\section{Moment Propagation through Transformer Components}
\label{appendix:section:Moment_prop_components}

We provide detailed proofs of the closed-form expression for each of the transformer component -- Linear layer, Dropout, ReLU, GeLU, LayerNorm, and Softmax. 

For any component, input is represented as $\mathbf{x_{\text{in}}}$ and $\mathbf{x_{\text{out}}}$ is the output. The gradient flowing in into the component from the output side is represented as $\mathbf{g_{\text{out}}}$ and the backpropagated gradient towards the input is $\mathbf{g_{\text{in}}}$. We switch from vector to matrix notation ($\mathbf{X_{\text{in}}}$, $\mathbf{X_{\text{out}}}$) whenever needed. We assume that the input is distributed normally $\mathcal{N} (0, \sigma_{x_{in}})$. No assumptions are made regarding the covariance of the input -- it is not assumed to be IID, and it may/may-not have covariance both along the sequence length and hidden dimension. Additional assumptions needed to derive the proofs for softmax and attention can be found in the respective proofs. A detailed list of terms/notations used in the proofs is provided at the end of this work in \cref{sec:notations}.


\subsection{Embeddings}
\label{proof: embeddings}

The BERT model's embedding component consists of 3 look up tables - token embeddings, position embeddings, and segment embeddings. For a given input token, each of these 3 embeddings are added before being passed to the transformer model. Other transformer models, such as decoder-only GPT lack some (eg. segment) of these, but the derivations remain similar. In the general case, these theoretical derivations can be replaced by the empirically observed moments of the inputs fed to the transformer model (as we did for Speech-to-Text translation). We derive formulae for each of these embedding types below.

\paragraph{Token Embeddings} We do not assume the input embeddings to be IID. Repetition of same token introduces correlation across the sequence length. We assume that the input tokens have been sampled from a multinomial distribution. The words / token ids are distributed almost according to Zipf's law~\citep{zipf}. Assuming we initialize all the embeddings with variance $\sigma_{w_{embd}}^2$, the relevant statistics for word embeddings output ${x_{\text{out}_{we}}}$ are as follows

\begin{align*}
\mu_{x_{\text{out}_{we}}} &= 0  \\
\sigma_{x_{\text{out}_{we}}}^2 &= \sigma_{w_{\text{embd}}}^2 \\
\mathrm{Cov}^l({x_{\text{out}_{we}}}) &= \sum{\frac{N_i * (N_i - 1)}{L*(L-1)}} * \sigma_{w_{\text{embd}}}^2 \\
\mathrm{r}^l({x_{\text{out}_{we}}}) &= \sum{\frac{N_i * (N_i - 1)}{L*(L-1)}} \\
\mathrm{Cov}^d({x_{\text{out}_{we}}}) &= 0 \\
 \end{align*}

 Assume $i$th word occurs $N_i$ times, it contributes $ \frac{N_i * (N_i - 1)}{L*(L-1))}$ to the covariance along sequence length. Similarly, we can calculate the correlation for segment-type embeddings output ${x_{\text{out}_{se}}}$. Zipf's law states that the probability for each token is inversely proportional to its rank. For the word with rank $i$, $p_i = \frac{c}{i}$, where $c = \frac{1}{\sum_i \frac{1}{i}} = \frac{1}{\gamma + \mathrm{log}(|V|)}$, where $\gamma \approx 0.58$ is the Euler's constant. 
 
 For a sentence of length $L$, the token with probability $p_i$ is expected to occur $p_i . L$ times. Hence, for a given vocabulary size $|V|$, we can calculate the correlation as follows

\begin{align*}
r^l({x_{\text{out}_{we}}}) &= \sum{\frac{N_i * (N_i - 1)}{L*(L-1)}} \\
&= \sum_i^{|V|}{\frac{p_i L * (p_i L - 1)}{L*(L-1)}} \\
&= \frac{\sum_i p_i^2 * L - 1}{L-1} \\
&= \frac{\sum_i \frac{c^2}{i^2} * L - 1}{L-1} \\
&\approx \frac{\frac{L \pi^2}{6.(\gamma + \mathrm{log}(|V|))^2} - 1}{L-1} \\
&\approx  \frac{\pi^2}{6.\mathrm{log}(|V|)^2} ~~ \text{,~assuming $\gamma \approx 0.58 << \mathrm{log}(|V|) \approx 10.4$, $L >> 1$}
 \end{align*}
  
\paragraph{Segment Type Embeddings} Similarly, the segment type embeddings have two possible values denoting the sentence order. If first sentence has length $x$, we can consider this as a special case of the analysis performed above with two possible tokens, where $N_1 = x$ and $N_2 = L-x$. Assuming $x$ is distributed uniformly between $0$ to $L$, $L-x$ also has the same distribution. Hence,  
\begin{align*}
r^l({x_{\text{out}_{se}}}, N_1, N_2) &= \frac{ N_1^2 + N_2^2  - L}{L*(L-1)} \\
 \end{align*}
 Taking expectation, we get
 \begin{align*}
r^l({x_{\text{out}_{se}}}) &= \frac{ \frac{2}{3} * L^2   - L}{L*(L-1)} \\
&\approx \frac{2}{3}
 \end{align*}

\paragraph{Position Embeddings} Since learnt position embeddings are lookup tables with unique inputs, the correlation from position embeddings is $0$.

\paragraph{Final Model Input Embeddings}
Each of the above embeddings are added before being passed to the transformer model. Since the variance is same for all embedding types, the final correlation is the average of the three. Hence:

 \begin{align*}
r^l({x_{\text{out}}}) &= \frac{1}{3}(r^l({x_{\text{out}_{we}}}) + r^l({x_{\text{out}_{se}}})) \\
&= \frac{\pi^2}{18*\mathrm{log}(|V|)^2} + \frac{2}{9}
 \end{align*}
 
For our case, $|V| = 32000$ and sequence length $L=256$, the theoretically predicted correlation $r^l_{x_{in}} = 0.227$ which is within $3\%$ of the empirically observed correlation ($0.221$). 

Hence, the final moments for the embedding output are

\begin{empheq}[box=\widefbox]{align*}
  \mu_{x_{\text{out}}} &= 0  \\
 \sigma_{x_{\text{out}}}^2 &= 3 * \sigma_{w_{\text{embd}}}^2 \\
 \mathrm{Cov}^l_{x_{\text{out}}} &= (\frac{\pi^2}{18*\mathrm{log}(|V|)^2} + \frac{2}{9}) \sigma_{x_{\text{out}}}^2 \\
 \mathrm{Cov}^d_{x_{\text{out}}} &= 0 
\end{empheq}


\subsection{Linear}
\label{proof: linear}
For linear layer with $d_{in}$ dimensional input $\mathbf{x_{\text{in}}}$, and $d_{out}$ dimensional output $\mathbf{x_{\text{out}}}$, we can define the forward pass mathematically as,
\begin{align*}
    \mathbf{x_{\text{out}}} & = \mathbf{x_{\text{in}}}\mathbf{W} \\
    \implies {x_{\text{out}_{j}}} & = \displaystyle\sum_{i=1}^{d_{\text{in}}} {x_{\text{in}_i}}W_{i,j} \\
\end{align*}
Similarly, we define the backward pass as,
\begin{align*}
    \mathbf{g_{{\text{in}}}} & = \mathbf{g_{{\text{out}}}}\mathbf{W^T} \\
    \implies g_{{\text{in}_j}} & = \displaystyle\sum_{i=1}^{d_{\text{out}}} g_{{{\text{out}_{i}}}}W_{j,i}
\end{align*}
\\
For expectation of output we have,
\begin{align*}
    \mathbb{E}[{x_{\text{out}_{j}}}] &= \mathbb{E}[\displaystyle\sum_{i=1}^{d_{\text{in}}} {x_{\text{in}_i}}W_{i,j}] = \displaystyle\sum_{i=1}^{d_{\text{in}}} \mathbb{E}[{x_{\text{in}_i}}W_{i,j}] \\
    &= \displaystyle\sum_{i=1}^{d_{\text{in}}} \mathbb{E}[{x_{\text{in}_i}}]\mathbb{E}[W_{i,j}] = \mu_{x_{\text{in}}}\mu_w \\
    \tag*{(As weights and input are independent of each other)} \\
    \Aboxed{\mu_{x_{\text{out}}} & = 0} \tag{ $\forall j$ }
\end{align*}
To get variance of the output of forward pass we have,
\begin{align*}
    \mathrm{Var}({x_{\text{out}_{j}}}) & = \mathrm{Var}(\displaystyle\sum_{i=1}^{d_{\text{in}}} {x_{\text{in}_i}}W_{i,j})\\
    \intertext{As the weights are initialized independently each term in summation is independent of each other}
    & = \displaystyle\sum_{i=1}^{d_{\text{in}}}(\mathrm{Var}({x_{\text{in}_i}}W_{i,j})) \\ 
    & = \displaystyle\sum_{i=1}^{d_{\text{in}}}((\sigma^2_{x_{\text{in}}} + \mu^2_{x_{\text{in}}})(\sigma_w^2+\mu_w^2) - \mu^2_{x_{\text{in}}}\mu_w^2) \\
    \tag*{(As weights and input are independent of each other)} \\
    & = \displaystyle\sum_{i=1}^{d_{\text{in}}}(\sigma^2_{x_{\text{in}}} + \mu^2_{x_{\text{in}}})\sigma_w^2 \\
    \mathrm{Var}({x_{\text{out}_{j}}}) & = d_{\text{in}}(\sigma^2_{x_{\text{in}}} + \mu^2_{x_{\text{in}}})\sigma_w^2 \tag{ $\forall j$ } \\
    \Aboxed{\sigma_{x_{\text{out}}}^2 &= d_{\text{in}}(\sigma^2_{x_{\text{in}}} + \mu^2_{x_{\text{in}}})\sigma_w^2}
\end{align*}
If we have two inputs $\mathbf{x_{\text{in}}}$ and $\mathbf{y_{\text{in}}}$ such that for all $i$ we have $\mathrm{Corr}(x_{\text{in}_{i}},y_{\text{in}_{i}}) = r^l_{x_{\text{in}}}$, and $\mathbf{{x_{\text{out}}}} = \mathbf{x_{\text{in}}}\mathbf{W}$ and $\mathbf{{y_{\text{out}}}} = \mathbf{y_{\text{in}}}\mathbf{W}$. Then for any $j$ we have 
\begin{align*}
    \mathrm{Corr}({x_{\text{out}_{j}}},{y_{\text{out}_{j}}}) & = \frac{\mathbb{E}[{x_{\text{out}_{j}}}{y_{\text{out}_{j}}}] - \mathbb{E}[{x_{\text{out}_{j}}}]\mathbb{E}[{y_{\text{out}_{j}}}]}{\sqrt{\mathrm{Var}({x_{\text{out}_{j}}})\mathrm{Var}({y_{\text{out}_{j}}})}}\\
    & = \frac{\mathbb{E}[{x_{\text{out}_{j}}}{y_{\text{out}_{j}}}]}{\sqrt{\sigma_{x_{\text{out}}}^2\sigma_{x_{\text{out}}}^2}} \\
    & = \frac{\mathbb{E}[\sum_{i=1}^{d_{\text{in}}} {x_{\text{in}_i}}W_{i,j}\sum_{k=1}^{d_{in}} {y_{\text{in}_k}}W_{k,j}]}{\sigma_{x_{\text{out}}}^2} \\
    & = \frac{\mathbb{E}[\sum_{i=1}^{d_{\text{in}}} {x_{\text{in}_i}}{y_{\text{in}_i}}W_{i,j}^2 + \sum_{k=1, k\neq i}^{d_{in}}\sum_{i=1}^{d_{\text{in}}} {x_{\text{in}_i}}{y_{\text{in}_k}}W_{i,j}W_{k,j}]}{\sigma_{x_{\text{out}}}^2} \\
    \intertext{In second summation all terms are independent of each other and as the expectation of weights is 0 we have}
    \mathrm{Corr}({x_{\text{out}_{j}}},{y_{\text{out}_{j}}}) &= \frac{\mathbb{E}[\sum_{i=1}^{d_{\text{in}}} {x_{\text{in}_i}}{y_{\text{in}_i}}W_{i,j}^2]}{\sigma_{x_{\text{out}}}^2} \\
    &= \frac{\sum_{i=1}^{d_{\text{in}}}\mathbb{E}[{x_{\text{in}_i}}{y_{\text{in}_i}}W_{i,j}^2]}{\sigma_{x_{\text{out}}}^2} \tag*{(Independence of weight initialization)}\\ 
    &= \frac{\sum_{i=1}^{d_{\text{in}}}\mathbb{E}[{x_{\text{in}_i}}{y_{\text{in}_i}}]\mathbb{E}[W_{i,j}^2]}{\sigma_{x_{\text{out}}}^2} \\
    &= \frac{\sum_{i=1}^{d_{\text{in}}}(r^l_{x_{\text{in}}}\sigma_{x_{\text{in}}}^2 + \mu_{x_{\text{in}}}^2)\sigma_w^2}{\sigma_{x_{\text{out}}}^2} \text{~~~~~~~~~~~~~(Definition of correlation)}\\
    &= \frac{d_{\text{in}}(r^l_{x_{\text{in}}}\sigma_{x_{\text{in}}}^2 + \mu_{x_{\text{in}}}^2)\sigma_w^2}{d_{\text{in}}(\sigma^2_{x_{\text{in}}} + \mu^2_{x_{\text{in}}})\sigma_w^2} \\
    \mathrm{Corr}({x_{\text{out}_{j}}},{y_{\text{out}_{j}}}) &= \frac{r^l_{x_{\text{in}}}\sigma_{x_{\text{in}}}^2 + \mu_{x_{\text{in}}}^2}{\sigma^2_{x_{\text{in}}} + \mu^2_{x_{\text{in}}}} \\
    \Aboxed{r^l_{x_{\text{out}}} &= \frac{r^l_{x_{\text{in}}}\sigma_{x_{\text{in}}}^2 + \mu_{x_{\text{in}}}^2}{\sigma^2_{x_{\text{in}}} + \mu^2_{x_{\text{in}}}}}
\end{align*}

As the backward pass has similar structure, assuming $\mu_{g_{{\text{out}}}} = 0$  we can use the same analysis to get,
\begin{empheq}[box=\widefbox]{align*}
    \mu_{g_{{\text{in}}}} & = 0 \\
    \sigma_{g_{{\text{in}}}}^2 & = {d_{\text{out}}}\sigma_{g_{{\text{out}}}}^2\sigma_w^2 
\end{empheq}

\subsection{Dropout}
\label{proof: dropout}

We can define Dropout mathematically as,
\begin{align*}
    \mathbf{x_{\text{out}}} & = \mathrm{Dropout}(\mathbf{x_{\text{in}}}) \\
    \implies {x_{\text{out}_i}} &= \begin{cases}
        \frac{x_{\text{in}_i}}{(1-p)}  & \text{with probability } 1-p \\
		0 & \text{else}
    \end{cases}
\end{align*}
To calculate expectation of dropout,
\begin{align*}
    \mathbb{E}[{x_{\text{out}_i}}] &= 0*p + (1-p)*\mathbb{E}[\frac{{x_{\text{in}_i}}}{(1-p)}] \\
    \Aboxed{\mu_{x_{\text{out}}} &= \mu_{x_{\text{in}}}}
\end{align*}
For variance,
\begin{align*}
    \mathrm{Var}({x_{\text{out}_i}}) &= \mathbb{E}[{x^2_{\text{out}_i}}] - \mathbb{E}[{x_{\text{out}_i}}]^2 \\
    &= 0*p + (1-p)*\mathbb{E}[\frac{x^2_{\text{in}_i}}{(1-p)^2}] - \mu_{x_{\text{in}}}^2 \\
    &= \frac{\mathbb{E}[{x^2_{\text{in}_i}}]}{(1-p)} - \mu_x^2 \\
    &= \frac{\sigma_{x_{\text{in}}}^2 + \mu_{x_{\text{in}}}^2}{(1-p)} - \mu_{x_{\text{in}}}^2 \\
    \Aboxed{\sigma_{x_{\text{out}}}^2 &= \frac{\sigma_{x_{\text{in}}}^2 + p\mu_{x_{\text{in}}}^2}{(1-p)}}
\end{align*}
If we have two inputs $\mathbf{x_{\text{in}}}$ and $\mathbf{y_{\text{in}}}$ such that for all $i$ we have $\mathrm{Corr}(x_{\text{in}_{i}},y_{\text{in}_{i}}) = r^l_{x_{\text{in}}}$, and $\mathbf{{x_{\text{out}}}} = \mathrm{Dropout}(\mathbf{x_{\text{in}}})$ and $\mathbf{{y_{\text{out}}}} = \mathrm{Dropout}(\mathbf{y_{\text{in}}})$. Then for any $j$ we have 
\begin{align*}
    \mathrm{Corr}({x_{\text{out}_{j}}},{y_{\text{out}_{j}}}) & = \frac{\mathbb{E}[{x_{\text{out}_{j}}}{y_{\text{out}_{j}}}] - \mathbb{E}[{x_{\text{out}_{j}}}]\mathbb{E}[{y_{\text{out}_{j}}}]}{\sqrt{\mathrm{Var}({x_{\text{out}_{j}}})\mathrm{Var}({y_{\text{out}_{j}}})}}\\
    & = \frac{\mathbb{E}[{x_{\text{out}_{j}}}{y_{\text{out}_{j}}}] - \mu_{x_{\text{out}}}\mu_{x_{\text{out}}}}{\sqrt{\sigma_{x_{\text{out}}}^2\sigma_{x_{\text{out}}}^2}} \\
     & = \frac{p^2*0 + 2*p*(1-p)*0 + (1-p)^2 *\mathbb{E}[\frac{{x_{\text{in}_{j}}}{y_{\text{in}_{j}}}}{(1-p)*(1-p)}] - \mu_{x_{\text{out}}}^2}{\sigma_{x_{\text{out}}}^2} \\ 
    & = \frac{\mathbb{E}[{x_{\text{in}_{j}}}{y_{\text{in}_{j}}}] - \mu_{x_{\text{out}}}^2}{\sigma_{x_{\text{out}}}^2} \\
   \Aboxed{\mathrm{Corr}({x_{\text{out}_{j}}},{y_{\text{out}_{j}}}) & = \frac{(r^l_{x_{\text{in}}}\sigma_{x_{\text{in}}}^2)(1-p)}{\sigma_{x_{\text{in}}}^2 + p\mu_{x_{\text{in}}}^2} = r^l_{x_{\text{out}}}}
\end{align*}
We can define the backward pass of Dropout as,
\begin{align*}
    g_{{{\text{in}}}_i} &= \begin{cases}
        \frac{g_{{{\text{out}}}_i}}{(1-p)}  & \text{if $x_i$ isn't dropped out (which has probability $(1 -p)$)} \\
		0 & \text{else}
    \end{cases}
\end{align*}
Again we can see that backward has similar definition to that of forward pass. Assuming $\mu_{g_{x_{\text{out}}}} = 0$ and using similar analysis we get,
\begin{empheq}[box=\widefbox]{align*}
    \mu_{g_{{\text{in}}}} & = 0 \\
    \sigma_{g_{{\text{in}}}}^2 & = \frac{\sigma_{g_{{\text{out}}}}^2}{(1-p)} \\
\end{empheq}

\subsection{ReLU}
\label{proof: relu}

Formuale functionally equivalent to ours for $\mu_x$, $\sigma^2_x$, and $\sigma^2_g$ have also been derived in \citet{NormalizationPropagation}.

We can define ReLU mathematically as,
\begin{align*}
    \mathbf{x_{\text{out}}} & = \mathrm{ReLU}(\mathbf{x_{\text{in}}}) \\
    \implies {x_{\text{out}_i}} &= \begin{cases}
        {x_{\text{in}_i}}  & \text{if }{x_{\text{in}_i}} > 0 \\
		0 & \text{else}
    \end{cases}
\end{align*}
For getting expectation of output of ReLU for normally distributed input we have,
\begin{align*}
    \mathbb{E}[{x_{\text{out}_i}}] &= \int_{-\infty}^{\infty}\frac{\mathrm{ReLU}({x_{\text{in}_i}})\exp{(\frac{-{x^2_{\text{in}_i}}}{2\sigma_{x_{\text{in}}}^2})}}{\sqrt{2\pi}\sigma_{x_{\text{in}}}}d{x_{\text{in}_i}} \\
    &= \int_{-\infty}^{0}\frac{0*\exp{(\frac{-{x^2_{\text{in}_i}}}{2\sigma_{x_{\text{in}}}^2})}}{\sqrt{2\pi}\sigma_{x_{\text{in}}}}d{x_{\text{in}_i}} + \int_{0}^{\infty}\frac{{x_{\text{in}_i}}\exp{(\frac{-{x^2_{\text{in}_i}}}{2\sigma_{x_{\text{in}}}^2})}}{\sqrt{2\pi}\sigma_{x_{\text{in}}}}d{x_{\text{in}_i}} \\
    &= \int_{0}^{\infty}\frac{{x_{\text{in}_i}}\exp{(\frac{-{x^2_{\text{in}_i}}}{2\sigma_{x_{\text{in}}}^2})}}{\sqrt{2\pi}\sigma_{x_{\text{in}}}}d{x_{\text{in}_i}} \\
    \intertext{Substituting $t = \displaystyle\frac{{x^2_{\text{in}_i}}}{2\sigma_{x_{\text{in}}}^2}$ we have $dt = \displaystyle\frac{{x_{\text{in}_i}}d{x_{\text{in}_i}}}{\sigma_{x_{\text{in}}}^2}$ we get,}
    \mathbb{E}[{x_{\text{out}_i}}] &= \int_{0}^{\infty}\frac{\sigma_{x_{\text{in}}}\exp{(-t)}dt}{\sqrt{2\pi}}\\
    &= \frac{\sigma_{x_{\text{in}}}}{\sqrt{2\pi}}[-\exp{(-t)}]_0^{\infty} = \frac{\sigma_{x_{\text{in}}}}{\sqrt{2\pi}}
\end{align*}
Hence, the mean of output
\begin{align}
    \Aboxed{\mu_{x_{\text{out}}} = \frac{\sigma_{x_{\text{in}}}}{\sqrt{2\pi}} }
\end{align}
Variance of output can be calculated by,
\begin{align*}
    \mathrm{Var}({x_{\text{out}_i}}) &= \mathbb{E}[{x_{\text{out}_i}}^2] - \mathbb{E}[{x_{\text{out}_i}}]^2\\
    &= \int_{-\infty}^{\infty}\frac{(\mathrm{ReLU}({x_{\text{in}_i}}))^2\exp{(\frac{-{x^2_{\text{in}_i}}}{2\sigma_{x_{\text{in}}}^2})}}{\sqrt{2\pi}\sigma_{x_{\text{in}}}}d{x_{\text{in}_i}} - \frac{\sigma_{x_{\text{in}}}^2}{2\pi}\\
    &= \int_{-\infty}^{0}\frac{0*\exp{(\frac{-{x^2_{\text{in}_i}}}{2\sigma_{x_{\text{in}}}^2})}}{\sqrt{2\pi}\sigma_{x_{\text{in}}}}d{x_{\text{in}_i}} + \int_{0}^{\infty}\frac{{x^2_{\text{in}_i}}\exp{(\frac{-{x^2_{\text{in}_i}}}{2\sigma_{x_{\text{in}}}^2})}}{\sqrt{2\pi}\sigma_{x_{\text{in}}}}d{x_{\text{in}_i}} - \frac{\sigma_{x_{\text{in}}}^2}{2\pi}\\
    &= \int_{0}^{\infty}\frac{{x^2_{\text{in}_i}}\exp{(\frac{-{x^2_{\text{in}_i}}}{2\sigma_{x_{\text{in}}}^2})}}{\sqrt{2\pi}\sigma_{x_{\text{in}}}}d{x_{\text{in}_i}} - \frac{\sigma_{x_{\text{in}}}^2}{2\pi}\\
    \intertext{Let $I = \displaystyle\int_{0}^{\infty}\frac{{x^2_{\text{in}_i}}\exp{(\frac{-{x^2_{\text{in}_i}}}{2\sigma_{x_{\text{in}}}^2})}}{\sqrt{2\pi}\sigma_{x_{\text{in}}}}d{x_{\text{in}_i}}$, then substituting $t = -{x_{\text{in}_i}}$ we have,}
    I &= \int_{0}^{-\infty}\frac{-t^2\exp{(\frac{-t^2}{2\sigma_{x_{\text{in}}}^2})}}{\sqrt{2\pi}\sigma_{x_{\text{in}}}}dt\\
    &= \int_{-\infty}^{0}\frac{t^2\exp{(\frac{-t^2}{2\sigma_{x_{\text{in}}}^2})}}{\sqrt{2\pi}\sigma_{x_{\text{in}}}}dt\\
    \implies I+I & = \int_{-\infty}^{0}\frac{t^2\exp{(\frac{-t^2}{2\sigma_{x_{\text{in}}}^2})}}{\sqrt{2\pi}\sigma_{x_{\text{in}}}}dt + \int_{0}^{\infty}\frac{{x^2_{\text{in}_i}}\exp{(\frac{-{x^2_{\text{in}_i}}}{2\sigma_{x_{\text{in}}}^2})}}{\sqrt{2\pi}\sigma_{x_{\text{in}}}}d{x_{\text{in}_i}} \\
    2I &= \int_{-\infty}^{\infty}\frac{{x^2_{\text{in}_i}}\exp{(\frac{-{x^2_{\text{in}_i}}}{2\sigma_{x_{\text{in}}}^2})}}{\sqrt{2\pi}\sigma_{x_{\text{in}}}}d{x_{\text{in}_i}} = \sigma_{x_{\text{in}}}^2\\
    \implies \mathrm{Var}({x_{\text{out}_i}}) &= \frac{\sigma_{x_{\text{in}}}^2}{2} - \frac{\sigma_{x_{\text{in}}}^2}{2\pi} = \frac{\sigma_{x_{\text{in}}}^2}{2}(1-\frac{1}{\pi}) \\ \Aboxed{\sigma_{x_{\text{out}}}^2 &= \frac{\sigma_{x_{\text{in}}}^2}{2}(1-\frac{1}{\pi})} 
\end{align*}
Now for two inputs $\mathbf{{x_{\text{in}}}}$ and $\mathbf{{y_{\text{in}}}}$ such that for all $i$ we have $\mathrm{Corr({x_{\text{in}_i}},{y_{\text{in}_i}})} = r^l_{x_\text{in}}$, and $\mathbf{{x_{\text{out}}}} = \mathrm{ReLU}(\mathbf{{x_{\text{in}}}})$ and $\mathbf{{y_{\text{out}}}} = \mathrm{ReLU}(\mathbf{{y_{\text{in}}}})$. Then for any $j$ we have,

\begin{align*}
    & \mathrm{Corr}({x_{\text{out}_{j}}},{y_{\text{out}_{j}}}) = \frac{\mathbb{E}[{x_{\text{out}_{j}}}{y_{\text{out}_{j}}}] - \mathbb{E}[{x_{\text{out}_{j}}}]\mathbb{E}[{y_{\text{out}_{j}}}]}{\sqrt{\mathrm{Var}({x_{\text{out}_{j}}})\mathrm{Var}({y_{\text{out}_{j}}})}}\\
    & \mathbb{E}[{x_{\text{out}_{j}}}{y_{\text{out}_{j}}}] = \displaystyle\int_{0}^{\infty}\int_{0}^{\infty}\frac{{x_{\text{in}_j}}{y_{\text{in}_j}}}{2\pi\sigma_{x_{\text{in}}}^2\sqrt{1-({r^l_{x_{\text{in}}}})^2}}\exp{(\frac{-({x^2_{\text{in}_j}} + {y^2_{\text{in}_j}} - 2r^l_{x_{\text{in}}}{x_{\text{in}_j}}{y_{\text{in}_j}})}{2\sigma_{x_{\text{in}}}^2(1-({r^l_{x_{\text{in}}}})^2)})}d{x_{\text{in}_j}}d{y_{\text{in}_j}}\\
    &= \displaystyle\int_{0}^{\infty}\int_{0}^{\infty}\frac{{x_{\text{in}_j}}{y_{\text{in}_j}}}{2\pi\sigma_{x_{\text{in}}}^2\sqrt{1-({r^l_{x_{\text{in}}}})^2}}\exp{(\frac{-({x_{\text{in}_j}} -r^l_{x_{\text{in}}}{y_{\text{in}_j}})^2}{2\sigma_{x_{\text{in}}}^2(1-({r^l_{x_{\text{in}}}})^2)})}\exp{(\frac{-{y^2_{\text{in}_j}}}{2\sigma_{x_{\text{in}}}^2})}d{x_{\text{in}_j}}d{y_{\text{in}_j}}
\end{align*}
Substituting $t = {x_{\text{in}_j}} - r^l_{x_{\text{in}}}{y_{\text{in}_j}}$, and assuming $y_{\text{in}_j}$ is constant for the inner integral,$dx_{\text{in}_j} = dt$
\begin{align*}
    & \mathbb{E}[{x_{\text{out}_j}}{y_{\text{out}_j}}] = \\
    &= \int_{0}^{\infty}\frac{{y_{\text{in}_j} \exp{(\frac{-{y^2_{\text{in}_j}}}{2\sigma_{x_{\text{in}}}^2})}}}{\sqrt{2\pi}\sigma_{x_{\text{in}}}} \displaystyle\int_{-r^l_{x_{\text{in}}}{y_{\text{in}_j}}}^{\infty}\frac{t+r^l_{x_{\text{in}}}{y_{\text{in}_j}}}{\sqrt{2\pi}\sigma_{x_{\text{in}}}\sqrt{1-({r^l_{x_{\text{in}}}})^2}}\exp{(\frac{-t^2}{2\sigma_{x_{\text{in}}}^2(1-({r^l_{x_{\text{in}}}})^2)})}dt  d{y_{\text{in}_j}} \\
    &= \int_{0}^{\infty}\frac{{y_{\text{in}_j}}}{\sqrt{2\pi}\sigma_{x_{\text{in}}}}\exp{(\frac{-{y^2_{\text{in}_j}}}{2\sigma_{x_{\text{in}}}^2})}\int_{-r^l_{x_{\text{in}}}{y_{\text{in}_j}}}^{\infty}\frac{t}{\sqrt{2\pi}\sigma_{x_{\text{in}}}\sqrt{1-({r^l_{x_{\text{in}}}})^2}}\exp{(\frac{-t^2}{2\sigma_{x_{\text{in}}}^2(1-({r^l_{x_{\text{in}}}})^2)})}dt  d{y_{\text{in}_j}} \\
    & +\int_{0}^{\infty}\frac{{y_{\text{in}_j}}}{\sqrt{2\pi}\sigma_x}\exp{(\frac{-{y^2_{\text{in}_j}}}{2\sigma_{x_{\text{in}}}^2})}\int_{-r^l_{x_{\text{in}}}{y_{\text{in}_j}}}^{\infty}\frac{r^l_{x_{\text{in}}}{y_{\text{in}_j}}}{\sqrt{2\pi}\sigma_{x_{\text{in}}}\sqrt{1-({r^l_{x_{\text{in}}}})^2}}\exp{(\frac{-t^2}{2\sigma_{x_{\text{in}}}^2(1-({r^l_{x_{\text{in}}}})^2)})}dt 
 d{y_{\text{in}_j}}
 \end{align*}
 Let us first define $I_1$ and $I_2$ as: 
 \begin{align*}
    I_1 &= \displaystyle\int_{0}^{\infty}\frac{{y_{\text{in}_j}}}{\sqrt{2\pi}\sigma_{x_{\text{in}}}}\exp{(\frac{-{y^2_{\text{in}_j}}}{2\sigma_{x_{\text{in}}}^2})}\int_{-r^l_{x_{\text{in}}}{y_{\text{in}_j}}}^{\infty}\frac{t}{\sqrt{2\pi}\sigma_{x_{\text{in}}}\sqrt{1-({r^l_{x_{\text{in}}}})^2}}\exp{(\frac{-t^2}{2\sigma_{x_{\text{in}}}^2(1-({r^l_{x_{\text{in}}}})^2)})}dt d{y_{\text{in}_j}} \\
    I_2 &= \displaystyle\int_{0}^{\infty}\frac{{y_{\text{in}_j}}}{\sqrt{2\pi}\sigma_x}\exp{(\frac{-{y^2_{\text{in}_j}}}{2\sigma_{x_{\text{in}}}^2})}\int_{-r^l_{x_{\text{in}}}{y_{\text{in}_j}}}^{\infty}\frac{r^l_{x_{\text{in}}}{y_{\text{in}_j}}}{\sqrt{2\pi}\sigma_{x_{\text{in}}}\sqrt{1-({r^l_{x_{\text{in}}}})^2}}\exp{(\frac{-t^2}{2\sigma_{x_{\text{in}}}^2(1-({r^l_{x_{\text{in}}}})^2)})}dt d{y_{\text{in}_j}} \\
    I_1 &= \displaystyle\int_{0}^{\infty}\frac{{y_{\text{in}_j}}}{\sqrt{2\pi}\sigma_{x_{\text{in}}}}\exp{(\frac{-{y^2_{\text{in}_j}}}{2\sigma_{x_{\text{in}}}^2})}\int_{-r^l_{x_{\text{in}}}{y_{\text{in}_j}}}^{\infty}\frac{t}{\sqrt{2\pi}\sigma_{x_{\text{in}}}\sqrt{1-({r^l_{x_{\text{in}}}})^2}}\exp{(\frac{-t^2}{2\sigma_{x_{\text{in}}}^2(1-({r^l_{x_{\text{in}}}})^2)})}dt d{y_{\text{in}_j}}\\
\end{align*}
Substituting $p = \displaystyle\frac{t^2}{2\sigma_{x_{\text{in}}}^2(1-({r^l_{x_{\text{in}}}})^2)} $ we have $dp = \displaystyle\frac{tdt}{\sigma_{x_{\text{in}}}^2(1-({r^l_{x_{\text{in}}}})^2)}$
\begin{align*}
    I_1 &= \int_{0}^{\infty}\frac{{y_{\text{in}_j}}}{\sqrt{2\pi}\sigma_{x_{\text{in}}}}\exp{(\frac{-{y^2_{\text{in}_j}}}{2\sigma_{x_{\text{in}}}^2})}\int_{\frac{(r^l_{x_{\text{in}}}{y_{\text{in}_j}})^2}{2\sigma_{x_{\text{in}}}^2(1-({r^l_{x_{\text{in}}}})^2)}}^{\infty}\frac{\sigma_{x_{\text{in}}}\sqrt{(1-({r^l_{x_{\text{in}}}})^2)}}{\sqrt{2\pi}}\exp{(-p)}dp d{y_{\text{in}_j}}\\
    &= \int_{0}^{\infty}\frac{{y_{\text{in}_j}}}{\sqrt{2\pi}\sigma_{x_{\text{in}}}}\exp{(\frac{-{y^2_{\text{in}_j}}}{2\sigma_{x_{\text{in}}}^2})}\frac{\sigma_{x_{\text{in}}}\sqrt{(1-({r^l_{x_{\text{in}}}})^2)}}{\sqrt{2\pi}}\exp{(\frac{-(r^l_{x_{\text{in}}}{y_{\text{in}_j}})^2}{2\sigma_{x_{\text{in}}}^2(1-({r^l_{x_{\text{in}}}})^2)})}d{y_{\text{in}_j}}\\
    &= \int_{0}^{\infty}\frac{{y_{\text{in}_j}}\sqrt{(1-({r^l_{x_{\text{in}}}})^2)}}{2\pi}\exp{(\frac{-{y^2_{\text{in}_j}}}{2\sigma_{x_{\text{in}}}^2(1-({r^l_{x_{\text{in}}}})^2)})}d{y_{\text{in}_j}}\\
\end{align*}
Substituting $m=\displaystyle\frac{{y^2_{\text{in}_j}}}{2\sigma_{x_{\text{in}}}^2(1-({r^l_{x_{\text{in}}}})^2)}$, $dm = \displaystyle\frac{{y_{\text{in}_j}}d{y_{\text{in}_j}}}{\sigma_{x_{\text{in}}}^2(1-({r^l_{x_{\text{in}}}})^2)}$,
\begin{align*}
    I_1 &= \int_{0}^{\infty}\frac{\sqrt{(1-({r^l_{x_{\text{in}}}})^2)}}{2\pi}(1-({r^l_{x_{\text{in}}}})^2)\sigma_{x_{\text{in}}}^2\exp{(-m)}dm\\
    &=\frac{(1-({r^l_{x_{\text{in}}}})^2)^{\frac{3}{2}}\sigma_{x_{\text{in}}}^2}{2\pi} \\
    I_2 &= \int_{0}^{\infty}\frac{{y_{\text{in}_j}}}{\sqrt{2\pi}\sigma_{x_{\text{in}}}}\exp{(\frac{-{y^2_{\text{in}_j}}}{2\sigma_{x_{\text{in}}}^2})}\int_{-r^l_{x_{\text{in}}}{y_{\text{in}_j}}}^{\infty}\frac{r^l_{x_{\text{in}}}{y_{\text{in}_j}}}{\sqrt{2\pi}\sigma_{x_{\text{in}}}\sqrt{1-({r^l_{x_{\text{in}}}})^2}}\exp{(\frac{-t^2}{2\sigma_{x_{\text{in}}}^2(1-({r^l_{x_{\text{in}}}})^2)})}dt d{y_{\text{in}_j}}\\
    &= \int_{0}^{\infty}\frac{r^l_{x_{\text{in}}}{y^2_{\text{in}_j}}}{\sqrt{2\pi}\sigma_{x_{\text{in}}}}\exp{(\frac{-{y^2_{\text{in}_j}}}{2\sigma_{x_{\text{in}}}^2})}\int_{-r^l_{x_{\text{in}}}{y_{\text{in}_j}}}^{\infty}\frac{1}{\sqrt{2\pi}\sigma_{x_{\text{in}}}\sqrt{1-({r^l_{x_{\text{in}}}})^2}}\exp{(\frac{-t^2}{2\sigma_{x_{\text{in}}}^2(1-({r^l_{x_{\text{in}}}})^2)})}dt d{y_{\text{in}_j}}\\
\end{align*}
Substituting $p = -t$, where $\Phi$ is CDF of Standard Normal Distribution
\begin{align*}
    I_2 &= \int_{0}^{\infty}\frac{r^l_{x_{\text{in}}}{y^2_{\text{in}_j}}}{\sqrt{2\pi}\sigma_{x_{\text{in}}}}\exp{(\frac{-{y^2_{\text{in}_j}}}{2\sigma_{x_{\text{in}}}^2})}\int_{r^l_{x_{\text{in}}}{y_{\text{in}_j}}}^{-\infty}\frac{-1}{\sqrt{2\pi}\sigma_{x_{\text{in}}}\sqrt{1-({r^l_{x_{\text{in}}}})^2}}\exp{(\frac{-p^2}{2\sigma_{x_{\text{in}}}^2(1-({r^l_{x_{\text{in}}}})^2)})}dp d{y_{\text{in}_j}}\\
    &= \int_{0}^{\infty}\frac{r^l_{x_{\text{in}}}{y^2_{\text{in}_j}}}{\sqrt{2\pi}\sigma_{x_{\text{in}}}}\exp{(\frac{-{y^2_{\text{in}_j}}}{2\sigma_{x_{\text{in}}}^2})}\int^{r^l_{x_{\text{in}}}{y_{\text{in}_j}}}_{-\infty}\frac{1}{\sqrt{2\pi}\sigma_{x_{\text{in}}}\sqrt{1-({r^l_{x_{\text{in}}}})^2}}\exp{(\frac{-p^2}{2\sigma_{x_{\text{in}}}^2(1-({r^l_{x_{\text{in}}}})^2)})}dp d{y_{\text{in}_j}}\\
    &= \int_{0}^{\infty}\frac{r^l_{x_{\text{in}}}{y^2_{\text{in}_j}}}{\sqrt{2\pi}\sigma_{x_{\text{in}}}}\exp{(\frac{-{y^2_{\text{in}_j}}}{2\sigma_{x_{\text{in}}}^2})}\Phi(\frac{r^l_{x_{\text{in}}}{y_{\text{in}_j}}}{\sigma_{x_{\text{in}}}\sqrt{1-({r^l_{x_{\text{in}}}})^2}})d{y_{\text{in}_j}}\\
    &= \int_{0}^{\infty}\frac{r^l_{x_{\text{in}}}{y^2_{\text{in}_j}}}{\sqrt{2\pi}\sigma_{x_{\text{in}}}}\exp{(\frac{-{y^2_{\text{in}_j}}}{2\sigma_{x_{\text{in}}}^2})}[\frac{1}{2}(1+ \mathrm{erf}(\frac{r^l_{x_{\text{in}}}{y_{\text{in}_j}}}{\sigma_{x_{\text{in}}}\sqrt{2(1-({r^l_{x_{\text{in}}}})^2)}}))]d{y_{\text{in}_j}}\\
    &= \frac{r^l_{x_{\text{in}}}}{2}\int_{0}^{\infty}\frac{{y^2_{\text{in}_j}}}{\sqrt{2\pi}\sigma_{x_{\text{in}}}}\exp{(\frac{-{y^2_{\text{in}_j}}}{2\sigma_{x_{\text{in}}}^2})}d{y_{\text{in}_j}} + 
\end{align*}
\begin{align*}
    \frac{r^l_{x_{\text{in}}}}{2\sqrt{2\pi}\sigma_{x_{\text{in}}}}\int_{0}^{\infty}{y^2_{\text{in}_j}}\exp{(\frac{-{y^2_{\text{in}_j}}}{2\sigma_{x_{\text{in}}}^2})}\mathrm{erf}(\frac{r^l_{x_{\text{in}}}{y_{\text{in}_j}}}{\sigma_{x_{\text{in}}}\sqrt{2(1-({r^l_{x_{\text{in}}}})^2)}})d{y_{\text{in}_j}}
\end{align*}
Let us define $I_{2,1}$ and $I_{2,2}$ as
\begin{align*}
    I_{2,1} &= \displaystyle\frac{r^l_{x_{\text{in}}}}{2}\int_{0}^{\infty}\frac{{y^2_{\text{in}_j}}}{\sqrt{2\pi}\sigma_{x_{\text{in}}}}\exp{(\frac{-{y^2_{\text{in}_j}}}{2\sigma_{x_{\text{in}}}^2})}d{y_{\text{in}_j}} \\
    I_{2,2} &= \displaystyle\frac{r^l_{x_{\text{in}}}}{2\sqrt{2\pi}\sigma_{x_{\text{in}}}}\int_{0}^{\infty}{y^2_{\text{in}_j}}\exp{(\frac{-{y^2_{\text{in}_j}}}{2\sigma_{x_{\text{in}}}^2})}\mathrm{erf}(\frac{r^l_{x_{\text{in}}}{y_{\text{in}_j}}}{\sigma_{x_{\text{in}}}\sqrt{2(1-({r^l_{x_{\text{in}}}})^2)}})d{y_{\text{in}_j}} \\
    I_{2,1} &= \displaystyle\frac{r^l_{x_{\text{in}}}}{2}\int_{0}^{\infty}\frac{{y^2_{\text{in}_j}}}{\sqrt{2\pi}\sigma_{x_{\text{in}}}}\exp{(\frac{-{y^2_{\text{in}_j}}}{2\sigma_{x_{\text{in}}}^2})}d{y_{\text{in}_j}} \\
    I_{2,1} &= \frac{r^l_{x_{\text{in}}}\sigma_{x_{\text{in}}}^2}{4} \tag*{(Same integral as in variance calculation)}\\
\end{align*}
From \citet{TableIntegralsErrora} we have $\displaystyle\int_{0}^{\infty}x^2\exp{(-b^2x^2)\mathrm{erf}(ax)dx} = \frac{\sqrt{\pi}}{4b^3} - \frac{\tan^{-1}(\frac{b}{a})}{2\sqrt{\pi}b^3} + \frac{a}{2\sqrt{\pi}b^2(a^2+b^2)}$. 

Hence, putting $a = \displaystyle\frac{r^l_{x_{\text{in}}}}{\sigma_{x_{\text{in}}}\sqrt{2(1-({r^l_{x_{\text{in}}}})^2)}}$ and $b = \displaystyle\frac{1}{\sigma_{x_{\text{in}}}\sqrt{2}}$ we get,
\begin{align*}
    & I_{2,2} = \frac{r^l_{x_{\text{in}}}}{2\sqrt{2\pi}\sigma_{x_{\text{in}}}}[\frac{2\sqrt{2}\sigma_{x_{\text{in}}}^3}{4} - \frac{\tan^{-1}(\frac{\sqrt{(1-({r^l_{x_{\text{in}}}})^2)}}{r^l_{x_{\text{in}}}})2\sqrt{2}\sigma_{x_{\text{in}}}^3}{2\sqrt{\pi}} + \frac{\sqrt{2}r^l_{x_{\text{in}}}\sigma_{x_{\text{in}}}^3\sqrt{(1-({r^l_{x_{\text{in}}}})^2)}}{\sqrt{\pi}}] \\
    &= \frac{r^l_{x_{\text{in}}}\sigma_{x_{\text{in}}}^2}{4} - \frac{r^l_{x_{\text{in}}}\cos^{-1}{(r^l_{x_{\text{in}}})}\sigma_{x_{\text{in}}}^2}{2\pi} + \frac{({r^l_{x_{\text{in}}}})^2\sqrt{(1-({r^l_{x_{\text{in}}}})^2)}\sigma_{x_{\text{in}}}^2}{2\pi}\\
    & \mathbb{E}[{x_{\text{out}_j}}{y_{\text{out}_j}}] = I_1 + I_{2,1} + I_{2,2} \\
    & = \frac{(1-({r^l_{x_{\text{in}}}})^2)^{\frac{3}{2}}\sigma_{x_{\text{in}}}^2}{2\pi} + 2*\frac{r^l_{x_{\text{in}}}\sigma_{x_{\text{in}}}^2}{4}  - \frac{r^l_{x_{\text{in}}}\cos^{-1}{(r^l_{x_{\text{in}}})}\sigma_{x_{\text{in}}}^2}{2\pi} + \frac{{(r^l_{x_{\text{in}}}})^2\sqrt{(1-({r^l_{x_{\text{in}}}})^2)}\sigma_{x_{\text{in}}}^2}{2\pi} \\
    &= \frac{r^l_{x_{\text{in}}}\sigma_{x_{\text{in}}}^2}{2} - \frac{r^l_{x_{\text{in}}}\cos^{-1}{(r^l_{x_{\text{in}}})}\sigma_x^2}{2\pi} + \frac{\sqrt{(1-({r^l_{x_{\text{in}}})}^2)}\sigma_{x_{\text{in}}}^2}{2\pi}\\
    & \mathrm{Corr}({x_{\text{out}_j}},{y_{\text{out}_j}})  = \frac{\mathbb{E}[{x_{\text{out}_j}}{y_{\text{out}_j}}] - \mathbb{E}[{x_{\text{out}_j}}]\mathbb{E}[{y_{\text{out}_j}}]}{\sqrt{\mathrm{Var}({x_{\text{out}_j}})\mathrm{Var}({y_{\text{out}_j}})}}\\
    &= \frac{\displaystyle\frac{r^l_{x_{\text{in}}}\sigma_{x_{\text{in}}}^2}{2} - \frac{r^l_{x_{\text{in}}}\cos^{-1}{(r^l_{x_{\text{in}}})}\sigma_{x_{\text{in}}}^2}{2\pi} + \frac{\sqrt{(1-({r^l_{x_{\text{in}}}})^2)}\sigma_x^2}{2\pi} - \frac{\sigma_{x_{\text{in}}}^2}{2\pi}}{\displaystyle\frac{\sigma_{x_{\text{in}}}^2}{2}(1-\frac{1}{\pi})}\\
    \Aboxed{r^l_{x_{\text{out}}} &= \frac{\frac{\pi r^l_{x_{\text{in}}}}{2}  + r^l_{x_{\text{in}}}\sin^{-1}{(r^l_{x_{\text{in}}})} + \sqrt{(1-({r^l_{x_{\text{in}}}})^2)} - 1}{\pi - 1}}
\end{align*}
Backward pass on ReLU can be defined as,
\begin{align*}
    g_{{\text{in}_i}} &= \begin{cases}
        g_{{\text{out}_i}}  & \text{if ${x_{\text{in}_i}} > 0$ (which has probability $\frac{1}{2}$)} \\
		0 & \text{else}
    \end{cases}
\end{align*}
Assuming $\mu_{g_{{\text{out}}}} = 0$,
\begin{align*}
    \mathbb{E}[g_{{{\text{in}_i}}}] &= \frac{1}{2}*0 + \frac{1}{2}*\mathbb{E}[g_{{{\text{out}_i}}}]\\
    \Aboxed{\mu_{g_{{\text{in}}}} &= 0}\\
    \mathrm{Var}(g_{{{\text{in}_i}}}) &= \mathbb{E}[g_{{{\text{in}_i}}}^2] - \mathbb{E}[g_{{\text{in}_i}}]^2 = \mathbb{E}[g_{{\text{in}_i}}^2] \\
    &= \frac{1}{2}*0 + \frac{1}{2}*\mathbb{E}[g_{{\text{out}}}^2] \\
    \Aboxed{\sigma_{g_{{\text{in}}}}^2 &= \frac{\sigma_{g_{{\text{out}}}}^2}{2}}
\end{align*}
If for two inputs $\mathbf{x}_{\text{in}}$ and $\mathbf{y}_{\text{in}}$ for all $i$ we have $\mathrm{Corr}(g_{\text{out}_{x_i}},g_{\text{out}_{y_i}}) = r^l_{g_{\text{out}}}$, and $g_{\text{in}_{x_i}}, g_{\text{in}_{y_i}}$ be the gradient after passing through ReLU layer. Then we have,
\begin{align*}
    \mathbb{E}[g_{\text{in}_{x_i}}g_{\text{in}_{y_i}}] &= \mathbb{P}(x_{\text{in}_i}>0,y_{\text{in}_i}>0)\mathbb{E}[g_{\text{out}_{x_i}}g_{\text{out}_{y_i}}]\\
    &= \mathbb{P}(x_{\text{in}_i}>0,y_{\text{in}_i}>0)r^l_{g_{\text{out}}}\sigma_{g_{{\text{out}}}}^2 \\
\end{align*}
\begin{align*}
    & \mathbb{P}(x_{\text{in}_i}>0,y_{\text{in}_i}>0) = \\
    &= \int_{0}^{\infty}\int_{0}^{\infty}\frac{{x_{\text{in}_i}}{y_{\text{in}_i}}}{2\pi\sigma_{x_{\text{in}}}^2\sqrt{1-({r^l_{x_{\text{in}}}})^2}}\exp{(\frac{-({x^2_{\text{in}_i}} + {y^2_{\text{in}_i}} - 2r^l_{x_{\text{in}}}{x_{\text{in}_i}}{y_{\text{in}_i}})}{2\sigma_{x_{\text{in}}}^2(1-({r^l_{x_{\text{in}}}})^2)})}d{x_{\text{in}_i}}d{y_{\text{in}_i}}\\
    &= \int_{0}^{\infty}\int_{0}^{\infty}\frac{{x_{\text{in}_i}}{y_{\text{in}_i}}}{2\pi\sigma_{x_{\text{in}}}^2\sqrt{1-({r^l_{x_{\text{in}}}})^2}}\exp{(\frac{-({x_{\text{in}_i}} -r^l_{x_{\text{in}}}{y_{\text{in}_i}})^2}{2\sigma_{x_{\text{in}}}^2(1-({r^l_{x_{\text{in}}}})^2)})}\exp{(\frac{-{y^2_{\text{in}_i}}}{2\sigma_{x_{\text{in}}}^2})}d{x_{\text{in}_i}}d{y_{\text{in}_i}}\\
\end{align*}
Substituting $t = {x_{\text{in}_i}} - r^l_{x_{\text{in}}}{y_{\text{in}_i}}$, and assuming $y_{\text{in}_i}$ is constant for the inner integral,$dx_{\text{in}_i} = dt$
\begin{align*}
    & \mathbb{P}(x_{\text{in}_i}>0,y_{\text{in}_i}>0) = \\
    & \int_{0}^{\infty}\frac{1}{\sqrt{2\pi}\sigma_{x_{\text{in}}}}\exp{(\frac{-{y^2_{\text{in}_i}}}{2\sigma_{x_{\text{in}}}^2})}\displaystyle\int_{-r^l_{x_{\text{in}}}{y_{\text{in}_i}}}^{\infty}\frac{1}{\sqrt{2\pi}\sigma_{x_{\text{in}}}\sqrt{1-({r^l_{x_{\text{in}}}})^2}}\exp{(\frac{-t^2}{2\sigma_{x_{\text{in}}}^2(1-({r^l_{x_{\text{in}}}})^2)})}dt  d{y_{\text{in}_i}} \\
\end{align*}
Substituting $p = -t$, where $\Phi$ is CDF of Standard Normal Distribution
\begin{align*}
    & \mathbb{P}(x_{\text{in}_i}>0,y_{\text{in}_i}>0) = \\
    & = \int_{0}^{\infty}\frac{1}{\sqrt{2\pi}\sigma_{x_{\text{in}}}}\exp{(\frac{-{y^2_{\text{in}_i}}}{2\sigma_{x_{\text{in}}}^2})}\int_{r^l_{x_{\text{in}}}{y_{\text{in}_i}}}^{-\infty}\frac{-1}{\sqrt{2\pi}\sigma_{x_{\text{in}}}\sqrt{1-({r^l_{x_{\text{in}}}})^2}}\exp{(\frac{-p^2}{2\sigma_{x_{\text{in}}}^2(1-({r^l_{x_{\text{in}}}})^2)})}dp d{y_{\text{in}_i}}\\
    &= \int_{0}^{\infty}\frac{1}{\sqrt{2\pi}\sigma_{x_{\text{in}}}}\exp{(\frac{-{y^2_{\text{in}_i}}}{2\sigma_{x_{\text{in}}}^2})}\int^{r^l_{x_{\text{in}}}{y_{\text{in}_i}}}_{-\infty}\frac{1}{\sqrt{2\pi}\sigma_{x_{\text{in}}}\sqrt{1-({r^l_{x_{\text{in}}}})^2}}\exp{(\frac{-p^2}{2\sigma_{x_{\text{in}}}^2(1-({r^l_{x_{\text{in}}}})^2)})}dp d{y_{\text{in}_i}}\\
    &= \int_{0}^{\infty}\frac{1}{\sqrt{2\pi}\sigma_{x_{\text{in}}}}\exp{(\frac{-{y^2_{\text{in}_i}}}{2\sigma_{x_{\text{in}}}^2})}\Phi(\frac{r^l_{x_{\text{in}}}{y_{\text{in}_i}}}{\sigma_{x_{\text{in}}}\sqrt{1-({r^l_{x_{\text{in}}}})^2}})d{y_{\text{in}_i}}\\
    &= \int_{0}^{\infty}\frac{1}{\sqrt{2\pi}\sigma_{x_{\text{in}}}}\exp{(\frac{-{y^2_{\text{in}_i}}}{2\sigma_{x_{\text{in}}}^2})}[\frac{1}{2}(1+ \mathrm{erf}(\frac{r^l_{x_{\text{in}}}{y_{\text{in}_i}}}{\sigma_{x_{\text{in}}}\sqrt{2(1-({r^l_{x_{\text{in}}}})^2)}}))]d{y_{\text{in}_i}}\\
    &= \frac{1}{2}\int_{0}^{\infty}\frac{1}{\sqrt{2\pi}\sigma_{x_{\text{in}}}}\exp{(\frac{-{y^2_{\text{in}_i}}}{2\sigma_{x_{\text{in}}}^2})}d{y_{\text{in}_i}} + \frac{1}{2\sqrt{2\pi}\sigma_{x_{\text{in}}}}\int_{0}^{\infty}\exp{(\frac{-{y^2_{\text{in}_i}}}{2\sigma_{x_{\text{in}}}^2})}\mathrm{erf}(\frac{r^l_{x_{\text{in}}}{y_{\text{in}_i}}}{\sigma_{x_{\text{in}}}\sqrt{2(1-({r^l_{x_{\text{in}}}})^2)}})d{y_{\text{in}_i}}\\
    &= \frac{1}{4} + \frac{1}{2\sqrt{2\pi}\sigma_{x_{\text{in}}}}\int_{0}^{\infty}\exp{(\frac{-{y^2_{\text{in}_i}}}{2\sigma_{x_{\text{in}}}^2})}\mathrm{erf}(\frac{r^l_{x_{\text{in}}}{y_{\text{in}_i}}}{\sigma_{x_{\text{in}}}\sqrt{2(1-({r^l_{x_{\text{in}}}})^2)}})d{y_{\text{in}_i}}
\end{align*}
From \citet{TableIntegralsErrora} we have $\displaystyle\int_{0}^{\infty}\exp{(-b^2x^2)\mathrm{erf}(ax)dx} = \frac{\sqrt{\pi}}{2b} - \frac{1}{b\sqrt{\pi}}\tan^{-1}(\frac{b}{a})$

Putting $a = \displaystyle\frac{r^l_{x_{\text{in}}}}{\sigma_{x_{\text{in}}}\sqrt{2(1-({r^l_{x_{\text{in}}}})^2)}}$ and $b = \displaystyle\frac{1}{\sigma_{x_{\text{in}}}\sqrt{2}}$ we get,
\begin{align*}
    \mathbb{P}(x_{\text{in}_i}>0,y_{\text{in}_i}>0) &=  \frac{1}{4} + \frac{1}{2\sqrt{2\pi}\sigma_{x_{\text{in}}}}[\frac{\sqrt{\pi}\sigma_{x_{\text{in}}}\sqrt{2}}{2} - \frac{\sigma_{x_{\text{in}}}\sqrt{2}}{\sqrt{\pi}}\tan^{-1}(\frac{\sqrt{(1-({r^l_{x_{\text{in}}}})^2)}}{r^l_{x_{\text{in}}}})]\\
    &= \frac{1}{4} + \frac{1}{2\pi}[\frac{\pi}{2} - \cos^{-1}{(r^l_{x_{\text{in}}})}]\\
    &= \frac{1}{4} + \frac{\sin^{-1}{(r^l_{x_{\text{in}}})}}{2\pi}\\
    \implies \mathbb{E}[g_{\text{in}_{x_i}}g_{\text{in}_{y_i}}] &= (\frac{1}{4} + \frac{\sin^{-1}{(r^l_{x_{\text{in}}})}}{2\pi})r^l_{g_{\text{out}}}\sigma_{g_{{\text{out}}}}^2\\
    \mathrm{Corr}(g_{\text{in}_{x_i}},g_{\text{in}_{y_i}}) &= \frac{(\frac{1}{4} + \frac{\sin^{-1}{(r^l_{x_{\text{in}}})}}{2\pi})r^l_{g_{\text{out}}}\sigma_{g_{{\text{out}}}}^2}{\frac{\sigma_{g_{{\text{out}}}}^2}{2}}\\
    \Aboxed{r^l_{g_{\text{out}}} &= (\frac{1}{2} + \frac{\sin^{-1}{(r^l_{x_{\text{in}}})}}{\pi})r^l_{g_{\text{out}}}}
\end{align*}

\subsection{GeLU}
\label{proof: gelu}

Forward pass through GeLU is defined as,
\begin{align*}
    \mathbf{x_{\text{out}}} & = \mathrm{GeLU}(\mathbf{x_{\text{in}}})\\
    \implies {x_{\text{out}_i}} & = {x_{\text{in}_i}}\Phi({x_{\text{in}_i}})
    \intertext{where $\Phi(x)$ is CDF of Standard Normal Distribution at $x$}
    &= \frac{{x_{\text{in}_i}}}{2}\left(1+\mathrm{erf}(\frac{{x_{\text{in}_i}}}{\sqrt{2}})\right)
\end{align*}

To get the mean of output of GeLU, we have

\begin{align*}
    \mathbb{E}[{x_{\text{out}_i}}] &= \int_{-\infty}^{\infty}\frac{{x_{\text{out}_i}}}{\sqrt{2\pi}\sigma_{x_{\text{in}}}}\exp{(\frac{-x_{\text{in}_i}^2}{2\sigma_{x_{\text{in}}}^2})}dx_{\text{in}_i}\\
    &= \int_{-\infty}^{\infty}\frac{{x_{\text{in}_i}}(1+\mathrm{erf}(\frac{{x_{\text{in}_i}}}{\sqrt{2}}))}{2\sqrt{2\pi}\sigma_{x_{\text{in}}}}\exp{(\frac{-x_{\text{in}_i}^2}{2\sigma_{x_{\text{in}}}^2})}dx_{\text{in}_i}\\
    &= \int_{-\infty}^{\infty}\frac{{x_{\text{in}_i}}}{2\sqrt{2\pi}\sigma_{x_{\text{in}}}}\exp{(\frac{-x_{\text{in}_i}^2}{2\sigma_{x_{\text{in}}}^2})}dx_{\text{in}_i} + \int_{-\infty}^{\infty}\frac{{x_{\text{in}_i}}\mathrm{erf}(\frac{{x_{\text{in}_i}}}{\sqrt{2}})}{2\sqrt{2\pi}\sigma_{x_{\text{in}}}}\exp{(\frac{-x_{\text{in}_i}^2}{2\sigma_{x_{\text{in}}}^2})}dx_{\text{in}_i}\\
    &= \int_{-\infty}^{\infty}\frac{{x_{\text{in}_i}}\mathrm{erf}(\frac{{x_{\text{in}_i}}}{\sqrt{2}})}{2\sqrt{2\pi}\sigma_{x_{\text{in}}}}\exp{(\frac{-x_{\text{in}_i}^2}{2\sigma_{x_{\text{in}}}^2})}dx_{\text{in}_i} \tag*{(Integral of odd function)}\\
    &= \frac{1}{2\sqrt{2\pi}\sigma_{x_{\text{in}}}}\int_{-\infty}^{\infty}{x_{\text{in}_i}}\mathrm{erf}(\frac{{x_{\text{in}_i}}}{\sqrt{2}})\exp{(\frac{-x_{\text{in}_i}^2}{2\sigma_{x_{\text{in}}}^2})}dx_{\text{in}_i}
\end{align*}
From 2.6.1.4 of \citet{integrals_book}, $\displaystyle\int_{-\infty}^{\infty}z\mathrm{erf}(az)\exp{(-a_1z^2)}dz = \frac{a}{a_1\sqrt{a^2+a_1}}$

Substituting, $\displaystyle a=\frac{1}{\sqrt{2}}, a_1 = \frac{1}{2\sigma_{x_{\text{in}}}^2}$, we have
\begin{align*}
    & \mathbb{E}[{x_{\text{out}_i}}] = \frac{1}{2\sqrt{2\pi}\sigma_{x_{\text{in}}}}\frac{\frac{1}{\sqrt{2}}}{\frac{1}{2\sigma_{x_{\text{in}}}^2}\sqrt{\frac{1}{2}+\frac{1}{2\sigma_{x_{\text{in}}}^2}}}\\
    &= \frac{1}{2\sqrt{2\pi}\sigma_{x_{\text{in}}}}\frac{2\sigma_{x_{\text{in}}}^3}{\sqrt{\sigma_{x_{\text{in}}}^2+1}}\\
     \Aboxed{\mu_{x_{\text{out}}} &= \frac{\sigma_{x_{\text{in}}}^2}{\sqrt{2\pi(\sigma_{x_{\text{in}}}^2+1)}}} 
\end{align*}
For calculating variance of output,
\begin{align*}
    \mathbb{E}[{x^2_{\text{out}_i}}] &= \int_{-\infty}^{\infty}\frac{{x^2_{\text{out}_i}}}{\sqrt{2\pi}\sigma_{x_{\text{in}}}}\exp{(\frac{-x_{\text{in}_i}^2}{2\sigma_{x_{\text{in}}}^2})}dx_{\text{in}_i}\\
    &= \int_{-\infty}^{\infty}\frac{{x^2_{\text{in}_i}}(1+\mathrm{erf}(\frac{{x_{\text{in}_i}}}{\sqrt{2}}))^2}{4\sqrt{2\pi}\sigma_{x_{\text{in}}}}\exp{(\frac{-x_{\text{in}_i}^2}{2\sigma_{x_{\text{in}}}^2})}dx_{\text{in}_i}\\
    &= \int_{-\infty}^{\infty}\frac{{x^2_{\text{in}_i}}}{4\sqrt{2\pi}\sigma_{x_{\text{in}}}}\exp{(\frac{-x_{\text{in}_i}^2}{2\sigma_{x_{\text{in}}}^2})}dx_{\text{in}_i} \\
    & + \int_{-\infty}^{\infty}\frac{{x^2_{\text{in}_i}\mathrm{erf}(\frac{{x_{\text{in}_i}}}{\sqrt{2}})}}{2\sqrt{2\pi}\sigma_{x_{\text{in}}}}\exp{(\frac{-x_{\text{in}_i}^2}{2\sigma_{x_{\text{in}}}^2})}dx_{\text{in}_i} + \int_{-\infty}^{\infty}\frac{{x^2_{\text{in}_i}}\mathrm{erf}^2(\frac{{x_{\text{in}_i}}}{\sqrt{2}})}{4\sqrt{2\pi}\sigma_{x_{\text{in}}}}\exp{(\frac{-x_{\text{in}_i}^2}{2\sigma_{x_{\text{in}}}^2})}dx_{\text{in}_i}\\
    &= \frac{\sigma_{x_{\text{in}}}^2}{4} + \int_{-\infty}^{\infty}\frac{{x^2_{\text{in}_i}}\mathrm{erf}^2(\frac{{x_{\text{in}_i}}}{\sqrt{2}})}{4\sqrt{2\pi}\sigma_{x_{\text{in}}}}\exp{(\frac{-x_{\text{in}_i}^2}{2\sigma_{x_{\text{in}}}^2})}dx_{\text{in}_i}\tag*{(Definition of variance, and integral of odd function)}\\
    &= \frac{\sigma_{x_{\text{in}}}^2}{4} + \frac{1}{{4\sqrt{2\pi}\sigma_{x_{\text{in}}}}}\int_{-\infty}^{\infty}{x^2_{\text{in}_i}}\mathrm{erf}^2(\frac{{x_{\text{in}_i}}}{\sqrt{2}})\exp{(\frac{-x_{\text{in}_i}^2}{2\sigma_{x_{\text{in}}}^2})}dx_{\text{in}_i}
\end{align*}
    From 2.7.3.3 of \citet{integrals_book} 
\begin{align*}
    & \int_{-\infty}^{\infty}z^2\exp{(-az^2)}\mathrm{erf}(a_1z)\mathrm{erf}(a_2z) = \\
    & \frac{1}{\sqrt{\pi}}(\frac{1}{a\sqrt{a}}\tan^{-1}{(\frac{a_1a_2}{\sqrt{a^2+aa_1^2+aa_2^2}})} + \frac{a_1a_2(2a+a_1^2+a_2^2)}{a\sqrt{a+a_1^2+a_2^2}(a^2+aa_1^2+aa_2^2+a_1^2a_2^2)})    
\end{align*}

    Substituting $a = \frac{1}{2\sigma_{x_{\text{in}}}^2}, a_1 = a_2 = \frac{1}{\sqrt{2}}$ 

\begin{align*}
    & \int_{-\infty}^{\infty}{x^2_{\text{in}_i}}\mathrm{erf}^2(\frac{{x_{\text{in}_i}}}{\sqrt{2}})\exp{(\frac{-x_{\text{in}_i}^2}{2\sigma_{x_{\text{in}}}^2})}dx_{\text{in}_i} \\
    &= \frac{1}{\sqrt{\pi}}(2\sqrt{2}\sigma_{x_{\text{in}}}^3\tan^{-1}{(\frac{\frac{1}{2}}{\sqrt{\frac{1}{4\sigma_{x_{\text{in}}}^4} + \frac{1}{2\sigma_{x_{\text{in}}}^2}}})} + \frac{\frac{1}{2}(\frac{1}{\sigma_{x_{\text{in}}}^2}+1)}{\frac{1}{2\sigma_{x_{\text{in}}}^2}\sqrt{\frac{1}{2\sigma_{x_{\text{in}}}^2}+1}(\frac{1}{4\sigma_{x_{\text{in}}}^4} + \frac{1}{2\sigma_{x_{\text{in}}}^2} + \frac{1}{4})})\\
    &= \frac{1}{\sqrt{\pi}}(2\sqrt{2}\sigma_{x_{\text{in}}}^3\tan^{-1}{(\frac{\sigma_{x_{\text{in}}}^2}{\sqrt{(\sigma_{x_{\text{in}}}^2+1)^2 - \sigma_{x_{\text{in}}}^4}})} + \frac{4\sqrt{2}\sigma_{x_{\text{in}}}^5(\sigma_{x_{\text{in}}}^2+1)}{\sqrt{2\sigma_{x_{\text{in}}}^2+1}(\sigma_{x_{\text{in}}}^4 + 2\sigma_{x_{\text{in}}}^2 + 1)})\\
    &= \frac{1}{\sqrt{\pi}}(2\sqrt{2}\sigma_{x_{\text{in}}}^3\sin^{-1}{(\frac{\sigma_{x_{\text{in}}}^2}{\sigma_{x_{\text{in}}}^2 + 1})} + \frac{4\sqrt{2}\sigma_{x_{\text{in}}}^5}{\sqrt{2\sigma_{x_{\text{in}}}^2+1}(\sigma_{x_{\text{in}}}^2+1)})\\
    &= \frac{2\sqrt{2}\sigma_{x_{\text{in}}}^3}{\sqrt{\pi}}(\sin^{-1}{(\frac{\sigma_{x_{\text{in}}}^2}{\sigma_{x_{\text{in}}}^2 + 1})} + \frac{2\sigma_{x_{\text{in}}}^2}{\sqrt{2\sigma_{x_{\text{in}}}^2+1}(\sigma_{x_{\text{in}}}^2+1)}))
\end{align*}
\begin{align*}
    \mathbb{E}[{x^2_{\text{out}_i}}] &= \frac{\sigma_{x_{\text{in}}}^2}{4} + \frac{1}{{4\sqrt{2\pi}\sigma_{x_{\text{in}}}}}\int_{-\infty}^{\infty}{x^2_{\text{in}_i}}\mathrm{erf}^2(\frac{{x_{\text{in}_i}}}{\sqrt{2}})\exp{(\frac{-x_{\text{in}_i}^2}{2\sigma_{x_{\text{in}}}^2})}dx_{\text{in}_i} \\
    &= \frac{\sigma_{x_{\text{in}}}^2}{4} + \frac{1}{{4\sqrt{2\pi}\sigma_{x_{\text{in}}}}}\frac{2\sqrt{2}\sigma_{x_{\text{in}}}^3}{\sqrt{\pi}}(\sin^{-1}{(\frac{\sigma_{x_{\text{in}}}^2}{\sigma_{x_{\text{in}}}^2 + 1})} + \frac{2\sigma_{x_{\text{in}}}^2}{\sqrt{2\sigma_{x_{\text{in}}}^2+1}(\sigma_{x_{\text{in}}}^2+1)}))\\
    \mathbb{E}[{x^2_{\text{out}_i}}] &= \frac{\sigma_{x_{\text{in}}}^2}{4} + \frac{\sigma_{x_{\text{in}}}^2}{2\pi}(\sin^{-1}{(\frac{\sigma_{x_{\text{in}}}^2}{\sigma_{x_{\text{in}}}^2 + 1})} + \frac{2\sigma_{x_{\text{in}}}^2}{\sqrt{2\sigma_{x_{\text{in}}}^2+1}(\sigma_{x_{\text{in}}}^2+1)}))\\
    \mathrm{Var}({x_{\text{out}_i}}) &= \mathbb{E}[{x^2_{\text{out}_i}}] - (\mathbb{E}[{x_{\text{out}_i}}])^2 \\
    \Aboxed{\sigma^2_{x_{out}} &= \frac{\sigma^2_{x_{in}}}{2 \pi} (\frac{\pi}{2} - \frac{\sigma^2_{x_{in}}}{1+\sigma^2_{x_{in}}} + \sin^{-1}(\frac{\sigma^2_{x_{in}}}{1+\sigma^2_{x_{in}}}) + \frac{2 \sigma^2_{x_{in}}}{(1+\sigma^2_{x_{in}})\sqrt{1+2\sigma^2_{x_{in}}}})}
\end{align*}

Now if we have two inputs $\mathbf{x_{\text{in}}}$ and $\mathbf{y_{\text{in}}}$ such that for all values of $i$, we have $\mathrm{Corr}({x_{\text{in}_i}}, {y_{\text{in}_i}}) = r^l_{x_{\text{in}}}$, then we can calculate the covariance  $\mathrm{Cov}({x_{\text{out}_j}}, {y_{\text{out}_j}})$ for any $j$ as,
\begin{align*}
    \mathrm{Cov}({x_{\text{out}_j}}, {y_{\text{out}_j}}) &= \mathbb{E}[{x_{\text{out}_j}}{y_{\text{out}_j}}] - \mathbb{E}[{x_{\text{out}_j}}]\mathbb{E}[{y_{\text{out}_j}}] 
\end{align*}
\begin{align*}
    & \mathbb{E}[{x_{\text{out}_j}}{y_{\text{out}_j}}] \\
    &= \iint_{-\infty}^{\infty}\frac{{x_{\text{out}_j}}{y_{\text{out}_j}}}{2\pi\sigma_{x_{\text{in}}}^2\sqrt{(1- (r^l_{x_{\text{in}}})^2)}}\exp{(\frac{-x^2_{\text{in}_j} + 2r^l_{x_{\text{in}}}x_{\text{in}_j}y_{\text{in}_j} - y^2_{\text{in}_j}}{2\sigma_{x_{\text{in}}}^2(1- (r^l_{x_{\text{in}}})^2)})}dx_{\text{in}_j}dy_{\text{in}_j} = I\\
    &= \iint_{-\infty}^{\infty}\frac{{x_{\text{in}_j}(1+\mathrm{erf}(\frac{x_{\text{in}_j}}{\sqrt{2}}))}{y_{\text{in}_j}(1+\mathrm{erf}(\frac{y_{\text{in}_j}}{\sqrt{2}}))}}{8\pi\sigma_{x_{\text{in}}}^2\sqrt{(1- (r^l_{x_{\text{in}}})^2)}}\exp{(\frac{-x^2_{\text{in}_j} + 2r^l_{x_{\text{in}}}x_{\text{in}_j}y_{\text{in}_j} - y^2_{\text{in}_j}}{2\sigma_{x_{\text{in}}}^2(1- (r^l_{x_{\text{in}}})^2)})}dx_{\text{in}_j}dy_{\text{in}_j}\\
    &= \int_{-\infty}^{\infty}\frac{{y_{\text{in}_j}(1+\mathrm{erf}(\frac{y_{\text{in}_j}}{\sqrt{2}}))}}{8\pi\sigma_{x_{\text{in}}}^2\sqrt{(1- (r^l_{x_{\text{in}}})^2)}}\exp{(\frac{-y^2_{\text{in}_j}}{2\sigma_{x_{\text{in}}}^2(1- (r^l_{x_{\text{in}}})^2)})}I_Xdy_{\text{in}_j}
\end{align*}
Where $I_X = \displaystyle\int_{-\infty}^{\infty}{x_{\text{in}_j}(1+\mathrm{erf}(\frac{x_{\text{in}_j}}{\sqrt{2}}))}\exp{(\frac{-x^2_{\text{in}_j} + 2r^l_{x_{\text{in}}}x_{\text{in}_j}y_{\text{in}_j}}{2\sigma_{x_{\text{in}}}^2(1- (r^l_{x_{\text{in}}})^2)})}dx_{\text{in}_j}$
\begin{align*}
    & I_X = \int_{-\infty}^{\infty}{x_{\text{in}_j}(1+\mathrm{erf}(\frac{x_{\text{in}_j}}{\sqrt{2}}))}\exp{(\frac{-x^2_{\text{in}_j} + 2r^l_{x_{\text{in}}}x_{\text{in}_j}y_{\text{in}_j}}{2\sigma_{x_{\text{in}}}^2(1- (r^l_{x_{\text{in}}})^2)})}dx_{\text{in}_j}\\
    &= \int_{-\infty}^{\infty}{x_{\text{in}_j}}\exp{(\frac{-x^2_{\text{in}_j} + 2r^l_{x_{\text{in}}}x_{\text{in}_j}y_{\text{in}_j}}{2\sigma_{x_{\text{in}}}^2(1- (r^l_{x_{\text{in}}})^2)})}dx_{\text{in}_j} + \\
    & \hspace{100pt} \int_{-\infty}^{\infty}{x_{\text{in}_j}\mathrm{erf}(\frac{x_{\text{in}_j}}{\sqrt{2}})}\exp{(\frac{-x^2_{\text{in}_j} + 2r^l_{x_{\text{in}}}x_{\text{in}_j}y_{\text{in}_j}}{2\sigma_{x_{\text{in}}}^2(1- (r^l_{x_{\text{in}}})^2)})}dx_{\text{in}_j}\\
\end{align*}
Let, $\displaystyle I_{X,1} = \int_{-\infty}^{\infty}{x_{\text{in}_j}}\exp{(\frac{-x^2_{\text{in}_j} + 2r^l_{x_{\text{in}}}x_{\text{in}_j}y_{\text{in}_j}}{2\sigma_{x_{\text{in}}}^2(1- (r^l_{x_{\text{in}}})^2)})}dx_{\text{in}_j}$

$\displaystyle I_{X,2} =  \int_{-\infty}^{\infty}{x_{\text{in}_j}\mathrm{erf}(\frac{x_{\text{in}_j}}{\sqrt{2}})}\exp{(\frac{-x^2_{\text{in}_j} + 2r^l_{x_{\text{in}}}x_{\text{in}_j}y_{\text{in}_j}}{2\sigma_{x_{\text{in}}}^2(1- (r^l_{x_{\text{in}}})^2)})}dx_{\text{in}_j}$

\begin{align*}
    & I_{X,1} = \int_{-\infty}^{\infty}{x_{\text{in}_j}}\exp{(\frac{-x^2_{\text{in}_j} + 2r^l_{x_{\text{in}}}x_{\text{in}_j}y_{\text{in}_j}}{2\sigma_{x_{\text{in}}}^2(1- (r^l_{x_{\text{in}}})^2)})}dx_{\text{in}_j} \\
    &= \int_{-\infty}^{\infty}{x_{\text{in}_j}}\exp{(\frac{-x^2_{\text{in}_j} + 2r^l_{x_{\text{in}}}x_{\text{in}_j}y_{\text{in}_j}}{2\sigma_{x_{\text{in}}}^2(1- (r^l_{x_{\text{in}}})^2)})}\exp{(\frac{-(r^l_{x_{\text{in}}})^2y^2_{\text{in}_j}}{2\sigma_{x_{\text{in}}}^2(1- (r^l_{x_{\text{in}}})^2)})}\exp{(\frac{(r^l_{x_{\text{in}}})^2y^2_{\text{in}_j}}{2\sigma_{x_{\text{in}}}^2(1- (r^l_{x_{\text{in}}})^2)})}dx_{\text{in}_j}\\
    &= \exp{(\frac{(r^l_{x_{\text{in}}})^2y^2_{\text{in}_j}}{2\sigma_{x_{\text{in}}}^2(1- (r^l_{x_{\text{in}}})^2)})}\int_{-\infty}^{\infty}{x_{\text{in}_j}}\exp{(\frac{-(x_{\text{in}_j} - r^l_{x_{\text{in}}}y_{\text{in}_j})^2}{2\sigma_{x_{\text{in}}}^2(1- (r^l_{x_{\text{in}}})^2)})}dx_{\text{in}_j}\\
    &= \int_{-\infty}^{\infty}\frac{x_{\text{in}_j}}{\sqrt{2\pi}\sigma_{x_{\text{in}}}\sqrt{(1- (r^l_{x_{\text{in}}})^2)}}\exp{(\frac{-(x_{\text{in}_j} - r^l_{x_{\text{in}}}y_{\text{in}_j})^2}{2\sigma_{x_{\text{in}}}^2(1- (r^l_{x_{\text{in}}})^2)})}dx_{\text{in}_j}\\
    &= r^l_{x_{\text{in}}}y_{\text{in}_j}\sqrt{2\pi}\sigma_{x_{\text{in}}}\sqrt{(1- (r^l_{x_{\text{in}}})^2)}\exp{(\frac{(r^l_{x_{\text{in}}})^2y^2_{\text{in}_j}}{2\sigma_{x_{\text{in}}}^2(1- (r^l_{x_{\text{in}}})^2)})}\\
    & I_{X,2} =  \int_{-\infty}^{\infty}{x_{\text{in}_j}\mathrm{erf}(\frac{x_{\text{in}_j}}{\sqrt{2}})}\exp{(\frac{-x^2_{\text{in}_j} + 2r^l_{x_{\text{in}}}x_{\text{in}_j}y_{\text{in}_j}}{2\sigma_{x_{\text{in}}}^2(1- (r^l_{x_{\text{in}}})^2)})}dx_{\text{in}_j} 
\end{align*}
From 2.7.2.4 of \citet{integrals_book}, 
\begin{align*}
\int_{-\infty}^{\infty}z\mathrm{erf}(a_1z)\exp{(-az^2+bz)}dz &= \\
= \frac{\sqrt{\pi}b}{2a\sqrt{a}}\exp{(\frac{b^2}{4a})}\mathrm{erf}(\frac{a_1b}{2\sqrt{a^2+aa_1^2}}) & + \frac{a_1}{a\sqrt{a+a_1^2}}\exp{(\frac{b^2}{4a+4a_1^2})}
\end{align*}
    
Substituting $a_1 = \frac{1}{\sqrt{2}}, a = \frac{1}{2\sigma_{x_{\text{in}}}^2(1- (r^l_{x_{\text{in}}})^2)}, b = \frac{r^l_{x_{\text{in}}}y_{\text{in}_j}}{\sigma_{x_{\text{in}}}^2(1- (r^l_{x_{\text{in}}})^2)}$, we get
\begin{align*}
    I_{X,2} &= \frac{\sqrt{\pi}\frac{r^l_{x_{\text{in}}}y_{\text{in}_j}}{\sigma_{x_{\text{in}}}^2(1- (r^l_{x_{\text{in}}})^2)}}{2\frac{1}{2\sqrt{2}\sigma_{x_{\text{in}}}^3(1- (r^l_{x_{\text{in}}})^2)^{\frac{3}{2}}}}\exp{(\frac{\frac{(r^l_{x_{\text{in}}})^2y^2_{\text{in}_j}}{\sigma_{x_{\text{in}}}^4(1- (r^l_{x_{\text{in}}})^2)^2}}{4\frac{1}{2\sigma_{x_{\text{in}}}^2(1- (r^l_{x_{\text{in}}})^2)}})}\mathrm{erf}(\frac{\frac{r^l_{x_{\text{in}}}y_{\text{in}_j}}{\sqrt{2}\sigma_{x_{\text{in}}}^2(1- (r^l_{x_{\text{in}}})^2)}}{2\sqrt{\frac{1}{4\sigma_{x_{\text{in}}}^4(1- (r^l_{x_{\text{in}}})^2)^2} + \frac{1}{4\sigma_{x_{\text{in}}}^2(1- (r^l_{x_{\text{in}}})^2)}}})\\
    &+ \frac{\frac{1}{\sqrt{2}}}{\frac{1}{2\sigma_{x_{\text{in}}}^2(1- (r^l_{x_{\text{in}}})^2)}\sqrt{\frac{1}{2\sigma_{x_{\text{in}}}^2(1- (r^l_{x_{\text{in}}})^2)} + \frac{1}{2}}}\exp{(\frac{\frac{(r^l_{x_{\text{in}}})^2y^2_{\text{in}_j}}{\sigma_{x_{\text{in}}}^4(1- (r^l_{x_{\text{in}}})^2)^2}}{4\frac{1}{2\sigma_{x_{\text{in}}}^2(1- (r^l_{x_{\text{in}}})^2)} + \frac{4}{2}})}\\
    &= r^l_{x_{\text{in}}}y_{\text{in}_j}\sqrt{2\pi}\sigma_{x_{\text{in}}}\sqrt{(1- (r^l_{x_{\text{in}}})^2)}\exp{(\frac{(r^l_{x_{\text{in}}})^2y^2_{\text{in}_j}}{2\sigma_{x_{\text{in}}}^2(1- (r^l_{x_{\text{in}}})^2)})}\mathrm{erf}(\frac{r^l_{x_{\text{in}}}y_{\text{in}_j}}{\sqrt{2(\sigma_{x_{\text{in}}}^2(1-(r^l_{x_{\text{in}}})^2) + 1)}})\\
    &+ \frac{2\sigma_{x_{\text{in}}}^3(1-(r^l_{x_{\text{in}}})^2)^\frac{3}{2}}{\sqrt{\sigma_{x_{\text{in}}}^2(1-(r^l_{x_{\text{in}}})^2) + 1}}\exp{(\frac{(r^l_{x_{\text{in}}})^2y^2_{\text{in}_j}}{2(\sigma_{x_{\text{in}}}^2(1-(r^l_{x_{\text{in}}})^2) + 1)\sigma_{x_{\text{in}}}^2(1-(r^l_{x_{\text{in}}})^2)})}
\end{align*}
Let us define $I_{X,2,1}$ and $I_{X,2,2}$  as:
\begin{align*}
    I_{X,2,1} &= r^l_{x_{\text{in}}}y_{\text{in}_j}\sqrt{2\pi}\sigma_{x_{\text{in}}}\sqrt{(1- (r^l_{x_{\text{in}}})^2)}\exp{(\frac{(r^l_{x_{\text{in}}})^2y^2_{\text{in}_j}}{2\sigma_{x_{\text{in}}}^2(1- (r^l_{x_{\text{in}}})^2)})}\mathrm{erf}(\frac{r^l_{x_{\text{in}}}y_{\text{in}_j}}{\sqrt{2(\sigma_{x_{\text{in}}}^2(1-(r^l_{x_{\text{in}}})^2) + 1)}})\\ 
    I_{X,2,2} &= \frac{2\sigma_{x_{\text{in}}}^3(1-(r^l_{x_{\text{in}}})^2)^\frac{3}{2}}{\sqrt{\sigma_{x_{\text{in}}}^2(1-(r^l_{x_{\text{in}}})^2) + 1}}\exp{(\frac{(r^l_{x_{\text{in}}})^2y^2_{\text{in}_j}}{2(\sigma_{x_{\text{in}}}^2(1-(r^l_{x_{\text{in}}})^2) + 1)\sigma_{x_{\text{in}}}^2(1-(r^l_{x_{\text{in}}})^2)})}
\end{align*}

\begin{align*}
    I &= \int_{-\infty}^{\infty}\frac{{y_{\text{in}_j}(1+\mathrm{erf}(\frac{y_{\text{in}_j}}{\sqrt{2}}))}}{8\pi\sigma_{x_{\text{in}}}^2\sqrt{(1- (r^l_{x_{\text{in}}})^2)}}\exp{(\frac{-y^2_{\text{in}_j}}{2\sigma_{x_{\text{in}}}^2(1- (r^l_{x_{\text{in}}})^2)})}I_Xdy_{\text{in}_j}\\
    &= \int_{-\infty}^{\infty}\frac{{y_{\text{in}_j}(1+\mathrm{erf}(\frac{y_{\text{in}_j}}{\sqrt{2}}))}}{8\pi\sigma_{x_{\text{in}}}^2\sqrt{(1- (r^l_{x_{\text{in}}})^2)}}\exp{(\frac{-y^2_{\text{in}_j}}{2\sigma_{x_{\text{in}}}^2(1- (r^l_{x_{\text{in}}})^2)})}(I_{X,1} + I_{X,2,1} + I_{X,2,2})dy_{\text{in}_j}\\
    I_1 &= \int_{-\infty}^{\infty}\frac{{y_{\text{in}_j}(1+\mathrm{erf}(\frac{y_{\text{in}_j}}{\sqrt{2}}))}}{8\pi\sigma_{x_{\text{in}}}^2\sqrt{(1- (r^l_{x_{\text{in}}})^2)}}\exp{(\frac{-y^2_{\text{in}_j}}{2\sigma_{x_{\text{in}}}^2(1- (r^l_{x_{\text{in}}})^2)})}I_{X,1}dy_{\text{in}_j}\\
    I_2 &= \int_{-\infty}^{\infty}\frac{{y_{\text{in}_j}(1+\mathrm{erf}(\frac{y_{\text{in}_j}}{\sqrt{2}}))}}{8\pi\sigma_{x_{\text{in}}}^2\sqrt{(1- (r^l_{x_{\text{in}}})^2)}}\exp{(\frac{-y^2_{\text{in}_j}}{2\sigma_{x_{\text{in}}}^2(1- (r^l_{x_{\text{in}}})^2)})}I_{X,2,1}dy_{\text{in}_j}\\
    I_3 &= \int_{-\infty}^{\infty}\frac{{y_{\text{in}_j}(1+\mathrm{erf}(\frac{y_{\text{in}_j}}{\sqrt{2}}))}}{8\pi\sigma_{x_{\text{in}}}^2\sqrt{(1- (r^l_{x_{\text{in}}})^2)}}\exp{(\frac{-y^2_{\text{in}_j}}{2\sigma_{x_{\text{in}}}^2(1- (r^l_{x_{\text{in}}})^2)})}I_{X,2,2}dy_{\text{in}_j}\\
\end{align*}
We have $I = I_1 + I_2 + I_3$
\begin{align*}
    I_1 &= \int_{-\infty}^{\infty}\frac{{y_{\text{in}_j}(1+\mathrm{erf}(\frac{y_{\text{in}_j}}{\sqrt{2}}))}}{8\pi\sigma_{x_{\text{in}}}^2\sqrt{(1- (r^l_{x_{\text{in}}})^2)}}\exp{(\frac{-y^2_{\text{in}_j}}{2\sigma_{x_{\text{in}}}^2(1- (r^l_{x_{\text{in}}})^2)})}r^l_{x_{\text{in}}}y_{\text{in}_j}\\
    &\hspace{20pt}\sqrt{2\pi}\sigma_{x_{\text{in}}}\sqrt{(1- (r^l_{x_{\text{in}}})^2)}\exp{(\frac{(r^l_{x_{\text{in}}})^2y^2_{\text{in}_j}}{2\sigma_{x_{\text{in}}}^2(1- (r^l_{x_{\text{in}}})^2)})}dy_{\text{in}_j}\\
    &= \frac{r^l_{x_{\text{in}}}}{4}\int_{-\infty}^{\infty}\frac{{y^2_{\text{in}_j}(1+\mathrm{erf}(\frac{y_{\text{in}_j}}{\sqrt{2}}))}}{\sqrt{2\pi}\sigma_{x_{\text{in}}}^2}\exp{(\frac{-y^2_{\text{in}_j}}{2\sigma_{x_{\text{in}}}^2})}dy_{\text{in}_j}\\
    &= \frac{r^l_{x_{\text{in}}}}{4}\int_{-\infty}^{\infty}\frac{{y^2_{\text{in}_j}}}{\sqrt{2\pi}\sigma_{x_{\text{in}}}^2}\exp{(\frac{-y^2_{\text{in}_j}}{2\sigma_{x_{\text{in}}}^2})}dy_{\text{in}_j} + \frac{r^l_{x_{\text{in}}}}{4}\int_{-\infty}^{\infty}\frac{{y^2_{\text{in}_j}\mathrm{erf}(\frac{y_{\text{in}_j}}{\sqrt{2}})}}{\sqrt{2\pi}\sigma_{x_{\text{in}}}^2}\exp{(\frac{-y^2_{\text{in}_j}}{2\sigma_{x_{\text{in}}}^2})}dy_{\text{in}_j}\\
    &= \frac{r^l_{x_{\text{in}}}\sigma_{x_{\text{in}}}^2}{4} \tag*{(Definition of variance, and integral of odd function)}\\
    I_2 &= \int_{-\infty}^{\infty}\frac{{y_{\text{in}_j}(1+\mathrm{erf}(\frac{y_{\text{in}_j}}{\sqrt{2}}))}}{8\pi\sigma_{x_{\text{in}}}^2\sqrt{(1- (r^l_{x_{\text{in}}})^2)}}\exp{(\frac{-y^2_{\text{in}_j}}{2\sigma_{x_{\text{in}}}^2(1- (r^l_{x_{\text{in}}})^2)})}r^l_{x_{\text{in}}}y_{\text{in}_j}\\
    &\hspace{20pt}\sqrt{2\pi}\sigma_{x_{\text{in}}}\sqrt{(1- (r^l_{x_{\text{in}}})^2)}\exp{(\frac{(r^l_{x_{\text{in}}})^2y^2_{\text{in}_j}}{2\sigma_{x_{\text{in}}}^2(1- (r^l_{x_{\text{in}}})^2)})}\mathrm{erf}(\frac{r^l_{x_{\text{in}}}y_{\text{in}_j}}{\sqrt{2(\sigma_{x_{\text{in}}}^2(1-(r^l_{x_{\text{in}}})^2) + 1)}})dy_{\text{in}_j}\\
    &= \frac{r^l_{x_{\text{in}}}}{4\sqrt{2\pi}\sigma_{x_{\text{in}}}}\int_{-\infty}^{\infty}{y^2_{\text{in}_j}(1+\mathrm{erf}(\frac{y_{\text{in}_j}}{\sqrt{2}}))}\exp{(\frac{-y^2_{\text{in}_j}}{2\sigma_{x_{\text{in}}}^2})}\mathrm{erf}(\frac{r^l_{x_{\text{in}}}y_{\text{in}_j}}{\sqrt{2(\sigma_{x_{\text{in}}}^2(1-(r^l_{x_{\text{in}}})^2) + 1)}})dy_{\text{in}_j}\\
    &= \frac{r^l_{x_{\text{in}}}}{4\sqrt{2\pi}\sigma_{x_{\text{in}}}}\int_{-\infty}^{\infty}{y^2_{\text{in}_j}}\exp{(\frac{-y^2_{\text{in}_j}}{2\sigma_{x_{\text{in}}}^2})}\mathrm{erf}(\frac{r^l_{x_{\text{in}}}y_{\text{in}_j}}{\sqrt{2(\sigma_{x_{\text{in}}}^2(1-(r^l_{x_{\text{in}}})^2) + 1)}})dy_{\text{in}_j} \\
    &+ \frac{r^l_{x_{\text{in}}}}{4\sqrt{2\pi}\sigma_{x_{\text{in}}}}\int_{-\infty}^{\infty}{y^2_{\text{in}_j}\mathrm{erf}(\frac{y_{\text{in}_j}}{\sqrt{2}})}\exp{(\frac{-y^2_{\text{in}_j}}{2\sigma_{x_{\text{in}}}^2})}\mathrm{erf}(\frac{r^l_{x_{\text{in}}}y_{\text{in}_j}}{\sqrt{2(\sigma_{x_{\text{in}}}^2(1-(r^l_{x_{\text{in}}})^2) + 1)}})dy_{\text{in}_j}\\
    &= \frac{r^l_{x_{\text{in}}}}{4\sqrt{2\pi}\sigma_{x_{\text{in}}}}\int_{-\infty}^{\infty}{y^2_{\text{in}_j}\mathrm{erf}(\frac{y_{\text{in}_j}}{\sqrt{2}})}\exp{(\frac{-y^2_{\text{in}_j}}{2\sigma_{x_{\text{in}}}^2})}\mathrm{erf}(\frac{r^l_{x_{\text{in}}}y_{\text{in}_j}}{\sqrt{2(\sigma_{x_{\text{in}}}^2(1-(r^l_{x_{\text{in}}})^2) + 1)}})dy_{\text{in}_j} \tag*{(Integral of Odd function)}
\end{align*}
    From 2.7.3.3 of \citet{integrals_book}, 
\begin{align*}
    & \int_{-\infty}^{\infty}z^2\exp{(-az^2)}\mathrm{erf}(a_1z)\mathrm{erf}(a_2z) = \\ 
    & \frac{1}{\sqrt{\pi}}(\frac{1}{a\sqrt{a}}\tan^{-1}{(\frac{a_1a_2}{\sqrt{a^2+aa_1^2+aa_2^2}})} + \frac{a_1a_2(2a+a_1^2+a_2^2)}{a\sqrt{a+a_1^2+a_2^2}(a^2+aa_1^2+aa_2^2+a_1^2a_2^2)})
\end{align*}
    Substituting $a = \frac{1}{2\sigma_{x_{\text{in}}}^2}, a_1 = \frac{1}{\sqrt{2}}, a_2 = \frac{r^l_{x_{\text{in}}}}{\sqrt{2(\sigma_{x_{\text{in}}}^2(1-(r^l_{x_{\text{in}}})^2) + 1)}}$
\begin{align*}
    a_1a_2 &= \frac{r^l_{x_{\text{in}}}}{2\sqrt{(\sigma_{x_{\text{in}}}^2(1-(r^l_{x_{\text{in}}})^2) + 1)}}\\
    a^2+aa_1^2+aa_2^2 &= \frac{1}{4\sigma_{x_{\text{in}}}^4} + \frac{1}{4\sigma_{x_{\text{in}}}^2} + \frac{(r^l_{x_{\text{in}}})^2}{4\sigma_{x_{\text{in}}}^2(\sigma_{x_{\text{in}}}^2(1-(r^l_{x_{\text{in}}})^2) + 1)}\\
    &= \frac{\sigma_{x_{\text{in}}}^2(1-(r^l_{x_{\text{in}}})^2) + 1 + \sigma_{x_{\text{in}}}^4(1-(r^l_{x_{\text{in}}})^2) + \sigma_{x_{\text{in}}}^2 + (r^l_{x_{\text{in}}})^2\sigma_{x_{\text{in}}}^2}{4\sigma_{x_{\text{in}}}^4(\sigma_{x_{\text{in}}}^2(1-(r^l_{x_{\text{in}}})^2) + 1)}\\
    &= \frac{\sigma_{x_{\text{in}}}^4 + 2\sigma_{x_{\text{in}}}^2 + 1 - (r^l_{x_{\text{in}}})^2\sigma_{x_{\text{in}}}^4}{4\sigma_{x_{\text{in}}}^4(\sigma_{x_{\text{in}}}^2(1-(r^l_{x_{\text{in}}})^2) + 1)} = \frac{(\sigma_{x_{\text{in}}}^2 + 1)^2 - (r^l_{x_{\text{in}}}\sigma_{x_{\text{in}}}^2)^2}{4\sigma_{x_{\text{in}}}^4(\sigma_{x_{\text{in}}}^2(1-(r^l_{x_{\text{in}}})^2) + 1)}\\
    a+a_1^2+a_2^2 &= \frac{a^2+aa_1^2+aa_2^2}{a} = \frac{(\sigma_{x_{\text{in}}}^2 + 1)^2 - (r^l_{x_{\text{in}}}\sigma_{x_{\text{in}}}^2)^2}{4\sigma_{x_{\text{in}}}^4(\sigma_{x_{\text{in}}}^2(1-(r^l_{x_{\text{in}}})^2) + 1)}*2\sigma_{x_{\text{in}}}^2 \\
    & = \frac{(\sigma_{x_{\text{in}}}^2 + 1)^2 - (r^l_{x_{\text{in}}}\sigma_{x_{\text{in}}}^2)^2}{2\sigma_{x_{\text{in}}}^2(\sigma_{x_{\text{in}}}^2(1-(r^l_{x_{\text{in}}})^2) + 1)}\\
    a^2+aa_1^2+aa_2^2+a_1^2a_2^2 &= \frac{(\sigma_{x_{\text{in}}}^2 + 1)^2 - (r^l_{x_{\text{in}}}\sigma_{x_{\text{in}}}^2)^2}{4\sigma_{x_{\text{in}}}^4(\sigma_{x_{\text{in}}}^2(1-(r^l_{x_{\text{in}}})^2) + 1)} + \frac{(r^l_{x_{\text{in}}})^2}{4(\sigma_{x_{\text{in}}}^2(1-(r^l_{x_{\text{in}}})^2) + 1)}\\
    &= \frac{(\sigma_{x_{\text{in}}}^2 + 1)^2 - (r^l_{x_{\text{in}}}\sigma_{x_{\text{in}}}^2)^2 + (r^l_{x_{\text{in}}})^2\sigma_{x_{\text{in}}}^4}{4\sigma_{x_{\text{in}}}^4(\sigma_{x_{\text{in}}}^2(1-(r^l_{x_{\text{in}}})^2) + 1)} = \frac{(\sigma_{x_{\text{in}}}^2 + 1)^2}{4\sigma_{x_{\text{in}}}^4(\sigma_{x_{\text{in}}}^2(1-(r^l_{x_{\text{in}}})^2) + 1)}\\
    2a+a_1^2+a_2^2 &= \frac{1}{2\sigma_{x_{\text{in}}}^2} + \frac{(\sigma_{x_{\text{in}}}^2 + 1)^2 - (r^l_{x_{\text{in}}}\sigma_{x_{\text{in}}}^2)^2}{2\sigma_{x_{\text{in}}}^2(\sigma_{x_{\text{in}}}^2(1-(r^l_{x_{\text{in}}})^2) + 1)}\\
    &= \frac{(\sigma_{x_{\text{in}}}^2 + 1)^2 - (r^l_{x_{\text{in}}}\sigma_{x_{\text{in}}}^2)^2 + \sigma_{x_{\text{in}}}^2(1-(r^l_{x_{\text{in}}})^2) + 1}{2\sigma_{x_{\text{in}}}^2(\sigma_{x_{\text{in}}}^2(1-(r^l_{x_{\text{in}}})^2) + 1)}\\
    &= \frac{(\sigma_{x_{\text{in}}}^2 + 1)^2 + \sigma_{x_{\text{in}}}^2 + 1 - (r^l_{x_{\text{in}}}\sigma_{x_{\text{in}}}^2)^2 - \sigma_{x_{\text{in}}}^2(r^l_{x_{\text{in}}})^2}{2\sigma_{x_{\text{in}}}^2(\sigma_{x_{\text{in}}}^2(1-(r^l_{x_{\text{in}}})^2) + 1)}\\
    &= \frac{(\sigma_{x_{\text{in}}}^2 + 1)(\sigma_{x_{\text{in}}}^2 + 2) - (r^l_{x_{\text{in}}})^2\sigma_{x_{\text{in}}}^2(\sigma_{x_{\text{in}}}^2 + 1)}{2\sigma_{x_{\text{in}}}^2(\sigma_{x_{\text{in}}}^2(1-(r^l_{x_{\text{in}}})^2) + 1)}\\
    &= \frac{(\sigma_{x_{\text{in}}}^2 + 1)(\sigma_{x_{\text{in}}}^2(1 - (r^l_{x_{\text{in}}})^2) + 2)}{2\sigma_{x_{\text{in}}}^2(\sigma_{x_{\text{in}}}^2(1-(r^l_{x_{\text{in}}})^2) + 1)}\\
\end{align*}
\begin{align*}
    I_2 = \frac{r^l_{x_{\text{in}}}}{4\sqrt{2}\pi\sigma_{x_{\text{in}}}}( & 2\sqrt{2}\sigma_{x_{\text{in}}}^3\tan^{-1}({\frac{\frac{r^l_{x_{\text{in}}}}{2\sqrt{(\sigma_{x_{\text{in}}}^2(1-(r^l_{x_{\text{in}}})^2) + 1)}}}{\sqrt{\frac{(\sigma_{x_{\text{in}}}^2 + 1)^2 - (r^l_{x_{\text{in}}}\sigma_{x_{\text{in}}}^2)^2}{4\sigma_{x_{\text{in}}}^4(\sigma_{x_{\text{in}}}^2(1-(r^l_{x_{\text{in}}})^2) + 1)}}}}))\\
    &+ \frac{r^l_{x_{\text{in}}}}{4\sqrt{2}\pi\sigma_{x_{\text{in}}}}(\frac{\frac{r^l_{x_{\text{in}}}}{2\sqrt{(\sigma_{x_{\text{in}}}^2(1-(r^l_{x_{\text{in}}})^2) + 1)}}\frac{(\sigma_{x_{\text{in}}}^2 + 1)(\sigma_{x_{\text{in}}}^2(1 - (r^l_{x_{\text{in}}})^2) + 2)}{2\sigma_{x_{\text{in}}}^2(\sigma_{x_{\text{in}}}^2(1-(r^l_{x_{\text{in}}})^2) + 1)}}{\frac{1}{2\sigma_{x_{\text{in}}}^2}\sqrt{\frac{(\sigma_{x_{\text{in}}}^2 + 1)^2 - (r^l_{x_{\text{in}}}\sigma_{x_{\text{in}}}^2)^2}{2\sigma_{x_{\text{in}}}^2(\sigma_{x_{\text{in}}}^2(1-(r^l_{x_{\text{in}}})^2) + 1)}}\frac{(\sigma_{x_{\text{in}}}^2 + 1)^2}{4\sigma_{x_{\text{in}}}^4(\sigma_{x_{\text{in}}}^2(1-(r^l_{x_{\text{in}}})^2) + 1)}})\\
    = \frac{r^l_{x_{\text{in}}}}{4\sqrt{2}\pi\sigma_{x_{\text{in}}}}(& 2\sqrt{2}\sigma_{x_{\text{in}}}^3\tan^{-1}{(\frac{r^l_{x_{\text{in}}}\sigma_{x_{\text{in}}}^2}{\sqrt{(\sigma_{x_{\text{in}}}^2 + 1)^2 - (r^l_{x_{\text{in}}}\sigma_{x_{\text{in}}}^2)^2}})})\\
    &+ \frac{r^l_{x_{\text{in}}}}{4\sqrt{2}\pi\sigma_{x_{\text{in}}}}(\frac{2\sqrt{2}r^l_{x_{\text{in}}}\sigma_{x_{\text{in}}}^5(\sigma_{x_{\text{in}}}^2(1 - (r^l_{x_{\text{in}}})^2) + 2)}{(\sigma_{x_{\text{in}}}^2 + 1){\sqrt{(\sigma_{x_{\text{in}}}^2 + 1)^2 - (r^l_{x_{\text{in}}}\sigma_{x_{\text{in}}}^2)^2}}})
 \end{align*}
 \begin{align*}
    & I_2 = \frac{r^l_{x_{\text{in}}}\sigma_{x_{\text{in}}}^2}{2\pi}(\sin^{-1}{(\frac{r^l_{x_{\text{in}}}\sigma_{x_{\text{in}}}^2}{\sigma_{x_{\text{in}}}^2 + 1})} + \frac{r^l_{x_{\text{in}}}\sigma_{x_{\text{in}}}^2(\sigma_{x_{\text{in}}}^2(1 - (r^l_{x_{\text{in}}})^2) + 2)}{(\sigma_{x_{\text{in}}}^2 + 1){\sqrt{(\sigma_{x_{\text{in}}}^2 + 1)^2 - (r^l_{x_{\text{in}}}\sigma_{x_{\text{in}}}^2)^2}}})\\
    & I_3 = \int_{-\infty}^{\infty}\frac{{y_{\text{in}_j}(1+\mathrm{erf}(\frac{y_{\text{in}_j}}{\sqrt{2}}))}}{8\pi\sigma_{x_{\text{in}}}^2\sqrt{(1- (r^l_{x_{\text{in}}})^2)}}\exp{(\frac{-y^2_{\text{in}_j}}{2\sigma_{x_{\text{in}}}^2(1- (r^l_{x_{\text{in}}})^2)})}\\
    &\hspace{20pt}\frac{2\sigma_{x_{\text{in}}}^3(1-(r^l_{x_{\text{in}}})^2)^\frac{3}{2}}{\sqrt{\sigma_{x_{\text{in}}}^2(1-(r^l_{x_{\text{in}}})^2) + 1}}\exp{(\frac{(r^l_{x_{\text{in}}})^2y^2_{\text{in}_j}}{2(\sigma_{x_{\text{in}}}^2(1-(r^l_{x_{\text{in}}})^2) + 1)\sigma_{x_{\text{in}}}^2(1-(r^l_{x_{\text{in}}})^2)})}dy_{\text{in}_j}\\
    &= \int_{-\infty}^{\infty}\frac{\sigma_{x_{\text{in}}}(1-(r^l_{x_{\text{in}}})^2){y_{\text{in}_j}(1+\mathrm{erf}(\frac{y_{\text{in}_j}}{\sqrt{2}}))}}{4\pi\sqrt{\sigma_{x_{\text{in}}}^2(1-(r^l_{x_{\text{in}}})^2) + 1}}\exp{(\frac{-y^2_{\text{in}_j}(\sigma_{x_{\text{in}}}^2(1-(r^l_{x_{\text{in}}})^2) + 1 - (r^l_{x_{\text{in}}})^2)}{2(\sigma_{x_{\text{in}}}^2(1-(r^l_{x_{\text{in}}})^2) + 1)\sigma_{x_{\text{in}}}^2(1-(r^l_{x_{\text{in}}})^2)})}dy_{\text{in}_j}\\
    &= \int_{-\infty}^{\infty}\frac{\sigma_{x_{\text{in}}}(1-(r^l_{x_{\text{in}}})^2){y_{\text{in}_j}(1+\mathrm{erf}(\frac{y_{\text{in}_j}}{\sqrt{2}}))}}{4\pi\sqrt{\sigma_{x_{\text{in}}}^2(1-(r^l_{x_{\text{in}}})^2) + 1}}\exp{(\frac{-y^2_{\text{in}_j}(\sigma_{x_{\text{in}}}^2+1)(1-(r^l_{x_{\text{in}}})^2)}{2(\sigma_{x_{\text{in}}}^2(1-(r^l_{x_{\text{in}}})^2) + 1)\sigma_{x_{\text{in}}}^2(1-(r^l_{x_{\text{in}}})^2)})}dy_{\text{in}_j}\\
    &= \frac{\sigma_{x_{\text{in}}}(1-(r^l_{x_{\text{in}}})^2)}{4\pi\sqrt{\sigma_{x_{\text{in}}}^2(1-(r^l_{x_{\text{in}}})^2) + 1}}\int_{-\infty}^{\infty}{y_{\text{in}_j}(1+\mathrm{erf}(\frac{y_{\text{in}_j}}{\sqrt{2}}))}\exp{(\frac{-y^2_{\text{in}_j}(\sigma_{x_{\text{in}}}^2+1)}{2(\sigma_{x_{\text{in}}}^2(1-(r^l_{x_{\text{in}}})^2) + 1)\sigma_{x_{\text{in}}}^2})}dy_{\text{in}_j}\\
    &= \frac{\sigma_{x_{\text{in}}}(1-(r^l_{x_{\text{in}}})^2)}{4\pi\sqrt{\sigma_{x_{\text{in}}}^2(1-(r^l_{x_{\text{in}}})^2) + 1}}\int_{-\infty}^{\infty}{y_{\text{in}_j}}\exp{(\frac{-y^2_{\text{in}_j}(\sigma_{x_{\text{in}}}^2+1)}{2(\sigma_{x_{\text{in}}}^2(1-(r^l_{x_{\text{in}}})^2) + 1)\sigma_{x_{\text{in}}}^2})}dy_{\text{in}_j} \\
    &+ \frac{\sigma_{x_{\text{in}}}(1-(r^l_{x_{\text{in}}})^2)}{4\pi\sqrt{\sigma_{x_{\text{in}}}^2(1-(r^l_{x_{\text{in}}})^2) + 1}}\int_{-\infty}^{\infty}{y_{\text{in}_j}\mathrm{erf}(\frac{y_{\text{in}_j}}{\sqrt{2}})}\exp{(\frac{-y^2_{\text{in}_j}(\sigma_{x_{\text{in}}}^2+1)}{2(\sigma_{x_{\text{in}}}^2(1-(r^l_{x_{\text{in}}})^2) + 1)\sigma_{x_{\text{in}}}^2})}dy_{\text{in}_j}\\
    &= \frac{\sigma_{x_{\text{in}}}(1-(r^l_{x_{\text{in}}})^2)}{4\pi\sqrt{\sigma_{x_{\text{in}}}^2(1-(r^l_{x_{\text{in}}})^2) + 1}}\int_{-\infty}^{\infty}{y_{\text{in}_j}\mathrm{erf}(\frac{y_{\text{in}_j}}{\sqrt{2}})}\exp{(\frac{-y^2_{\text{in}_j}(\sigma_{x_{\text{in}}}^2+1)}{2(\sigma_{x_{\text{in}}}^2(1-(r^l_{x_{\text{in}}})^2) + 1)\sigma_{x_{\text{in}}}^2})}dy_{\text{in}_j} \tag*{(Integral of Odd function)}
\end{align*}

From 2.6.1.4 of \citet{integrals_book}, $\displaystyle\int_{-\infty}^{\infty}z\mathrm{erf}(az)\exp{(-a_1z^2)}dz = \frac{a}{a_1\sqrt{a^2+a_1}}$

Substituting, $\displaystyle a=\frac{1}{\sqrt{2}}, a_1 = \frac{(\sigma_{x_{\text{in}}}^2+1)}{2\sigma_{x_{\text{in}}}^2(\sigma_{x_{\text{in}}}^2(1-(r^l_{x_{\text{in}}})^2) + 1)}$, we have
\begin{align*}
    I_3 &= \frac{\sigma_{x_{\text{in}}}(1-(r^l_{x_{\text{in}}})^2)}{4\pi\sqrt{\sigma_{x_{\text{in}}}^2(1-(r^l_{x_{\text{in}}})^2) + 1}}(\frac{\frac{1}{\sqrt{2}}}{\frac{(\sigma_{x_{\text{in}}}^2+1)}{2\sigma_{x_{\text{in}}}^2(\sigma_{x_{\text{in}}}^2(1-(r^l_{x_{\text{in}}})^2) + 1)}\sqrt{\frac{1}{2}+\frac{(\sigma_{x_{\text{in}}}^2+1)}{2\sigma_{x_{\text{in}}}^2(\sigma_{x_{\text{in}}}^2(1-(r^l_{x_{\text{in}}})^2) + 1)}}})\\
    &= \frac{\sigma_{x_{\text{in}}}(1-(r^l_{x_{\text{in}}})^2)}{4\pi\sqrt{\sigma_{x_{\text{in}}}^2(1-(r^l_{x_{\text{in}}})^2) + 1}}\frac{2\sigma_{x_{\text{in}}}^3(\sigma_{x_{\text{in}}}^2(1-(r^l_{x_{\text{in}}})^2) + 1)^{\frac{3}{2}}}{(\sigma_{x_{\text{in}}}^2+1)\sqrt{\sigma_{x_{\text{in}}}^4(1-(r^l_{x_{\text{in}}})^2)}+\sigma_{x_{\text{in}}}^2+\sigma_{x_{\text{in}}}^2+1}\\
    I_3 &= \frac{\sigma_{x_{\text{in}}}^4(\sigma_{x_{\text{in}}}^2(1-(r^l_{x_{\text{in}}})^2) + 1)(1-(r^l_{x_{\text{in}}})^2)}{2\pi(\sigma_{x_{\text{in}}}^2+1)\sqrt{(\sigma_{x_{\text{in}}}^2 + 1)^2 - (r^l_{x_{\text{in}}}\sigma_{x_{\text{in}}}^2)^2}}\\
\end{align*}
Finally we have,
\begin{align*}
    I &= I_1 + I_2 + I_3\\
    &= \frac{r^l_{x_{\text{in}}}\sigma_{x_{\text{in}}}^2}{4} + \frac{r^l_{x_{\text{in}}}\sigma_{x_{\text{in}}}^2}{2\pi}(\sin^{-1}{(\frac{r^l_{x_{\text{in}}}\sigma_{x_{\text{in}}}^2}{\sigma_{x_{\text{in}}}^2 + 1})} + \frac{r^l_{x_{\text{in}}}\sigma_{x_{\text{in}}}^2(\sigma_{x_{\text{in}}}^2(1 - (r^l_{x_{\text{in}}})^2) + 2)}{(\sigma_{x_{\text{in}}}^2 + 1){\sqrt{(\sigma_{x_{\text{in}}}^2 + 1)^2 - (r^l_{x_{\text{in}}}\sigma_{x_{\text{in}}}^2)^2}}}) \\
    & \hspace{170pt} + \frac{\sigma_{x_{\text{in}}}^4(\sigma_{x_{\text{in}}}^2(1-(r^l_{x_{\text{in}}})^2) + 1)(1-(r^l_{x_{\text{in}}})^2)}{2\pi(\sigma_{x_{\text{in}}}^2+1)\sqrt{(\sigma_{x_{\text{in}}}^2 + 1)^2 - (r^l_{x_{\text{in}}}\sigma_{x_{\text{in}}}^2)^2}}
\end{align*}
\begin{align*}
    I &= \frac{r^l_{x_{\text{in}}}\sigma_{x_{\text{in}}}^2}{4} + \frac{r^l_{x_{\text{in}}}\sigma_{x_{\text{in}}}^2}{2\pi}\sin^{-1}{(\frac{r^l_{x_{\text{in}}}\sigma_{x_{\text{in}}}^2}{\sigma_{x_{\text{in}}}^2 + 1})} + \frac{\sigma_{x_{\text{in}}}^4(\sigma_{x_{\text{in}}}^2(1-(r^l_{x_{\text{in}}})^2) + 1 + (r^l_{x_{\text{in}}})^2)}{2\pi(\sigma_{x_{\text{in}}}^2+1)\sqrt{(\sigma_{x_{\text{in}}}^2 + 1)^2 - (r^l_{x_{\text{in}}}\sigma_{x_{\text{in}}}^2)^2}}\\
    I &= \frac{\sigma_{x_{\text{in}}}^2}{4}\left[r^l_{x_{\text{in}}} + \frac{2r^l_{x_{\text{in}}}}{\pi}\sin^{-1}{(\frac{r^l_{x_{\text{in}}}\sigma_{x_{\text{in}}}^2}{\sigma_{x_{\text{in}}}^2 + 1})} + \frac{2\sigma_{x_{\text{in}}}^2(\sigma_{x_{\text{in}}}^2(1-(r^l_{x_{\text{in}}})^2) + 1 + (r^l_{x_{\text{in}}})^2)}{\pi(\sigma_{x_{\text{in}}}^2+1)\sqrt{(\sigma_{x_{\text{in}}}^2 + 1)^2 - (r^l_{x_{\text{in}}}\sigma_{x_{\text{in}}}^2)^2}}\right] 
\end{align*}
We have, 
\begin{align*}
    \mathrm{Cov}({x_{\text{out}_j}}, {y_{\text{out}_j}}) &= I - \mathbb{E}[{x_{\text{out}_j}}]\mathbb{E}[{y_{\text{out}_j}}] \\
    \mathrm{Cov}({x_{\text{out}_j}}, {y_{\text{out}_j}}) &= I - \frac{\sigma_{x_{\text{in}}}^4}{2\pi(\sigma_{x_{\text{in}}}^2 + 1)}  
\end{align*}
\begin{empheq}[box=\widefbox]{align*}
    \mathrm{Cov}({x_{\text{out}_j}}, {y_{\text{out}_j}}) =  \frac{\sigma_{x_{\text{in}}}^2}{4 \pi} (\pi r^l_{x_{\text{in}}} &+ 2r^l_{x_{\text{in}}}\sin^{-1}{(\frac{r^l_{x_{\text{in}}}\sigma_{x_{\text{in}}}^2}{\sigma_{x_{\text{in}}}^2 + 1})} \\ &+ \frac{2\sigma_{x_{\text{in}}}^2(\sigma_{x_{\text{in}}}^2(1-(r^l_{x_{\text{in}}})^2) + 1 + (r^l_{x_{\text{in}}})^2)}{(\sigma_{x_{\text{in}}}^2+1)\sqrt{(\sigma_{x_{\text{in}}}^2 + 1)^2 - (r^l_{x_{\text{in}}}\sigma_{x_{\text{in}}}^2)^2}} - \frac{2\sigma_{x_{\text{in}}}^2}{(\sigma_{x_{\text{in}}}^2 + 1)} ) 
\end{empheq}
The backward pass through GeLU is defined as,
\begin{align*}
    g_{\text{in}_i} &= (\Phi(x_{\text{in}_i}) + \frac{x_{\text{in}_i}}{\sqrt{2\pi}}\exp{(\frac{-x^2_{\text{in}_i}}{2})})g_{\text{out}_i}\\
    &= (\frac{1}{2}(1 + \mathrm{erf}(\frac{x_{\text{in}_i}}{\sqrt{2}})) + \frac{x_{\text{in}_i}}{\sqrt{2\pi}}\exp{(\frac{-x^2_{\text{in}_i}}{2})})g_{\text{out}_i}
\end{align*}
So the mean of gradient is obtained as following,
\begin{align*}
    \mathbb{E}[g_{\text{in}_i}] &= \mathbb{E}[(\frac{1}{2}(1 + \mathrm{erf}(\frac{x_{\text{in}_i}}{\sqrt{2}})) + \frac{x_{\text{in}_i}}{\sqrt{2\pi}}\exp{(\frac{-x^2_{\text{in}_i}}{2})})g_{\text{out}_i}]\\
    &= \mathbb{E}[(\frac{1}{2}(1 + \mathrm{erf}(\frac{x_{\text{in}_i}}{\sqrt{2}})) + \frac{x_{\text{in}_i}}{\sqrt{2\pi}}\exp{(\frac{-x^2_{\text{in}_i}}{2})})]\mathbb{E}[g_{\text{out}_i}] = 0 \\
    \Aboxed{\mu_{g_{\text{in}}} &= 0}
\end{align*}
Similarly for variance,
\begin{align*}
    \mathbb{E}[g^2_{\text{in}_i}] &= \mathbb{E}[(\frac{1}{2}(1 + \mathrm{erf}(\frac{x_{\text{in}_i}}{\sqrt{2}})) + \frac{x_{\text{in}_i}}{\sqrt{2\pi}}\exp{(\frac{-x^2_{\text{in}_i}}{2})})^{2}g^{2}_{\text{out}_i}]\\
    &= \mathbb{E}[(\frac{1}{2}(1 + \mathrm{erf}(\frac{x_{\text{in}_i}}{\sqrt{2}})) + \frac{x_{\text{in}_i}}{\sqrt{2\pi}}\exp{(\frac{-x^2_{\text{in}_i}}{2})})^{2}]\mathbb{E}[g^{2}_{\text{out}_i}]\\
    &= \mathbb{E}[(\frac{1}{2}(1 + \mathrm{erf}(\frac{x_{\text{in}_i}}{\sqrt{2}})) + \frac{x_{\text{in}_i}}{\sqrt{2\pi}}\exp{(\frac{-x^2_{\text{in}_i}}{2})})^{2}]\sigma_{g_{\text{out}}}^2
\end{align*}
\begin{align*}
    I &= \mathbb{E}[(\frac{1}{2}(1 + \mathrm{erf}(\frac{x_{\text{in}_i}}{\sqrt{2}})) + \frac{x_{\text{in}_i}}{\sqrt{2\pi}}\exp{(\frac{-x^2_{\text{in}_i}}{2})})^{2}] \\
    &= \int_{-\infty}^{\infty}(\frac{1}{2}(1 + \mathrm{erf}(\frac{x_{\text{in}_i}}{\sqrt{2}})) + \frac{x_{\text{in}_i}}{\sqrt{2\pi}}\exp{(\frac{-x^2_{\text{in}_i}}{2})})^{2}\frac{\exp{(\frac{-x_{\text{in}_i}^2}{2\sigma_{x_{\text{in}}}^2})}}{\sqrt{2\pi}\sigma_{x_{\text{in}}}}dx_{\text{in}_i} 
\end{align*}
\begin{align*}
    I = \int_{-\infty}^{\infty}(\frac{1}{4}+ \frac{\mathrm{erf}^2(\frac{x_{\text{in}_i}}{\sqrt{2}})}{4} &+ \frac{x_{\text{in}_i}^2\exp{(-x^2_{\text{in}_i})}}{2\pi} + \frac{\mathrm{erf}(\frac{x_{\text{in}_i}}{\sqrt{2}})}{2} + \\
    & \frac{x_{\text{in}_i}\exp{(\frac{-x^2_{\text{in}_i}}{2})}}{\sqrt{2\pi}} + \frac{x_{\text{in}_i}\exp{(\frac{-x^2_{\text{in}_i}}{2})}\mathrm{erf}(\frac{x_{\text{in}_i}}{\sqrt{2}})}{\sqrt{2\pi}})\frac{\exp{(\frac{-x_{\text{in}_i}^2}{2\sigma_{x_{\text{in}}}^2})}}{\sqrt{2\pi}\sigma_{x_{\text{in}}}}dx_{\text{in}_i}
\end{align*}
\begin{align*}
    I_1 &= \int_{-\infty}^{\infty}\frac{1}{4}\frac{\exp{(\frac{-x_{\text{in}_i}^2}{2\sigma_{x_{\text{in}}}^2})}}{\sqrt{2\pi}\sigma_{x_{\text{in}}}}dx_{\text{in}_i}\\
    I_1 &= \frac{1}{4} \\
    I_2 &= \int_{-\infty}^{\infty}\frac{\mathrm{erf}^2(\frac{x_{\text{in}_i}}{\sqrt{2}})}{4}\frac{\exp{(\frac{-x_{\text{in}_i}^2}{2\sigma_{x_{\text{in}}}^2})}}{\sqrt{2\pi}\sigma_{x_{\text{in}}}}dx_{\text{in}_i}\\
    &= \frac{1}{4\sqrt{2\pi}\sigma_{x_{\text{in}}}}\int_{-\infty}^{\infty}\mathrm{erf}^2(\frac{x_{\text{in}_i}}{\sqrt{2}})\exp{(\frac{-x_{\text{in}_i}^2}{2\sigma_{x_{\text{in}}}^2})}dx_{\text{in}_i}
\end{align*}
    From 2.7.1.3 of \citet{integrals_book}, 
    \begin{align*}
    \int_{-\infty}^{\infty}\mathrm{erf}(a_1z)\mathrm{erf}(a_2z)\exp{(-az^2)}dz = \frac{2}{\sqrt{\pi a}}\tan^{-1}{(\frac{a_1a_2}{\sqrt{a^2+aa_1^2+aa_2^2}})}    
    \end{align*}
        
    Substituting $a = \frac{1}{2\sigma_{x_{\text{in}}}^2}, a_1 = a_2 = \frac{1}{\sqrt{2}}$
\begin{align*}
    I_2 &= \frac{1}{4\sqrt{2\pi}\sigma_{x_{\text{in}}}}\frac{2}{\sqrt{\pi  \frac{1}{2\sigma_{x_{\text{in}}}^2}}}\tan^{-1}{(\frac{\frac{1}{2}}{\sqrt{\frac{1}{4\sigma_{x_{\text{in}}}^4} + \frac{1}{4\sigma_{x_{\text{in}}}^2} + \frac{1}{4\sigma_{x_{\text{in}}}^2}}})}\\
    &= \frac{1}{2\pi}\tan^{-1}{(\frac{\sigma_{x_{\text{in}}}^2}{\sqrt{2\sigma_{x_{\text{in}}}^2 + 1}})} = \frac{1}{2\pi}\tan^{-1}{(\frac{\sigma_{x_{\text{in}}}^2}{\sqrt{(\sigma_{x_{\text{in}}}^2 + 1)^2 -  \sigma_{x_{\text{in}}}^4}})}\\
    I_2 &= \frac{1}{2\pi}\sin^{-1}{(\frac{\sigma_{x_{\text{in}}}^2}{\sigma_{x_{\text{in}}}^2 + 1})}\\
    I_3 &= \int_{-\infty}^{\infty}\frac{x_{\text{in}_i}^2\exp{(-x^2_{\text{in}_i})}}{2\pi}\frac{\exp{(\frac{-x_{\text{in}_i}^2}{2\sigma_{x_{\text{in}}}^2})}}{\sqrt{2\pi}\sigma_{x_{\text{in}}}}dx_{\text{in}_i}\\
    &= \frac{1}{2\pi\sigma_{x_{\text{in}}}}\int_{-\infty}^{\infty}\frac{x_{\text{in}_i}^2}{\sqrt{2\pi}}\exp{(\frac{-x_{\text{in}_i}^2(2\sigma_{x_{\text{in}}}^2 + 1)}{2\sigma_{x_{\text{in}}}^2})}dx_{\text{in}_i}\\
    &= \frac{1}{2\pi\sigma_{x_{\text{in}}}}\frac{\sigma_{x_{\text{in}}}}{\sqrt{(2\sigma_{x_{\text{in}}}^2 + 1)}}\int_{-\infty}^{\infty}\frac{x_{\text{in}_i}^2}{\sqrt{2\pi}\frac{\sigma_{x_{\text{in}}}}{\sqrt{(2\sigma_{x_{\text{in}}}^2 + 1)}}}\exp{(\frac{-x_{\text{in}_i}^2(2\sigma_{x_{\text{in}}}^2 + 1)}{2\sigma_{x_{\text{in}}}^2})}dx_{\text{in}_i}\\
    &= \frac{1}{2\pi\sigma_{x_{\text{in}}}}\frac{\sigma_{x_{\text{in}}}}{\sqrt{(2\sigma_{x_{\text{in}}}^2 + 1)}}\frac{\sigma_{x_{\text{in}}}^2}{(2\sigma_{x_{\text{in}}}^2 + 1)} \tag*{(Definition of variance)}\\
    I_3 &= \frac{\sigma^2_{x_{\text{in}}}}{2\pi(2\sigma^2_{x_{\text{in}}} + 1)^{\frac{3}{2}}}\\
    I_4 &= \int_{-\infty}^{\infty}\frac{\mathrm{erf}(\frac{x_{\text{in}_i}}{\sqrt{2}})}{2}\frac{\exp{(\frac{-x_{\text{in}_i}^2}{2\sigma_{x_{\text{in}}}^2})}}{\sqrt{2\pi}\sigma_{x_{\text{in}}}}dx_{\text{in}_i} = 0 \tag*{(Integral of odd function)}\\
    I_5 &= \int_{-\infty}^{\infty}\frac{x_{\text{in}_i}\exp{(\frac{-x^2_{\text{in}_i}}{2})}}{\sqrt{2\pi}}\frac{\exp{(\frac{-x_{\text{in}_i}^2}{2\sigma_{x_{\text{in}}}^2})}}{\sqrt{2\pi}\sigma_{x_{\text{in}}}}dx_{\text{in}_i} = 0 \tag*{(Integral of odd function)}\\
    I_6 &= \int_{-\infty}^{\infty}\frac{x_{\text{in}_i}\exp{(\frac{-x^2_{\text{in}_i}}{2})}\mathrm{erf}(\frac{x_{\text{in}_i}}{\sqrt{2}})}{\sqrt{2\pi}}\frac{\exp{(\frac{-x_{\text{in}_i}^2}{2\sigma_{x_{\text{in}}}^2})}}{\sqrt{2\pi}\sigma_{x_{\text{in}}}}dx_{\text{in}_i}\\
    &= \frac{1}{2\pi\sigma_{x_{\text{in}}}}\int_{-\infty}^{\infty}x_{\text{in}_i}\mathrm{erf}(\frac{x_{\text{in}_i}}{\sqrt{2}})\exp{(\frac{-x_{\text{in}_i}^2(\sigma_{x_{\text{in}}}^2 + 1)}{2\sigma_{x_{\text{in}}}^2})}dx_{\text{in}_i}
    \intertext{From 2.6.1.4 of \citet{integrals_book}, $\int_{-\infty}^{\infty}z\mathrm{erf}(az)\exp{(-a_1z^2)}dz = \frac{a}{a_1\sqrt{a^2+a_1}}$}
    \intertext{Substituting, $a=\frac{1}{\sqrt{2}}, a_1 = \frac{(\sigma_{x_{\text{in}}}^2+1)}{2\sigma_{x_{\text{in}}}^2}$, we have}
    I_6 &= \frac{1}{2\pi\sigma_{x_{\text{in}}}} \frac{\frac{1}{\sqrt{2}}}{\frac{(\sigma_{x_{\text{in}}}^2+1)}{2\sigma_{x_{\text{in}}}^2}\sqrt{\frac{1}{2}+ \frac{(\sigma_{x_{\text{in}}}^2+1)}{2\sigma_{x_{\text{in}}}^2}}}\\
    &= \frac{1}{2\pi\sigma_{x_{\text{in}}}} \frac{2\sigma_{x_{\text{in}}}^3}{(\sigma_{x_{\text{in}}}^2 + 1)\sqrt{2\sigma_{x_{\text{in}}}^2 + 1}}\\
    I_6 &= \frac{\sigma_{x_{\text{in}}}^2}{\pi(\sigma_{x_{\text{in}}}^2 + 1)\sqrt{2\sigma_{x_{\text{in}}}^2 + 1}}\\
    I &= I_1 + I_2 + I_3 + I_4 + I_5 + I_6\\
    &= \frac{1}{4} + \frac{1}{2\pi}\sin^{-1}{(\frac{\sigma_{x_{\text{in}}}^2}{\sigma_{x_{\text{in}}}^2 + 1})} + \frac{\sigma^2_{x_{\text{in}}}}{2\pi(2\sigma^2_{x_{\text{in}}} + 1)^{\frac{3}{2}}} + \frac{\sigma_{x_{\text{in}}}^2}{\pi(\sigma_{x_{\text{in}}}^2 + 1)\sqrt{2\sigma_{x_{\text{in}}}^2 + 1}}\\
    &= \frac{1}{4} + \frac{1}{2\pi}\sin^{-1}{(\frac{\sigma_{x_{\text{in}}}^2}{\sigma_{x_{\text{in}}}^2 + 1})} + \frac{\sigma^2_{x_{\text{in}}}(4\sigma^2_{x_{\text{in}}} + 2 + \sigma^2_{x_{\text{in}}}+1)}{2\pi(\sigma_{x_{\text{in}}}^2 + 1)(2\sigma^2_{x_{\text{in}}} + 1)^{\frac{3}{2}}}\\
    I &= \frac{1}{4} + \frac{1}{2\pi}\sin^{-1}{(\frac{\sigma_{x_{\text{in}}}^2}{\sigma_{x_{\text{in}}}^2 + 1})} + \frac{\sigma^2_{x_{\text{in}}}(5\sigma^2_{x_{\text{in}}} + 3)}{2\pi(\sigma_{x_{\text{in}}}^2 + 1)(2\sigma^2_{x_{\text{in}}} + 1)^{\frac{3}{2}}}
\end{align*}
So the variance of gradient of input of GeLU comes out to be
\begin{align*}
    \mathbb{E}[g^2_{\text{in}_i}] &= I\sigma_{g_{\text{out}}}^2\\
    \Aboxed{\sigma_{g_{\text{in}}}^2 &= \left[\frac{1}{4} + \frac{1}{2\pi}\sin^{-1}{(\frac{\sigma_{x_{\text{in}}}^2}{\sigma_{x_{\text{in}}}^2 + 1})} + \frac{\sigma^2_{x_{\text{in}}}(5\sigma^2_{x_{\text{in}}} + 3)}{2\pi(\sigma_{x_{\text{in}}}^2 + 1)(2\sigma^2_{x_{\text{in}}} + 1)^{\frac{3}{2}}}\right]\sigma_{g_{\text{out}}}^2}
\end{align*}
If for two inputs $\mathbf{x}_{\text{in}}$ and $\mathbf{y}_{\text{in}}$ for all $i$ we have $\mathrm{Corr}(g_{\text{out}_{x_i}},g_{\text{out}_{y_i}}) = r^l_{g_{\text{out}}}$, and $g_{\text{in}_{x_i}}, g_{\text{in}_{y_i}}$ be the gradient after passing through GeLU layer. Then we have,
\begin{align*}
    & \mathbb{E}[g_{\text{in}_{x_i}}g_{\text{in}_{y_i}}] =\\
    & = \mathbb{E}[(\frac{1}{2}(1 + \mathrm{erf}(\frac{x_{\text{in}_i}}{\sqrt{2}})) + \frac{x_{\text{in}_i}}{\sqrt{2\pi}}\exp{(\frac{-x^2_{\text{in}_i}}{2})})g_{\text{out}_{x_i}}(\frac{1}{2}(1 + \mathrm{erf}(\frac{y_{\text{in}_i}}{\sqrt{2}})) + \frac{y_{\text{in}_i}}{\sqrt{2\pi}}\exp{(\frac{-y^2_{\text{in}_i}}{2})})g_{\text{out}_{y_i}}]\\
\end{align*}
\begin{align*}
    \mathbb{E}[g_{\text{in}_{x_i}}g_{\text{in}_{y_i}}] = \mathbb{E}[(\frac{1}{2}&(1 + \mathrm{erf}(\frac{x_{\text{in}_i}}{\sqrt{2}})) + \\
    & \frac{x_{\text{in}_i}}{\sqrt{2\pi}}\exp{(\frac{-x^2_{\text{in}_i}}{2})})(\frac{1}{2}(1 + \mathrm{erf}(\frac{y_{\text{in}_i}}{\sqrt{2}})) + \frac{y_{\text{in}_i}}{\sqrt{2\pi}}\exp{(\frac{-y^2_{\text{in}_i}}{2})})]\mathbb{E}[g_{\text{out}_{x_i}}g_{\text{out}_{y_i}}]\\
    = \mathbb{E}[(\frac{1}{2}&(1 + \mathrm{erf}(\frac{x_{\text{in}_i}}{\sqrt{2}})) +\\
    & \frac{x_{\text{in}_i}}{\sqrt{2\pi}}\exp{(\frac{-x^2_{\text{in}_i}}{2})})(\frac{1}{2}(1 + \mathrm{erf}(\frac{y_{\text{in}_i}}{\sqrt{2}})) + \frac{y_{\text{in}_i}}{\sqrt{2\pi}}\exp{(\frac{-y^2_{\text{in}_i}}{2})})]r^l_{g_{\text{out}}}\sigma_{g_{{\text{out}}}}^2 \\
    I = \mathbb{E}[(\frac{1}{2}&(1 + \mathrm{erf}(\frac{x_{\text{in}_i}}{\sqrt{2}})) + \\
    & \frac{x_{\text{in}_i}}{\sqrt{2\pi}}\exp{(\frac{-x^2_{\text{in}_i}}{2})})(\frac{1}{2}(1 + \mathrm{erf}(\frac{y_{\text{in}_i}}{\sqrt{2}})) + \frac{y_{\text{in}_i}}{\sqrt{2\pi}}\exp{(\frac{-y^2_{\text{in}_i}}{2})})]\\
    = \int_{-\infty}^{\infty}( \frac{1}{2}&(1 + \mathrm{erf}(\frac{x_{\text{in}_i}}{\sqrt{2}})) + \\
    & \frac{x_{\text{in}_i}}{\sqrt{2\pi}}\exp{(\frac{-x^2_{\text{in}_i}}{2})})(\frac{1}{2}(1 + \mathrm{erf}(\frac{y_{\text{in}_i}}{\sqrt{2}})) + \frac{y_{\text{in}_i}}{\sqrt{2\pi}}\exp{(\frac{-y^2_{\text{in}_i}}{2})})p_{x_{\text{in}_i},y_{\text{in}_i}}dx_{\text{in}_i}dy_{\text{in}_i}
\end{align*}
Where $p_{x_{\text{in}_i},y_{\text{in}_i}} = \displaystyle\frac{1}{2\pi\sigma_{x_{\text{in}}}^2\sqrt{(1- (r^l_{x_{\text{in}}})^2)}}\exp{(\frac{-x^2_{\text{in}_i} + 2r^l_{x_{\text{in}}}x_{\text{in}_i}y_{\text{in}_i} - y^2_{\text{in}_i}}{2\sigma_{x_{\text{in}}}^2(1- (r^l_{x_{\text{in}}})^2)})}$
\begin{align*}
    I &= \int_{-\infty}^{\infty}\frac{(\frac{1}{2}(1 + \mathrm{erf}(\frac{y_{\text{in}_i}}{\sqrt{2}})) + \frac{y_{\text{in}_i}}{\sqrt{2\pi}}\exp{(\frac{-y^2_{\text{in}_i}}{2})})}{2\pi\sigma_{x_{\text{in}}}^2\sqrt{(1- (r^l_{x_{\text{in}}})^2)}}\exp{(\frac{ - y^2_{\text{in}_i}}{2\sigma_{x_{\text{in}}}^2(1- (r^l_{x_{\text{in}}})^2)})}I_Xdy_{\text{in}_i}
\end{align*}
    Where,
\begin{align*}
    I_X &= \int_{-\infty}^{\infty}(\frac{1}{2}(1 + \mathrm{erf}(\frac{x_{\text{in}_i}}{\sqrt{2}})) + \frac{x_{\text{in}_i}}{\sqrt{2\pi}}\exp{(\frac{-x^2_{\text{in}_i}}{2})})\exp{(\frac{-x^2_{\text{in}_i} + 2r^l_{x_{\text{in}}}x_{\text{in}_i}y_{\text{in}_i}}{2\sigma_{x_{\text{in}}}^2(1- (r^l_{x_{\text{in}}})^2)})}dx_{\text{in}_i}\\
    I_{X,1} &= \int_{-\infty}^{\infty}\frac{1}{2}\exp{(\frac{-x^2_{\text{in}_i} + 2r^l_{x_{\text{in}}}x_{\text{in}_i}y_{\text{in}_i}}{2\sigma_{x_{\text{in}}}^2(1- (r^l_{x_{\text{in}}})^2)})}dx_{\text{in}_i}\\
    &= \frac{1}{2}\int_{-\infty}^{\infty}\exp{(\frac{-x^2_{\text{in}_i} + 2r^l_{x_{\text{in}}}x_{\text{in}_i}y_{\text{in}_i}}{2\sigma_{x_{\text{in}}}^2(1- (r^l_{x_{\text{in}}})^2)})}\exp{(\frac{-(r^l_{x_{\text{in}}})^2y^2_{\text{in}_i}}{2\sigma_{x_{\text{in}}}^2(1- (r^l_{x_{\text{in}}})^2)})}\exp{(\frac{(r^l_{x_{\text{in}}})^2y^2_{\text{in}_i}}{2\sigma_{x_{\text{in}}}^2(1- (r^l_{x_{\text{in}}})^2)})}dx_{\text{in}_i}\\
    &= \frac{1}{2}\exp{(\frac{(r^l_{x_{\text{in}}})^2y^2_{\text{in}_i}}{2\sigma_{x_{\text{in}}}^2(1- (r^l_{x_{\text{in}}})^2)})}\int_{-\infty}^{\infty}\exp{(\frac{-(x_{\text{in}_i} - r^l_{x_{\text{in}}}y_{\text{in}_i})^2}{2\sigma_{x_{\text{in}}}^2(1- (r^l_{x_{\text{in}}})^2)})}dx_{\text{in}_i}\\
    &= \frac{1}{2}\exp{(\frac{(r^l_{x_{\text{in}}})^2y^2_{\text{in}_i}}{2\sigma_{x_{\text{in}}}^2(1- (r^l_{x_{\text{in}}})^2)})}\sqrt{2\pi}\sigma_{x_{\text{in}}}\sqrt{(1- (r^l_{x_{\text{in}}})^2)}\int_{-\infty}^{\infty}\frac{\exp{(\frac{-(x_{\text{in}_i} - r^l_{x_{\text{in}}}y_{\text{in}_i})^2}{2\sigma_{x_{\text{in}}}^2(1- (r^l_{x_{\text{in}}})^2)})}}{\sqrt{2\pi}\sigma_{x_{\text{in}}}\sqrt{(1- (r^l_{x_{\text{in}}})^2)}}dx_{\text{in}_i}\\
    I_{X,1} &= \frac{\sqrt{2\pi}\sigma_{x_{\text{in}}}\sqrt{(1- (r^l_{x_{\text{in}}})^2)}}{2}\exp{(\frac{(r^l_{x_{\text{in}}})^2y^2_{\text{in}_i}}{2\sigma_{x_{\text{in}}}^2(1- (r^l_{x_{\text{in}}})^2)})}\\
    I_{X,2} &= \int_{-\infty}^{\infty}\frac{\mathrm{erf}(\frac{x_{\text{in}_i}}{\sqrt{2}})}{2}\exp{(\frac{-x^2_{\text{in}_i} + 2r^l_{x_{\text{in}}}x_{\text{in}_i}y_{\text{in}_i}}{2\sigma_{x_{\text{in}}}^2(1- (r^l_{x_{\text{in}}})^2)})}dx_{\text{in}_i}\\
    &= \frac{1}{2}\int_{-\infty}^{\infty}\mathrm{erf}(\frac{x_{\text{in}_i}}{\sqrt{2}})\exp{(\frac{-x^2_{\text{in}_i} + 2r^l_{x_{\text{in}}}x_{\text{in}_i}y_{\text{in}_i}}{2\sigma_{x_{\text{in}}}^2(1- (r^l_{x_{\text{in}}})^2)})}dx_{\text{in}_i}
\end{align*}
    From 2.7.1.6 of \citet{integrals_book}, 
    \begin{align*}
        \int_{-\infty}^{\infty}\mathrm{erf}(a_1z)\exp{(-az^2+bz)}dz = \sqrt{\frac{\pi}{a}}\exp{(\frac{b^2}{4a})}\mathrm{erf}(\frac{a_1b}{2\sqrt{a^2+aa_1^2}})
    \end{align*}
    
    Substituting $a_1 = \frac{1}{\sqrt{2}}, a = \frac{1}{2\sigma_{x_{\text{in}}}^2(1- (r^l_{x_{\text{in}}})^2)}, b = \frac{r^l_{x_{\text{in}}}y_{\text{in}_i}}{\sigma_{x_{\text{in}}}^2(1- (r^l_{x_{\text{in}}})^2)}$
\begin{align*}
    I_{X,2} &= \frac{1}{2}\sqrt{\frac{\pi}{\frac{1}{2\sigma_{x_{\text{in}}}^2(1- (r^l_{x_{\text{in}}})^2)}}}\exp{(\frac{\frac{(r^l_{x_{\text{in}}})^2y^2_{\text{in}_i}}{\sigma_{x_{\text{in}}}^4(1- (r^l_{x_{\text{in}}})^2)^2}}{4\frac{1}{2\sigma_{x_{\text{in}}}^2(1- (r^l_{x_{\text{in}}})^2)}})}\mathrm{erf}(\frac{\frac{r^l_{x_{\text{in}}}y_{\text{in}_i}}{\sqrt{2}\sigma_{x_{\text{in}}}^2(1- (r^l_{x_{\text{in}}})^2)}}{2\sqrt{\frac{1}{4\sigma_{x_{\text{in}}}^4(1- (r^l_{x_{\text{in}}})^2)^2} + \frac{1}{4\sigma_{x_{\text{in}}}^2(1- (r^l_{x_{\text{in}}})^2)}}})\\
    I_{X,2} &=  \frac{\sqrt{2\pi}\sigma_{x_{\text{in}}}\sqrt{(1- (r^l_{x_{\text{in}}})^2)}}{2}\exp{(\frac{(r^l_{x_{\text{in}}})^2y^2_{\text{in}_i}}{2\sigma_{x_{\text{in}}}^2(1- (r^l_{x_{\text{in}}})^2)})}\mathrm{erf}(\frac{r^l_{x_{\text{in}}}y_{\text{in}_i}}{\sqrt{2(\sigma_{x_{\text{in}}}^2(1- (r^l_{x_{\text{in}}})^2) + 1)}})\\
    I_{X,3} &= \int_{-\infty}^{\infty}\frac{x_{\text{in}_i}}{\sqrt{2\pi}}\exp{(\frac{-x^2_{\text{in}_i}}{2})}\exp{(\frac{-x^2_{\text{in}_i} + 2r^l_{x_{\text{in}}}x_{\text{in}_i}y_{\text{in}_i}}{2\sigma_{x_{\text{in}}}^2(1- (r^l_{x_{\text{in}}})^2)})}dx_{\text{in}_i}\\
    &= \int_{-\infty}^{\infty}\frac{x_{\text{in}_i}}{\sqrt{2\pi}}\exp{(\frac{-x^2_{\text{in}_i}(\sigma_{x_{\text{in}}}^2(1- (r^l_{x_{\text{in}}})^2) + 1) + 2r^l_{x_{\text{in}}}x_{\text{in}_i}y_{\text{in}_i}}{2\sigma_{x_{\text{in}}}^2(1- (r^l_{x_{\text{in}}})^2)})}dx_{\text{in}_i}\\
    &= \int_{-\infty}^{\infty}\frac{x_{\text{in}_i}}{\sqrt{2\pi}}\exp{(\frac{-x^2_{\text{in}_i} + \frac{2r^l_{x_{\text{in}}}x_{\text{in}_i}y_{\text{in}_i}}{(\sigma_{x_{\text{in}}}^2(1- (r^l_{x_{\text{in}}})^2) + 1)}}{\frac{2\sigma_{x_{\text{in}}}^2(1- (r^l_{x_{\text{in}}})^2)}{(\sigma_{x_{\text{in}}}^2(1- (r^l_{x_{\text{in}}})^2) + 1)}})}dx_{\text{in}_i}
\end{align*}
\begin{align*}
    = \int_{-\infty}^{\infty}\frac{x_{\text{in}_i}}{\sqrt{2\pi}}\exp{(\frac{-x^2_{\text{in}_i} + \frac{2r^l_{x_{\text{in}}}x_{\text{in}_i}y_{\text{in}_i}}{(\sigma_{x_{\text{in}}}^2(1- (r^l_{x_{\text{in}}})^2) + 1)}}{\frac{2\sigma_{x_{\text{in}}}^2(1- (r^l_{x_{\text{in}}})^2)}{(\sigma_{x_{\text{in}}}^2(1- (r^l_{x_{\text{in}}})^2) + 1)}})}&\exp{(\frac{\frac{-(r^l_{x_{\text{in}}})^2y^2_{\text{in}_i}}{(\sigma_{x_{\text{in}}}^2(1- (r^l_{x_{\text{in}}})^2) + 1)^2}}{\frac{2\sigma_{x_{\text{in}}}^2(1- (r^l_{x_{\text{in}}})^2)}{(\sigma_{x_{\text{in}}}^2(1- (r^l_{x_{\text{in}}})^2) + 1)}})}*\\ & \hspace{50pt} \exp{(\frac{\frac{(r^l_{x_{\text{in}}})^2y^2_{\text{in}_i}}{(\sigma_{x_{\text{in}}}^2(1- (r^l_{x_{\text{in}}})^2) + 1)^2}}{\frac{2\sigma_{x_{\text{in}}}^2(1- (r^l_{x_{\text{in}}})^2)}{(\sigma_{x_{\text{in}}}^2(1- (r^l_{x_{\text{in}}})^2) + 1)}})}dx_{\text{in}_i}
\end{align*}
\begin{align*}
    &=\exp{(\frac{(r^l_{x_{\text{in}}})^2y^2_{\text{in}_i}}{2\sigma_{x_{\text{in}}}^2(1- (r^l_{x_{\text{in}}})^2)(\sigma_{x_{\text{in}}}^2(1- (r^l_{x_{\text{in}}})^2) + 1)})} *\\
    & \hspace{100pt} \int_{-\infty}^{\infty}\frac{x_{\text{in}_i}}{\sqrt{2\pi}}\exp{(\frac{-(x_{\text{in}_i} - \frac{r^l_{x_{\text{in}}}y_{\text{in}_i}}{(\sigma_{x_{\text{in}}}^2(1- (r^l_{x_{\text{in}}})^2) + 1)})^2}{\frac{2\sigma_{x_{\text{in}}}^2(1- (r^l_{x_{\text{in}}})^2)}{(\sigma_{x_{\text{in}}}^2(1- (r^l_{x_{\text{in}}})^2) + 1)}})}dx_{\text{in}_i}\\
    &=\exp{(\frac{(r^l_{x_{\text{in}}})^2y^2_{\text{in}_i}}{2\sigma_{x_{\text{in}}}^2(1- (r^l_{x_{\text{in}}})^2)(\sigma_{x_{\text{in}}}^2(1- (r^l_{x_{\text{in}}})^2) + 1)})}\frac{\sigma_{x_{\text{in}}}\sqrt{1 - (r^l_{x_{\text{in}}})^2}}{\sqrt{(\sigma_{x_{\text{in}}}^2(1- (r^l_{x_{\text{in}}})^2) + 1)}}\\
    & \hspace{100pt} \int_{-\infty}^{\infty}\frac{x_{\text{in}_i}}{\sqrt{2\pi}\frac{\sigma_{x_{\text{in}}}\sqrt{1 - (r^l_{x_{\text{in}}})^2}}{\sqrt{(\sigma_{x_{\text{in}}}^2(1- (r^l_{x_{\text{in}}})^2) + 1)}}}\exp{(\frac{-(x_{\text{in}_i} - \frac{r^l_{x_{\text{in}}}y_{\text{in}_i}}{(\sigma_{x_{\text{in}}}^2(1- (r^l_{x_{\text{in}}})^2) + 1)})^2}{\frac{2\sigma_{x_{\text{in}}}^2(1- (r^l_{x_{\text{in}}})^2)}{(\sigma_{x_{\text{in}}}^2(1- (r^l_{x_{\text{in}}})^2) + 1)}})}dx_{\text{in}_i}\\
    &= \exp{(\frac{(r^l_{x_{\text{in}}})^2y^2_{\text{in}_i}}{2\sigma_{x_{\text{in}}}^2(1- (r^l_{x_{\text{in}}})^2)(\sigma_{x_{\text{in}}}^2(1- (r^l_{x_{\text{in}}})^2) + 1)})}.\\
    & \hspace{170pt} \frac{\sigma_{x_{\text{in}}}\sqrt{1 - (r^l_{x_{\text{in}}})^2}}{\sqrt{(\sigma_{x_{\text{in}}}^2(1- (r^l_{x_{\text{in}}})^2) + 1)}}\frac{r^l_{x_{\text{in}}}y_{\text{in}_i}}{(\sigma_{x_{\text{in}}}^2(1- (r^l_{x_{\text{in}}})^2) + 1)}
\end{align*}
\begin{align*}
    I_{X,3} = \frac{r^l_{x_{\text{in}}}y_{\text{in}_i}\sigma_{x_{\text{in}}}\sqrt{1 - (r^l_{x_{\text{in}}})^2}}{(\sigma_{x_{\text{in}}}^2(1- (r^l_{x_{\text{in}}})^2) + 1)^{\frac{3}{2}}}\exp{(\frac{(r^l_{x_{\text{in}}})^2y^2_{\text{in}_i}}{2\sigma_{x_{\text{in}}}^2(1- (r^l_{x_{\text{in}}})^2)(\sigma_{x_{\text{in}}}^2(1- (r^l_{x_{\text{in}}})^2) + 1)})}
\end{align*}
\begin{align*}
    &I = \\
    & \int_{-\infty}^{\infty}\frac{(\frac{1}{2}(1 + \mathrm{erf}(\frac{y_{\text{in}_i}}{\sqrt{2}})) + \frac{y_{\text{in}_i}}{\sqrt{2\pi}}\exp{(\frac{-y^2_{\text{in}_i}}{2})})}{2\pi\sigma_{x_{\text{in}}}^2\sqrt{(1- (r^l_{x_{\text{in}}})^2)}}\exp{(\frac{ - y^2_{\text{in}_i}}{2\sigma_{x_{\text{in}}}^2(1- (r^l_{x_{\text{in}}})^2)})}(I_{X,1}+I_{X,2}+I_{X,3})dy_{\text{in}_i}
\end{align*}
\begin{align*}
    I_1 = \int_{-\infty}^{\infty}\frac{(\frac{1}{2}(1 + \mathrm{erf}(\frac{y_{\text{in}_i}}{\sqrt{2}})) + \frac{y_{\text{in}_i}}{\sqrt{2\pi}}\exp{(\frac{-y^2_{\text{in}_i}}{2})})}{2\pi\sigma_{x_{\text{in}}}^2\sqrt{(1- (r^l_{x_{\text{in}}})^2)}}\exp{(\frac{ - y^2_{\text{in}_i}}{2\sigma_{x_{\text{in}}}^2(1- (r^l_{x_{\text{in}}})^2)})}I_{X,1}dy_{\text{in}_i}\\
\end{align*}
\begin{align*}
    &= \int_{-\infty}^{\infty}\frac{(\frac{1}{2}(1 + \mathrm{erf}(\frac{y_{\text{in}_i}}{\sqrt{2}})) + \frac{y_{\text{in}_i}}{\sqrt{2\pi}}\exp{(\frac{-y^2_{\text{in}_i}}{2})})}{2\pi\sigma_{x_{\text{in}}}^2\sqrt{(1- (r^l_{x_{\text{in}}})^2)}}\exp{(\frac{ - y^2_{\text{in}_i}}{2\sigma_{x_{\text{in}}}^2(1- (r^l_{x_{\text{in}}})^2)})}\\
    &\hspace{20pt}\frac{\sqrt{2\pi}\sigma_{x_{\text{in}}}\sqrt{(1- (r^l_{x_{\text{in}}})^2)}}{2}\exp{(\frac{(r^l_{x_{\text{in}}})^2y^2_{\text{in}_i}}{2\sigma_{x_{\text{in}}}^2(1- (r^l_{x_{\text{in}}})^2)})}dy_{\text{in}_i}\\
    &= \frac{1}{2}\int_{-\infty}^{\infty}\frac{(\frac{1}{2}(1 + \mathrm{erf}(\frac{y_{\text{in}_i}}{\sqrt{2}})) + \frac{y_{\text{in}_i}}{\sqrt{2\pi}}\exp{(\frac{-y^2_{\text{in}_i}}{2})})}{\sqrt{2\pi}\sigma_{x_{\text{in}}}}\exp{(\frac{ - y^2_{\text{in}_i}}{2\sigma_{x_{\text{in}}}^2})}dy_{\text{in}_i}\\
    I_{1,1} &= \frac{1}{4}\int_{-\infty}^{\infty}\frac{1}{\sqrt{2\pi}\sigma_{x_{\text{in}}}}\exp{(\frac{ - y^2_{\text{in}_i}}{2\sigma_{x_{\text{in}}}^2})}dy_{\text{in}_i} = \frac{1}{4}\\
    I_{1,2} &= \frac{1}{4}\int_{-\infty}^{\infty}\frac{\mathrm{erf}(\frac{y_{\text{in}_i}}{\sqrt{2}})}{\sqrt{2\pi}\sigma_{x_{\text{in}}}}\exp{(\frac{ - y^2_{\text{in}_i}}{2\sigma_{x_{\text{in}}}^2})}dy_{\text{in}_i} = 0 \tag*{(Integral of odd function)}\\
    I_{1,3} &= \frac{1}{2}\int_{-\infty}^{\infty}\frac{y_{\text{in}_i}\exp{(\frac{-y^2_{\text{in}_i}}{2})}}{
    2\pi\sigma_{x_{\text{in}}}}\exp{(\frac{ - y^2_{\text{in}_i}}{2\sigma_{x_{\text{in}}}^2})}dy_{\text{in}_i} = 0 \tag*{(Integral of odd function)}\\
    I_2 &= \int_{-\infty}^{\infty}\frac{(\frac{1}{2}(1 + \mathrm{erf}(\frac{y_{\text{in}_i}}{\sqrt{2}})) + \frac{y_{\text{in}_i}}{\sqrt{2\pi}}\exp{(\frac{-y^2_{\text{in}_i}}{2})})}{2\pi\sigma_{x_{\text{in}}}^2\sqrt{(1- (r^l_{x_{\text{in}}})^2)}}\exp{(\frac{ - y^2_{\text{in}_i}}{2\sigma_{x_{\text{in}}}^2(1- (r^l_{x_{\text{in}}})^2)})}I_{X,2}dy_{\text{in}_i}\\
    &= \int_{-\infty}^{\infty}\frac{(\frac{1}{2}(1 + \mathrm{erf}(\frac{y_{\text{in}_i}}{\sqrt{2}})) + \frac{y_{\text{in}_i}}{\sqrt{2\pi}}\exp{(\frac{-y^2_{\text{in}_i}}{2})})}{2\pi\sigma_{x_{\text{in}}}^2\sqrt{(1- (r^l_{x_{\text{in}}})^2)}}\exp{(\frac{ - y^2_{\text{in}_i}}{2\sigma_{x_{\text{in}}}^2(1- (r^l_{x_{\text{in}}})^2)})}\\
    &\hspace{20pt}\frac{\sqrt{2\pi}\sigma_{x_{\text{in}}}\sqrt{(1- (r^l_{x_{\text{in}}})^2)}}{2}\exp{(\frac{(r^l_{x_{\text{in}}})^2y^2_{\text{in}_i}}{2\sigma_{x_{\text{in}}}^2(1- (r^l_{x_{\text{in}}})^2)})}\mathrm{erf}(\frac{r^l_{x_{\text{in}}}y_{\text{in}_i}}{\sqrt{2(\sigma_{x_{\text{in}}}^2(1- (r^l_{x_{\text{in}}})^2) + 1)}})dy_{\text{in}_i}\\
    &= \frac{1}{2}\int_{-\infty}^{\infty}\frac{(\frac{1}{2}(1 + \mathrm{erf}(\frac{y_{\text{in}_i}}{\sqrt{2}})) + \frac{y_{\text{in}_i}}{\sqrt{2\pi}}\exp{(\frac{-y^2_{\text{in}_i}}{2})})}{\sqrt{2\pi}\sigma_{x_{\text{in}}}}\exp{(\frac{ - y^2_{\text{in}_i}}{2\sigma_{x_{\text{in}}}^2})}. \\
    & \hspace{150pt} \mathrm{erf}(\frac{r^l_{x_{\text{in}}}y_{\text{in}_i}}{\sqrt{2(\sigma_{x_{\text{in}}}^2(1- (r^l_{x_{\text{in}}})^2) + 1)}})dy_{\text{in}_i}\\
    I_{2,1} &= \frac{1}{4}\int_{-\infty}^{\infty}\frac{1}{\sqrt{2\pi}\sigma_{x_{\text{in}}}}\exp{(\frac{ - y^2_{\text{in}_i}}{2\sigma_{x_{\text{in}}}^2})}\mathrm{erf}(\frac{r^l_{x_{\text{in}}}y_{\text{in}_i}}{\sqrt{2(\sigma_{x_{\text{in}}}^2(1- (r^l_{x_{\text{in}}})^2) + 1)}})dy_{\text{in}_i} = 0 \tag*{(Integral of odd function)}\\
    I_{2,2} &= \frac{1}{4\sqrt{2\pi}\sigma_{x_{\text{in}}}}\int_{-\infty}^{\infty}\mathrm{erf}(\frac{y_{\text{in}_i}}{\sqrt{2}})\exp{(\frac{ - y^2_{\text{in}_i}}{2\sigma_{x_{\text{in}}}^2})}\mathrm{erf}(\frac{r^l_{x_{\text{in}}}y_{\text{in}_i}}{\sqrt{2(\sigma_{x_{\text{in}}}^2(1- (r^l_{x_{\text{in}}})^2) + 1)}})dy_{\text{in}_i}\\
    \intertext{From 2.7.1.3 of \citet{integrals_book},}
    \intertext{$\int_{-\infty}^{\infty}\mathrm{erf}(a_1z)\mathrm{erf}(a_2z)\exp{(-az^2)}dz = \frac{2}{\sqrt{\pi a}}\tan^{-1}{(\frac{a_1a_2}{\sqrt{a^2+aa_1^2+aa_2^2}})}$}
    \intertext{Substituting $a = \frac{1}{2\sigma_{x_{\text{in}}}^2}, a_1 = \frac{1}{\sqrt{2}}, a_2 = \frac{r^l_{x_{\text{in}}}}{\sqrt{2(\sigma_{x_{\text{in}}}^2(1- (r^l_{x_{\text{in}}})^2) + 1)}}$}
    I_{2,2} &= \frac{1}{4\sqrt{2\pi}\sigma_{x_{\text{in}}}}\frac{2}{\sqrt{\pi\frac{1}{2\sigma_{x_{\text{in}}}^2}}}\tan^{-1}{(\frac{\frac{r^l_{x_{\text{in}}}}{2\sqrt{(\sigma_{x_{\text{in}}}^2(1- (r^l_{x_{\text{in}}})^2) + 1)}}}{\sqrt{\frac{1}{4\sigma_{x_{\text{in}}}^4}+\frac{1}{4\sigma_{x_{\text{in}}}^2}+\frac{(r^l_{x_{\text{in}}})^2}{4\sigma_{x_{\text{in}}}^2(\sigma_{x_{\text{in}}}^2(1- (r^l_{x_{\text{in}}})^2) + 1)}}})}\\
    I_{2,2} &= \frac{1}{2\pi}\tan^{-1}{(\frac{r^l_{x_{\text{in}}}\sigma_{x_{\text{in}}}^2}{\sqrt{\sigma_{x_{\text{in}}}^4 + 2\sigma_{x_{\text{in}}}^2 + 1 - (r^l_{x_{\text{in}}})^2\sigma_{x_{\text{in}}}^4}})} = \frac{1}{2\pi}\tan^{-1}{(\frac{r^l_{x_{\text{in}}}\sigma_{x_{\text{in}}}^2}{\sqrt{(\sigma_{x_{\text{in}}}^2 + 1)^2 - (r^l_{x_{\text{in}}}\sigma_{x_{\text{in}}}^2)^2}})}\\
    I_{2,2} &= \frac{1}{2\pi}\sin^{-1}{(\frac{r^l_{x_{\text{in}}}\sigma_{x_{\text{in}}}^2}{\sigma_{x_{\text{in}}}^2 + 1})}\\
    I_{2,3} &= \frac{1}{4\pi\sigma_{x_{\text{in}}}}\int_{-\infty}^{\infty}y_{\text{in}_i}\exp{(\frac{-y^2_{\text{in}_i}}{2})}
    \exp{(\frac{ - y^2_{\text{in}_i}}{2\sigma_{x_{\text{in}}}^2})}\mathrm{erf}(\frac{r^l_{x_{\text{in}}}y_{\text{in}_i}}{\sqrt{2(\sigma_{x_{\text{in}}}^2(1- (r^l_{x_{\text{in}}})^2) + 1)}})dy_{\text{in}_i}\\
    &= \frac{1}{4\pi\sigma_{x_{\text{in}}}}\int_{-\infty}^{\infty}y_{\text{in}_i}
    \exp{(\frac{ - y^2_{\text{in}_i}(\sigma_{x_{\text{in}}}^2 + 1)}{2\sigma_{x_{\text{in}}}^2})}\mathrm{erf}(\frac{r^l_{x_{\text{in}}}y_{\text{in}_i}}{\sqrt{2(\sigma_{x_{\text{in}}}^2(1- (r^l_{x_{\text{in}}})^2) + 1)}})dy_{\text{in}_i}
\end{align*}
    From 2.6.1.4 of \citet{integrals_book}, 
    $\int_{-\infty}^{\infty}z\mathrm{erf}(az)\exp{(-a_1z^2)}dz = \frac{a}{a_1\sqrt{a^2+a_1}}$
    
    Substituting, $a=\frac{r^l_{x_{\text{in}}}}{\sqrt{2(\sigma_{x_{\text{in}}}^2(1- (r^l_{x_{\text{in}}})^2) + 1)}}, a_1 = \frac{(\sigma_{x_{\text{in}}}^2+1)}{2\sigma_{x_{\text{in}}}^2}$, we have
\begin{align*}
    I_{2,3} &= \frac{1}{4\pi\sigma_{x_{\text{in}}}}\frac{\frac{r^l_{x_{\text{in}}}}{\sqrt{2(\sigma_{x_{\text{in}}}^2(1- (r^l_{x_{\text{in}}})^2) + 1)}}}{\frac{(\sigma_{x_{\text{in}}}^2+1)}{2\sigma_{x_{\text{in}}}^2}\sqrt{\frac{(r^l_{x_{\text{in}}})^2}{2(\sigma_{x_{\text{in}}}^2(1- (r^l_{x_{\text{in}}})^2) + 1)}+\frac{(\sigma_{x_{\text{in}}}^2+1)}{2\sigma_{x_{\text{in}}}^2}}}\\
    &= \frac{r^l_{x_{\text{in}}}\sigma_{x_{\text{in}}}^2}{2\pi(\sigma_{x_{\text{in}}}^2+1)\sqrt{\sigma_{x_{\text{in}}}^4 + 2\sigma_{x_{\text{in}}}^2 + 1 - (r^l_{x_{\text{in}}})^2\sigma_{x_{\text{in}}}^4}}\\
    I_{2,3} &= \frac{r^l_{x_{\text{in}}}\sigma_{x_{\text{in}}}^2}{2\pi(\sigma_{x_{\text{in}}}^2+1)\sqrt{(\sigma_{x_{\text{in}}}^2 + 1)^2 - (r^l_{x_{\text{in}}}\sigma_{x_{\text{in}}}^2)^2}}\\
    I_3 &= \int_{-\infty}^{\infty}\frac{(\frac{1}{2}(1 + \mathrm{erf}(\frac{y_{\text{in}_i}}{\sqrt{2}})) + \frac{y_{\text{in}_i}}{\sqrt{2\pi}}\exp{(\frac{-y^2_{\text{in}_i}}{2})})}{2\pi\sigma_{x_{\text{in}}}^2\sqrt{(1- (r^l_{x_{\text{in}}})^2)}}\exp{(\frac{ - y^2_{\text{in}_i}}{2\sigma_{x_{\text{in}}}^2(1- (r^l_{x_{\text{in}}})^2)})}I_{X,3}dy_{\text{in}_i}\\
    &= \int_{-\infty}^{\infty}\frac{(\frac{1}{2}(1 + \mathrm{erf}(\frac{y_{\text{in}_i}}{\sqrt{2}})) + \frac{y_{\text{in}_i}}{\sqrt{2\pi}}\exp{(\frac{-y^2_{\text{in}_i}}{2})})}{2\pi\sigma_{x_{\text{in}}}^2\sqrt{(1- (r^l_{x_{\text{in}}})^2)}}\exp{(\frac{ - y^2_{\text{in}_i}}{2\sigma_{x_{\text{in}}}^2(1- (r^l_{x_{\text{in}}})^2)})}\\
    &\hspace{20pt}\frac{r^l_{x_{\text{in}}}y_{\text{in}_i}\sigma_{x_{\text{in}}}\sqrt{1 - (r^l_{x_{\text{in}}})^2}}{(\sigma_{x_{\text{in}}}^2(1- (r^l_{x_{\text{in}}})^2) + 1)^{\frac{3}{2}}}\exp{(\frac{(r^l_{x_{\text{in}}})^2y^2_{\text{in}_i}}{2\sigma_{x_{\text{in}}}^2(1- (r^l_{x_{\text{in}}})^2)(\sigma_{x_{\text{in}}}^2(1- (r^l_{x_{\text{in}}})^2) + 1)})}dy_{\text{in}_i}\\
    &= \frac{r^l_{x_{\text{in}}}}{2\pi\sigma_{x_{\text{in}}}(\sigma_{x_{\text{in}}}^2(1- (r^l_{x_{\text{in}}})^2) + 1)^{\frac{3}{2}}}\int_{-\infty}^{\infty}y_{\text{in}_i}(\frac{1}{2}(1 + \mathrm{erf}(\frac{y_{\text{in}_i}}{\sqrt{2}})) + \frac{y_{\text{in}_i}}{\sqrt{2\pi}}\exp{(\frac{-y^2_{\text{in}_i}}{2})})\\
    &\hspace{20pt}\exp{(\frac{ - y^2_{\text{in}_i}}{2\sigma_{x_{\text{in}}}^2(1- (r^l_{x_{\text{in}}})^2)})}\exp{(\frac{(r^l_{x_{\text{in}}})^2y^2_{\text{in}_i}}{2\sigma_{x_{\text{in}}}^2(1- (r^l_{x_{\text{in}}})^2)(\sigma_{x_{\text{in}}}^2(1- (r^l_{x_{\text{in}}})^2) + 1)})}dy_{\text{in}_i}\\
    &= \frac{r^l_{x_{\text{in}}}}{2\pi\sigma_{x_{\text{in}}}(\sigma_{x_{\text{in}}}^2(1- (r^l_{x_{\text{in}}})^2) + 1)^{\frac{3}{2}}}\int_{-\infty}^{\infty}y_{\text{in}_i}(\frac{1}{2}(1 + \mathrm{erf}(\frac{y_{\text{in}_i}}{\sqrt{2}})) + \frac{y_{\text{in}_i}}{\sqrt{2\pi}}\exp{(\frac{-y^2_{\text{in}_i}}{2})})\\
    &\hspace{20pt}\exp{(\frac{-y^2_{\text{in}_i}(\sigma_{x_{\text{in}}}^2(1-(r^l_{x_{\text{in}}})^2) + 1 - (r^l_{x_{\text{in}}})^2)}{2(\sigma_{x_{\text{in}}}^2(1-(r^l_{x_{\text{in}}})^2) + 1)\sigma_{x_{\text{in}}}^2(1-(r^l_{x_{\text{in}}})^2)})}dy_{\text{in}_i}\\
    &= \frac{r^l_{x_{\text{in}}}}{2\pi\sigma_{x_{\text{in}}}(\sigma_{x_{\text{in}}}^2(1- (r^l_{x_{\text{in}}})^2) + 1)^{\frac{3}{2}}}\int_{-\infty}^{\infty}y_{\text{in}_i}(\frac{1}{2}(1 + \mathrm{erf}(\frac{y_{\text{in}_i}}{\sqrt{2}})) + \frac{y_{\text{in}_i}}{\sqrt{2\pi}}\exp{(\frac{-y^2_{\text{in}_i}}{2})})\\
    &\hspace{20pt}\exp{(\frac{-y^2_{\text{in}_i}(\sigma_{x_{\text{in}}}^2 + 1)}{2\sigma_{x_{\text{in}}}^2(\sigma_{x_{\text{in}}}^2(1-(r^l_{x_{\text{in}}})^2) + 1)})}dy_{\text{in}_i}\\
    I_{3,1} &= \frac{r^l_{x_{\text{in}}}}{4\pi\sigma_{x_{\text{in}}}(\sigma_{x_{\text{in}}}^2(1- (r^l_{x_{\text{in}}})^2) + 1)^{\frac{3}{2}}}\int_{-\infty}^{\infty}y_{\text{in}_i}\exp{(\frac{-y^2_{\text{in}_i}(\sigma_{x_{\text{in}}}^2 + 1)}{2\sigma_{x_{\text{in}}}^2(\sigma_{x_{\text{in}}}^2(1-(r^l_{x_{\text{in}}})^2) + 1)})}dy_{\text{in}_i} = 0 \tag*{(Integral of odd function)}\\
    I_{3,2} &= \frac{r^l_{x_{\text{in}}}}{4\pi\sigma_{x_{\text{in}}}(\sigma_{x_{\text{in}}}^2(1- (r^l_{x_{\text{in}}})^2) + 1)^{\frac{3}{2}}}\int_{-\infty}^{\infty}y_{\text{in}_i}\mathrm{erf}(\frac{y_{\text{in}_i}}{\sqrt{2}})\exp{(\frac{-y^2_{\text{in}_i}(\sigma_{x_{\text{in}}}^2 + 1)}{2\sigma_{x_{\text{in}}}^2(\sigma_{x_{\text{in}}}^2(1-(r^l_{x_{\text{in}}})^2) + 1)})}dy_{\text{in}_i}
\end{align*}
    From 2.6.1.4 of \citet{integrals_book}, 
    $\int_{-\infty}^{\infty}z\mathrm{erf}(az)\exp{(-a_1z^2)}dz = \frac{a}{a_1\sqrt{a^2+a_1}}$
    
    Substituting, $a=\frac{1}{\sqrt{2}}, a_1 = \frac{(\sigma_{x_{\text{in}}}^2+1)}{2\sigma_{x_{\text{in}}}^2(\sigma_{x_{\text{in}}}^2(1-(r^l_{x_{\text{in}}})^2) + 1)}$, we have
\begin{align*}
    I_{3,2} &= \frac{r^l_{x_{\text{in}}}}{4\pi\sigma_{x_{\text{in}}}(\sigma_{x_{\text{in}}}^2(1- (r^l_{x_{\text{in}}})^2) + 1)^{\frac{3}{2}}}\frac{\frac{1}{\sqrt{2}}}{\frac{(\sigma_{x_{\text{in}}}^2+1)}{2\sigma_{x_{\text{in}}}^2(\sigma_{x_{\text{in}}}^2(1-(r^l_{x_{\text{in}}})^2) + 1)}\sqrt{\frac{1}{2}+\frac{(\sigma_{x_{\text{in}}}^2+1)}{2\sigma_{x_{\text{in}}}^2(\sigma_{x_{\text{in}}}^2(1-(r^l_{x_{\text{in}}})^2) + 1)}}}\\
    &= \frac{r^l_{x_{\text{in}}}\sigma_{x_{\text{in}}}^2}{2\pi(\sigma_{x_{\text{in}}}^2+1)\sqrt{\sigma_{x_{\text{in}}}^4 + 2\sigma_{x_{\text{in}}}^2 + 1 - (r^l_{x_{\text{in}}})^2\sigma_{x_{\text{in}}}^4}}\\
    I_{3,2} &= \frac{r^l_{x_{\text{in}}}\sigma_{x_{\text{in}}}^2}{2\pi(\sigma_{x_{\text{in}}}^2+1)\sqrt{(\sigma_{x_{\text{in}}}^2 + 1)^2 - (r^l_{x_{\text{in}}}\sigma_{x_{\text{in}}}^2)^2}}
\end{align*}
\begin{align*}
    I_{3,3} &=  \frac{r^l_{x_{\text{in}}}}{2\pi\sigma_{x_{\text{in}}}(\sigma_{x_{\text{in}}}^2(1- (r^l_{x_{\text{in}}})^2) + 1)^{\frac{3}{2}}}. \\
    & \hspace{120pt} \int_{-\infty}^{\infty}\frac{y^2_{\text{in}_i}}{\sqrt{2\pi}}\exp{(\frac{-y^2_{\text{in}_i}}{2})}\exp{(\frac{-y^2_{\text{in}_i}(\sigma_{x_{\text{in}}}^2 + 1)}{2\sigma_{x_{\text{in}}}^2(\sigma_{x_{\text{in}}}^2(1-(r^l_{x_{\text{in}}})^2) + 1)})}dy_{\text{in}_i}\\
    &= \frac{r^l_{x_{\text{in}}}}{2\pi\sigma_{x_{\text{in}}}(\sigma_{x_{\text{in}}}^2(1- (r^l_{x_{\text{in}}})^2) + 1)^{\frac{3}{2}}}. \\
    & \hspace{120pt} \int_{-\infty}^{\infty}\frac{y^2_{\text{in}_i}}{\sqrt{2\pi}}\exp{(\frac{-y^2_{\text{in}_i}(\sigma_{x_{\text{in}}}^4 + 2\sigma_{x_{\text{in}}}^2 + 1 - (r^l_{x_{\text{in}}})^2\sigma_{x_{\text{in}}}^4)}{2\sigma_{x_{\text{in}}}^2(\sigma_{x_{\text{in}}}^2(1-(r^l_{x_{\text{in}}})^2) + 1)})}dy_{\text{in}_i}\\
    &= \frac{r^l_{x_{\text{in}}}}{2\pi\sigma_{x_{\text{in}}}(\sigma_{x_{\text{in}}}^2(1- (r^l_{x_{\text{in}}})^2) + 1)^{\frac{3}{2}}}\int_{-\infty}^{\infty}\frac{y^2_{\text{in}_i}}{\sqrt{2\pi}}\exp{(\frac{-y^2_{\text{in}_i}((\sigma_{x_{\text{in}}}^2 + 1)^2 - (r^l_{x_{\text{in}}}\sigma_{x_{\text{in}}}^2)^2)}{2\sigma_{x_{\text{in}}}^2(\sigma_{x_{\text{in}}}^2(1-(r^l_{x_{\text{in}}})^2) + 1)})}dy_{\text{in}_i}\\
    &= \frac{r^l_{x_{\text{in}}}}{2\pi\sigma_{x_{\text{in}}}(\sigma_{x_{\text{in}}}^2(1- (r^l_{x_{\text{in}}})^2) + 1)^{\frac{3}{2}}}\frac{\sigma_{x_{\text{in}}}\sqrt{(\sigma_{x_{\text{in}}}^2(1-(r^l_{x_{\text{in}}})^2) + 1)}}{\sqrt{(\sigma_{x_{\text{in}}}^2 + 1)^2 - (r^l_{x_{\text{in}}}\sigma_{x_{\text{in}}}^2)^2}}\\
    &\hspace{20pt}\int_{-\infty}^{\infty}\frac{y^2_{\text{in}_i}}{\sqrt{2\pi}\frac{\sigma_{x_{\text{in}}}\sqrt{(\sigma_{x_{\text{in}}}^2(1-(r^l_{x_{\text{in}}})^2) + 1)}}{\sqrt{(\sigma_{x_{\text{in}}}^2 + 1)^2 - (r^l_{x_{\text{in}}}\sigma_{x_{\text{in}}}^2)^2}}}\exp{(\frac{-y^2_{\text{in}_i}((\sigma_{x_{\text{in}}}^2 + 1)^2 - (r^l_{x_{\text{in}}}\sigma_{x_{\text{in}}}^2)^2)}{2\sigma_{x_{\text{in}}}^2(\sigma_{x_{\text{in}}}^2(1-(r^l_{x_{\text{in}}})^2) + 1)})}dy_{\text{in}_i}\\
    &= \frac{r^l_{x_{\text{in}}}}{2\pi\sigma_{x_{\text{in}}}(\sigma_{x_{\text{in}}}^2(1- (r^l_{x_{\text{in}}})^2) + 1)^{\frac{3}{2}}}\frac{\sigma_{x_{\text{in}}}^3(\sigma_{x_{\text{in}}}^2(1-(r^l_{x_{\text{in}}})^2) + 1)^{\frac{3}{2}}}{((\sigma_{x_{\text{in}}}^2 + 1)^2 - (r^l_{x_{\text{in}}}\sigma_{x_{\text{in}}}^2)^2)^{\frac{3}{2}}}\\
    I_{3,3} &= \frac{r^l_{x_{\text{in}}}\sigma_{x_{\text{in}}}^2}{2\pi((\sigma_{x_{\text{in}}}^2 + 1)^2 - (r^l_{x_{\text{in}}}\sigma_{x_{\text{in}}}^2)^2)^{\frac{3}{2}}}
\end{align*}
\begin{align*}
    I &= I_1 + I_2 + I_3 \\
    &= I_{1,1} + I_{1,2} + I_{1,3} + I_{2,1} + I_{2,2} + I_{2,3} + I_{3,1} + I_{3,2} + I_{3,3}
\end{align*}
\begin{align*}
    I = \frac{1}{4} &+ \frac{1}{2\pi}\sin^{-1}{(\frac{r^l_{x_{\text{in}}}\sigma_{x_{\text{in}}}^2}{\sigma_{x_{\text{in}}}^2 + 1})} + \\
    & \frac{2r^l_{x_{\text{in}}}\sigma_{x_{\text{in}}}^2}{2\pi(\sigma_{x_{\text{in}}}^2+1)\sqrt{(\sigma_{x_{\text{in}}}^2 + 1)^2 - (r^l_{x_{\text{in}}}\sigma_{x_{\text{in}}}^2)^2}} + \frac{r^l_{x_{\text{in}}}\sigma_{x_{\text{in}}}^2}{2\pi((\sigma_{x_{\text{in}}}^2 + 1)^2 - (r^l_{x_{\text{in}}}\sigma_{x_{\text{in}}}^2)^2)^{\frac{3}{2}}}\\
    I = \frac{1}{4} &+ \frac{1}{2\pi}\sin^{-1}{(\frac{r^l_{x_{\text{in}}}\sigma_{x_{\text{in}}}^2}{\sigma_{x_{\text{in}}}^2 + 1})}  + \frac{r^l_{x_{\text{in}}}\sigma_{x_{\text{in}}}^2((2\sigma_{x_{\text{in}}}^2 + 3)(\sigma_{x_{\text{in}}}^2 + 1) - 2(r^l_{x_{\text{in}}}\sigma_{x_{\text{in}}}^2)^2)}{2\pi(\sigma_{x_{\text{in}}}^2 + 1)((\sigma_{x_{\text{in}}}^2 + 1)^2 - (r^l_{x_{\text{in}}}\sigma_{x_{\text{in}}}^2)^2)^{\frac{3}{2}}}
\end{align*}
We defined $\mathrm{Cov}(g_{\text{in}_{x_i}},g_{\text{in}_{y_i}})$, as
\begin{align*}
    \mathrm{Cov}(g_{\text{in}_{x_i}},g_{\text{in}_{y_i}}) &= Ir^l_{g_{\text{out}}}\sigma_{g_{{\text{out}}}}^2
\end{align*}
\begin{empheq}[box=\widefbox]{align*}
    & \mathrm{Cov}(g_{\text{in}_{x_i}},g_{\text{in}_{y_i}}) = \\
    &  \left[\frac{1}{4} + \frac{1}{2\pi}\sin^{-1}{(\frac{r^l_{x_{\text{in}}}\sigma_{x_{\text{in}}}^2}{\sigma_{x_{\text{in}}}^2 + 1})}  + \frac{r^l_{x_{\text{in}}}\sigma_{x_{\text{in}}}^2((2\sigma_{x_{\text{in}}}^2 + 3)(\sigma_{x_{\text{in}}}^2 + 1) - 2(r^l_{x_{\text{in}}}\sigma_{x_{\text{in}}}^2)^2)}{2\pi(\sigma_{x_{\text{in}}}^2 + 1)((\sigma_{x_{\text{in}}}^2 + 1)^2 - (r^l_{x_{\text{in}}}\sigma_{x_{\text{in}}}^2)^2)^{\frac{3}{2}}}\right]r^l_{g_{\text{out}}}\sigma_{g_{{\text{out}}}}^2
\end{empheq}

\subsection{LayerNorm}
\label{proof: layernorm}

The affine transformation for layernorm are typically initialized with $1$ scale and $0$ bias, so they do not change any of our derivations below and are ignored henceforth. 
For an input $\mathbf{x_{\text{in}}}$ the forward pass of LayerNorm is,
\begin{align*}
    \mathbf{x_{\text{out}}} & = \mathrm{LayerNorm}(\mathbf{x_{\text{in}}}) \\
    \implies {x_{\text{out}_i}} &= \frac{{x_{\text{in}_i}} - {\bar{x}_{\text{in}}}}{\hat{\sigma}_{x_{\text{in}}}}
    \intertext{Where}
    {\bar{x}_{\text{in}}} &= \frac{\sum_{i=1}^{d_{\text{in}}} {x_{\text{in}_i}}}{{d_{\text{in}}}}\\
    {\hat{\sigma}_{x_{\text{in}}}} &= \sqrt{\frac{\sum_{i=1}^{d_{\text{in}}} ({x_{\text{in}_i}} - {\bar{x}_{\text{in}}})^2}{{d_{\text{in}}}}}
\end{align*}
To get expectation of output of LayerNorm,
\begin{align*}
    \mathbb{E}[x_{\text{out}_i}] &= \mathbb{E}[\frac{{x_{\text{in}_i}} - {\bar{x}_{\text{in}}}}{\hat{\sigma}_{x_{\text{in}}}}]\\
    \sum_{i=1}^{d_{\text{in}}}\mathbb{E}[x_{\text{out}_i}] &= \sum_{i=1}^{d_{\text{in}}}\mathbb{E}[\frac{{x_{\text{in}_i}} - {\bar{x}_{\text{in}}}}{\hat{\sigma}_{x_{\text{in}}}}]\\
    &= \mathbb{E}[\sum_{i=1}^{d_{\text{in}}}\frac{{x_{\text{in}_i}} - {\bar{x}_{\text{in}}}}{\hat{\sigma}_{x_{\text{in}}}}]\\
    &= \mathbb{E}[\frac{\sum_{i=1}^{d_{\text{in}}}({x_{\text{in}_i}} - {\bar{x}_{\text{in}}})}{\hat{\sigma}_{x_{\text{in}}}}] \\
    \sum_{i=1}^{d_{\text{in}}}\mathbb{E}[x_{\text{out}_i}] &= 0\\
    \intertext{By symmetry for any $i,j$ and $i \neq j$ we have $\mathbb{E}[x_{\text{out}_i}] = \mathbb{E}[x_{\text{out}_j}] = \mu_{x_{\text{out}}}$}
    \implies d_{\text{in}}\mu_{x_{\text{out}}} = 0\\
    \Aboxed{\mu_{x_{\text{out}}} = 0}
\end{align*}
Similarly we calculate variance of output by,
\begin{align*}
    \mathrm{Var}({x_{\text{out}_i}}) &= \mathbb{E}[{x^2_{\text{out}_i}}] - \mathbb{E}[{x_{\text{out}_i}}]^2 = \mathbb{E}[{x^2_{\text{out}_i}}] \\
    \mathbb{E}[{x^2_{\text{out}_i}}] &= \mathbb{E}[\frac{({x_{\text{in}_i}} - {\bar{x}_{\text{in}}})^2}{\hat{\sigma}^2_{x_{\text{in}}}}]\\
    \sum_{i=1}^{d_{\text{in}}}\mathbb{E}[x^2_{\text{out}_i}] &= \sum_{i=1}^{d_{\text{in}}}\mathbb{E}[\frac{({x_{\text{in}_i}} - {\bar{x}_{\text{in}}})^2}{\hat{\sigma}^2_{x_{\text{in}}}}]\\
    &= \mathbb{E}[\sum_{i=1}^{d_{\text{in}}}\frac{({x_{\text{in}_i}} - {\bar{x}_{\text{in}}})^2}{\hat{\sigma}^2_{x_{\text{in}}}}]\\
    &= \mathbb{E}[\frac{\sum_{i=1}^{d_{\text{in}}}({x_{\text{in}_i}} - {\bar{x}_{\text{in}}})^2}{\hat{\sigma}^2_{x_{\text{in}}}}] \\
    \sum_{i=1}^{d_{\text{in}}}\mathbb{E}[x^2_{\text{out}_i}] &= {d_{\text{in}}}\\
    \intertext{By symmetry for any $i,j$ and $i \neq j$ we have $\mathbb{E}[x^2_{\text{out}_i}] = \mathbb{E}[x^2_{\text{out}_j}] = \sigma^2_{x_{\text{out}}}$}
    \implies d_{\text{in}}\sigma^2_{x_{\text{out}}} &= d_{\text{in}}\\
    \Aboxed{\sigma^2_{x_{\text{out}}} &= 1}
\end{align*}

Now we have $\hat{\sigma}_{x_{\text{in}}} \overset{a.s}{\longrightarrow} \sigma_{x_{\text{in}}}$ for large $d_\text{in}$. So for large values of ${d_{\text{in}}}$ we can treat $\hat{\sigma}_{x_{\text{in}}}$ as a constant which has value $\sigma_{x_{\text{in}}}$. We use this approximation to get the following results.
For two inputs $\mathbf{x_{\text{in}}}$ and $\mathbf{y_{\text{in}}}$ such that for all $i$, $\mathrm{Corr}({x_{\text{in}_i}},{y_{\text{in}_i}}) = r^l_{x_{\text{in}}}$. For all $j$ we have,
\begin{align*}
    \mathrm{Corr}({x_{\text{out}_j}},{y_{\text{out}_j}}) &= \frac{\mathbb{E}[{x_{\text{out}_j}}{y_{\text{out}_j}}] - \mathbb{E}[{x_{\text{out}_j}}]\mathbb{E}[{y_{\text{out}_j}}]}{\sqrt{\mathrm{Var}({x_{\text{out}_j}})\mathrm{Var}({y_{\text{out}_j}})}} \\
    &= \frac{\mathbb{E}[{x_{\text{out}_j}}{y_{\text{out}_j}}] - \mu_{x_{\text{out}}}\mu_{x_{\text{out}}}}{\sqrt{\sigma^2_{x_{\text{out}}}\sigma^2_{x_{\text{out}}}}} \\
    &= \frac{\mathbb{E}[{x_{\text{out}_j}}{y_{\text{out}_j}}] - 0}{\sqrt{1}}\\
    &= \mathbb{E}[{x_{\text{out}_j}}{y_{\text{out}_j}}]\\
    &= \mathbb{E}[\frac{({x_{\text{in}_j}} - {\bar{x}_{\text{in}}})({y_{\text{in}_j}} - {\bar{y}_{\text{in}}})}{\hat{\sigma}_{x_{\text{in}}}\hat{\sigma}_{y_{\text{in}}}}]\\
    &\approx \mathbb{E}[\frac{({x_{\text{in}_j}} - {\bar{x}_{\text{in}}})({y_{\text{in}_j}} - {\bar{y}_{\text{in}}})}{\sigma_{x_{\text{in}}}\sigma_{x_{\text{in}}}}] \\
    &= \frac{\mathbb{E}[({x_{\text{in}_j}} - {\bar{x}_{\text{in}}})({y_{\text{in}_j}} - {\bar{y}_{\text{in}}})]}{\sigma_{x_{\text{in}}}^2}\\
    &= \frac{\mathbb{E}[({x_{\text{in}_j}} - \frac{\sum_{k=1}^{d_{\text{in}}} {x_{\text{in}_k}}}{{d_{\text{in}}}})({y_{\text{in}_j}} - \frac{\sum_{l=1}^{d_{\text{in}}} {y_{\text{in}_l}}}{{d_{\text{in}}}})]}{\sigma_{x_{\text{in}}}^2}\\
    &= \frac{\mathbb{E}[{x_{\text{in}_j}}{y_{\text{in}_j}} - {y_{\text{in}_j}}\frac{\sum_{k=1}^{d_{\text{in}}} {x_{\text{in}_k}}}{{d_{\text{in}}}} - {x_{\text{in}_j}}\frac{\sum_{l=1}^{d_{\text{in}}} {y_{\text{in}_l}}}{{d_{\text{in}}}} + \frac{\sum_{k=1}^{d_{\text{in}}} {x_{\text{in}_k}}}{{d_{\text{in}}}}\frac{\sum_{l=1}^{d_{\text{in}}} {y_{\text{in}_l}}}{{d_{\text{in}}}}]}{\sigma_{x_{\text{in}}}^2}\\
\end{align*}
    Elements belonging to different dimensions from $\mathbf{x_{\text{in}}}$ and $\mathbf{y_{\text{in}}}$ are independent of each other and hence for $i,j$ and $i \neq j$ we have $\mathbb{E}[{x_{\text{in}_i}}{y_{\text{in}_j}}] = \mu^2_{x_{\text{in}}}$.
\begin{align*}
    &= \frac{\mathbb{E}[{x_{\text{in}_j}}{y_{\text{in}_j}}] - \mathbb{E}[{y_{\text{in}_j}}\frac{\sum_{k=1}^{d_{\text{in}}} {x_{\text{in}_k}}}{{d_{\text{in}}}}] - \mathbb{E}[{x_{\text{in}_j}}\frac{\sum_{l=1}^{d_{\text{in}}} {y_{\text{in}_l}}}{{d_{\text{in}}}}] + \mathbb{E}[\frac{\sum_{k=1}^{d_{\text{in}}} {x_{\text{in}_k}}}{{d_{\text{in}}}}\frac{\sum_{l=1}^{d_{\text{in}}} {y_{\text{in}_l}}}{{d_{\text{in}}}}]}{\sigma_{x_{\text{in}}}^2} \\
    &= \frac{r^l_{x_{\text{in}}}\sigma^2_{x_{\text{in}}} + \mu^2_{x_{\text{in}}} - \frac{r^l_{x_{\text{in}}}\sigma^2_{x_{\text{in}}} + d_{\text{in}} \mu^2_{x_{\text{in}}}}{d_{\text{in}}} - \frac{r^l_{x_{\text{in}}}\sigma^2_{x_{\text{in}}} + d_{\text{in}} \mu^2_{x_{\text{in}}}}{d_{\text{in}}} + \frac{r^l_{x_{\text{in}}}d_{\text{in}}\sigma^2_{x_{\text{in}}} + d^2_{\text{in}} \mu^2_{x_{\text{in}}}}{d^2_{\text{in}}}}{\sigma_{x_{\text{in}}}^2} \\
    &= \frac{r^l_{x_{\text{in}}}\sigma^2_{x_{\text{in}}}(1-\frac{1}{d_\text{in}})}{\sigma_{x_{\text{in}}}^2}\\
\end{align*}
\begin{align*}
    \Aboxed{\mathrm{Corr}({x_{\text{out}_j}},{y_{\text{out}_j}}) = r^l_{x_{\text{in}}}(1-\frac{1}{d_\text{in}}) \approx r^l_{x_{\text{in}}} = r^l_{x_{\text{out}}}}
\end{align*}
From \citet{xu2019understanding} (Eq. 17), the backward pass through LayerNorm is,
\begin{align*}
    \mathbf{g_{{\text{in}}}} &= \frac{\mathbf{g_{{\text{out}}}}}{\hat{\sigma}_{x_{\text{in}}}}(\mathbf{I_{d_{\text{in}}}} - \frac{\mathbf{1^T_{d_{\text{in}}}}\mathbf{1_{d_{\text{in}}}} + \mathbf{x^T_{\text{out}}}\mathbf{x_{\text{out}}}}{d_{\text{in}}})\\
    &\approx \frac{\mathbf{g_{{\text{out}}}}}{\sigma_{x_{\text{in}}}}(\mathbf{I_{d_{\text{in}}}} - \frac{\mathbf{1^T_{d_{\text{in}}}}\mathbf{1_{d_{\text{in}}}} + \mathbf{x^T_{\text{out}}}\mathbf{x_{\text{out}}}}{d_{\text{in}}})
    \intertext{We have $\displaystyle\lim_{{d_{\text{in}}} \to \infty} \frac{\mathbf{1^T_{d_{\text{in}}}}\mathbf{1_{d_{\text{in}}}} + \mathbf{x^T_{\text{out}}}\mathbf{x_{\text{out}}}}{d_{\text{in}}} = \mathbf{O_{d_{\text{in}},d_{\text{in}}}}$ where $\mathbf{O_{d_{\text{in}},d_{\text{in}}}}$ is zero matrix with shape $d_{\text{in}} \times d_{\text{in}}$}
    \mathbf{g_{{\text{in}}}} &\approx \frac{\mathbf{g_{{\text{out}}}}}{\sigma_{x_{\text{in}}}}(\mathbf{I_{d_{\text{in}}}}) \\
    &= \frac{\mathbf{g_{{\text{out}}}}}{\sigma_{x_{\text{in}}}}\\
    \implies g_{{\text{in}_i}} &= \frac{g_{{\text{out}_i}}}{\sigma_{x_{\text{in}}}}
\end{align*}
If $\mu_{g_{{\text{out}}}} = 0$,
\begin{empheq}[box=\widefbox]{align*}
    \mu_{g_{{\text{in}}}} &= 0\\
    \sigma^2_{g_{{\text{in}}}} &= \frac{\sigma^2_{g_{{\text{out}}}}}{\sigma^2_{x_{\text{in}}}}
\end{empheq}
\subsection{Softmax}
\label{proof: softmax_ln}
\label{proof: softmax}

\textbf{Assumption}: Other than assuming normally distributed inputs, we also assume that L is large $L>>1$ to derive softmax variance. 

The forward pass of Softmax can be defined as
\begin{align*}
    \mathbf{x_{\text{out}}} &= \mathrm{Softmax}(\mathbf{x_{\text{in}}})\\
    {x_{\text{out}_i}} &= \frac{e^{x_{\text{in}_i}}}{\sum_{j=1}^{L}e^{x_{\text{in}_j}}}\\
\end{align*}
For calculating mean we can easily see that,
\begin{align*}
    \sum_{i=1}^{L}{x_{\text{out}_i}} &= 1\\
    \intertext{Taking expectation both sides, we get}
    \mathbb{E}[\sum_{i=1}^{L}{x_{\text{out}_i}}] &= 1\\
    \sum_{i=1}^{L}\mathbb{E}[{x_{\text{out}_i}}] &= 1\\
    \intertext{By symmetry we can assume that for any $i,j, i\neq j$, we have $\mathbb{E}[{x_{\text{out}_i}}] = \mathbb{E}[{x_{\text{out}_j}}]$}
    L\mathbb{E}[{x_{\text{out}_i}}] &= 1\\
    \Aboxed{\mu_{x_{\text{out}}} &= \frac{1}{L}}
\end{align*}

Let us define $z = \sum_j e^{y_j}$ where $y_j = x_j-x_i$ is normally distributed $\mathcal{N}(0, \sigma_j)$. Hence, each $e^{y_j}$ is log-normally distributed, and $z$ is a sum of correlated log-normals. Following \citep{LognormalApprox}, this sum of log-normals can be approximated as another log-normal random variable, $Log\mathcal{N}(\mu_z, \sigma_z)$, where $\mu_z$ and $\sigma_z$ are as follows -

\begin{align*}
    S_+ &= E [\sum_j y_j] = \sum_j e^{\frac{\sigma_j^2}{2}} \\ 
    \sigma^2_z &= \frac{1}{S_+^2} \sum_{j,k} corr_{j,k} \sigma_j \sigma_k e^{\frac{1}{2}(\sigma_j^2 + \sigma_k^2)}\\
    \mu_z &= ln(S_+) - \frac{\sigma_z^2}{2}\\
\end{align*}

Since the difference of two normals $x_j$ and $x_i$ is also normal, from the M.G.F. of normal distribution, we have $\sigma_j^2 = 2 \sigma^2_{x_{\text{in}}} (1 - r_{x_{\text{in}}})$  if $j \neq i$, and $\sigma_j^2 = 0$ if $j=i$. 

Also, $corr_{j,k} = 0$ if $j=i$ or $k=i$, else $corr_{j,k} = \frac{1}{2}$.

We can substitute these values in the above equations, to get

\begin{align*}
    S_+ &= (L-1) e^{\sigma^2_{x_{\text{in}}} (1 - r_{x_{\text{in}}})} + 1 \\ 
    \sigma^2_z &=  \sigma^2_{x_{\text{in}}} (1 - r_{x_{\text{in}}}) \frac{L}{L-1} \\
    \mu_z &= ln(S_+) - \frac{\sigma_z^2}{2} \\
\end{align*}

Since $z$ is log-normal, $x_{\text{out}} = \frac{1}{z}$ is also log-normal with $Log\mathcal{N}(-\mu_z, \sigma_z)$. The variance of log-normal distribution can be obtained from standard formulae for log-normal distribution as $(e^{\sigma_z^2} - 1) e^{\sigma_z^2 - 2 \mu_z}$.

Substituting the values of $\mu_z$ and $\sigma_z$ from above, we get

\begin{align*}
    \sigma_{x_{\text{out}}}^2 &= \frac{(e^{\sigma_z^2} - 1) e^{2 * \sigma_z^2}}{S_+^2} \\
    &= \frac{(e^{\sigma^2_{x_{\text{in}}} (1 - r_{x_{\text{in}}}) \frac{L}{L-1}} - 1)e^{2\sigma^2_{x_{\text{in}}} (1 - r_{x_{\text{in}}}) \frac{L}{L-1}}}{((L-1) e^{\sigma^2_{x_{\text{in}}} (1 - r_{x_{\text{in}}})} + 1)^2}
\end{align*}

For large L, we can ignore the 1 in the denominator -

\begin{align*}
    \sigma_{x_{\text{out}}}^2 = \frac{(e^{\sigma^2_{x_{\text{in}}} (1 - r_{x_{\text{in}}}) \frac{L}{L-1}} - 1)}{(L-1)^2}
\end{align*}

If $L >> 1$ and $\sigma^2_{x_{\text{in}}}$ is small, we get the more simplified formula as -

\begin{align*}
    \Aboxed{\sigma_{x_{\text{out}}}^2 &\approx \frac{(e^{(1 - r^d_{x_{\text{in}}})\sigma^2_{x_{\text{in}}}} - 1)}{L^2}} \tag*{\text{(Assuming $L>>1$)}}
\end{align*}

Using the mean and variances, we can calculate the scale of softmax output as follows-

\begin{align*}
     E[{x^2_{\text{out}}}] &= \sigma_{x_{\text{out}}}^2 + \mu_{x_{\text{out}}}^2 \\
     & =   \frac{(e^{(1 - r^d_{x_{\text{in}}})\sigma^2_{x_{\text{in}}}})}{L^2}
\end{align*}

The Jacobian of Softmax can be calculated as (\citep{LipschitzConstantSelfAttention}):
\begin{align*}
    J_{i,j} &= \begin{cases}
        x_{\text{out}_i}(1-x_{\text{out}_i}) & \text{if }i=j\\
        -x_{\text{out}_i}x_{\text{out}_j} & \text{else}
    \end{cases}
    \intertext{For large values of $L$ this approximately becomes}
    \mathbf{J} &\approx \mathrm{diag}(\mathbf{x_{\text{out}}})\\
    \mathbf{g_{\text{in}}} &= \mathbf{g_{\text{out}}}\mathbf{J} \\
    {g_{\text{in}_i}} &\approx {g_{\text{out}_i}}{x_{\text{out}_i}}\\
    \mathbb{E}[{g_{\text{in}_i}}] &\approx \mathbb{E}[{g_{\text{out}_i}}{x_{\text{out}_i}}]\\
    &= \mathbb{E}[{g_{\text{out}_i}}]\mathbb{E}[{x_{\text{out}_i}}] = 0 = \mu_{g_{\text{in}}}\\
    \mathbb{E}[{g^2_{\text{in}_i}}] &\approx \mathbb{E}[{g^2_{\text{out}_i}}{x^2_{\text{out}_i}}]\\
    &= \mathbb{E}[{g^2_{\text{out}_i}}]\mathbb{E}[{x^2_{\text{out}_i}}]\\
    \Aboxed{\sigma^2_{g_{\text{in}}} &= \sigma^2_{g_{\text{out}}}\frac{(e^{(1 - r^d_{x_{\text{in}}})\sigma^2_{x_{\text{in}}}})}{L^2}}
\end{align*}

\subsection{Scaled Dot-Product Attention}
\label{proof: sdpa}

\textbf{Inapplicability of Direct Usage of Softmax Derivations for SHA}: One may be tempted to assume attention scores to be independent of values. This then enables the use of our previous LogNormal-based softmax derivation, to easily derive the forward variances.

But the theoretically calculated moments strongly disagree with empirical simulations. This is because SHA is 
$\mathbf{X_{\text{out}}} = \mathrm{Dropout}(\mathrm{SoftMax}(\dfrac{\mathbf{X_{\text{in}}}\mathbf{W_Q}\mathbf{{W_K}^T}\mathbf{X^T_{\text{in}}}}{\sqrt{d_k}}))\mathbf{X_{\text{in}}}\mathbf{W_V}$, and the $\mathbf{{W_K}^T}\mathbf{X^T_{\text{in}}}$ term cannot be treated independently of the $\mathbf{X_{\text{in}}}\mathbf{W_V}$ term. A simple verification of this can be checked by simply simulating $\mathbf{(X{W}^T})X$, and verifying that the variances of the results do not match that of $L*\sigma^2(\mathbf{(XW)})$, but do if the second $\mathbf{X}$ is replaced by another random tensor.

This necessitates an alternate methodology to derive SHA, where the components are treated as a unified whole.

\textbf{Assumption}: We assume that $L$ and $d_{\text{in}}$ are very large when compared to scale of scores being passed to the Softmax. These approximations hold true for small values of $\sigma_q$ and $\sigma_k$, and the resulting formulae are fairly accurate, as shown in the numerical verification section.

The forward pass of Scaled Dot-Product Attention is
\begin{align*}
    \mathbf{X_{\text{out}}} &= \mathrm{Dropout}(\mathrm{SoftMax}(\frac{\mathbf{Q}\mathbf{K^T}}{\sqrt{d_{i,k}}}))\mathbf{V}
    \intertext{Where,}
    \mathbf{Q} &= \mathbf{X_{\text{in}}}\mathbf{W_Q}\\
    \mathbf{K} &= \mathbf{X_{\text{in}}}\mathbf{W_K}\\
    \mathbf{V} &= \mathbf{X_{\text{in}}}\mathbf{W_V}\\
    \mathbf{X_{\text{out}}} &= \mathrm{Dropout}(\mathrm{SoftMax}(\frac{\mathbf{X_{\text{in}}}\mathbf{W_Q}\mathbf{{W_K}^T}\mathbf{X^T_{\text{in}}}}{\sqrt{d_{i,k}}}))\mathbf{X_{\text{in}}}\mathbf{W_V}
    \intertext{Let,}
    \mathbf{O} &= \mathrm{Dropout}(\mathrm{SoftMax}(\frac{\mathbf{X_{\text{in}}}\mathbf{W_Q}\mathbf{{W_K}^T}\mathbf{X^T_{\text{in}}}}{\sqrt{d_{i,k}}}))\mathbf{X_{\text{in}}}\\
    \mathbf{W} &= \frac{\mathbf{X_{\text{in}}}\mathbf{W_Q}\mathbf{{W_K}^T}}{\sqrt{d_{i,k}}}\\
    \mathbf{O} &= \mathrm{Dropout}(\mathrm{SoftMax}(\mathbf{W}\mathbf{X^T_{\text{in}}}))\mathbf{X_{\text{in}}}
\end{align*}
Using results from Linear Layer we have $\sigma^2_w = d_{\text{in}}\sigma^2_{x_{\text{in}}}\sigma^2_{q}\sigma^2_{k} = d_{\text{in}}\sigma^2_{x_{\text{in}}}\sigma^2_{qk}$
\begin{align*}
    O_{i,j} &= \sum_{k=1}^{L} \mathrm{Dropout}(\mathrm{SoftMax}(\mathbf{W}\mathbf{X^T_{\text{in}}}))_{i,k}{X_{\text{in}_{k,j}}}\\
    &= \sum_{k=1}^{L} \mathrm{Dropout}(\frac{\exp{((\mathbf{W}\mathbf{X^T_{\text{in}}})_{i,k}})}{\displaystyle\sum_{m=1}^{L}\exp{((\mathbf{W}\mathbf{X^T_{\text{in}}})_{i,m}})}){X_{\text{in}_{k,j}}}\\
    &= \sum_{k=1}^{L} \frac{\mathrm{Dropout}(\exp{((\mathbf{W}\mathbf{X^T_{\text{in}}})_{i,k}}))}{\displaystyle\sum_{m=1}^{L}\exp{((\mathbf{W}\mathbf{X^T_{\text{in}}})_{i,m}})}{X_{\text{in}_{k,j}}}\\
    &= \frac{\displaystyle\sum_{k=1}^{L}\mathrm{Dropout}(\exp{((\mathbf{W}\mathbf{X^T_{\text{in}}})_{i,k}})){X_{\text{in}_{k,j}}}}{\displaystyle\sum_{m=1}^{L}\exp{((\mathbf{W}\mathbf{X^T_{\text{in}}})_{i,m}})}\\
    &= \frac{\displaystyle\sum_{k=1}^{L}\mathrm{Dropout}(\exp{(\sum_{l=1}^{d_{\text{in}}} W_{i,l}X_{\text{in}_{k,l}})}){X_{\text{in}_{k,j}}}}{\displaystyle\sum_{m=1}^{L}\exp{(\sum_{n=1}^{d_{\text{in}}} W_{i,n}X_{\text{in}_{m,n}})}}\\
    &= \frac{\displaystyle\sum_{k=1}^{L}\mathrm{Dropout}(\exp{(\sum_{l=1}^{d_{\text{in}}} W_{i,l}X_{\text{in}_{k,l}})}{X_{\text{in}_{k,j}}})}{\displaystyle\sum_{m=1}^{L}\exp{(\sum_{n=1}^{d_{\text{in}}} W_{i,n}X_{\text{in}_{m,n}})}}\\
 \end{align*}
Each $X_{\text{in}_{i,j}}$ can be written as:
\begin{align*}
    X_{\text{in}_{i,j}} &= \epsilon_j + \delta_{i,j}
    \intertext{Where $\epsilon_j$ and $\delta_{i,j}$ are all independent and defined as}
    \epsilon_j &\sim \mathcal{N}(0, r^l_{x_{\text{in}}}\sigma_{x_{\text{in}}}^2)\\
    \delta_{i,j} &\sim \mathcal{N}(0, (1-r^l_{x_{\text{in}}})\sigma_{x_{\text{in}}}^2)\\
    O_{i,j} &= \frac{\displaystyle\sum_{k=1}^{L}\mathrm{Dropout}(\exp{(\sum_{l=1}^{d_{\text{in}}} W_{i,l}X_{\text{in}_{k,l}})}{X_{\text{in}_{k,j}}})}{\displaystyle\sum_{k=1}^{L}\exp{(\sum_{l=1}^{d_{\text{in}}} W_{i,l}X_{\text{in}_{k,l}})}}\\
    &= \frac{\displaystyle\sum_{k=1}^{L}(1-d_{i,k})(\exp{(\sum_{l=1}^{d_{\text{in}}} W_{i,l}X_{\text{in}_{k,l}})}{X_{\text{in}_{k,j}}})}{(1-p)\displaystyle\sum_{k=1}^{L}\exp{(\sum_{l=1}^{d_{\text{in}}} W_{i,l}X_{\text{in}_{k,l}})}}\\
    \intertext{Where $d_{i,k}$ is Bernoulli random variable which is 1 with probability $p$}
    &= \frac{\sum_{k=1}^L(1-d_{i,k})\exp{(\sum_{l=1}^{d_{\text{in}}}W_{i,l}(\epsilon_l + \delta_{k,l}))}(\epsilon_j + \delta_{k,j})}{(1-p)\sum_{k=1}^L\exp{(\sum_{l=1}^{d_{\text{in}}}W_{i,l}(\epsilon_l + \delta_{k,l}))}}\\
    &= \epsilon_j\frac{\sum_{k=1}^L(1-d_{i,k})\exp{(\sum_{l=1}^{d_{\text{in}}}W_{i,l}\epsilon_l)}\exp{(\sum_{l=1}^{d_{\text{in}}}W_{i,l}\delta_{k,l})}}{(1-p)\sum_{k=1}^L\exp{(\sum_{l=1}^{d_{\text{in}}}W_{i,l}\epsilon_l)}\exp{(\sum_{l=1}^{d_{\text{in}}}W_{i,l}\delta_{k,l})}} \\
    &\quad + \frac{\sum_{k=1}^L(1-d_{i,k})\exp{(\sum_{l=1}^{d_{\text{in}}}W_{i,l}\epsilon_l)}\exp{(\sum_{l=1}^{d_{\text{in}}}W_{i,l}\delta_{k,l})}\delta_{k,j}}{(1-p)\sum_{k=1}^L\exp{(\sum_{l=1}^{d_{\text{in}}}W_{i,l}\epsilon_l)}\exp{(\sum_{l=1}^{d_{\text{in}}}W_{i,l}\delta_{k,l})}}\\
    &= \epsilon_j\frac{\sum_{k=1}^L(1-d_{i,k})\exp{(\sum_{l=1}^{d_{\text{in}}}W_{i,l}\delta_{k,l})}}{(1-p)\sum_{k=1}^L\exp{(\sum_{l=1}^{d_{\text{in}}}W_{i,l}\delta_{k,l})}} + \frac{\sum_{k=1}^L(1-d_{i,k})\exp{(\sum_{l=1}^{d_{\text{in}}}W_{i,l}\delta_{k,l})}\delta_{k,j}}{(1-p)\sum_{k=1}^L\exp{(\sum_{l=1}^{d_{\text{in}}}W_{i,l}\delta_{k,l})}}
\end{align*}
Let $v_1 = \epsilon_j\frac{\sum_{k=1}^L(1-d_{i,k})\exp{(\sum_{l=1}^{d_{\text{in}}}W_{i,l}\delta_{k,l})}}{(1-p)\sum_{k=1}^L\exp{(\sum_{l=1}^{d_{\text{in}}}W_{i,l}\delta_{k,l})}}$ and $v_2 = \frac{\sum_{k=1}^L(1-d_{i,k})\exp{(\sum_{l=1}^{d_{\text{in}}}W_{i,l}\delta_{k,l})}\delta_{k,j}}{(1-p)\sum_{k=1}^L\exp{(\sum_{l=1}^{d_{\text{in}}}W_{i,l}\delta_{k,l})}}$. We have,
\begin{align*}
    O_{i,j} &= v_1 + v_2\\
    \intertext{Given a fixed $\epsilon, W$, we have}
    v_1 | \epsilon,W &= \epsilon_j\frac{\sum_{k=1}^L(1-d_{i,k})\exp{(\sum_{l=1}^{d_{\text{in}}}W_{i,l}\delta_{k,l})}}{(1-p)\sum_{k=1}^L\exp{(\sum_{l=1}^{d_{\text{in}}}W_{i,l}\delta_{k,l})}}\\
    &= \epsilon_j\frac{\frac{\sum_{k=1}^L(1-d_{i,k})\exp{(\sum_{l=1}^{d_{\text{in}}}W_{i,l}\delta_{k,l})}}{L}}{(1-p)\frac{\sum_{k=1}^L\exp{(\sum_{l=1}^{d_{\text{in}}}W_{i,l}\delta_{k,l})}}{L}}\\
    \intertext{By WLLN, $\frac{\sum_{k=1}^L(1-d_{i,k})\exp{(\sum_{l=1}^{d_{\text{in}}}W_{i,l}\delta_{k,l})}}{L} {\overset{p}\to} (1-p)\mathbb{E}_{\delta}[\exp{(\sum_{l=1}^{d_{\text{in}}}W_{i,l}\delta_{k,l})}]$, and}
    \intertext{$(1-p)\frac{\sum_{k=1}^L\exp{(\sum_{l=1}^{d_{\text{in}}}W_{i,l}\delta_{k,l})}}{L} {\overset{p}\to} (1-p)\mathbb{E}_{\delta}[\exp{(\sum_{l=1}^{d_{\text{in}}}W_{i,l}\delta_{k,l})}]$}
    \intertext{Thus, we have $v_1 | \epsilon,W {\overset{p}\to} \epsilon_j$}
    v_2|\epsilon, W &= \frac{\sum_{k=1}^L(1-d_{i,k})\exp{(\sum_{l=1}^{d_{\text{in}}}W_{i,l}\delta_{k,l})}\delta_{k,j}}{(1-p)\sum_{k=1}^L\exp{(\sum_{l=1}^{d_{\text{in}}}W_{i,l}\delta_{k,l})}}\\
    &= \frac{\frac{1}{\sqrt{L}}\sqrt{L}\sum_{k=1}^L\frac{(1-d_{i,k})\exp{(\sum_{l=1}^{d_{\text{in}}}W_{i,l}\delta_{k,l})}\delta_{k,j}}{L}}{(1-p)\sum_{k=1}^L\frac{\exp{(\sum_{l=1}^{d_{\text{in}}}W_{i,l}\delta_{k,l})}}{L}}\\
    \intertext{Let $\mu_{\text{num}} = \mathbb{E}_{\delta,d}[(1-d_{i,k})\exp{(\sum_{l=1}^{d_{\text{in}}}W_{i,l}\delta_{k,l})}\delta_{k,j}]$, $\sigma_{\text{num}}^2 = \mathrm{Var}_{\delta,d}((1-d_{i,k})\exp{(\sum_{l=1}^{d_{\text{in}}}W_{i,l}\delta_{k,l})}\delta_{k,j})$. By central limit theorem for large $L$,}
    \sqrt{L}\frac{\sum_{k=1}^L(1-d_{i,k})\exp{(\sum_{l=1}^{d_{\text{in}}}W_{i,l}\delta_{k,l})}\delta_{k,j}}{L} &= \sqrt{L}\frac{\sum_{k=1}^L(1-d_{i,k})(\exp{(\sum_{l=1}^{d_{\text{in}}}W_{i,l}\delta_{k,l})}\delta_{k,j} - \mu_{\text{num}})}{L} + \sqrt{L}\mu_{\text{num}}\\
    \sqrt{L}\frac{\sum_{k=1}^L(1-d_{i,k})\exp{(\sum_{l=1}^{d_{\text{in}}}W_{i,l}\delta_{k,l})}\delta_{k,j}}{L} &{\overset{d}\to} \mathcal{N}(0, \sigma_{\text{num}}^2) + \sqrt{L}\mu_{\text{num}}\\
    \frac{\sum_{k=1}^L(1-d_{i,k})\exp{(\sum_{l=1}^{d_{\text{in}}}W_{i,l}\delta_{k,l})}\delta_{k,j}}{L} &{\overset{d}\to} \mathcal{N}(\mu_{\text{num}}, \frac{\sigma_{\text{num}}^2}{L})\\
    \mu_{\text{num}} &= \mathbb{E}_d[1-d_{i,k}](\prod_{l=1, l \neq j}^{l = d}\mathbb{E}_{\delta}[\exp{(W_{i,l}\delta_{k,l})}])\mathbb{E}_{\delta}[\exp{(W_{i,j}\delta_{k,j})}\delta_{k,j}]\\
    \mathbb{E}_{\delta}[\exp{(W_{i,l}\delta_{k,l})}] &= \exp{(\frac{W_{i,l}^2\sigma_{\delta}^2}{2})} \tag{MGF of gaussian}\\
    \mathbb{E}_{\delta}[\exp{(W_{i,j}\delta_{k,j})}\delta_{k,j}] &= \int_{-\infty}^{\infty}\frac{\exp{(W_{i,j}\delta_{k,j})}\delta_{k,j}}{\sqrt{2\pi}\sigma_{\delta}}\exp{(-\frac{\delta_{k,j}^2}{2\sigma_{\delta}^2})}d\delta_{k,j}\\
    &= \int_{-\infty}^{\infty}\exp{(\frac{W^2_{i,j}\sigma_{\delta}^2}{2})}\frac{\delta_{k,j}}{\sqrt{2\pi}\sigma_{\delta}}\exp{(-\frac{(\delta_{k,j} - W_{i,j}\sigma_{\delta}^2)^2}{2\sigma_{\delta}^2})}d\delta_{k,j}\\
    &= \exp{(\frac{W^2_{i,j}\sigma_{\delta}^2}{2})}\int_{-\infty}^{\infty}\frac{\delta_{k,j}}{\sqrt{2\pi}\sigma_{\delta}}\exp{(-\frac{(\delta_{k,j} - W_{i,j}\sigma_{\delta}^2)^2}{2\sigma_{\delta}^2})}d\delta_{k,j}\\
    &= \exp{(\frac{W^2_{i,j}\sigma_{\delta}^2}{2})}W_{i,j}\sigma_{\delta}^2\\
    \mu_{\text{num}} &= (1-p)\exp{(\frac{\sum_{l=1}^{d_{\text{in}}} W^2_{i,l}\sigma_{\delta}^2}{2})}W_{i,j}\sigma_{\delta}^2\\
    \sigma_{\text{num}}^2 = \mathbb{E}_d[(1-d_{i,k})^2]&(\prod_{l=1, l \neq j}^{l = d}\mathbb{E}_{\delta}[\exp{(2W_{i,l}\delta_{k,l})}]\mathbb{E}_{\delta}[\exp{(2W_{i,j}\delta_{k,j})}\delta_{k,j}^2] - \mu_{\text{num}}^2)\\
    \mathbb{E}_{\delta_{k,l}}[\exp{(2W_{i,l}\delta_{k,l})}] &= \exp{(2W_{i,l}^2\sigma_{\delta}^2)} \tag{MGF of gaussian}\\
    \mathbb{E}_{\delta_{k,j}}[\exp{(2W_{i,j}\delta_{k,j})}\delta_{k,j}^2] &= \int_{-\infty}^{\infty}\frac{\exp{(2W_{i,j}\delta_{k,j})}\delta_{k,j}^2}{\sqrt{2\pi}\sigma_{\delta}}\exp{(-\frac{\delta_{k,j}^2}{2\sigma_{\delta}^2})}d\delta_{k,j}\\
    &= \int_{-\infty}^{\infty}\exp{(2W^2_{i,j}\sigma_{\delta}^2)}\frac{\delta_{k,j}^2}{\sqrt{2\pi}\sigma_{\delta}}\exp{(-\frac{(\delta_{k,j} - 2W_{i,j}\sigma_{\delta}^2)^2}{2\sigma_{\delta}^2})}d\delta_{k,j}\\
    &= \exp{(2W^2_{i,j}\sigma_{\delta}^2)}\int_{-\infty}^{\infty}\frac{\delta_{k,j}^2}{\sqrt{2\pi}\sigma_{\delta}}\exp{(-\frac{(\delta_{k,j} - 2W_{i,j}\sigma_{\delta}^2)^2}{2\sigma_{\delta}^2})}d\delta_{k,j}\\
    &= \exp{(2W^2_{i,j}\sigma_{\delta}^2)}(4W_{i,j}^2\sigma_{\delta}^4 + \sigma_{\delta}^2)\\
    \sigma_{\text{num}}^2 = (1-p)\exp{(2\sum_{l=1}^{d_{\text{in}}} W^2_{i,l}\sigma_{\delta}^2)}&(4W_{i,j}^2\sigma_{\delta}^4 + \sigma_{\delta}^2) - (1-p)^2\exp{(\sum_{l=1}^{d_{\text{in}}} W^2_{i,l}\sigma_{\delta}^2)}W_{i,j}^2\sigma_{\delta}^4 
    \intertext{Similarly, $\sum_{k=1}^L\exp{(\sum_{l=1}^{d_{\text{in}}}W_{i,l}\delta_{k,l})}$ is also a sum of $L$ i.i.d. random variables for fixed $W$. By WLLN we have,}
    (1-p)\frac{\sum_{k=1}^L\exp{(\sum_{l=1}^{d_{\text{in}}}W_{i,l}\delta_{k,l})}}{L} &{\overset{p}\to} (1-p)\mathbb{E}_{\delta}[\exp{(\sum_{l=1}^{d_{\text{in}}}W_{i,l}\delta_{k,l})}]\\
    &{\overset{p}\to} (1-p)(\prod_{l=1}^{l = d}\mathbb{E}_{\delta}[\exp{(W_{i,l}\delta_{k,l})}])\\
    (1-p)\frac{\sum_{k=1}^L\exp{(\sum_{l=1}^{d_{\text{in}}}W_{i,l}\delta_{k,l})}}{L} &{\overset{p}\to} (1-p)\exp{(\frac{\sum_{l=1}^{d_{\text{in}}} W^2_{i,l}\sigma_{\delta}^2}{2})}\\
    v_2 &= \frac{\frac{\sum_{k=1}^L\exp{(\sum_{l=1}^{d_{\text{in}}}W_{i,l}\delta_{k,l})}\delta_{k,j}}{L}}{\frac{\sum_{k=1}^L\exp{(\sum_{l=1}^{d_{\text{in}}}W_{i,l}\delta_{k,l})}}{L}}
    \intertext{As for a given $W, \epsilon$, both the numerator and denominator converge in distribution and denominator is converging to a constant by Slutskys theorem,}
    v_2|W,\epsilon &{\overset{d}\to} \mathcal{N}(\frac{\mu_{\text{num}}}{(1-p)\exp{(\frac{\sum_{l=1}^{d_{\text{in}}}W^2_{i,l}\sigma_{\delta}^2}{2})}}, \frac{\sigma_{\text{num}}^2}{L(1-p)^2\exp{(\sum_{l=1}^{d_{\text{in}}}W^2_{i,l}\sigma_{\delta}^2)}})\\
    v_2|W,\epsilon &{\overset{d}\to} \mathcal{N}(W_{i,j}\sigma_{\delta}^2, \frac{\frac{\exp{(\sum_{l=1}^{d_{\text{in}}}W^2_{i,l}\sigma_{\delta}^2)}(4W_{i,j}^2\sigma_{\delta}^4 + \sigma_{\delta}^2)}{(1-p)} - W_{i,j}^2\sigma_{\delta}^4}{L})\\
\end{align*}
Thus we have, 
\begin{align*}
    O_{i,j}|W, \epsilon &\sim \mathcal{N}(W_{i,j}\sigma_{\delta}^2, \frac{\frac{\exp{(\sum_{l=1}^{d_{\text{in}}} W^2_{i,l}\sigma_{\delta}^2)}(4W_{i,j}^2\sigma_{\delta}^4 + \sigma_{\delta}^2)}{(1-p)} - W_{i,j}^2\sigma_{\delta}^4}{L}) + \epsilon_j\\
    \intertext{We have,}
    \mathbb{E}[O_{i,j}|W] &= W_{i,j}\sigma_{\delta}^2 + 0 = W_{i,j}\sigma_{\delta}^2\\
    \mathbb{E}[O^2_{i,j}|W] &= \frac{\frac{\exp{(\sum_{l=1}^{d_{\text{in}}} W^2_{i,l}\sigma_{\delta}^2)}(4W_{i,j}^2\sigma_{\delta}^4 + \sigma_{\delta}^2)}{(1-p)} - W_{i,j}^2\sigma_{\delta}^4}{L} + 
    \sigma_{\epsilon}^2\\    
    \mathbb{E}[O_{i,j}] &= \mathbb{E}_{W}[O_{i,j}|W] = \mathbb{E}_{W}[W_{i,j}\sigma_{\delta}^2] = 0\\
    \mathbb{E}[O_{i,j}^2] &= \mathbb{E}_{W}[O_{i,j}^2|W]\\
    &= \mathbb{E}_{W}[W^2_{i,j}\sigma_{\delta}^4 + \frac{\frac{\exp{(\sum_{l=1}^{d_{\text{in}}} W^2_{i,l}\sigma_{\delta}^2)}(4W_{i,j}^2\sigma_{\delta}^4 + \sigma_{\delta}^2)}{(1-p)} - W_{i,j}^2\sigma_{\delta}^4}{L} + \sigma_{\epsilon}^2]\\
    \intertext{For large $d_{\text{in}}$ by WLLN and continuous mapping theorem $\exp{(\sum_{l=1}^{d_{\text{in}}} W^2_{i,l}\sigma_{\delta}^2)} \approx \exp{(d_{\text{in}}\sigma_w^2\sigma_{\delta}^2)}$}
    &= \frac{(L-1)\sigma_w^2\sigma_{\delta}^4+ \frac{\exp{(d_{\text{in}}\sigma_w^2\sigma_{\delta}^2)}(4\sigma_w^2\sigma_\delta^4 + \sigma_\delta^2)}{(1-p)}}{L} + \sigma_\epsilon^2\\
    &= \frac{(1-r^l_{x_{\text{in}}})^2(L-1)d_{\text{in}}\sigma^6_{x_{\text{in}}}\sigma^2_{qk}+ \frac{\exp{((1-r^l_{x_{\text{in}}})d^2_{\text{in}}\sigma^4_{x_{\text{in}}}\sigma^2_{qk})}(4(1-r^l_{x_{\text{in}}})^2 d_{\text{in}}\sigma^6_{x_{\text{in}}}\sigma^2_{qk} + (1-r^l_{x_{\text{in}}})\sigma_{x_{\text{in}}}^2)}{(1-p)}}{L} + r^l_{x_{\text{in}}}\sigma^2_{x_{\text{in}}}
\end{align*}

Hence,
\begin{empheq}[box=\fbox]{align*}
    \mu_{x_{out}} &= 0 \\
    \sigma^2_{x_{out}}    &= \frac{(1-r^l_{x_{\text{in}}})^2(L-1)d_{\text{in}}\sigma^6_{x_{\text{in}}}\sigma^2_{qk}+ \frac{\exp{((1-r^l_{x_{\text{in}}})d^2_{\text{in}}\sigma^4_{x_{\text{in}}}\sigma^2_{qk})}(4(1-r^l_{x_{\text{in}}})^2 d_{\text{in}}\sigma^6_{x_{\text{in}}}\sigma^2_{qk} + (1-r^l_{x_{\text{in}}})\sigma_{x_{\text{in}}}^2)}{(1-p)}}{L} + r^l_{x_{\text{in}}}\sigma^2_{x_{\text{in}}}
\end{empheq}

Now to get covariance we make two approximations. As the term $\frac{\sum_{k=1}^L(1-d_{i,k})\exp{(\sum_{l=1}^{d_{\text{in}}}W_{i,l}\delta_{k,l})}}{(1-p)\sum_{k=1}^L\exp{(\sum_{l=1}^{d_{\text{in}}}W_{i,l}\delta_{k,l})}}$ converges to 1, we approximate $v_{1_{i,j}} \approx \epsilon_j$. Also we will treat $\sum_{k=1}^L\exp{(\sum_{l=1}^{d_{\text{in}}}W_{i,l}\delta_{k,l})} \approx \exp{(\frac{\sum_{l=1}^{d_{\text{in}}} W^2_{i,l}\sigma_{\delta}^2}{2})}$. Then, we have
\begin{align*}
    v_{1_{i,j}} &\approx \epsilon_j\\
    v_{2_{i,j}} &\approx \frac{\sum_{k=1}^L\frac{(1-d_{i,k})\exp{(\sum_{l=1}^{d_{\text{in}}}W_{i,l}\delta_{k,l})}\delta_{k,j}}{L}}{(1-p)\exp{(\frac{\sum_{l=1}^{d_{\text{in}}} W^2_{i,l}\sigma_{\delta}^2}{2})}}
    \intertext{This makes $v_{1_{i,j}}$ and $v_{2_{i,j}}$ independent. For covariance}
    \mathbb{E}[O_{i,j}O_{m,j}] &= \mathbb{E}_W[\mathbb{E}[O_{i,j}O_{m,j}|W]]\\
    O_{i,j}O_{m,j}|W &= (v_{1_{i,j}} + v_{2_{i,j}})(v_{1_{m,j}}+v_{2_{m,j}})\\
    &= v_{1_{i,j}}v_{1_{m,j}} + v_{1_{i,j}}v_{2_{m,j}} + v_{2_{i,j}}v_{1_{m,j}} + v_{2_{i,j}}v_{2_{m,j}}\\
    v_{1_{i,j}}v_{1_{m,j}} &= \epsilon_j^2\\
    \mathbb{E}[v_{1_{i,j}}v_{1_{m,j}}|W] &= \sigma_\epsilon^2\\
    \intertext{As $v_{1_{i,j}} = v_{1_{m,j}} = \epsilon_j, v_{1_{i,j}}v_{2_{m,j}} + v_{2_{i,j}}v_{1_{m,j}} = \epsilon_j(v_{2_{i,j}}+v_{2_{m,j}})$, and $\epsilon_j$ is independent of $(v_{2_{i,j}}+v_{2_{m,j}})$. Thus, we have}
    \mathbb{E}[v_{1_{i,j}}v_{2_{m,j}} + v_{2_{i,j}}v_{1_{m,j}}|W] &= \mathbb{E}[\epsilon_j|W]\mathbb{E}[(v_{2_{i,j}}+v_{2_{m,j}})|W] = 0*\mathbb{E}[(v_{2_{i,j}}+v_{2_{m,j}})|W] = 0\\
    v_{2_{i,j}}v_{2_{m,j}} &= \frac{\sum_{k_1=1}^L\frac{(1-d_{i,k_1})\exp{(\sum_{l=1}^{d_{\text{in}}}W_{i,l}\delta_{k_1,l})}\delta_{k_1,j}}{L}\sum_{k_2=1}^L\frac{(1-d_{m,k_2})\exp{(\sum_{l=1}^{d_{\text{in}}}W_{m,l}\delta_{k_2,l})}\delta_{k_2,j}}{L}}{(1-p)^2\exp{(\frac{\sum_{l=1}^{d_{\text{in}}} (W^2_{i,l}+W^2_{m,l})\sigma_{\delta}^2}{2})}}
\end{align*}
\begin{align*}
    \mathbb{E}[v_{2_{i,j}}v_{2_{m,j}}|W] = \frac{\mathbb{E}[\sum_{k_1=1}^L(1-d_{i,k_1})\exp{(\sum_{l=1}^{d_{\text{in}}}W_{i,l}\delta_{k_1,l})}\delta_{k_1,j}\sum_{k_2=1}^L(1-d_{m,k_2})\exp{(\sum_{l=1}^{d_{\text{in}}}W_{m,l}\delta_{k_2,l})}\delta_{k_2,j}]}{L^2(1-p)^2\exp{(\frac{\sum_{l=1}^{d_{\text{in}}} (W^2_{i,l}+W^2_{m,l})\sigma_{\delta}^2}{2})}}
\end{align*}
\begin{align*}
    \intertext{Breaking summation into two parts: $k_1 = k_2 = k$ and $k_1 \neq k_2$, we get}
    &= \frac{\mathbb{E}[\sum_{k=1}^L(1-d_{i,k})(1-d_{m,k})\exp{(\sum_{l=1}^{d_{\text{in}}}(W_{i,l}+W_{m,l})\delta_{k,l})}\delta^2_{k,j}]}{L^2(1-p)^2\exp{(\frac{\sum_{l=1}^{d_{\text{in}}} (W^2_{i,l}+W^2_{m,l})\sigma_{\delta}^2}{2})}}\\
    &\quad +\frac{\mathbb{E}[\sum_{k_1=1}^L\sum_{k_2=1,k_2\neq k_1}^L(1-d_{i,k_1})(1-d_{m,k_2})\exp{(\sum_{l=1}^{d_{\text{in}}}W_{i,l}\delta_{k_1,l})}\exp{(\sum_{l=1}^{d_{\text{in}}}W_{m,l}\delta_{k_2,l})}\delta_{k_1,j}\delta_{k_2,j}]}{L^2(1-p)^2\exp{(\frac{\sum_{l=1}^{d_{\text{in}}} (W^2_{i,l}+W^2_{m,l})\sigma_{\delta}^2}{2})}}\\
    &= \frac{\sum_{k=1}^L\mathbb{E}[(1-d_{i,k})(1-d_{m,k})\exp{(\sum_{l=1}^{d_{\text{in}}}(W_{i,l}+W_{m,l})\delta_{k,l})}\delta^2_{k,j}]}{L^2(1-p)^2\exp{(\frac{\sum_{l=1}^{d_{\text{in}}} (W^2_{i,l}+W^2_{m,l})\sigma_{\delta}^2}{2})}}\\
    &\quad +\frac{\sum_{k_1=1}^L\sum_{k_2=1,k_2\neq k_1}^L\mathbb{E}[(1-d_{i,k_1})(1-d_{m,k_2})\exp{(\sum_{l=1}^{d_{\text{in}}}W_{i,l}\delta_{k_1,l})}\exp{(\sum_{l=1}^{d_{\text{in}}}W_{m,l}\delta_{k_2,l})}\delta_{k_1,j}\delta_{k_2,j}]}{L^2(1-p)^2\exp{(\frac{\sum_{l=1}^{d_{\text{in}}} (W^2_{i,l}+W^2_{m,l})\sigma_{\delta}^2}{2})}}\\
\end{align*}
\begin{align*}
    \mathbb{E}[(1-d_{i,k})(1-d_{m,k})\exp{(\sum_{l=1}^{d_{\text{in}}}(W_{i,l}+W_{m,l})\delta_{k,l})}\delta^2_{k,j}] &= \\
    = \mathbb{E}[(1-d_{i,k})]&\mathbb{E}[(1-d_{m,k})]\mathbb{E}[\exp{(\sum_{l=1}^{d_{\text{in}}}(W_{i,l}+W_{m,l})\delta_{k,l})}\delta^2_{k,j}]\\
    = (1-p)^2&\exp{(\frac{\sum_{l=1}^{d_{\text{in}}} (W_{i,l}+W_{m,l})^2\sigma_{\delta}^2}{2})}((W_{i,j}+W_{m,j})^2\sigma_{\delta}^4 + \sigma_{\delta}^2)
\end{align*}
\begin{align*}
    \mathbb{E}[(1-d_{i,k_1})(1-d_{m,k_2})\exp{(\sum_{l=1}^{d_{\text{in}}}W_{i,l}\delta_{k_1,l})}&\exp{(\sum_{l=1}^{d_{\text{in}}}W_{m,l}\delta_{k_2,l})}\delta_{k_1,j}\delta_{k_2,j}] = \\ 
    \mathbb{E}[(1-d_{i,k_1})]\mathbb{E}[(1-d_{m,k_2})]& \mathbb{E}[\exp{(\sum_{l=1}^{d_{\text{in}}}W_{i,l}\delta_{k_1,l})}\delta_{k_1,j}]\mathbb{E}[\exp{(\sum_{l=1}^{d_{\text{in}}}W_{m,l}\delta_{k_2,l})}\delta_{k_2,j}]\\
    &= (1-p)^2\exp{(\frac{\sum_{l=1}^{d_{\text{in}}} (W_{i,l}^2+W_{m,l}^2)\sigma_{\delta}^2}{2})}W_{i,j}W_{m,j}\sigma_\delta^4
\end{align*}
\begin{align*}
    \mathbb{E}[v_{2_{i,j}}v_{2_{m,j}}|W] &= \frac{\exp{(\frac{\sum_{l=1}^{d_{\text{in}}} (W_{i,l}+W_{m,l})^2\sigma_{\delta}^2}{2})}((W_{i,j}+W_{m,j})^2\sigma_{\delta}^4 + \sigma_{\delta}^2)}{L\exp{(\frac{\sum_{l=1}^{d_{\text{in}}} (W^2_{i,l}+W^2_{m,l})\sigma_{\delta}^2}{2})}}\\
    &\quad + \frac{(L-1)\exp{(\frac{\sum_{l=1}^{d_{\text{in}}} (W_{i,l}^2+W_{m,l}^2)\sigma_{\delta}^2}{2})}W_{i,j}W_{m,j}\sigma_\delta^4}{L\exp{(\frac{\sum_{l=1}^{d_{\text{in}}} (W^2_{i,l}+W^2_{m,l})\sigma_{\delta}^2}{2})}}\\
    &= \frac{\exp{(\sum_{l=1}^{d_{\text{in}}} W_{i,l}W_{m,l}\sigma_{\delta}^2)}((W_{i,j}+W_{m,j})^2\sigma_{\delta}^4 + \sigma_{\delta}^2)}{L} + \frac{(L-1)W_{i,j}W_{m,j}\sigma_\delta^4}{L}
\end{align*}
So, we have
\begin{align*}
    \mathbb{E}[O_{i,j}O_{m,j}|W] &= \sigma_\epsilon^2 + \frac{\exp{(\sum_{l=1}^{d_{\text{in}}} W_{i,l}W_{m,l}\sigma_{\delta}^2)}((W_{i,j}+W_{m,j})^2\sigma_{\delta}^4 + \sigma_{\delta}^2)}{L} + \frac{(L-1)W_{i,j}W_{m,j}\sigma_\delta^4}{L}\\
    \mathbb{E}[O_{i,j}O_{m,j}] &= \mathbb{E}_W[\mathbb{E}[O_{i,j}O_{m,j}|W]]\\
    &= \mathbb{E}_W[\sigma_\epsilon^2 + \frac{\exp{(\sum_{l=1}^{d_{\text{in}}} W_{i,l}W_{m,l}\sigma_{\delta}^2)}((W_{i,j}+W_{m,j})^2\sigma_{\delta}^4 + \sigma_{\delta}^2)}{L} + \frac{(L-1)W_{i,j}W_{m,j}\sigma_\delta^4}{L}]\\
    &= \sigma_\epsilon^2 + \mathbb{E}_W[\frac{\exp{(\sum_{l=1}^{d_{\text{in}}} W_{i,l}W_{m,l}\sigma_{\delta}^2)}((W_{i,j}+W_{m,j})^2\sigma_{\delta}^4 + \sigma_{\delta}^2)}{L}]
    \intertext{For large values of $d_{\text{in}}$ by WLLN and continuous mapping theorem we have $\exp{(\sum_{l=1}^{d_{\text{in}}} W_{i,l}W_{m,l}\sigma_{\delta}^2)} \approx 1$. Thus, we have}
    \mathbb{E}[O_{i,j}O_{m,j}] &= \sigma_\epsilon^2 + \frac{(2\sigma_w^2\sigma_\delta^4 + \sigma_\delta^2)}{L}\\
    \mathbb{E}[O_{i,j}O_{m,j}] &= r^l_{x_{\text{in}}}\sigma^2_{x_{\text{in}}} + \frac{(2(1-r^l_{x_{\text{in}}})^2 d_{\text{in}}\sigma^6_{x_{\text{in}}}\sigma^2_{qk} + (1-r^l_{x_{\text{in}}})\sigma_{x_{\text{in}}}^2)}{L}\\
\end{align*}



The convergence arguments we have made require the scale of the variables to be small when compared to $L$ and $d_{\text{in}}$. The growth in scale can be controlled easily by controlling $\sigma_{qk}$, and we observe that if we let $\sigma_{qk}$ become arbitrarily large the scores passed to Softmax diverge leading to degenerate attention only attending to one token which has the highest score. To avoid this degenerate attention, we choose smaller values of $\sigma_{q}, \sigma_{k}$ and in that scenario, the approximate value for variance and covariance are,
\begin{empheq}[box=\fbox]{align*}
    \sigma^2_{x_{out}}  &\approx  r^l_{x_{\text{in}}}\sigma^2_{x_{\text{in}}} \\
    \mathrm{Cov}^l_{x_{out}} &\approx  r^l_{x_{\text{in}}}\sigma^2_{x_{\text{in}}}
\end{empheq}
To get the final variance and covariance we can use results of Linear layer to account for $\mathbf{W_V}$.
If we initialize $\sigma_{q}$ and $\sigma_{k}$ to be small, in initial phase of training the output of Softmax layer can be treated as being a constant $= \mathbf{\frac{1^T_{L}1_{L}}{L}}$. Using this assumption we have,
\begin{align*}
    \mathbf{X_{\text{out}}} &\approx \mathrm{Dropout}(\mathbf{\frac{1^T_{L}1_{L}}{L}})\mathbf{X_{\text{in}}}\mathbf{W_V}\\
    \implies \mathbf{g_{X_{\text{in}}}} &\approx \mathrm{Dropout}(\mathbf{\frac{1^T_{L}1_{L}}{L}})^T\mathbf{g_{X_{\text{out}}}}\mathbf{W_V}^T\\
    &= \mathrm{Dropout}(\mathbf{\frac{1^T_{L}1_{L}}{L}})\mathbf{g_{X_{\text{out}}}}\mathbf{W_V}^T
\end{align*}
\begin{empheq}[box=\fbox]{align*}
    \mu_{g_{{\text{in}}}} &= 0\\
    \sigma^2_{g_{{\text{in}}}} &= \frac{\sigma^2_{g_{{\text{out}}}}d\sigma^2_{v}}{L(1-p)}(1+(L-1)r^l_{g_{{\text{out}}}}(1-p))\\
    \mathrm{Cov}^l_{g_{{\text{in}}}} &= \frac{\sigma^2_{g_{{\text{out}}}}d\sigma^2_{v}}{L}(1+(L-1)r^l_{g_{{\text{out}}}})
\end{empheq}

\section{Moment Propagation through Transformer Blocks}
\label{appendix:section:Moment_prop_blocks}

\subsection{Transformer Attention Block}
\label{proof: attn_block}
A forward pass through attention block consists of LayerNorm, followed by Scaled Dot-Product Attention, followed by an output projection layer (a Linear Layer), and finally a Dropout. Using the results from above we get,
\begin{align*}
    & \mu_{x_{\text{out}}} = 0*0*0*0 = 0 \\
    & \sigma_{x_{\text{out}}}^2 \\
    &= \left(\frac{(1-r^l_{x_{\text{in}}})^2(L-1)d_{\text{in}}\sigma^6_{x_{\text{in}}}\sigma^2_q\sigma^2_k+ \frac{\exp{((1-r^l_{x_{\text{in}}})d^2_{\text{in}}\sigma^4_{x_{\text{in}}}\sigma^2_q\sigma^2_k)}(4(1-r^l_{x_{\text{in}}})^2 d_{\text{in}}\sigma^6_{x_{\text{in}}}\sigma^2_q\sigma^2_k + (1-r^l_{x_{\text{in}}})\sigma_{x_{\text{in}}}^2)}{(1-p)}}{L} + r^l_{x_{\text{in}}}\sigma^2_{x_{\text{in}}}\right).d_{\text{in}}\sigma^2_v.\frac{d_{\text{in}}\sigma^2_o}{(1-p)}\\
    &= \dfrac{d_{\text{in}}^2\sigma^2_o\sigma^2_v\sigma_{x_{\text{in}}}^2}{(1-p)}\left(\frac{(1-r^l_{x_{\text{in}}})^2(L-1)d_{\text{in}}\sigma^4_{x_{\text{in}}}\sigma^2_q\sigma^2_k+ \frac{\exp{((1-r^l_{x_{\text{in}}})d^2_{\text{in}}\sigma^4_{x_{\text{in}}}\sigma^2_q\sigma^2_k)}(4(1-r^l_{x_{\text{in}}})^2 d_{\text{in}}\sigma^4_{x_{\text{in}}}\sigma^2_q\sigma^2_k + (1-r^l_{x_{\text{in}}}))}{(1-p)}}{L} + r^l_{x_{\text{in}}}\right)\\ \\
    & \mathrm{Cov}^l_{x_{\text{out}}} \\
    &= \left(r^l_{x_{\text{in}}}\sigma^2_{x_{\text{in}}} + \frac{(2(1-r^l_{x_{\text{in}}})^2 d_{\text{in}}\sigma^6_{x_{\text{in}}}\sigma^2_q\sigma^2_k + (1-r^l_{x_{\text{in}}})\sigma_{x_{\text{in}}}^2)}{L}\right).d_{\text{in}}\sigma^2_v.d_{\text{in}}\sigma^2_o.1\\
    &= d_{\text{in}}^2\sigma^2_o\sigma^2_v\sigma_{x_{\text{in}}}^2\left(r^l_{x_{\text{in}}} + \frac{(2(1-r^l_{x_{\text{in}}})^2 d_{\text{in}}\sigma^4_{x_{\text{in}}}\sigma^2_q\sigma^2_k + (1-r^l_{x_{\text{in}}}))}{L}\right)
\end{align*}
\begin{flalign*}
    \sigma^2_{g_{\text{in}}} &= \sigma^2_{g_{\text{out}}}*\frac{1}{(1-p)}*d_{\text{in}}\sigma_o^2*\frac{d_{\text{in}}\sigma^2_{v}}{L(1-p)}(1+(L-1)r^l_{g_{{\text{out}}}}(1-p)) &&\\
    & = \dfrac{d_{\text{in}}^2\sigma^2_{g_{{\text{out}}}}\sigma^2_{v}\sigma^2_{o}}{L(1-p)^2}(1+(L-1)r^l_{g_{{\text{out}}}}(1-p))&&\\
\end{flalign*}
\begin{flalign*}
    \mathrm{Cov}^l_{g_{\text{in}}} &= \sigma^2_{g_{\text{out}}}*1*d_{\text{in}}\sigma_o^2*\frac{d_{\text{in}}\sigma^2_{v}}{L}(1+(L-1)r^l_{g_{{\text{out}}}}) &&\\
    & = \dfrac{d_{\text{in}}^2\sigma^2_{g_{{\text{out}}}}\sigma^2_{v}\sigma^2_{o}}{L}(1+(L-1)r^l_{g_{{\text{out}}}})&&
\end{flalign*}

\subsection{Transformer FFN Block}
\label{proof: ffn_block}
A forward pass through the FFN block of a transfer has a LayerNorm, then a Linear layer from $d$ to $4d$, which is then passed through a ReLU gate, the output of which is the projected back to $d$ dimension using another Linear layer, and eventually passed through a Dropout. Again using the results from above we get,
\begin{align*}
    \mu_{x_{\text{out}}} &= 0 \tag*{(Last Linear Layer makes it 0)}\\
    \sigma^2_{x_{\text{out}}} &= 1*d_{\text{in}}\sigma_{w_1}^2*(\frac{\pi - 1}{2\pi} + \frac{1}{2\pi})*4d_{\text{in}}\sigma_{w_2}^2*\frac{1}{(1-p)}*\sigma_{x_{\text{in}}}^2\\
    &= \frac{2d_{\text{in}}^2\sigma_{w_1}^2\sigma_{w_2}^2}{(1-p)} \sigma_{x_{\text{in}}}^2\\
    \mathrm{Cov}^l_{x_{\text{out}}} &= d_{\text{in}}\sigma_{w_1}^2*(\dfrac{r^l_{x_{\text{in}}}}{4} + \dfrac{(1-(r^l_{x_{\text{in}}})^2)^{0.5}}{2\pi} + \dfrac{r^l_{x_{\text{in}}}  \sin^{-1}(r^l_{x_{\text{in}}})}{2\pi} - \dfrac{1}{2\pi} + \dfrac{1}{2\pi})*4d_{\text{in}}\sigma_{w_2}^2*\sigma_{x_{\text{in}}}^2\\
    &= 4d_{\text{in}}^2\sigma_{w_1}^2\sigma_{w_2}^2\sigma_{x_{\text{in}}}^2(\dfrac{r^l_{x_{\text{in}}}}{4} + \dfrac{(1-(r^l_{x_{\text{in}}})^2)^{0.5}}{2\pi} + \dfrac{r^l_{x_{\text{in}}}  \sin^{-1}(r^l_{x_{\text{in}}})}{2\pi}) \\
    \mathrm{r}^l_{x_{\text{out}}} &= 2*(1-p)*(\dfrac{r^l_{x_{\text{in}}}}{4} + \dfrac{(1-(r^l_{x_{\text{in}}})^2)^{0.5}}{2\pi} + \dfrac{r^l_{x_{\text{in}}}  \sin^{-1}(r^l_{x_{\text{in}}})}{2\pi})\\
    & \approx  (1-p) * (\dfrac{r^l_{x_{\text{in}}}}{2} + \dfrac{1}{\pi} + (\dfrac{1}{2} - \dfrac{1}{\pi}){r^l_{x_{\text{in}}}}^2)  \tag*{(Fitting a 2-nd order polynomial)} \\
    \sigma^2_{g_{\text{in}}} &= \sigma^2_{g_{\text{out}}}*\frac{1}{(1-p)}*d_{\text{in}}\sigma_{w_2}^2*\frac{1}{2}*4d_{\text{in}}\sigma_{w_1}^2\\
     &=\dfrac{2d_{\text{in}}^2\sigma_{w_1}^2\sigma_{w_2}^2\sigma_{g_{\text{out}}}^2}{(1-p)}\\
     \mathrm{Cov}^l_{g_{\text{in}}} &= \mathrm{Cov}^l_{g_{\text{out}}}*1*d_{\text{in}}\sigma_{w_2}^2*(\dfrac{1}{4} + \dfrac{\sin^{-1}(r^l_{x_{\text{in}}})}{2\pi})*4d_{\text{in}}\sigma_{w_1}^2\\
     &= 4d_{\text{in}}^2\sigma_{w_1}^2\sigma_{w_2}^2\mathrm{Cov}^l_{g_{\text{out}}} (\dfrac{1}{4} + \dfrac{\sin^{-1}(r^l_{x_{\text{in}}})}{2\pi})
\end{align*}

\section{Summary Table of Moment Propagation through Transformer Components}
\label{appendix:section:summary_table}

In \autoref{table:moment_forward_mean}, \autoref{table:moment_forward}, \autoref{table:moment_back_var}, \autoref{table:moment_cov_l}, \autoref{table:moment_cov_d} and \autoref{table:moment_cov_g}, we summarize the signal propagation formulae for all the transformer components.

\begin{table}[H]
\caption{Moment Propagation (mean) during forward pass through components of transformer model.}
\begin{center}
\begin{adjustbox}{max width=\textwidth}
\begin{tabular}{l c } 
 \toprule
 Component & $\mu_{x_{\text{out}}}$ \\ [1.0ex] 
 \midrule
    Embeddings &  0  \\ [2.5ex] 
    
    FFN ($d_1.d_2$)&  0  \\ [2.5ex] 
    
    ReLU &  $\dfrac{\sigma_{x_{\text{in}}}}{\sqrt{(2\pi)}}$   \\ [2.5ex] 

    GeLU &  $\dfrac{\sigma_{x_{\text{in}}}^2}{\sqrt{2\pi(\sigma_{x_{\text{in}}}^2+1)}}$  \\ [2.5ex] 
    
    LayerNorm ($d$)&  0  \\ [2.5ex]

    Dropout ($p$) & $\mu_{x_{\text{in}}}$    \\ [2.5ex] 

    Softmax &  $\frac{1}{L}$  \\ [2.5ex] 

    SHA Block (without V) &  0  \\ [2.5ex] 

    Attn Block &  0  \\ [2.5ex]
    
    FFN Block &  0  \\ [2.5ex]

\bottomrule
\end{tabular}
\end{adjustbox}
\end{center}
\label{table:moment_forward_mean}
\end{table}

\begin{table}[H]
\caption{Moment Propagation (variance) during forward pass through components of transformer model.}
\begin{center}
\begin{adjustbox}{max width=\textwidth}
\begin{tabular}{l c} 
 \toprule
 Component &  $\sigma^2_{x_{\text{out}}}$ \\ [1.0ex] 
 \midrule
    Embeddings & 0 \\ [2.5ex] 
    
    FFN ($d_1.d_2$) & $d_1  \sigma_w^2  (\sigma_{x_{\text{in}}}^2 + \mu_{x_{\text{in}}}^2)$  \\ [2.5ex] 
    
    ReLU & $\dfrac{(\pi-1)}{(2\pi)}  \sigma_{x_{\text{in}}}^2$  \\ [2.5ex] 

    GeLU & $\frac{\sigma^2_{x_{\text{in}}}}{2 \pi} (\frac{\pi}{2} - \frac{\sigma^2_{x_{\text{in}}}}{1+\sigma^2_{x_{\text{in}}}} + \sin^{-1}(\frac{\sigma^2_{x_{\text{in}}}}{1+\sigma^2_{x_{\text{in}}}}) + \frac{2 \sigma^2_{x_{\text{in}}}}{(1+\sigma^2_{x_{\text{in}}})\sqrt{1+2\sigma^2_{x_{\text{in}}}}})$ \\ [2.5ex] 
    
    Layer              Norm ($d$) & 1 \\ [2.5ex]

    Dropout ($p$) & $\dfrac{\sigma_{x_{\text{in}}}^2 + p  \mu_{x_{\text{in}}}^2}{1-p}$   \\ [2.5ex] 

    Softmax &  $\frac{(e^{\sigma^2_{x_{\text{in}}} (1 - r^l_{x_{\text{in}}}) \frac{L}{L-1}} - 1)e^{2\sigma^2_{x_{\text{in}}} (1 - r^l_{x_{\text{in}}}) \frac{L}{L-1}}}{((L-1) e^{\sigma^2_{x_{\text{in}}} (1 - r^l_{x_{\text{in}}})} + 1)^2}$ \\ [2.5ex] 

    SHA (without V) & $\dfrac{d_{\text{in}}\sigma_{x_{\text{in}}}^2}{(1-p)}\left(\frac{(1-r^l_{x_{\text{in}}})^2(L-1)d_{\text{in}}\sigma^4_{x_{\text{in}}}\sigma^2_q\sigma^2_k+ \frac{\exp{((1-r^l_{x_{\text{in}}})d^2_{\text{in}}\sigma^4_{x_{\text{in}}}\sigma^2_q\sigma^2_k)}(4(1-r^l_{x_{\text{in}}})^2 d_{\text{in}}\sigma^4_{x_{\text{in}}}\sigma^2_q\sigma^2_k + (1-r^l_{x_{\text{in}}}))}{(1-p)}}{L} + r^l_{x_{\text{in}}}\right)$  \\ [2.5ex] 

    Attn Block (Approx) &  $\dfrac{d_{\text{in}}^2\sigma^2_o\sigma^2_v\sigma_{x_{\text{in}}}^2}{(1-p)}\left(\frac{(1-r^l_{x_{\text{in}}})^2(L-1)d_{\text{in}}\sigma^4_{x_{\text{in}}}\sigma^2_q\sigma^2_k+ \frac{\exp{((1-r^l_{x_{\text{in}}})d^2_{\text{in}}\sigma^4_{x_{\text{in}}}\sigma^2_q\sigma^2_k)}(4(1-r^l_{x_{\text{in}}})^2 d_{\text{in}}\sigma^4_{x_{\text{in}}}\sigma^2_q\sigma^2_k + (1-r^l_{x_{\text{in}}}))}{(1-p)}}{L} + r^l_{x_{\text{in}}}\right)$ \\ [2.5ex]
    
    FFN Block &  $\dfrac{2d_{\text{in}}^2\sigma_{w_1}^2\sigma_{w_2}^2\sigma^2_{x_{\text{in}}}}{(1-p)}$ \\ [2.5ex]

\bottomrule
\end{tabular}
\end{adjustbox}
\end{center}
\label{table:moment_forward}
\end{table}

\begin{table}[H]
\caption{Moment Propagation (variance) during backwards pass through components of transformer model.}
\begin{center}
\begin{adjustbox}{max width=\textwidth}
\begin{tabular}{l c} 
 \toprule
 Component & $\sigma^2_{g_{\text{in}}}$\\ [1.0ex] 
 \midrule
    Embeddings &  - \\ [2.5ex] 
    
    FFN ($d_1.d_2$)&  $d_2  \sigma_w^2  \sigma_{g_{\text{out}}}^2 $ \\ [2.5ex] 
    
    ReLU &  $\dfrac{1}{2}\sigma_{g_{\text{out}}}^2$ \\ [2.5ex] 

    GeLU & $\left[\frac{1}{4} + \frac{1}{2\pi}\sin^{-1}{(\frac{\sigma_{x_{\text{in}}}^2}{\sigma_{x_{\text{in}}}^2 + 1})} + \frac{\sigma^2_{x_{\text{in}}}(5\sigma^2_{x_{\text{in}}} + 3)}{2\pi(\sigma_{x_{\text{in}}}^2 + 1)(2\sigma^2_{x_{\text{in}}} + 1)^{\frac{3}{2}}}\right]\sigma_{g_{\text{out}}}^2$ \\ [2.5ex] 
    
    LayerNorm ($d$)&   $\dfrac{\sigma_{g_{\text{out}}}^2}{\sigma_{x_{\text{in}}}^2}$ \\ [2.5ex]

    Dropout ($p$) & $\dfrac{1}{1-p}\sigma_{g_{\text{out}}}^2$ \\ [2.5ex] 

    Softmax & $ (\frac{(e^{\sigma^2_{x_{\text{in}}} (1 - r^l_{x_{\text{in}}}) \frac{L}{L-1}} - 1)e^{2\sigma^2_{x_{\text{in}}} (1 - r^l_{x_{\text{in}}}) \frac{L}{L-1}}}{((L-1) e^{\sigma^2_{x_{\text{in}}} (1 - r^l_{x_{\text{in}}})} + 1)^2} + \frac{1}{L^2})\sigma_{g_{\text{out}}}^2 $ \\ [2.5ex] 

    SHA Block (without V) & $\dfrac{d_{\text{in}}\sigma^2_{g_{{\text{out}}}}}{L(1-p)^2}(1+(L-1)r^l_{g_{{\text{out}}}}(1-p))$ \\ [2.5ex] 

    Attn Block (Approx) & $\dfrac{d_{\text{in}}^2\sigma^2_{g_{{\text{out}}}}\sigma^2_{v}\sigma^2_{o}}{L(1-p)^2}(1+(L-1)r^l_{g_{{\text{out}}}}(1-p))$ \\ [2.5ex]
    
    FFN Block & $\dfrac{2d_{\text{in}}^2\sigma_{w_1}^2\sigma_{w_2}^2\sigma_{g_{\text{out}}}^2}{(1-p)}$ \\ [2.5ex]

\bottomrule
\end{tabular}
\end{adjustbox}
\end{center}
\label{table:moment_back_var}
\end{table}

\begin{table}[H]
\caption{Covariance (along sequence length) propagation through the components of transformer model.}
\begin{center}
\begin{adjustbox}{max width=\textwidth}
\begin{tabular}{l c} 
 \toprule
 Component & $\mathrm{Cov}^l_{x_{\text{out}}}$ \\ [1.0ex] 
 \midrule
    Embeddings & $\sum{\dfrac{N_i * (N_i - 1)}{L*(L-1))}} * \sigma_{w_{\text{embd}}}^2$  \\ [2.5ex] 
    
    FFN ($d_1.d_2$) & $d_1 \sigma_w^2  (\mathrm{Cov}^l_{x_{\text{in}}} + \mu_{x_{\text{in}}}^2)$ \\ [2.5ex] 
    
    ReLU & $ (\dfrac{1}{4} + \dfrac{\sin^{-1}{(r^l_{x_{\text{in}}})}}{2\pi})\mathrm{Cov}^l_{x_{\text{in}}} - (1 - \sqrt{(1-(r^l_{x_{\text{in}}})^2)})\dfrac{\sigma^2_{x_{\text{in}}}}{2\pi} $ \\ [2.5ex]

    GeLU & $\frac{\sigma_{x_{\text{in}}}^2}{4 \pi} (\pi r^l_{x_{\text{in}}} + 2r^l_{x_{\text{in}}}\sin^{-1}{(\frac{r^l_{x_{\text{in}}}\sigma_{x_{\text{in}}}^2}{\sigma_{x_{\text{in}}}^2 + 1})} + \frac{2\sigma_{x_{\text{in}}}^2(\sigma_{x_{\text{in}}}^2(1-(r^l_{x_{\text{in}}})^2) + 1 + (r^l_{x_{\text{in}}})^2)}{(\sigma_{x_{\text{in}}}^2+1)\sqrt{(\sigma_{x_{\text{in}}}^2 + 1)^2 - (r^l_{x_{\text{in}}}\sigma_{x_{\text{in}}}^2)^2}} - \frac{2\sigma_{x_{\text{in}}}^2}{(\sigma_{x_{\text{in}}}^2 + 1)} )$ \\ [2.5ex]
    
    LayerNorm ($d$)& $ (1-\dfrac{1}{d})\dfrac{\mathrm{Cov}^l_{x_{\text{in}}}}{\sigma_{x_{\text{in}}}^2} $ \\ [2.5ex]

    Dropout ($p$) & $\mathrm{Cov}^l_{x_{\text{in}}}$ \\ [2.5ex] 


    SHA (without V) & $d_{\text{in}}\sigma_{x_{\text{in}}}^2\left(r^l_{x_{\text{in}}} + \frac{(2(1-r^l_{x_{\text{in}}})^2 d_{\text{in}}\sigma^4_{x_{\text{in}}}\sigma^2_q\sigma^2_k + (1-r^l_{x_{\text{in}}}))}{L}\right)$  \\ [2.5ex] 

    Attn Block (Approx) & $d_{\text{in}}^2\sigma^2_o\sigma^2_v\sigma_{x_{\text{in}}}^2\left(r^l_{x_{\text{in}}} + \frac{(2(1-r^l_{x_{\text{in}}})^2 d_{\text{in}}\sigma^4_{x_{\text{in}}}\sigma^2_q\sigma^2_k + (1-r^l_{x_{\text{in}}}))}{L}\right)$  \\ [2.5ex] 
    
    FFN Block & $4d_{\text{in}}\sigma_{w_1}^2\sigma_{w_2}^2\sigma_{x_{\text{in}}}^2(\dfrac{r^l_{x_{\text{in}}}}{4} + \dfrac{\sqrt{(1-(r^l_{x_{\text{in}}})^2}}{2\pi} + \dfrac{r^l_{x_{\text{in}}}  \sin^{-1}(r^l_{x_{\text{in}}})}{2\pi}) $ \\ [2.5ex] 
\bottomrule
\end{tabular}
\end{adjustbox}
\end{center}
\label{table:moment_cov_l}
\end{table}

\begin{table}[H]
\caption{Covariance (hidden dimension) propagation through the components of transformer model.}
\begin{center}
\begin{adjustbox}{max width=\textwidth}
\begin{tabular}{l c} 
 \toprule
 Component & $\mathrm{Cov}^d_{x_{\text{out}}}$  \\ [1.0ex] 
 \midrule
    Embeddings & 0 \\ [2.5ex] 
    
    FFN ($d_1.d_2$) & 0 \\ [2.5ex] 
    
    ReLU & $(\dfrac{1}{4} + \dfrac{\sin^{-1}{(r^d_{x_{\text{in}}})}}{2\pi})\mathrm{Cov}^d_{x_{\text{in}}} - (1 - \sqrt{(1-(r^d_{x_{\text{in}}})^2)})\dfrac{\sigma^2_{x_{\text{in}}}}{2\pi} $ \\ [2.5ex] 

    GeLU & \\ [2.5ex] 
    
    LayerNorm ($d$) & $-\dfrac{1}{d-1}$  \\ [2.5ex]

    Dropout ($p$) & $\mathrm{Cov}^d_{x_{\text{in}}}$ \\ [2.5ex] 


    SHA Block(without V ) & 0 \\ [2.5ex]
    
    Attn Block & 0 \\ [2.5ex] 
    
    FFN Block &  0 \\ [2.5ex] 
\bottomrule
\end{tabular}
\end{adjustbox}
\end{center}
\label{table:moment_cov_d}
\end{table}

\begin{table}[H]
\caption{Gradient covariance (along sequence length) propagation through the components of transformer model.}
\begin{center}
\begin{adjustbox}{max width=\textwidth}
\begin{tabular}{l c} 
 \toprule
 Component & $\mathrm{Cov}^l_{g_{\text{in}}}$ \\ [1.0ex] 
 \midrule
    Embeddings & - \\ [2.5ex] 
    
    FFN ($d_1.d_2$) & $d_2 \sigma_w^2  \mathrm{Cov}^l_{g_{\text{out}}}$ \\ [2.5ex] 
    
    ReLU & $(\dfrac{1}{4} + \dfrac{\sin^{-1}{(r^l_{x_{\text{in}}})}}{2\pi})\mathrm{Cov}^l_{g_{\text{out}}}$\\ [2.5ex] 

    GeLU & $\left[\frac{1}{4} + \frac{1}{2\pi}\sin^{-1}{(\frac{r^l_{x_{\text{in}}}\sigma_{x_{\text{in}}}^2}{\sigma_{x_{\text{in}}}^2 + 1})}  + \frac{r^l_{x_{\text{in}}}\sigma_{x_{\text{in}}}^2((2\sigma_{x_{\text{in}}}^2 + 3)(\sigma_{x_{\text{in}}}^2 + 1) - 2(r^l_{x_{\text{in}}}\sigma_{x_{\text{in}}}^2)^2)}{2\pi(\sigma_{x_{\text{in}}}^2 + 1)((\sigma_{x_{\text{in}}}^2 + 1)^2 - (r^l_{x_{\text{in}}}\sigma_{x_{\text{in}}}^2)^2)^{\frac{3}{2}}}\right]r^l_{g_{\text{out}}}\sigma_{g_{{\text{out}}}}^2$ \\ [2.5ex]
    
    LayerNorm ($d$) & $\dfrac{\mathrm{Cov}^l_{g_{\text{out}}}}{\sigma_{x_{\text{in}}}^2}$\\ [2.5ex]

    Dropout ($p$) & $\mathrm{Cov}^l_{g_{\text{out}}}$\\ [2.5ex] 


    SHA Block (without V)  & $\dfrac{d_{\text{in}}\sigma^2_{g_{{\text{out}}}}}{L}(1+(L-1)r^l_{g_{{\text{out}}}})$\\ [2.5ex] 
    
    Attn Block (Approx) & $\dfrac{d_{\text{in}}^2\sigma^2_{g_{{\text{out}}}}\sigma^2_{v}\sigma^2_{o}}{L}(1+(L-1)r^l_{g_{{\text{out}}}})$\\ [2.5ex] 
    
    FFN Block & $4d_{\text{in}}^2\sigma_{w_1}^2\sigma_{w_2}^2\mathrm{Cov}^l_{g_{\text{out}}} (\dfrac{1}{4} + \dfrac{\sin^{-1}(r^l_{x_{\text{in}}})}{2\pi})$\\ [2.5ex] 
\bottomrule
\end{tabular}
\end{adjustbox}
\end{center}
\label{table:moment_cov_g}
\end{table}

\section{Numerical Verification}
\label{appendix:section:numerical_verification}

We perform numerical verification for the formulae reported in \autoref{table:moment_forward_mean}, \autoref{table:moment_forward}, \autoref{table:moment_back_var}, \autoref{table:moment_cov_l}, \autoref{table:moment_cov_d} and \autoref{table:moment_cov_g}. The parameter ranges have been provided in \autoref{table:verify_range}. For each parameter, 3-5 values were sampled uniformly (or log uniformly) across the range for numerical simulation. \autoref{table:verify_errors} provides the percentage error corresponding to the $50_{th}$, $90_{th}$ and $99_{th}$ percentile. These simulation results are all fully reproducible using our released code. Even at 99 percentile, no error (other than SHA backwards) is larger than $10\%$, verifying our assumptions.

\begin{table}[H]
\caption{Percentage Errors [50th, 90th, 99th percentile] for the theoretical formulas corresponding to forward and backward pass through components of the transformer model.}
\begin{center}
\begin{adjustbox}{max width=\textwidth}
\begin{tabular}{l c c c c c} 
 \toprule
 Component & $\mu_{x_{\text{out}}}$ & $\sigma^2_{x_{\text{out}}}$ & $\sigma^2_{g_{\text{in}}}$ & $\mathrm{Cov}^l_{x_{\text{out}}}$ & $\mathrm{Cov}^l_{g_{\text{in}}}$ \\ [1.0ex] 
 \midrule
    FFN & [0.0, 0.4, 1.3] & [0.4, 1.4, 2.8] & [0.2, 1.0, 2.2] & [0.4, 1.4, 2.8] & [0.2, 1.0, 2.2] \\ [1.0ex]

    ReLU & [0.3, 1.3, 2.3] & [0.5, 1.9, 3.4] & [0.6, 1.5, 2.6] & [0.3, 1.6, 3.1] & [0.2, 1.1, 2.3] \\ [1.0ex]
    
    GeLU & [0.1, 1.0, 2.4] & [0.2, 0.6, 1.3] & [0.2, 0.6, 1.1] & [0.1, 0.5, 1.2] & [0.1, 0.4, 0.9] \\ [1.0ex]

    LayerNorm & [0.0 , 0.0, 0.0] & [0.0, 0.0, 0.0] & [0.4, 1.5, 3.2] & [0.1, 0.5, 1.0] & [0.2, 0.9, 2.2] \\ [1.0ex]

    Dropout & [0.0, 0.1, 0.5] & [0.1, 0.5, 1.5] & [0.1, 0.7, 1.5] & [0.0, 0.4, 1.3] & [0.1, 0.5, 1.2] \\ [1.0ex]

    Softmax & [0.0 , 0.0, 0.0] &  [0.2, 0.9, 4.0] & [0.1, 0.6, 4.5] & - & - \\ [1.0ex]
    
    Single-Head Atten. & [0.2, 1.0, 2.5] & [1.4, 4.1, 7.8] & [2.2, 13.3, 44.5] & [1.3, 3.9, 7.4] & [1.6, 4.5, 8.2] \\

\bottomrule
\end{tabular}
\end{adjustbox}
\end{center}
\label{table:verify_errors}
\end{table}
\begin{table}[H]
\caption{Range of input variance/correlations used for theoretical formula verification reported in \autoref{table:verify_errors} for the theoretical formulas corresponding to forward and backward pass through components of the transformer model. The dropout probability range was $[0,1)$ for Dropout and Single-Head Attention, and $\sigma^2_{w}$ for FFN was $[10^{-2}, 10^2]/d_\text{in}$.}
\begin{center}
\begin{adjustbox}{max width=\textwidth}
\begin{tabular}{l c c c c c c c c} 
 \toprule
 Component & $\mu_{x_{\text{in}}}$ & $\sigma^2_{x_{\text{in}}}$ & $\sigma^2_{g_{\text{out}}}$ & $\mathrm{Corr}^l_{x_{\text{in}}}$ & $\mathrm{Corr}^l_{g_{\text{out}}}$ & $d_\text{in}$ & $d_\text{out}$ & $L$  \\ [1.0ex] 
 \midrule
    FFN & [-10, 10] & [0.1, 10] & [0.1, 10] & [0, 1.0) & [0, 1.0) &  [$10^1$, $10^3$] & [$10^1$, $10^3$] & [$10^2$, $10^3$] \\ [1.0ex]

    ReLU & [0] & [0.1, 10] & [0.1, 10] & [0, 1.0) & [0, 1.0) & - & - & [$10^2$, $10^3$]\\ [1.0ex]
    
    GeLU & [0] & [0.1, 10] & [0.1, 10] & [0, 1.0) & [0, 1.0) & - & - & [$10^2$, $10^3$]\\ [1.0ex]

    LayerNorm & [-10, 10] & [0.1, 10] & [0.1, 10] & [0, 1.0) & [0, 1.0) & [$10^2$, $10^3$] & - & [$10^2$, $10^3$]\\ [1.0ex]

    Dropout & [-10, 10] & [0.1, 10] & [0.1, 10] & [0, 1.0) & [0, 1.0) & [$10^2$, $10^3$] & - & [$10^2$, $10^3$]\\ [1.0ex]

    Softmax & [0] & [$10^{-4}$, 1] & [0.1, 10] & [0, 1.0) & - & - & - & [300, $10^4$] \\ [1.0ex]
    
    Single-Head Atten. & [0] & [1] & [0.1, 10] & [0, 1.0) & [0, 1.0) & [$10^2$, $10^3$] & [32, 64, 128, 256] & [300, $10^4$] \\

\bottomrule
\end{tabular}
\end{adjustbox}
\end{center}
\label{table:verify_range}
\end{table}

\section{Moment Propagation through the Entire Transformer Model}
\label{appendix:section:Moment_prop_model}

\subsection{Vanilla Pre-LN}
\label{proof: vanilla preLN}

We will use the approximations listed in \cref{table:block_moments} here.

\subsubsection{Forward Pass}

For forward pass, a Transformer Pre-LN has LayerNorm followed by the Attention block, residual connection, LayerNorm, and then the FFN block. Let $\sigma^2_{\text{layer}}$ be the output variance after $1$ such layer, and $\sigma^2_{\text{model}}$ be the output variance after the entire model of $N$ layers.

\begin{align*}
\sigma^2_{x_\text{attn}} &= \frac{d^2\sigma_o^2\sigma_v^2 *r^l_{x_{\text{in}}}}{(1-p)} \\
\sigma^2_{x_\text{ffn}} &= \frac{2d^2\sigma_{w_1}^2\sigma_{w_2}^2}{(1-p)} \\
\sigma^2_{x_\text{layer}} &= \sigma^2_{x_{\text{in}}} + \sigma^2_{x_\text{attn}} + \sigma^2_{x_\text{ffn}} \\
 &= \sigma^2_{x_{\text{in}}} + \frac{d^2\sigma_o^2\sigma_v^2 *r^l_{x_{\text{in}}}}{(1-p)} + \frac{2d^2\sigma_{w_1}^2\sigma_{w_2}^2}{(1-p)} \\
\text{Let, } C_1 &= \frac{d^2\sigma_o^2\sigma_v^2}{(1-p)}, C_2=\frac{2d^2\sigma_{w_1}^2\sigma_{w_2}^2}{(1-p)bu}, \\
\text{Then, } \sigma^2_{x_\text{layer}} &= \sigma^2_{x_{\text{in}}} + C_1*r^l_{x_{\text{in}}} + C_2 \\
\end{align*}

As we discuss in \cref{rank_collapse}, the correlation $r^l_{x_{\text{in}}}$ quickly reaches a stable constant maximum value $r^l_{x_{\text{max}}}$, which can be found using the calculations in \cref{appendix:section:correlation_analysis}. Let $r^l_{x_{\text{min}}} > 0$ be the minimum value of this correlation, let $C_3 = C_1*r^l_{x_{\text{max}}} + C_2 $, and $C_4 = C_1*r^l_{x_{\text{min}}} + C_2 $. Then,

\begin{align*}
\sigma^2_{x_{\text{in}}} + C_4 \leq \sigma^2_{x_\text{layer}} & \leq \sigma^2_{x_{x_\text{in}}} + C_3
\end{align*}

Hence after N layers,
\begin{align*}
\sigma^2_{x_{\text{in}}} + N*C_4 \leq \sigma^2_{x_\text{model}} &\leq \sigma^2_{x_{\text{in}}} + N*C_3
\end{align*}
\begin{align}
\implies \sigma^2_{x_\text{model}} &= \Theta(N)
\end{align}

This shows that output variance of Pre-LN will increase linearly with number of layers $N$.

In practice, because the correlation quickly reaches $r^l_{x_{\text{max}}}$, the variance of the entire model $\sigma^2_{x_\text{model}} \approx \sigma^2_{x_{\text{in}}} + N*C_3$.

\paragraph{Discussion:} This has the effect that transformer blocks near the output can affect the model output much less, as the skip connection variance increases but block output variance is constant. We conjecture that parameters in these are hence not being utilized to their full potential. Specifically in case of Xavier initialization, $C_1 = 2.2, C_2=0.4, r^l_{x_{\text{max}}}=0.85$. For large $d$, $\sigma^2_{x_{\text{in}}}$ will be negligibly small compared to $\sigma^2_{x_\text{layer}}$, so we have - 

\begin{align*}
\sigma^2_{x_\text{model}} \approx C_3*N \approx (2.2*0.85 + 0.4)N \approx 2.2N
\end{align*}


\subsubsection{Backward Pass}

For the backward pass, a Transformer Pre-LN gradient will first backpropagate through the FFN block, then gets rescaled by Layernorm, and added with the skip connection. It then backpropagates through the Attention block, gets rescaled by Layernorm, and finally added with the skip connection. Let $\sigma^2_{g, n}$ be the gradient variance backpropagating from the $n^{th}$ layer, and $\sigma^2_{g_\text{model}}$ be the gradient variance after the entire model of $N$ layers.

For the Attention block, let $\sigma^2_{g_\text{attn}, n-1}$ be the gradient backpropagating from the block. Then for long sequence length $L$ we have - 

\begin{align*}
\sigma^2_{g_\text{attn}, n-1} &= \frac{d^2\sigma_o^2\sigma_v^2*\sigma^2_{g_{\text{out},n}}}{L(1-p)}(1+(L-1)r^l_{g_{\text{out}}, n}) \\
&\approx \frac{d^2\sigma_o^2\sigma_v^2 * r^l_{g_{\text{out}}, l}*\sigma^2_{g_{\text{out}},n}}{(1-p)}
\end{align*}

$\sigma^2_{g_\text{attn}, n-1}$ is then rescaled by the Layernorm to give $\sigma^2_{g_\text{attn-layernorm}, n-1}$. As Layernorm scales gradient by the inverse of the input variance $\sigma^2_{x_{\text{in}}, n-1}$, which from the section above, we know is approximately $\sigma^2_{x_{\text{in}}, n-1} = C_3*(n-1)$. Then 

\begin{align*}
    \sigma^2_{g_\text{attn}, n-1} &= C_1 * r^l_{g_{\text{out}}, n} *\sigma^2_{g_{\text{out}},n} \\
    \sigma^2_{g_\text{attn-layernorm}, n-1} &= \frac{C_1 * r^l_{g_{\text{out}}, n} *\sigma^2_{g_{\text{out}},n}}{\sigma^2_{x_{\text{in}}, n-1}} \\
    &\approx \frac{C_1 * r^l_{g_{\text{out}}, n} *\sigma^2_{g_{\text{out}},n}}{C_3*(n-1)} \\
\end{align*}

Therefore, the final gradient $\sigma^2_{g_\text{attn-layer}, n-1}$ after addition with the skip connection is 

\begin{align*}
    \sigma^2_{g_\text{attn-layer}, n-1} =( 1 + \frac{C_1 * r^l_{g_{\text{out}}, n} }{C_3*(n-1)}) \sigma^2_{g_{\text{out}},n}
\end{align*}

Similarly, we can get $\sigma^2_{g_\text{ffn-layer}, n-1}$ for the ffn block. Then to get the gradient backpropagated through the entire layer $\sigma^2_{g_{\text{out}},n-1}$, we have,

\begin{align*}
    \sigma^2_{g_\text{ffn-layer}, n-1} &= ( 1 + \frac{C_2}{C_3*(n-1)}) \sigma^2_{g_{\text{out}},n} \\
    \sigma^2_{g_{\text{out}},n-1} &=( 1 + \frac{C_1 * r^l_{g_{\text{out}}, n} }{C_3*(n-1)})( 1 + \frac{C_2}{C_3*(n-1)}) \sigma^2_{g_{\text{out}},n} \\
    \sigma^2_{g_{\text{out}},n-1} &\approx ( 1 + \frac{C_1 * r^l_{g_{\text{out}}, n} }{C_3*(n-1)} + \frac{C_2}{C_3*(n-1)}) \sigma^2_{g_{\text{out}},n} \\
    &= ( 1 + \frac{C_1 * r^l_{g_{\text{out}}, n} +C_2}{C_3*(n-1)}) \sigma^2_{g_{\text{out}},n} \\
    &= ( 1 + \frac{C_1 * r^l_{g_{\text{out}}, n} +C_2}{(C_1 * r^l_{x_{\text{in}}, n} +C_2)*(n-1)}) \sigma^2_{g_{\text{out}},n} \\
    &= (1 + \frac{C_{g_{pre},n}}{n-1})  \sigma^2_{g_{\text{out}},n}
\end{align*}


Where we ignore higher order terms for large $n$, and define $C_{g_{pre},n} = \frac{C_1 * r^l_{g_{\text{out}}, n} +C_2}{C_1 * r^l_{x_{\text{in}}, n} +C_2}$. 

Since $C_{g_{pre},n}>0$, we will witness an increase in gradient going backward, and this increase is inversely proportional to the current layer $n$, matching with empirically observed growth~(\cref{fig:var_bck_pre}).

Let $C_{g_{pre},min}=\frac{C_2}{C_1 +C_2}=0.15$ be the minimum value of $C_{g_{pre},n}$, and $C_{g_{pre},max}=\frac{C_1+C_2}{C_2}=6.5$ be the maximum. Then the above equation is bounded by:
\begin{align*}
(1 + \frac{C_{g_{pre},min}}{n-1})  \sigma^2_{g_{\text{out}},n} \leq \sigma^2_{g_{\text{out}},n-1} \leq (1 + \frac{C_{g_{pre},min}}{n-1})  \sigma^2_{g_{\text{out}},max}
\end{align*}

Applying the above equation repeatedly until the final layer $N$, this recurrence can be approximately solved by treating $\sigma^2_{g_{\text{out}},n}$ as a continuous function of $n$, taking logarithm of both sides, and integrating. This gives the following solution for $\sigma^2_{g_{\text{out}},n}$:

\begin{align*}
 \sigma^2_{g_{\text{out}},N} * {(\frac{N}{n})}^{C_{g_{pre},min}} \leq\sigma^2_{g_{\text{out}},n} \leq \sigma^2_{g_{\text{out}},N} * {(\frac{N}{n})}^{C_{g_{pre},max}}
\end{align*}

If the correlation $r^l_{g_{\text{out}}, n}$ quickly reaches a stable constant maximum value $r^l_{g_{\text{max}}}$ (approximately equal to but slightly less than $r^l_{x_{\text{max}}}$ (\cref{appendix:section:correlation_analysis})), $C_{g_{pre}}\approx1$, and we get exactly hyperbolic growth as shown below:
\begin{align*}
\sigma^2_{g_{\text{out}},n} = \sigma^2_{g_{\text{out}},N} * (\frac{N}{n})
\end{align*}

The gradient variance will increase hyberbolically with number of layers $N$ while going backwards.

\paragraph{Discussion:} This has the effect that much lower learning rate is required for the entire model, because the gradients near the input layers are much higher, slowing down learning and making the model unstable.
\subsection{Vanilla Post-LN}
\label{proof: vanilla postLN}

\subsubsection{Forward Pass}

The forward pass of Post-LN is trivially always 1 at initialization, because the skip connection does not cross the LayerNorm.

\subsubsection{Backward Pass}

Following an analysis similar to that for Pre-LN, we get

\begin{align*}
    \sigma^2_{g_\text{ffn-layer}, n-1} &= \frac{1+C_2}{1+C_1 * r^l_{x_{\text{out}}, n-1}} \sigma^2_{g_{\text{out}},n} \\
    \sigma^2_{g_\text{attn-layer}, n-1} &= \frac{1+C_1 * r^l_{g_{\text{out}}, n} }{1+C_2} \sigma^2_{g_{\text{out}},n} \\
    \sigma^2_{g_{\text{out}},n-1} &= \frac{1+C_1 * r^l_{g_{\text{out}}, n} }{1+C_2} * \frac{1+C_2}{1+C_1 * r^l_{x_{\text{out}}, n-1}} * \sigma^2_{g_{\text{out}},n} \\
    &= \frac{1+C_1 * r^l_{g_{\text{out}}, n} }{1+C_1 * r^l_{x_{\text{out}}, n-1}}\sigma^2_{g_{\text{out}},n}
\end{align*}

Let $C_{5,n} = \frac{1+C_1 * r^l_{g_{\text{out}, n}} }{1+C_1 * r^l_{x_{\text{out}}, n-1}}$. As we discuss in \cref{appendix:section:correlation_analysis}, the correlations both quickly reach a maximum stable value. But the $r^l_{g_{\text{out}}, n}$'s maximum value $r^l_{g_{\text{max}}}$ is slightly different than $r^l_{x_{\text{max}}}$. Let $C_5 = \frac{1+C_1 * r^l_{g_{\text{max}}} }{1+C_1 * r^l_{x_{\text{max}}}}$, then $C_5$ can be either greater or smaller than $1$. Hence, we get 

\begin{align*}
    \sigma^2_{g_\text{attn-layer}, n-1} &= C_{5,n}\sigma^2_{g_{\text{out}},n} \\
    &= \prod_{i=n}^{N}C_{5,i}\sigma^2_{g_{\text{out}},N} \\
    &\approx C_5^{(N-n)}\sigma^2_{g_{\text{out}},N}
\end{align*}
\begin{align}
\sigma^2_{g_\text{attn-layer}, n-1} &= C_5^{(N-n)}\sigma^2_{g_{\text{out}},N}
\end{align}

This shows that gradient variance of Post-LN will decrease/increase exponentially with number of layers $N$ while going backwards.  Even very slightly different value of $C_5$ from $1$, such as $0.96$, will cause a $2000x$ fall in gradient after $200$ layers.

\paragraph{Discussion:} This shows why Post-LN transformer is much more difficult to train for deeper models than Pre-LN. While for Pre-LN the backwards gradient increases hyber-bolically to a maximum of $N$, in Post-LN the gradient can increase or decrease exponentially, stopping the model from converging.

\subsection{DeepScaleLM Pre-LN}
\label{proof: DeepScaleLM-static Pre-LN}

\subsubsection{Forward Pass}

In DeepScaleLM, the weight initialization are chosen specifically so that $\sigma^2_{x_\text{attn}}$ and $\sigma^2_{x_\text{ffn}}$ are both equal to $1$ for all layers, by iteratively calculating $r^l_{x_{\text{in}}}$ as detailed in \cref{sec: sup_pseudocode}. Also, the embeddings are initialized so that $\sigma^2_{x_{\text{in}}}$ is also $1$. Hence,
\begin{align*}
    \sigma^2_{\text{layer}} &= \lambda^2*\sigma^2_{\text{skip}} + \beta^2*\sigma^2_{\text{block}} \\
     &= \lambda^2 + \beta^2 = 1
\end{align*}
Hence the forward pass variance remains $1$ throughout the model.

\subsubsection{Backward Pass}
\label{dslmstat-pre-back}
For the FFN-block, we have $\sigma^2_{x_{\text{in}}, n-1}=\sigma^2_{x_{\text{out}}, n-1}=1$, as per equations in Table 2 of the main paper.

Similar to Vanilla-PreLN, we arrive at 
\begin{align*}
    \sigma^2_{g_\text{attn-layernorm}, n-1} &= \frac{C_1 * r^l_{g_{\text{out}}, n} *\sigma^2_{g_{\text{out}},n}}{\sigma^2_{x_{\text{in}}, n-1}}
\end{align*}

Here, $\sigma^2_{x_{\text{in}}, n-1}=1$ as shown above, and since weights are initialized so that $C1*r^l_{x_{\text{in}}}=1$. Let $C_{6,n}=\frac{r^l_{g_{\text{out}},n}}{r^l_{x_{\text{out}},n-1}}$:

\begin{align*}
    \sigma^2_{g_\text{attn-layernorm}, n-1} &= \frac{r^l_{g_{\text{out},n}}}{r^l_{x_{\text{in}, n-1}}} * \sigma^2_{g_{\text{out}},n} \\
    &= C_{6,n} * \sigma^2_{g_{\text{out}},n} \\
\end{align*}

Therefore, assuming no covariance between block gradients and skip connection (which will be true at initialization), the final gradient $\sigma^2_{g_\text{attn-layer}, n-1}$ after addition with the skip connection is 

\begin{align*}
    \sigma^2_{g_\text{attn-layer}, n-1} &= \lambda^2 \sigma^2_{g_{\text{out}},n} + \beta^2 \sigma^2_{g_\text{attn-layernorm}, n-1} \\
    &= \lambda^2 \sigma^2_{g_{\text{out}},n} + \beta^2 C_{6,n}\sigma^2_{g_{\text{out}},n} \\ 
    &= (\lambda^2 + \beta^2C_{6,n}) * \sigma^2_{g_{\text{out}},n} \\
    &= (1 + \frac{C_{6,n} - 1}{N}) * \sigma^2_{g_{\text{out}},n}
\end{align*}

Similarly for the FFN layer, $\sigma^2_{g_\text{ffn-layer}, n-1} = \sigma^2_{g_{\text{out}},n}$, as $\sigma^2_{x_{\text{in}}, n-1}=\sigma^2_{x_{\text{out}}, n-1}=1$.


Hence,
\begin{align*}
    \sigma^2_{g_\text{out}, n-1} &= (1 + \frac{C_{6,n} - 1}{N}) * \sigma^2_{g_{\text{out}},n}, \\
    \sigma^2_{g_\text{out}, 1} &=  \prod_{i=1}^{N} (1 + \frac{C_{6,n} - 1}{N})* \sigma^2_{g_{\text{out}},N}, \\
    &\approx \prod_{i=1}^{N} (1 + \frac{C_6 - 1}{N})* \sigma^2_{g_{\text{out}},N}, \\
    &\approx {(1 + \frac{C_6 - 1}{N})}^{N-1}* \sigma^2_{g_{\text{out}},N}, \\
    &=e^{C_6 - 1}*\sigma^2_{g_{\text{out}},N} \\
    &\approx \sigma^2_{g_{\text{out}},N}
\end{align*}

, where we applied $(1-\frac{k}{N})^N \approx e^{-k}$, and $C_6\approx1$.

\paragraph{Discussion:} Hence for DeepScaleLM, the backward variance of gradient remains constant (bounded by a constant)  across all layers.


\subsection{DeepScaleLM Post-LN}
\label{proof: DeepScaleLM-static Post-LN}

\subsubsection{Forward Pass}

Same as vanilla Post-LN, this will remain preserved at $1$.

\subsubsection{Backward Pass}

Following an analysis similar to that for Vanilla Post-LN, we get

\begin{align*}
    \sigma^2_{g_\text{ffn-layer}, n-1} &= \sigma^2_{g_{\text{out}},n} \\
    \sigma^2_{g_\text{attn-layer}, n-1} &= (\lambda^2*1+\beta^2*C_1 * r^l_{g_{\text{out}}, n} ) \sigma^2_{g_{\text{out}},n} \\
    &= (\lambda^2+\beta^2* \frac{r^l_{g_{\text{out}}, n}}{r^l_{x_{\text{in}}, n}} )  \sigma^2_{g_{\text{out}},n} \\
    \sigma^2_{g_{\text{out}},n-1} &= (\lambda^2+\beta^2* \frac{r^l_{g_{\text{out}}, n}}{r^l_{x_{\text{in}}, n}} )  \sigma^2_{g_{\text{out}},n} 
\end{align*}

Similar to Pre-LN, we use the maximum value of these correlations, and assume $C_6=1$. We get
\begin{align*}
    \sigma^2_{g_{\text{out}},n-1} &= (\lambda^2+\beta^2* \frac{r^l_{g_{\text{max}}}}{r^l_{x_{\text{max}}}} )  \sigma^2_{g_{\text{out}},n} \\
    &= (\lambda^2+\beta^2 C_6 )  \sigma^2_{g_{\text{out}},n} \\
    &\approx (\lambda^2+\beta^2 )  \sigma^2_{g_{\text{out}},n} \\
    &= \sigma^2_{g_{\text{out}},n}
\end{align*}

Hence for DeepScaleLM, the backward variance of gradient remains constant across all layers.

\paragraph{Discussion:} Similar to DeepScale-LM Pre-LN, the assumption $C_6=1$ is not required, and yields the same constant bound if we do not assume it to be 1.

\subsection{DeepScaleLM (Simplified) Pre-LN}
\label{proof: DeepScaleLM Pre-LN}

\subsubsection{Forward Pass}

For simplified DeepScaleLM, the initialization for the FFN block does not change, so its output remains $1$ same as DeepScaleLM.
For the Attention block, we changed its initialization to mimic that of the FFN block. We will show that initially, simplified DeepScaleLM's forward pass is bounded.

$\sigma^2_{x_\text{ffn}}=1$ as DeepScaleLM, $\sigma^2_{x_\text{attn}}=\frac{r^l_{x_{\text{in}}}}{2}$. Therefore, the output variance after layer $n$ will be
\begin{align*}
\sigma^2_{x_\text{attn-skip},n} &= \lambda^2*\sigma^2_{x_{\text{layer},n-1}} + \beta^2*\sigma^2_{x_\text{attn}} \\
&= (1-\frac{2}{N})*\sigma^2_{x_{\text{layer},n-1}} + \frac{1}{N}*r^l_{x_{\text{in}}}
\end{align*}
Similarly after the FFN block, the output skip will be - 
\begin{align*}
\sigma^2_{x_\text{layer},n} &= \lambda^2*\sigma^2_{x_\text{attn-skip},n} + \beta^2*\sigma^2_{x_\text{ffn}} \\
&= (1-\frac{2}{N})*((1-\frac{2}{N})*\sigma^2_{x_{\text{layer},n-1}} + \frac{1}{N}*r^l_{x_{\text{in}}}) + \frac{2}{N}*1 \\
&=(1-\frac{2}{N})^2*\sigma^2_{x_{\text{layer},n-1}} + (1-\frac{2}{N})* \frac{1}{N} * r^l_{x_{\text{in}}} + \frac{2}{N}
\end{align*}

As correlation coefficient $r^l_{x_{\text{in}}} \leq 1$, we get,

\begin{align*}
\sigma^2_{x_\text{layer},n} &\leq (1-\frac{2}{N})^2*\sigma^2_{x_{\text{layer},n-1}} + (1-\frac{2}{N})* \frac{1}{N} * 1 + \frac{2}{N} \\
&= (1-\frac{2}{N})^2*\sigma^2_{x_{\text{layer},n-1}} + \frac{3}{N} - \frac{2}{N^2} \\
&\leq (1-\frac{2}{N})^2*\sigma^2_{x_{\text{layer},n-1}} + \frac{3}{N}
\end{align*}

Applying the above recurrence equation $N$ times, we get
\begin{align*}
\sigma^2_{x_\text{layer},N} &\leq (1-\frac{2}{N})^{2N}*\sigma^2_{x_{\text{layer},0}} + \frac{3}{N}*\sum_{i=0}^N (1-\frac{2}{N})^{2i} \\
&= (1-\frac{2}{N})^{2N}*\sigma^2_{x_{\text{layer},0}} +  \frac{3}{N} * \frac{1-(1-\frac{2}{N})^{2N}}{1 - (1-\frac{2}{N})^2}
\end{align*}
Since $\lambda^2 + \beta^2 = 1$ and $\beta^2$ is small for large N.  We can rewrite the above equations completely in terms of $\beta$ as follows
\begin{align}
\label{dslm-pre-back}
\sigma^2_{x_\text{layer},N} 
&= (1-\beta^2)^{2N}*\sigma^2_{x_{\text{layer},0}} +  \frac{3}{2} \beta^2 * \frac{1-(1-\beta^2)^{2N}}{1 - (1-\beta^2)^2} \\
&\approx (1-\beta^2)^{2N} * \sigma^2_{x_{\text{layer},0}} + \frac{3}{4}(1-(1-\beta^2)^{2N})
\end{align}

For large $N$, we know $(1-\frac{k}{N})^N \approx e^{-k}$. So the above becomes - 
\begin{align*}
\sigma^2_{x_\text{layer},N} &\approx e^{-4}*\sigma^2_{x_{\text{layer},0}} +  \frac{3}{N} * \frac{1-e^{-4}}{\frac{4}{N} - \frac{4}{N^2}} \\
&\leq e^{-4}*\sigma^2_{x_{\text{layer},0}} +  \frac{3}{N} * \frac{1-e^{-4}}{\frac{4}{N}} \\
&= e^{-4}*1 +  \frac{3}{4} * (1-e^{-4}) \\
&= \frac{3}{4} + \frac{1}{4e^4}
\end{align*}

This gives us an upper bound on the output variance after $N$ layers. By setting $r^l_{x_{\text{in}}}=0$ instead of 1 in the equation above, and proceeding similarly, we can also arrive at a lower bound of $\frac{1}{2} + \frac{1}{2e^4}$.

\begin{align}
\frac{1}{2} + \frac{1}{2e^4} \leq \sigma^2_{x_\text{layer},N} \leq \frac{3}{4} + \frac{1}{4e^4}
\end{align}

\paragraph{Discussion}  Informally, this is because the attention block output variance will be between $0$ and $0.5$, and ffn block output always $1$. Because of our $\lambda,\beta$ scaling, the output will slowly converge to be in between the two outputs. 

Note that the above derivation assumes no correlation between the block output and the skip connection. As we mentioned in our main paper, we do observe correlation between the input and the output. As such, theoretically, after every block, the variance $\sigma^2_{x_{\text{layer},n}}$ can increase by $\sigma^2_{x_{\text{block}}} + \sqrt{\sigma^2_{x_{\text{layer},n}}}$. This will cause the final output variance to increase by factors of $2*\sqrt{N}$. In practice however, we observe the output variances to not grow too large.

\subsubsection{Backward Pass}

Similar to DeepScaleLM Pre-LN, we arrive at 

\begin{align*}
    \sigma^2_{g_\text{attn-layernorm}, n-1} &= \frac{C_1 * r^l_{g_{\text{out}}, n} *\sigma^2_{g_{\text{out}},n}}{\sigma^2_{x_{\text{in}}, n-1}} \\
    &\approx \frac{0.5*C_6}{\sigma^2_{x_{\text{in}}, n-1}} * \sigma^2_{g_{\text{out}},n}
\end{align*}

\begin{align*}
    \sigma^2_{g_\text{attn-layer}, n-1} &= \lambda^2 \sigma^2_{g_{\text{out}},n} + \beta^2 \sigma^2_{g_\text{attn-layernorm}, n-1} \\
    &= (\lambda^2 + \beta^2 * \frac{0.5*C_6}{\sigma^2_{x_{\text{in}}, n-1}}) * \sigma^2_{g_{\text{out}},n} \\
    &= (1 + \frac{2}{N} * (\frac{0.5*C_6}{\sigma^2_{x_{\text{in}}, n-1}} -1)) * \sigma^2_{g_{\text{out}},n}
\end{align*}

Similarly, for the FFN layer, we get 
\begin{align*}
    \sigma^2_{g_\text{ffn-layer}, n-1} &= (1 + \frac{2}{N} * (\frac{1}{\sigma^2_{x_{\text{in}}, n-1}} -1)) * \sigma^2_{g_{\text{out}},n}
\end{align*}

Multiplying these, we get 
\begin{align*}
\sigma^2_{g_{\text{out}},n-1} &= (1 + \frac{2}{N} * (\frac{0.5*C_6}{\sigma^2_{x_{\text{in}}, n-1}} -1)) * (1 + \frac{2}{N} * (\frac{1}{\sigma^2_{x_{\text{in}}, n-1}} -1)) * \sigma^2_{g_{\text{out}},n} \\
&\approx (1 + \frac{2}{N} * (\frac{0.5*C_6}{\sigma^2_{x_{\text{in}}, n-1}} + \frac{1}{\sigma^2_{x_{\text{in}}, n-1}} -2)) * \sigma^2_{g_{\text{out}},n} 
\end{align*}

As $0.5 \leq \sigma^2_{x_{\text{in}}, n-1}$, we get $-4 \leq (\frac{C_6}{\sigma^2_{x_{\text{in}}, n-1}} + \frac{2}{\sigma^2_{x_{\text{in}}, n-1}} -4) \leq 2C_6+2$. Hence, on applying the above recurrence N times, we get 

\begin{align*}
e^{-4} * \sigma^2_{g_{\text{out}},N} \leq \sigma^2_{g_{\text{out}},n-1} \leq e^{2C_6+2} * \sigma^2_{g_{\text{out}},N} 
\end{align*}

Hence, we show that even for simplified DeepScaleLM Pre-LN, the maximum relative increase/fall in gradient variance is bounded across layers.

\paragraph{Discussion:} The above derivations will also be valid if there is correlation in the input. Correlation will cause $\sigma^2_{x_{\text{in}}, n-1}$ to increase, effectively decreasing the backpropagated gradient through the block to decrease (as Layernorm will scale by inverse of $\sigma^2_{x_{\text{in}}, n-1}$). However, even in that case, our gradient will still be bounded by the above lower-bound. 

Intuitively, as the gradient can flow freely through the skip connection, hence, $\sigma^2_{g_{\text{out}},n-1} \geq \lambda^4*\sigma^2_{g_{\text{out}},n} $, which when applied $N$ times, yields $\sigma^2_{g_{\text{out}},1} \geq e^{-4} * \sigma^2_{g_{\text{out}},N}$
\subsection{DeepScaleLM (Simplified) Post-LN}
\label{proof: DeepScaleLM Post-LN}

\subsubsection{Forward Pass}
The forward pass variance for Post-LN is trivially bounded.

\subsubsection{Backward Pass}

Following an analysis similar to that for DeepScaleLM Post-LN, we get

\begin{align*}
    \sigma^2_{g_{\text{out}},n-1} &= \frac{\lambda^2+ 0.5*\beta^2* r^l_{g_{\text{out}}, n}}{\lambda^2+ 0.5*\beta^2* r^l_{x_{\text{in}}, n}} \sigma^2_{g_{\text{out}},n} \\
    &= \frac{1+\frac{2}{N} (0.5r^l_{g_{\text{out}}, n} - 1)}{1+ \frac{2}{N} (0.5r^l_{x_{\text{in}}, n} - 1)} \sigma^2_{g_{\text{out}},n} 
\end{align*}
Applying taylor expansion, we get,
\begin{align*}
    \sigma^2_{g_{\text{out}},n-1} &\approx (1+\frac{2}{N}( (0.5r^l_{g_{\text{out}}, n} - 1) - (0.5r^l_{x_{\text{in}}, n} - 1))) \sigma^2_{g_{\text{out}},n} \\
    &= (1+\frac{1}{N}(r^l_{g_{\text{out}}, n} - r^l_{x_{\text{in}}, n})) \sigma^2_{g_{\text{out}},n}
\end{align*}
The above equation can be rewritten in terms of $\beta$ as follows 
\begin{align}
\label{dslm-post-back}
    \sigma^2_{g_{\text{out}},n-1} &= (1+\frac{\beta^2}{2}(r^l_{g_{\text{out}}, n} - r^l_{x_{\text{in}}, n})) \sigma^2_{g_{\text{out}},n}
\end{align}

As $-2 \leq (r^l_{g_{\text{out}}, n} - r^l_{x_{\text{in}}, n}) \leq 2$, applying the above recurrence $N$ times we get 
\begin{align*}
e^{-2} * \sigma^2_{g_{\text{out}},N} \leq \sigma^2_{g_{\text{out}},n-1} \leq e^2 * \sigma^2_{g_{\text{out}},N} 
\end{align*}

\paragraph{Discussion:} The above derivations assume no correlation in the input, and hence is only correct at initialization. However, if there is correlation between the block output and skip connection ($r_x$), the layernorm will cause $\sigma^2_{g_{\text{out}},n-1}$ to be down-scaled by a factor of $1+\frac{2*r_x}{\sqrt{N}}$, where c is some constant, as opposed to $1+\frac{2}{N}$ above. However, if there is also correlation in the gradients of the block and skip connection ($r_g$), the numerator in the equations above for $\sigma^2_{g_{\text{out}},n-1}$ will also be increased, by a factor of $1+\frac{2*r_g}{\sqrt{N}}$. Hence if the correlations among the gradients and among the output are similar, the above bounds will remain.
If $\beta^2$ is set as $\frac{1}{N^2}$, then even if input correlations exist, the backward gradient will be bounded, following a similar derivation as above. However, we conjecture that this decreases the ability of the transformer layers to modify the skip connection too strongly, decreasing the ``expressivity'' of the model. This is similar to the approach of DSInit, which we show in our main paper does indeed decrease model performance.

\section{Rank Collapse and Correlation Analysis}
\label{appendix:section:correlation_analysis}
In the previous sections, we derived the formulas that determine how the correlation will change through the Attention and FFN blocks both for forward and backward pass. Both attention and FFN blocks modify the correlation as shown in the \cref{table:block_moments}.

Simplifying the formulae in the table above, we rewrite the output variance for the attention block as $\sigma^2_{x_\text{attn}} = C_1*r^l_{x_{\text{in}}}*\sigma^2_{x_{\text{in}}}$, and the output of the FFN block is $\sigma^2_{x_\text{ffn}} = C_2*\sigma^2_{x_{\text{in}}}$, where $C_1$ and $C_2$ are defined as follows.

\begin{align*}
C_1 &= \frac{d^2\sigma_o^2\sigma_v^2}{(1-p)}, C_2=\frac{2d^2\sigma_{w_1}^2\sigma_{w_2}^2}{(1-p)}, \\
\end{align*}

This also helps us to rewrite the backward pass as  the $\sigma^2_{g_\text{attn}} = C_1*r^l_{g_{\text{out}}} * \sigma^2_{g_\text{out}} $ and $\sigma^2_{g_\text{ffn}} = C_2* \sigma^2_{g_\text{out}}$.

Specifically in case of Xavier initialization with $0.1$ dropout, $C_1 = 2.2, C_2=0.4$.

Assuming a dropout of $0.1$, the FFN block (with the ReLU) will reduce the correlation if it rises above $0.64$ (where $r^l_{x_{\text{out}}} < r^l_{x_{\text{in}}}$ for FFN block). And the attention block will never output a correlation higher than $0.9$. Hence correlation will never reach $1$, but rather a steady, stable value between ReLU's maximum correlation and that of the attention block. Dropout's effect in preventing rank collapse was also observed in \cite{DropEdgeDeepGraph}.

We can approximate the stable value of correlation after many layers based on the weightage average of the correlation in the Attention output and FFN output. When the attention output is added to the skip connection, the new correlation will be a weighted (by variance) average of the correlation among the tokens of attention output and among the tokens in the skip connection. And the same will happen after the FFN block. 

A weighted average of the correlations of FFN and attention blocks gives the stable asymptotic correlation $r^l_{x_{\text{max}}}$

\begin{align*}
r^l_{x_{\text{max}}} &= \frac{C_1*(1-p) + C_2*(1-p)(\dfrac{1}{\pi} + \dfrac{r^l_{x_{\text{max}}}}{2} + (\dfrac{1}{2}-\dfrac{1}{\pi}){r^l_{x_{\text{max}}}}^2)}{C_1 + C_2}
\end{align*}

Specifically for the case of xavier initialization, solving the above equation with $C_1 = 2.2, C_2=0.4$, gives $r^l_{x_{\text{max}}} \approx 0.88$, as empirically verified in \cref{fig:rank_collapse}.

Similarly, the correlation for backward gradient will also converge at a stable value $r^l_{g_{\text{max}}}$, obtained by solving the below equation - 

\begin{align*}
r^l_{g_{\text{max}}} &= \frac{C_1*(1-p) + C_2*(1-p)(\frac{1}{2} + \frac{\sin^{-1}{(r^l_{x_{\text{max}}})}}{\pi})r^l_{g_{\text{max}}}}{C_1 + C_2}
\end{align*}

Specifically for the case of xavier initialization, this gives $r^l_{g_{\text{max}}} = 0.87$. Note how $r^l_{g_{\text{max}}} \approx r^l_{x_{\text{max}}}$.

\paragraph{Discussion on rank collapse observed in \citet{noci2022signal}} \citet{noci2022signal} focuses primarily on linear activation, we theoretically analyze the change in output correlation caused by ReLU. We find that ReLU (or any asymmetric non-linearity in general) critically affects correlation. As our closed form expressions suggest, both FFN block (because of ReLU) and dropout reduce the correlation. While \citet{noci2022signal} mentions the use of dropout, as we show above and observe empirically in \cref{fig:rank_collapse}, rank will not collapse with dropout, and perhaps \citet{noci2022signal} did not use dropout. 

We replicated the experimental settings of \citet{noci2022signal} without dropout, and observed that the rank collapse occurs due to incorrect initialization. They use a rather non-standard version of xavier initialization - instead of $\frac{2}{fan_{in} + fan_{out}}$, they use $\frac{1}{fan_{out}}$. Hence, they initialize a much higher value for V as $fan_{in}$ is much greater than $fan_{out}$ (``Number of heads'' times greater), and this results in variance of the output of the attention block $C1$ being much higher than FFN $C2$. As attention block outputs a much higher correlation than the FFN block, increasing its output variance without using dropout will result in rank collapse. This highlights the criticality of correct initialization, as well as the explainability power of our theoretical framework proposed in the paper.

\section{Discussion of Relative Strength}
\label{appendix:section:rel_strength}

In \autoref{dslm-pre-back}, we discussed that the backward recurrence equation for PreLN can be written as

\begin{align*}
\sigma^2_{x_\text{layer},N} 
&\approx (1-\beta^2)^{2N}*\sigma^2_{x_{\text{layer},0}} +  \frac{3}{4}  (1-(1-\beta^2)^{2N})
\end{align*}

Replacing $\beta^2=\frac{k}{N^\alpha}$ and using $(1+\frac{k}{N^\alpha})^N = e^{kN^{1-\alpha}}$, we get

\begin{align*}
\sigma^2_{x_\text{layer},N} 
&\approx e^{2cN^{1-\alpha}}*\sigma^2_{x_{\text{layer},0}} +  \frac{3}{4} (1-e^{2cN^{1-\alpha}}) \\
&= e^{2cN^{1-\alpha}} * (\sigma^2_{x_{\text{layer},0}} - \frac{3}{4}) + \frac{3}{4}
\end{align*}

Hence, the fall in gradient for $\beta^2 = \frac{k}{N^\alpha}$ is \ord{e^{kN^{1-\alpha}}}. 

Similarly for PostLN, we can use \autoref{dslm-post-back}
\begin{align*}
    \sigma^2_{g_{\text{out}},n-1} = (1+\frac{\beta^2}{2}(r^l_{g_{\text{out}}, n} - r^l_{x_{\text{in}}, n})) \sigma^2_{g_{\text{out}},n} \\
    (1-\beta^2) * \sigma^2_{g_{\text{out}},N} \leq \sigma^2_{g_{\text{out}},n-1} \leq (1+\beta^2) * \sigma^2_{g_{\text{out}},N} 
\end{align*}

Hence, for N layers, the gradient fall/growth is again \ord{e^{\pm kN^{1-\alpha}}}.

\section{Applying DeepscaleLM to Vision Transformers}
\label{appendix:subsection:vision}

Applying our method to vision transformers (for eg. ViT~\citep{vit} or DeiT~\cite{deit}) will only require handling the input embeddings section~\cref{proof: embeddings} - For ViT, this is a simple linear projection. Given normalized image inputs, our Linear section~\cref{proof: linear} provides formulae to calculate the variance and correlation of the embeddings which are input to the model.

We empirically verified that for images from ImageNet, the embeddings after the linear projection do indeed follow the normal distribution, with an $R^2$ of $0.95$. Furthermore, normalizing images to have approximately unit variance, given linear weights initialized by $\sqrt{\frac{1}{d}}$, the output variance was observed as $1.02$ (within $2\%$ error). While we used Zipf’s law to estimate input embedding correlation for text, this could simply be empirically measured for vision after the embedding layer -- we measured this to be $0.46$ using the code provided by \citet{vitbaseline}.

\begin{figure}[H]
    
    
    \begin{center}
    
    \includegraphics[page=14]{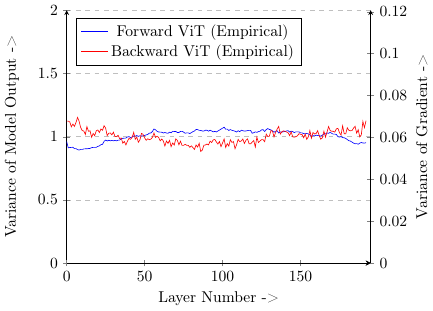}
    \caption{DeepScaleLM: The variances remain conserved for both backward and forwards for ViT, using ImageNet data, after even 192 transformer layers}
    \label{fig:appendix:vit_plot}
    \end{center}
\end{figure}

Using this measured value of input correlation, we can apply our DSLM method to ViT. As we show in \cref{fig:appendix:vit_plot}, our method successfully controls both the forward and backward moments for the ViT model with $100s$ of layers.

\section{Compute}
\label{appendix:section:compute}

\subsection{Theoretical compute}
\autoref{table:appendix:compute} provides the exact compute for the models reported in \cref{table: res1}. We follow the code provided by Electra \citep{electra} to calculate the each model's compute (FLOPs). We observe that up to $200$ layers, the extra compute is within $6-7$\% of the original shallow model.

\begin{table}[H]
\caption{Model compute with increasing depth (keeping $Nd^2$ constant).}
\begin{center}
\begin{tabular}{lcccc}
\toprule
    \textbf{Layers (N)} &  \textbf{Hidden Dim (d)} & \textbf{Params} & \textbf{Compute (Flops)} & \textbf{\% Extra}  \\
    \midrule
    12   & 1024 & 185M & 1.06e20 & - \\
    48   & 512  & 168M & 1.03e20 & -2.5\% \\
    192  & 256  & 160M & 1.12e20 & 6.3\%  \\
    784  & 128  & 156M & 1.38e20 & 30.6\% \\
    \midrule
    24   &  1024 & 336M & 1.92e20  & -  \\
    96   &  512 & 319M & 1.96e20 & 2.3\%  \\
    384  &  128 & 311M & 2.19e20 & 14.5\%  \\
    \bottomrule
\end{tabular}
\end{center}
\label{table:appendix:compute}
\end{table}

\subsection{Wall Clock times}

We also compared wall clock time overheads, and found them to not be too large. For example, the $48$-layer-$512$-d model has only $9.8\%$ overhead in wall clock time compared to $12$-layer-$1024$-d model. Even when larger number of layers, such as $96$-layer-$512$-d, the overhead is only $14.9\%$ compared to $24$-layer-$1024$-d model. Profiling revealed majority of the overhead was due to extra latency of added GPU kernel launches. Hence, approaches such as cudaGraphs (which batches kernel launches together) or graph compilation techniques may decrease this overhead further.

This overhead will decrease the bigger the original model size, and become much smaller. For example, for a $5$B params model with $24$-Layers-$4096$d (a reasonable shape in contemporary models, for example, LLaMA $7$B has $32$L-$4096$D) has much less compute overhead - only $6.6\%$ overhead at $96$ layers, and $13.6\%$ overhead at $192$ layers.

Despite this wall-clock time overhead, due to large performance gains from increasing depth, the $160$M params $192$-L model from \cref{table: res1} outperforms the vanilla $336$M BERT-large $24$-L model with $2x$ more params, even at equal wall times. 

Furthermore, a large fraction of the performance improvements mentioned happen when increasing the number of model layers by $4x$ - and as shown above, the wall clock time overhead is minimal. Making standard models $4x$ more deep to $50-100$ layers, will provide a large fraction of performance gains without much overhead.

Note that this performance overhead seems to be dependent on the framework used -- some frameworks may be less optimized for such deeper models and may incur additional overhead for small but deep models.

\section{Statistical Significance}
\label{appendix:section:stat_sig}

\subsection{Error Bars for Pre-Training Experiments}

In our initial experiments, we observed very little variation in performance across different runs -- we conjecture that the model is trained on a large enough number of tokens for differences in initialization/data seed to not matter. We provide mean and standard error for the $12$L-$1024$D Post-LN and DSLM models from \cref{table: res1} below:

\begin{table*}[h]
    \caption{Standard error across runs for pre-training.}
    \begin{center}
    \begin{tabular}{lcc}
         \toprule
         \textbf{Model} &  \textbf{Mean} & \textbf{Standard Error}\\
         \midrule
         Post-LN Baseline &  14.33 & 0.14\\
         DSLM &  15.56 & 0.08\\
         \bottomrule
    \end{tabular}
    \label{tab:std_err}
    \end{center}
\end{table*}

As the variation was so small, and due to compute limitations, we did not run multiple runs for other experiments thereafter. We also reported the best score for Baseline Post-LN, and the worst score for DSLM for the $12$L-$1024$D models Table 4 for a conservative comparison.



\subsection{Statistical Significance for Fine-tuning Experiments}

Mean and standard errors for all downstream fine-tuning experiments were reported in \cref{table:finetune_stderr}. The differences are statistically significant at $p<5\%$ for all datasets except QQP.

\section{Related Works}
\label{section:app_related}

\subsection{Initialization}
\label{section:related:initialization}
Several works, such as \citet{glorot, DelvingDeepRectifiers, CharacterizingSignalPropagationa} improved the initialization of ResNets/ReLU networks, but crucially these works do not consider the impact of correlation in the input, which is large in Transformer models. \citet{ExponentialExpressivity} takes correlation into account, and \citet{DeepInformationPropagation} initializes weights for networks with bounded activations so that correlation reaches $1$ asymptotically. 

Some works, such as \citet{AllYouNeed}, sequentially profile each layer empirically by running forward passes through the model, and scaling the weights and/or output to achieve unit variance, and \citet{UnderstandingDifficultyTraininga, verydeep} applied the same method for Transformers. \citet{UnitScalingOutoftheBoxa} also tries to achieve unit variance, but does not consider correlation in input or across tokens, and ignores the non-zero mean of ReLU. \citet{ReZeroAllYou} shows unit variance leads to faster convergence at the start of the training. 

We demonstrate that this profiling is unnecessary, and can instead be done theoretically in DeepScaleLM. Furthermore, where output or gradient increases in some prior works with more layers (eg. for ADMIN \citep{UnderstandingDifficultyTraininga}, grad decreases by \ord{N} (increases by \ord{log(N)} for Pre-LN)), our method allows maintaining both unit output and equal gradient across all layers at initialization, and bounded during training.

\citet{tensorprograms} proposed $\mu$P initialization such that updates to a layer are of the same order regardless of width. Their work was focused on enabling transfer of hyper-parameters across model widtd, and does not target solving pathologies inherent in deeper architectures -- they do not model the impact of ReLU and Attention on correlation, and hence are unable to prevent rank-collapse at large depths. When applied to $100$s of layers, $\mu$P diverges with rank collapse at initialization.

\subsection{Signal Propagation}

Signal propagation in Neural Networks has a long history, such as \citet{rel_neal, rel_lecun}. More recently, several works have focused on signal propagation for ResNets, such as \citet{DelvingDeepRectifiers, BatchNormalizationBiasesa, CharacterizingSignalPropagationa, DeepInformationPropagation, NormalisationDead, ProxyNormalizingActivationsMatcha, ScalingResNetsLargedeptha, SelfNormalizingNeuralNetworksb, Shattered}. 

For transformers, signal propagation was studied in \citet{xu2019understanding, AttentionNotAll, Catformer, noci2022signal}. Our work also considers previously neglected effects of dropout, input correlation between tokens, non-linearity, $QK$ initialization, and provides closed forms with verifiable correctness of this signal propagation. Ours is the first work to theoretically constrain the output and gradient to almost exactly unit without any profiling passes, showing the validity of our formulae and of our assumptions. 



\citet{DeepTransformersShortcutsa} extends neural kernel methods of DKS \citep{RapidTrainingDeep} to Transformers to model network behaviour, assuming the MLP to be linear in its effect on attention with respect to correlation. 
$Q/C$ maps in kernel methods are similar to signal propagation, as expected moments are equivalent to $q$ and $m$ values of kernels \citep{RapidTrainingDeep}. Our method relaxes these assumptions, and we show that considering the impact of ReLU/GeLU on correlation is critical to correctly modelling attention. In particular, our formulae show that an MLP block with GeLU will also increase correlation in the absence of dropout (the same setting as used in \citet{DeepTransformersShortcutsa} ). At large depths, \citet{DeepTransformersShortcutsa}'s method suffers from rank collapse (with their deeper models under-performing shallower ones), which our method successfully prevents.

We also account for cases with non-IID inputs that may occur due to segment/position embeddings or due to non-uniform token distributions in real data (that are distributed approximately per Zipf's law \citet{zipf}) -- and find that this strongly affects output variance of the attention block.

\subsection{Moment Control \& Residual Scaling}
\label{section:related:residual_scaling}

Bounded gradients, or normalizing per-layer gradients, have been shown to results in better/faster convergence \citep{PowerNormRethinkingBatcha, BlockNormalizedGradientMethoda, LargeBatchTrainingb, LargeBatchOptimizationb}. Woks such as \citet{LayerNormalizationsResiduala, NormFormerImprovedTransformerb, ImpactActivationFunction} also achieved improved training by empirically mitigating the gradient explosion.

Scaling with $\lambda^2 + \beta^2=1$ to control moments have often been used for ResNets \citep{Shattered, Inceptionv4, HowStartTraining, HowInitializeYour, fixup, NormalisationDead}. \citet{Inceptionv4} proposed to use any small $\beta$, \citet{Shattered} proposed to set $\beta^2 = 0.5$, \citet{ReZeroAllYou} sets $\beta=0$ and learnable. \citet{BatchNormalizationBiasesa} showed that $\lambda^2 = 0.5$ is not sufficient to solve vanishing gradients.

$\beta^2=\frac{k}{N}$ was used to control growth of moments in \citet{HowInitializeYour, CharacterizingSignalPropagationa, ScalingResNetsLargedeptha, StabilizeDeepResNet, noci2022signal, DeepTransformersShortcutsa, tensorprogramsvi}) . $\beta^2=\frac{k}{n}$, where $n$ is the current layer, was used in \citet{BatchNormalizationBiasesa, UnderstandingDifficultyTraininga, verydeep, Catformer, UnitScalingOutoftheBoxa}, but this results in logarithmic bounds instead of constant for forward propagation if $\lambda=1$ is used, and vanishing gradient for backward propagation otherwise.

Values of $\beta^2 < \frac{k}{N}$, such as (effectively) $\frac{1}{N^2}$ for DSInit~\citep{DSInit} or $\frac{1}{N^{1.5}}$ for DeepNet~\citep{deepnet} decrease sensitivity of the model, and may result in the model becoming ``too linear''. DeepNet shows performance improvements by making the model deeper, but keeping the hidden dimension constant. Our setting is much more strict -- we keep the number of parameters (and hence compute) constant, and our method still show performance improves on making the model deeper. For example, DeepNet`s $200$ layer model is $3.2$B params, whereas our $192$ layer model is $160$M params ($20$x smaller).
 
Sometimes, these $\beta$ values are used in conjunction with $\lambda=1$, such as in \citet{UnderstandingDifficultyTraininga, verydeep}, but as shown in \citet{DeepTransformersShortcutsa}, fully normalized residual connections with $\lambda^2 + \beta^2=1$ often perform better than those with $\lambda=1$. We also observed lower performance with $\lambda=1$ in our initial experiments, and hence we fully normalize the residual connections.

Our contribution goes beyond providing an optimal scaling scheme. Using the theoretical framework and closed-form expressions for moment propagation through both Pre-LN and Post-LN developed in this work, practitioners can make informed choices about using any of the scaling factors above based on the stability-performance tradeoffs, such as using a lower $\beta$ for scenarios with high correlation, or using higher $\beta$ with uncorrelated inputs.

\subsection{Other Network modifications for Deep Networks}
\label{section:related:other}

\citet{RevisitingOversmoothingBERT, DeepViTDeeperVision, AntiOversmoothingDeepVision, AttentionNotAll} showed that attention causes rank collapse in deeper models, and \citet{SimpleDeepGraph, AreMoreLayersa} showed the same for graphs. \citet{LayerNormalizationsResiduala} added some extra skip connections from the input of the model, \citet{TransformersTearsImprovinga} modified layernorm slightly, \citet{StabilizingTransformerTrainingc} normalized all linear layers by their spectral norm, and \citet{NormFormerImprovedTransformerb} added extra layer norms. Some works in particular, such as \citet{StabilizingTransformerTrainingc,DeepViTDeeperVision} can only prevent attention entropy collapse later during training, but our work will also prevent rank collapse at initialization caused by the very structure of the transformer model, in particular increase in correlation caused by both attention and ReLU/GeLU. The methods in these works are orthogonal to our approach, and our equations can be easily extended to cover the architectural modifications suggested in these.


\section{Discussion of Approximations and Assumptions}
\label{appendix:section:approx}

\subsection{Illustrative Approximations of Full Formulae in Main Paper}

Some values listed in \cref{table:moment_full_formulas1} are approximations/illustrative simplifications of their full closed forms in \cref{appendix:section:summary_table} and \cref{appendix:section:Moment_prop_components}. We discuss all of these below.

\begin{figure}[H]
\begin{minipage}[b]{0.45\linewidth}
    \begin{center}
    
    \includegraphics[page=12]{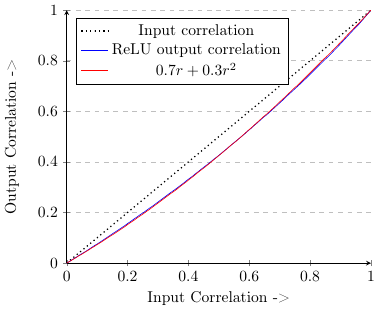}
    
    \caption{Approximation of the Relu forward correlation formula}
    \label{fig:appendix:approx_relu}
    \end{center}
\end{minipage}\hfill
\begin{minipage}[b]{0.45\linewidth}
    \begin{center}
    
    \includegraphics[page=13]{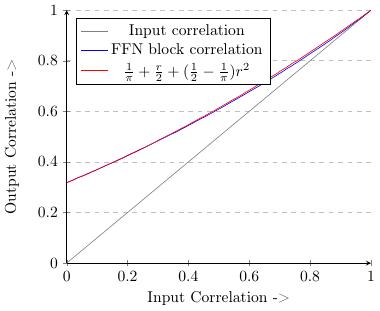}
    \caption{Approximation of the FFN forward correlation formula, without dropout. Dropout will reduce the above correlation by $1-p$.}
    \label{fig:appendix:approx_ffn}
    \end{center}
\end{minipage}
\end{figure}

\begin{itemize}
    \item For ReLU forward correlation, we used a simple polynomial regression of the closed form formula. This simple regression is a remarkably good fit, as shown in figure \cref{fig:appendix:approx_relu}, and can be reproduced in using our released code.
    \item For layernorm, we ignored the factor of 1 compared to $d$, or $1/d$ compared to $1$, assuming large enough hidden dimension $d$.
    \item For SHA without V, we used the final simplified formulae for $\sigma^2_{x_{\text{out}}}$ and output correlation from \cref{proof: sdpa}. For the gradient, we further simplified the formulae in \cref{proof: sdpa}, assuming $L \approx L-1$.
\end{itemize}

Furthermore, the formulae provided in \cref{table:block_moments} are approximate versions of the full formulae provided in \cref{appendix:section:summary_table}. In \cref{table:block_moments}, we applied a similar approximation as done in \cref{table:moment_full_formulas1} for ReLU, from the full formula in \cref{appendix:section:summary_table} for output correlation. This polynomial approximation is also a very good fit, as shown in \cref{fig:appendix:approx_ffn}, and can be reproduced using our released code. 

Our exact formulae for blocks and components also account for IID cases - as can be verified by our simulations, in which we do cover cases IID inputs with exactly $0$ correlation, as noted in  $\mathrm{Corr}^l_{x_{\text{in}}}$  column in \cref{table:verify_range}. In the simplified formulae, and in DeepScaleLM initialization and model, we simplified our formulae so that they only remain accurate for non-IID inputs. This was because of three considerations:

\begin{enumerate}
    \item In NLP domain, most text will inevitably be non-IID due to repeated common words. This was encountered in all our experiments.
    \item In Vision domain, for ViT in particular, there will be correlation among pixel intensities across patch embeddings, as discussed in common response section.
    \item In Speech domain, similar to text, most speech will inevitably be non-IID due to repeated common sounds. 
    \item Lastly, even if there is exactly $0$ correlation in input, the very first attention layer and the first FFN layer in particular, will add correlations to the output, ensuring our simplified formulae hold reasonably accurately.
\end{enumerate}

\subsection{Assumptions and Approximations in Derivations}
\label{section:appendix:assumptions}

\begin{itemize}
\item Except for attention, softmax and LayerNorm all other derivations of transformer components -- Embeddings, FFN, ReLU/GeLU, Dropout, FFN Block are fully exact, assuming only normal distribution of inputs, weights and gradients. We justify this normality assumption below: 
    \begin{enumerate}
        \item \textbf{Inputs:} As the embeddings are lookup tables of token-ids, and embedding weights are initialized from Normal distribution in Xavier, the inputs to the transformer are normally distributed.
        \item \textbf{Gradients:} As the model outputs are Normal, the softmax of the classification head results in a Log-Normal distribution for probabilities $p$, as shown in \cref{proof: softmax}. Since the cross-entropy loss is $-log(p)$, we expect the loss (and hence the final gradient being back-propagated) being log(Log-Normal distribution), to be a Normal distribution. We also verify this empirically by checking the normality of the backpropagated gradients to the deepest transformer layer, and the gradients match the best-fit Normal distribution with an $R^2$ of $0.999$, showing that the gradients are indeed Normally distributed. 
        \item \textbf{Weights:} Weights are initialized from Normal distribution in Xavier, and are hence Normal.
    \end{enumerate}
\item For attention, softmax and LayerNorm, we assume the sequence length $L$ and the hidden dimension $d$ are large. 
\item For embeddings, we  assumed Zipf's law to calculate initial input correlation in tokens, as well as assumed uniform distribution for segment lengths for next sentence prediction task of BERT. Note that this assumption is not strictly required, and can also be empirically observed and given as input to our method.
\end{itemize}

\section{DeepScaleLM Pseudocode}
\label{sec: sup_pseudocode}

\begin{figure}[H]
\begin{flushleft}

\end{flushleft}
\centering
\begin{lstlisting}[language=python,mathescape]
  ## Define constants of DeepScaleLM
  $\lambda^2$ = $1-\frac{2}{N}$ ; $\beta^2$ = $\frac{2}{N}$
  $\sigma_{e}^2$ = $\frac{1-p}{3}$ ; $\sigma_{qk}^2$ = $\frac{1}{d}$ ; $\sigma_{f}^2$ = $\frac{1}{d}*\sqrt{\frac{1-p}{2}}$

  ## Scale skip connection and block output 
  def add_skip(x, f(x)):
      return <@\textcolor{red}{$\boldsymbol{\mathbf{\lambda}}$}@> * x + <@\textcolor{red}{$\boldsymbol{\mathbf{\beta}}$}@> * f(x)

  ## Stable initialization of weights
  def init(w):
      if w is ['ffn', 'v_proj', 'out_proj']:
          nn.init.normal_(w, gain = <@\textcolor{red}{$\boldsymbol{\mathbf{\sigma_{f}}}$}@>)
      elif w is ['q_proj', 'k_proj']:
          nn.init.normal_(w, gain = <@\textcolor{red}{$\boldsymbol{\mathbf{\sigma_{qk}}}$}@>) 
      elif w is ['embd']:
          nn.init.normal_(w, gain = <@\textcolor{red}{$\boldsymbol{\mathbf{\sigma_{e}}}$}@>)
          
          
\end{lstlisting}
\label{fig:pseudo_impl}
\caption{Pseudo-code for simplified version of our DeepScaleLM method.}
\end{figure}

\begin{figure}[H]
\begin{lstlisting}[language=python,mathescape]
  ## Define constants for scaling residual and output
  $\lambda^2$ = $1-\frac{2}{N}$ ; $\beta^2$ = $\frac{2}{N}$
  ## Define constants for embedding and FFN block
  $\sigma_{e}^2$ = $\frac{1-p}{3}$ ;  $\sigma_{f}^2$ = $\frac{1}{d}*\sqrt{\frac{1-p}{2}}$
  
  ## Scale skip connection and block output 
  def add_skip(x, f(x)):
      return <@\textcolor{red}{$\boldsymbol{\mathbf{\lambda}}$}@> * x + <@\textcolor{red}{$\boldsymbol{\mathbf{\beta}}$}@> * f(x)

  ## Find layerwise input correlation upto N layers 
  def corr_input_layerwise(r, N):
      $\mathbf{r_{N}}$ = []
      for i in range(N):
          r = $\mathbf{\lambda}^2$ . r + $\mathbf{\beta}^2 (1-p)$
          r = $\mathbf{\lambda}^2$ . r + $\mathbf{\beta}^2 (1-p) ({r^l_{x_{\text{in}}}} + \dfrac{(1-(r^l_{x_{\text{in}}})^2)^{0.5}}{\pi} - \dfrac{r^l_{x_{\text{in}}}  cos^{-1}(r^l_{x_{\text{in}}})}{\pi})$
          $\mathbf{r_N}$.append(r)
      return $\mathbf{r_N}$

  ## Define constants for attention block
  $\sigma_{l,o}^2$ = $\frac{1}{d}*\sqrt{\frac{1-p}{r^{l,n}_{x_{in}}}}$; $\sigma_{qk}^2$ = $\frac{1}{d}$ ; $r = r^l_{x_{in}}$
  where $r^{l,n}_{x_{in}}$ = corr_input_layerwise(r, N)[n]

  
  ## Stable initialization of weights 
  def dslm_init(w, l):
      if w is ['ffn']:
          nn.init.normal_(w, gain = <@\textcolor{red}{$\boldsymbol{\mathbf{\sigma_{f}}}$}@>)
      elif w is ['v_proj', 'out_proj']:
          nn.init.normal_(w, gain = <@\textcolor{red}{$\boldsymbol{\mathbf{\sigma_{l,o}}}$}@>)
      elif w is ['q_proj', 'k_proj']:
          nn.init.normal_(w, gain = <@\textcolor{red}{$\boldsymbol{\mathbf{\sigma_{qk}}}$}@>) 
      elif w is ['embd']:
          nn.init.normal_(w, gain = <@\textcolor{red}{$\boldsymbol{\mathbf{\sigma_{e}}}$}@>)
          
\end{lstlisting}
\caption{Pseudo-code for our proposed method DeepScaleLM: We scale the block output and the skip connection before adding, and keep track of correlation across layers. We appropriately initialize the weights. ($N$: num of layers, $d$: model hidden dimension, $p$: dropout probability, $r^l_{x_{in}}$ is calculated based on expressions provided in \autoref{proof: embeddings}.)}
\label{fig:dslm_static_pseudo}

\end{figure}

\section{Hyper-parameters}
\label{section:hparams}
\paragraph{BERT Pretraining} We used Megatron-LM’s default BertWordPieceLowerCase tokenizer, with the original BERT lower-cased vocab, and with trainable position embeddings. The same hyper-parameters (including LR schedule, warmup) were used for all models, and LR search over the range below was performed for all models. The final best models always had optimal LR within the range and not at the boundary of the LR range for all of our experiments. Detailed hyper-params are provided in \cref{hparams_bert}.

\begin{table}[H]
\caption{Training Hyper-Parameters. We use all original hyper-parameters of BERT, except for learning-rate(LR).}
\begin{center}
\begin{tabular}{lc}
\hline
\toprule
\textbf{Parameters} & \textbf{Values} \\
\toprule
Optimizer & Adam \\
$\beta_1, \beta_2$ & 0.9, 0.999 \\
Effective Batch Size & 256 \\
Drop-out ($p$) & 0.1 \\
Sequence Length & 256 \\
Train Iters & 100,000  \\
Num GPUs & 8 \\
Learning rate & [1, 3, 5, 7, 10]*$10^{-4}$ \\
Schedule & Linear \\
LR Decay Iterations & 98\% \\
Warmup steps & 1\% \\
Min LR & $1*10^{-5}$ \\
Gradient clipping & 1.0 \\
Batch Size / GPU & 2 \\
Grad Accum Steps & 16 \\
\bottomrule
\end{tabular}
\label{hparams_bert}
\end{center}
\end{table}

\paragraph{Reproducible Longer Pre-training and Finetuning} Our released code provides exact scripts for both pre-training and all fine-tuning. We used all original/official hyper-params of BERT, except LR was increased for DSLM as mentioned previously.

\paragraph{Downstream Low Rank Finetuning} Following QLoRA~\citep{qlora}, we apply LoRA on all linear modules, with $r=32$, $\alpha=16$, and searched for LR. All other hyper-parameters were kept the same as finetuning. We used the same number of epochs as finetuning for LoRA, but perhaps more epochs may result in even better scores -- \citet{lora} used $30$ epochs for LoRA.

\paragraph{Vision ViT Training} We used ViT-S Baseline from \cite{vitbaseline} for ImageNet-1k along with its default hyper-parameters. It uses an MLP head, a Global AvgPool and a fixed 2D sin-cos position embeddings. The same hyper-parameters were used for all the models. Detailed hyper-parameters are provided in \cref{hparams_vit}

\begin{table}[H]
\caption{Training Hyper-Parameters for ViT Training. We use all original hyper-parameters of \cite{vitbaseline}, except for learning-rate LR.}
\begin{center}
\begin{tabular}{lc}
\hline
\toprule
\textbf{Parameters} & \textbf{Values} \\
\toprule
Optimizer & Adam \\
$\beta_1, \beta_2$ & 0.9, 0.999 \\
Weight Decay & $10^{-4}$ \\
Effective Batch Size & 1024 \\
Drop-out ($p$) & 0.0 \\
Patch Size & 16 \\
Training Image Size & 224x224 \\
Evaluation Image Size & 224x224 \\
Train Epochs & [90, 300] \\
Num GPUs & 8 \\
Learning rate & [1, 2, 3.5, 4]*$10^{-3}$ \\
Schedule & Linear \\
LR Decay Schedule & Cosine \\
Warmup steps & 10000 \\
Min LR & 0.0 \\
Gradient clipping & 1.0 \\
Batch Size / GPU & 128 \\
Augmentation & RandAug(n=2,mag=10)+MixUp(p=0.2) \\

\bottomrule
\end{tabular}
\label{hparams_vit}
\end{center}
\end{table}

\paragraph{Speech Fairseq Training} \cref{hparams_fairseq} provides the hyperparameters used to train the Speech translation models, following those of official fairseq. The same hyper-parameters were used for all the models. We report the BLEU by averaging the weights of the last 10 checkpoints at the end of training.

\begin{table}[H]
\caption{Training Hyper-Parameters for speech-to-text translation. We use all original hyper-parameters in Fairseq, except for effective batch size and learning-rate(LR).}
\begin{center}
\begin{tabular}{lc}
\hline
\toprule
\textbf{Parameters} & \textbf{Values} \\
\toprule
Optimizer & Adam \\
$\beta_1, \beta_2$ & 0.9, 0.999 \\
Source tokens per Batch  & [30k, 40k] \\
Drop-out ($p$) & 0.1 \\
Text Sequence Length & 1024 \\
Speech Sequence Length & 6000 \\
Train Iters & [66k, 100k]  \\
Num GPUs & 1 \\
Learning rate & [3, 5]*$10^{-4}$, [1, 2, 3, 4]*$10^{-3}$ \\
Schedule & Inverse Square-root \\
Warmup steps & 20\% \\
Gradient clipping & 10.0 \\
Batch Size / GPU & [52, 80] \\
Grad Accum Steps & 8,16 \\

\bottomrule
\end{tabular}
\label{hparams_fairseq}
\end{center}
\end{table}

\section{Notations}
\label{sec:notations}
Helpful definitions for notations used in this Manuscript.

$N$ - Number of layers in the transformer network

$L$ - Maximum sequence length for the transformer network

$d / d_{\text{in}}$ - Hidden dimension used to represent a token

$\mu_{x_{\text{in}}}$ - Expected value of single element in the input tensor to a layer/block

$\sigma_{x_{\text{in}}}^2$ - Variance of single element in the input tensor to a layer/block

$\mu_{x_{\text{out}}}$ - Expected value of single element in the output tensor of a layer/block

$\sigma_{x_{\text{out}}}^2$ - Variance of single element in the output tensor of a layer/block

$\mu_{g_{\text{in}}}$ - Expected value of single element in the gradient of input tensor to a layer/block

$\sigma_{g_{\text{in}}}^2$ - Variance of single element in the gradient of input tensor to a layer/block

$\mu_{g_{\text{out}}}$ - Expected value of single element in the gradient of output tensor of a layer/block

$\sigma_{g_{\text{out}}}^2$ - Variance of single element in the gradient of output tensor of a layer/block

$r^l_{x_{\text{in}}}$ - Correlation between two elements in the input tensor to a layer/block having same hidden dimension index but corresponding to different tokens

$r^l_{x_{\text{out}}}$ - Correlation between two elements in the output tensor of a layer/block having same hidden dimension index but corresponding to different tokens

$r^l_{g_{\text{in}}}$ - Correlation between two elements in the gradient of input tensor to a layer/block having same hidden dimension index but corresponding to different tokens

$r^l_{g_{\text{out}}}$ - Correlation between two elements in the gradient of output tensor of a layer/block having same hidden dimension index but corresponding to different tokens

$r^d_{x_{\text{in}}}$ - Correlation between two elements in the input tensor to a layer/block having different hidden dimension indices but corresponding to same token

$r^d_{x_{\text{out}}}$ - Correlation between two elements in the output tensor of a layer/block having different hidden dimension indices but corresponding to same token

$r^d_{g_{\text{in}}}$ - Correlation between two elements in the gradient of input tensor to a layer/block having different hidden dimension indices but corresponding to same token

$r^d_{g_{\text{out}}}$ - Correlation between two elements in the gradient of output tensor of a layer/block having different hidden dimension indices but corresponding to same token

    
    
    
    
 


\end{document}